\documentclass{article}

\PassOptionsToPackage{numbers,sort&compress}{natbib}
\usepackage[preprint]{neurips_2026}

\usepackage[utf8]{inputenc} 
\usepackage[T1]{fontenc}    
\usepackage{hyperref}       
\usepackage{url}            
\usepackage{booktabs}       
\usepackage{amsfonts}       
\usepackage{nicefrac}       
\usepackage{microtype}      
\usepackage{xcolor}         
\usepackage{tabularx}
\usepackage{array}
\usepackage{amsmath}
\usepackage{graphicx}
\usepackage{longtable}
\usepackage{caption}
\usepackage[table]{xcolor}
\usepackage{placeins}
\usepackage{multirow}
\usepackage{makecell}
\usepackage{wrapfig}
\usepackage[most]{tcolorbox}
\usepackage{pifont}

\newcommand{\xmark}{\ding{55}}
\newcommand{\ph}{\cellcolor{gray!12}\textcolor{gray}{--}}
\definecolor{headergray}{RGB}{242,244,247}
\definecolor{rowgray}{RGB}{248,249,251}
\definecolor{oursblue}{RGB}{226,239,255}
\definecolor{rulegray}{RGB}{190,190,190}
\definecolor{yesgreen}{RGB}{30,130,76}
\definecolor{nogray}{RGB}{150,150,150}
\definecolor{totalgray}{RGB}{235,238,242}

\newcommand{\yes}{\textcolor{yesgreen}{\checkmark}}
\newcommand{\no}{\textcolor{nogray}{\xmark}}

\newcolumntype{M}[1]{>{\centering\arraybackslash}m{#1}}
\newcolumntype{L}[1]{>{\raggedright\arraybackslash}m{#1}}

\title{ELDOR: A Dataset and Benchmark for Illegal Gold Mining in the Amazon Rainforest}

\author{%
{\bfseries
Kangning Cui$^{1,2,}$\thanks{Equal contribution.} \enspace 
Surendra Bohara$^{1,*}$ \enspace
Suraj Prasai$^{1}$ \enspace
Zishan Shao$^{3}$ \enspace
Wei Tang$^{4}$} \\
{\bfseries
Martin Pillaca$^{1,10}$ \enspace
Edwin Flores$^{1,10}$ \enspace
Zhen Yang$^{5}$ \enspace
Gregory Larsen$^{6}$ \enspace
Evan Dethier$^{7}$} \\
{\bfseries
David Lutz$^{8}$ \enspace
Jean-Michel Morel$^{9}$ \enspace
Miles Silman$^{1,10}$ \enspace
Victor Pauca$^{1}$ \enspace
Fan Yang$^{1}$} \\
\\[-0.5em]
$^{1}$Wake Forest University \enspace
$^{2}$City University of Hong Kong (Dongguan) \enspace
$^{3}$Duke University  \\
$^{4}$City University of Hong Kong \enspace
$^{5}$Yale University \enspace
$^{6}$Alaska Spatial Science \enspace 
$^{7}$Colby College  \\
$^{8}$Colby-Sawyer College \enspace
$^{9}$Lingnan University \enspace 
$^{10}$Centro de Innovación Cientifica Amazónica
}

\begin{document}

\maketitle

\begin{abstract}
  Illegal gold mining in the Amazon rainforest causes deforestation, water contamination, and long-term ecosystem disruption, yet remains difficult to monitor at fine spatial scales. Satellite imagery supports large-scale observation, but often misses small mining-related structures and subtle land-cover transitions, especially under frequent cloud cover. We introduce \textbf{ELDOR}, a large-scale UAV benchmark for monitoring environmental and landscape disturbance from illegal gold mining in the rainforest. ELDOR contains manually annotated orthomosaic imagery covering over 2{,}500 hectares, with pixel-level semantic labels for both mining-related activities and surrounding ecological structures. With this unified annotation source, we establish four benchmark tasks: semantic segmentation, segmentation-derived recognition, direct multi-label classification, and class-presence recognition with vision-language models. Across these tasks, we compare generic and remote-sensing-specific segmentation models, vision foundation model-related segmentation methods, direct multi-label classification methods, and vision-language models under a controlled closed-set protocol. Results show that current methods still struggle with rare small-scale mining structures and fine-grained recovery classes, suggesting the need for context-aware and multimodal modeling. To support domain analysis and practical use, we further build an interactive explorer for domain experts that provides a unified interface for data exploration and model inference.
\end{abstract}

\vspace{-0.5cm}
\begin{center}
\footnotesize
\texttt{Code:} \href{https://github.com/ckn3/GoldMiningMDD}{github.com/ckn3/GoldMiningMDD} \quad
\texttt{Checkpoints \& Dataset:} \href{https://huggingface.co/IRSC}{huggingface.co/IRSC}
\end{center}

\begin{center}
\small\itshape In sunshine and in shadow, in search of Eldorado.
\end{center}
\vspace{-0.5cm}
\begin{flushright}
\footnotesize --- Edgar Allan Poe, \textit{Eldorado}\hspace{2em}
\end{flushright}

\section{Introduction}
Persistent global demand for gold continues to drive gold mining activities, complicating prior stabilization efforts and intensifying environmental damage, particularly through deforestation and large-scale landscape disruption. This trend is particularly evident in Amazonian rainforest, especially in the Madre de Dios (MDD) region in Peru, which has been widely recognized as a major hotspot of illegal gold mining~\cite{caballero2018deforestation, camalan2022change}. 
The resulting impacts include deforestation~\cite{caballero2018deforestation, sonter2017mining}, water contamination~\cite{atwood2025landscape, dethier2023global, dethier2023operation}, and long-term ecosystem disruption~\cite{siqueira2022strategic, kahhat2019environmental, dethier2019heightened}, with severe consequences for biodiversity and ecosystem health~\cite{dossou2024artisanal}. Moreover, ecological restoration and future economic uses require fine-scale understanding of the types of degradation in an area, going \emph{beyond mined--not mined} mapping. In alluvial gold mining, specific structures such as heavy machinery, sluices, mining rafts, and tailings-like mounds further reveal the modality and intensity of land-use change, rather than only whether mining has occurred. These developments highlight an urgent need for accurate and scalable monitoring of mining activities and their environmental effects.

\begin{figure*}[t]
\centering
\includegraphics[width=\textwidth]{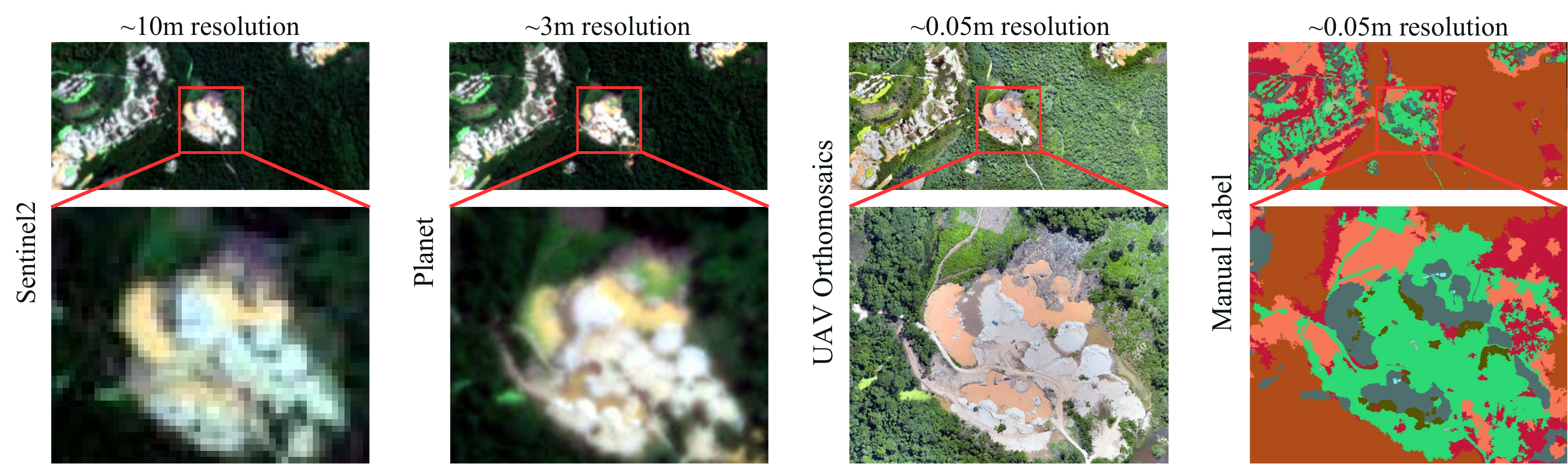}
\caption{
Comparison of spatial resolution across satellite and UAV imagery. 
Sentinel-2 ($\sim$10\,m) and Planet ($\sim$3\,m) imagery miss fine-scale mining structures and subtle land-cover transitions, while the UAV orthomosaic ($\sim$0.05\,m) reveals detailed patterns and their corresponding annotations. Zoom-in views highlight the progressive loss of detail at coarser resolutions. Color mapping follows Table~\ref{tab:class_mapping}.
}
\label{fig:resolution_comparison}
\end{figure*}

Satellite imagery supports large-scale observation, but it is often insufficient for small-scale mining monitoring because of limited spatial resolution and frequent cloud cover~\cite{nursamsi2024remote, obour2025assessing, cui2022semi, cheng2026classifying, zeng2026learning, guan2022classification}. As shown in Figure~\ref{fig:resolution_comparison}, small mining-related structures and subtle land-cover transitions, such as forest degradation and early regeneration, are difficult to identify reliably at satellite resolutions. In tropical regions, frequent cloud cover can further obscure the surface and reduce observation consistency~\cite{shrestha2019land}. In contrast, unmanned aerial vehicle (UAV) imagery provides centimeter-level resolution and reveals fine-grained patterns such as mining rafts, tailings piles, disturbed vegetation, and early-stage regeneration~\cite{nyamekye2021examining}. These are important indicators of mining activities and support the analysis of environmental changes at a local scale that may not be observable in satellite imagery~\cite{atwood2025landscape, cheng2026classifying}.

Existing resources remain inadequate for this problem. Most remote sensing benchmarks focus on coarse land-cover categories and lack annotations for mining-specific structures and disturbed land-cover types needed to distinguish active extraction from post-mining recovery~\cite{camalan2022change, wang2loveda, simionato2021identification, maus2022update}. The task is further complicated by class imbalance, geographic variation, and subtle visual differences among classes such as primary forest, regeneration, and disturbed vegetation~\cite{lacoste2023geo, dossou2024artisanal}. As a result, current computer vision and foundation models remain underexplored in realistic illegal mining settings, while their practical use by ecologists and biologists is still limited by the technical difficulty of deploying models, preparing inputs, and running inference.

To address these limitations, we introduce \textbf{ELDOR}, a large-scale UAV benchmark for monitoring \underline{\textbf{E}}nvironmental and \underline{\textbf{L}}andscape \underline{\textbf{D}}isturbance from illegal g\underline{\textbf{O}}ld mining in the \underline{\textbf{R}}ainforest. To the best of our knowledge, ELDOR is the first large-scale UAV benchmark for mining-related activities with fine-grained semantic classes. It is built from UAV images that we collected and manually annotated across more than 2{,}500 hectares, with pixel-level labels for both mining-related activities and surrounding ecological structures. Based on this unified annotation source, we define four benchmark tasks: semantic segmentation, segmentation-derived recognition, direct multi-label classification, and VLM-based class-presence recognition. We evaluate generic and remote-sensing-specific segmentation methods, methods related to vision foundation models (VFMs), direct multi-label classification methods, and adopt a controlled protocol for VLM-based evaluation. Our results show that existing methods remain challenged by rare small-scale mining structures and fine-grained recovery classes across tasks, highlighting a clear gap between current model capabilities and real-world monitoring needs. We also develop an interactive data explorer that integrates these orthomosaics with trained models for exploration and inference by domain experts. Our main contributions are as follows.

\begin{itemize}
    \item We introduce \textbf{ELDOR}, to the best of our knowledge the first large-scale UAV benchmark dataset with fine-grained semantic classes for illegal gold mining disturbance, built from self-collected imagery and manual annotations over more than 2{,}500 hectares.
    \item We establish a multi-task benchmark on ELDOR, including semantic segmentation, segmentation-derived recognition, direct multi-label classification, and VLM-based class-presence prediction under controlled evaluation protocols.
    \item We evaluate generic, remote-sensing-specific, and foundation-model-based methods, and show that current models remain limited on fine-grained and mining-related categories.
    \item We develop an interactive explorer that integrates these orthomosaics with trained models and allows domain experts to explore ELDOR and run model inference through an interface.
\end{itemize}

\section{Related Work}

\paragraph{Mining-related Remote Sensing Benchmarks}

\begin{wraptable}{r}{0.54\columnwidth}
\centering
\footnotesize
\setlength{\tabcolsep}{2.2pt}
\renewcommand{\arraystretch}{1.05}
\caption{Comparison of mining-related remote sensing datasets. S1: Sentinel-1; S2: Sentinel-2; L8: Landsat 8; L7: Landsat 7; GEP: Google Earth Pro; Res.: spatial resolution; \#Cls: number of classes; Veg.: Vegetation.}
\label{tab:mining_dataset_comparison}
\arrayrulecolor{rulegray}
\resizebox{\linewidth}{!}{%
\begin{tabular}{@{}c l c c c c c c@{}}
\toprule
\rowcolor{headergray}
\textbf{Year} & \textbf{Dataset} & \textbf{Modality} & \makecell[c]{\textbf{Res.}\\\textbf{(m)}} & \textbf{\#Cls} & \makecell[c]{\textbf{Veg.}\\\textbf{types}} & \makecell[c]{\textbf{Waste}\\\textbf{deposits}} & \makecell[c]{\textbf{Structures /}\\\textbf{equipment}} \\
\midrule
2021 & Simionato et al.~\cite{simionato2021identification} & S2       & 10      & 4  & \no & \no & \no \\
\rowcolor{rowgray}
2022 & Maus et al.~\cite{maus2022update}                  & S2       & 10      & 2  & \no & \no & \no \\
2022 & Camalan et al.~\cite{camalan2022change}            & S2       & 10      & 4  & \no & \no & \no \\
\rowcolor{rowgray}
2023 & MineSegSAT~\cite{macdonald2023minesegsat}          & S2       & 10      & 2  & \no & \no & \no \\
2023 & Tang et al.~\cite{tang2023global}                  & S2/GEP   & 1.5--10 & 2  & \no & \no & \no \\
\rowcolor{rowgray}
2024 & MineNetCD~\cite{yu2024minenetcd}                   & S2       & 10      & 2  & \no & \no & \no \\
2024 & Becerra et al.~\cite{becerra2024creating}          & S1       & 10      & 2  & \no & \no & \no \\
\rowcolor{rowgray}
2025 & SmallMinesDS~\cite{ofori2025smallminesds}          & S1/S2    & 10      & 2  & \no & \no & \no \\
2025 & Saputra et al.~\cite{saputra2025multi}             & S2/L8/L7 & 10--15  & 9  & \no & \yes & \yes \\
\rowcolor{rowgray}
2026 & Cheng et al.~\cite{cheng2026classifying}           & S2       & 10      & 7  & \no & \yes & \yes \\
\rowcolor{oursblue}
2026 & \textbf{ELDOR (ours)}                              & \textbf{UAV RGB} & \textbf{$\sim$ 0.05} & \textbf{14} & \yes & \yes & \yes \\
\bottomrule
\end{tabular}%
}
\arrayrulecolor{black}
\end{wraptable}

Most existing mining datasets are built from satellite imagery at around 10\,m resolution and fall into three groups: binary mining detection~\cite{maus2022update, macdonald2023minesegsat, tang2023global, becerra2024creating, ofori2025smallminesds}, coarse multi-class mapping~\cite{simionato2021identification, saputra2025multi, cheng2026classifying}, and temporal change detection~\cite{camalan2022change, yu2024minenetcd}. As summarized in Table~\ref{tab:mining_dataset_comparison}, these datasets remain limited in both spatial resolution and semantic granularity. Some include broader waste- or infrastructure-related categories~\cite{saputra2025multi, cheng2026classifying}, but satellite imagery does not support finer distinctions such as recovery stages, different material piles, or small mining-associated targets like mining rafts, buildings, and sluices. UAV imagery offers a natural way to address this gap because of its much higher spatial resolution and flexible deployment~\cite{ren2019review, perikleous2025aerial, cui2025detection, zhu2025orthomosaics}. However, existing UAV studies remain small-scale and rarely release orthomosaics or fine-grained annotations, which limits systematic evaluation and broader method development. In contrast, ELDOR is built from self-collected UAV orthomosaics at centimeter-level resolution and provides fine-grained semantic masks for these categories.

\paragraph{Paradigms for Aerial Scene Understanding}

Semantic segmentation remains the standard formulation for dense interpretation of aerial and remote sensing imagery, and has been widely studied for land-cover mapping and environmental monitoring~\cite{camalan2022change, polk2023unsupervised, cui2024superpixel, wang2026dinov3}. Existing methods span CNN-based, Transformer-based, and more recent state-space architectures~\cite{chen2018encoder, xie2021segformer, liu2024vmamba, jiao2023pearl}, with many general-purpose designs adapted to remote sensing settings~\cite{wang2022unetformer, cui2025efficient, zhang2026center, wang2025pyramidmamba}. Multi-label classification provides a complementary image-level formulation for patches containing multiple co-occurring categories, and includes both general visual recognition models and remote-sensing-specific variants~\cite{lanchantin2021general, zhu2021residual, ridnik2023ml, hua2020relation, kang2020graph, liu2024semantic, gupta2025mosaic}. More recently, VLM-based methods have introduced a third paradigm based on language-conditioned recognition and reasoning, although their applicability to remote sensing remains less established~\cite{lacoste2023geo, liu2024remoteclip, zhang2024rs5m, kuckreja2024geochat, bazi2024rs, pang2025vhm}. These paradigms differ in supervision and output format, from dense segmentation to image-level recognition and language-based prediction, which makes a unified benchmark necessary for systematic comparison under the same data and evaluation setting.

\section{ELDOR: A UAV Dataset for Illegal Gold Mining}
\label{sec:dataset}

\paragraph{Study Region and Data Collection}

To study spatial patterns of illegal gold mining and associated ecological transformations at fine spatial scales, we conducted UAV surveys across multiple sites in the MDD region of the Peruvian Amazon. All sites lie within the same mining region and are geographically close with no overlap. This supports cross-site comparison under broadly consistent regional conditions. The geographic extent of each site, including longitude and latitude ranges, is provided in the Appendix~\ref{app:data} (Table~\ref{tab:site_cards}). Field teams collected high-resolution RGB imagery using an EVO II Pro 6K platform (Autel, Shenzhen, China) equipped with a Sony IMX383 sensor (1-inch CMOS, $5472 \times 3648$ pixels). The resulting orthomosaics have a ground sampling distance ranging from approximately 0.03 m to 0.08 m per pixel, corresponding to centimeter-level spatial resolution. The resulting orthomosaics cover large contiguous areas, often spanning tens to hundreds of hectares per site. Flight missions followed standard photogrammetric survey practice to ensure sufficient coverage for reconstruction. Flight plans were created and executed using Mission Planner. Images were processed into orthomosaics using Agisoft Metashape with a standard structure-from-motion (SfM) and multi-view stereo (MVS) workflow. The orthomosaics were then georeferenced and aligned to a satellite basemap using ArcGIS Pro, so that all sites share a consistent spatial reference.

\begin{table*}[t]
\centering
\scriptsize
\caption{
Semantic classes in ELDOR with color codes, abbreviations, spatial coverage, and descriptions. 
Color indicates the annotation mask visualization. 
Area and Pixels denote total coverage across all sites, and (\%) represents the dataset proportion of each class.
}
\label{tab:class_mapping}
\setlength{\tabcolsep}{3pt}
\renewcommand{\arraystretch}{1.1}
\arrayrulecolor{rulegray}

\resizebox{\linewidth}{!}{%
\begin{tabular}{c c c l c c c c}
\toprule
\rowcolor{headergray}
\textbf{ID} & \textbf{Color} & \textbf{Abbr.} & \textbf{Class} & \textbf{Area (ha)} & \textbf{(\%)} & \textbf{Pixels} & \textbf{Brief Description} \\
\midrule
1  & \cellcolor[HTML]{8A6A3D} & BU  & Buildings           & 2.3745   & 0.09  & 6,367,488     & Mining camps and support structures. \\
\rowcolor{rowgray}
2  & \cellcolor[HTML]{7BEBFB} & MR  & Mining rafts        & 0.1467   & 0.01  & 461,925       & Floating platforms for active gold extraction. \\
3  & \cellcolor[HTML]{B04C18} & PF  & Primary forests     & 961.8239 & 38.45 & 3,308,805,552 & Intact forest, used as a baseline for disturbance. \\
\rowcolor{rowgray}
4  & \cellcolor[HTML]{EE92C6} & HM  & Heavy machinery     & 0.0161   & 0.00  & 43,313        & Excavation and transport equipment indicating active mining. \\
5  & \cellcolor[HTML]{4F6F6F} & WB  & Water bodies        & 321.2655 & 12.84 & 1,085,789,506 & Rivers, ponds, and pools often altered by mining. \\
\rowcolor{rowgray}
6  & \cellcolor[HTML]{84D08C} & AC  & Agricultural crops  & 33.5185  & 1.34  & 105,688,237   & Cultivated land reflecting surrounding human land use. \\
7  & \cellcolor[HTML]{23F3E3} & CM  & Compact mounds      & 66.2310  & 2.65  & 215,787,431   & Large processed material piles from heavy machinery. \\
\rowcolor{rowgray}
8  & \cellcolor[HTML]{585400} & GM  & Gravel mounds       & 13.2680  & 0.53  & 42,049,847    & Smaller residue piles left by washing processes. \\
9  & \cellcolor[HTML]{8DB51D} & GR  & Grass               & 28.3472  & 1.13  & 87,888,718    & Pasture or low vegetation in cleared areas. \\
\rowcolor{rowgray}
10 & \cellcolor[HTML]{C2163A} & T1R & Type 1 regeneration & 228.5563 & 9.14  & 660,993,175   & Early-stage recovery in recently disturbed areas. \\
11 & \cellcolor[HTML]{F77757} & T2R & Type 2 regeneration & 523.1002 & 20.91 & 1,710,062,485 & Later-stage recovery with shrubs and young trees. \\
\rowcolor{rowgray}
12 & \cellcolor[HTML]{2CD874} & BG  & Bare ground         & 322.9909 & 12.91 & 882,535,708   & Exposed soil strongly associated with mining disturbance. \\
13 & \cellcolor[HTML]{613991} & SL  & Sluices             & 0.0403   & 0.00  & 126,368       & Washing infrastructure for gold-bearing material. \\
\rowcolor{rowgray}
14 & \cellcolor[HTML]{969AAE} & VE  & Vehicles            & 0.0092   & 0.00  & 85,244        & Transport vehicles supporting mining logistics. \\
\midrule
\rowcolor{totalgray}
\textbf{Total} &  &  &  & \textbf{2,500.6} & \textbf{100.00} & \textbf{8,106,684,997} &  \\
\bottomrule
\end{tabular}%
}

\arrayrulecolor{black}
\end{table*}

\paragraph{Annotation Protocol and Semantic Taxonomy}

The orthomosaics were annotated using a polygon-based workflow overlaid on the UAV imagery, following standard geospatial annotation practice (e.g., ArcGIS). The labels were primarily produced by collaborators at the Centro de Innovación Cientifica Amazónica (CINCIA), who have domain expertise in Amazonian mining systems and remote sensing interpretation. Class assignment was guided by expert interpretation based on visual characteristics (e.g., texture, structure, and spatial context), together with domain knowledge of mining processes and ecological recovery, with efforts made to maintain consistency across sites, and annotation decisions were cross-checked among annotators to reduce systematic discrepancies.

In practice, the semantic taxonomy was refined iteratively through discussion among annotators and collaborators, especially for visually ambiguous categories such as early-stage regeneration and bare ground. The resulting 14 classes are defined with a focus on mining-related features and their associated landscape transformations. A summary is provided in Table~\ref{tab:class_mapping}, and detailed descriptions with representative examples are given in the Appendix~\ref{app:data} (Table~\ref{tab:class_cards}).

\paragraph{Dataset Statistics and Usage}

\begin{figure}[t]
\centering

\begin{minipage}[t]{0.7\linewidth}
\vspace{0pt}
\centering
\includegraphics[width=\linewidth]{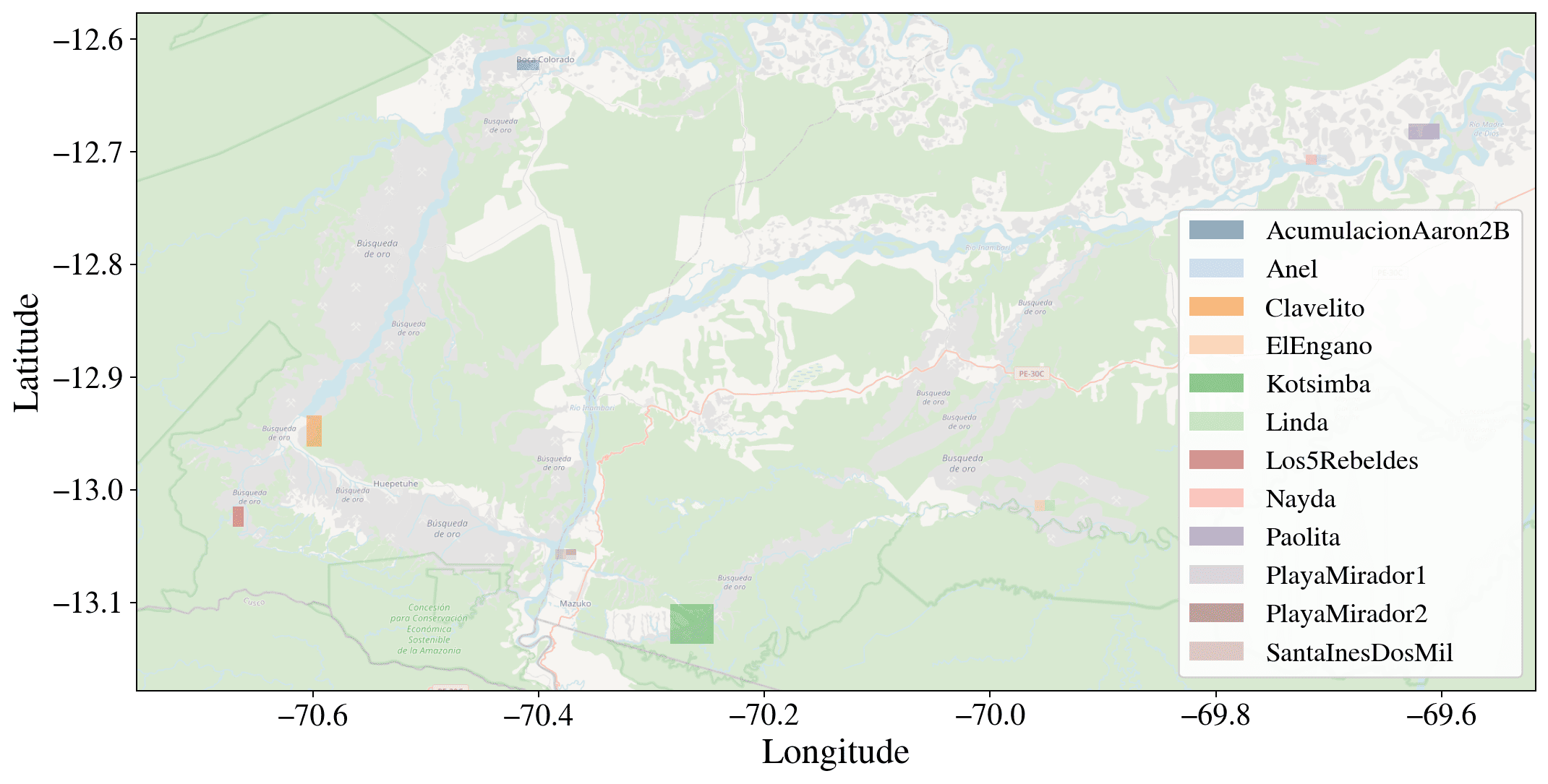}
\end{minipage}
\hfill
\begin{minipage}[t]{0.29\linewidth}
\vspace{0pt}
\centering

\includegraphics[width=\linewidth]{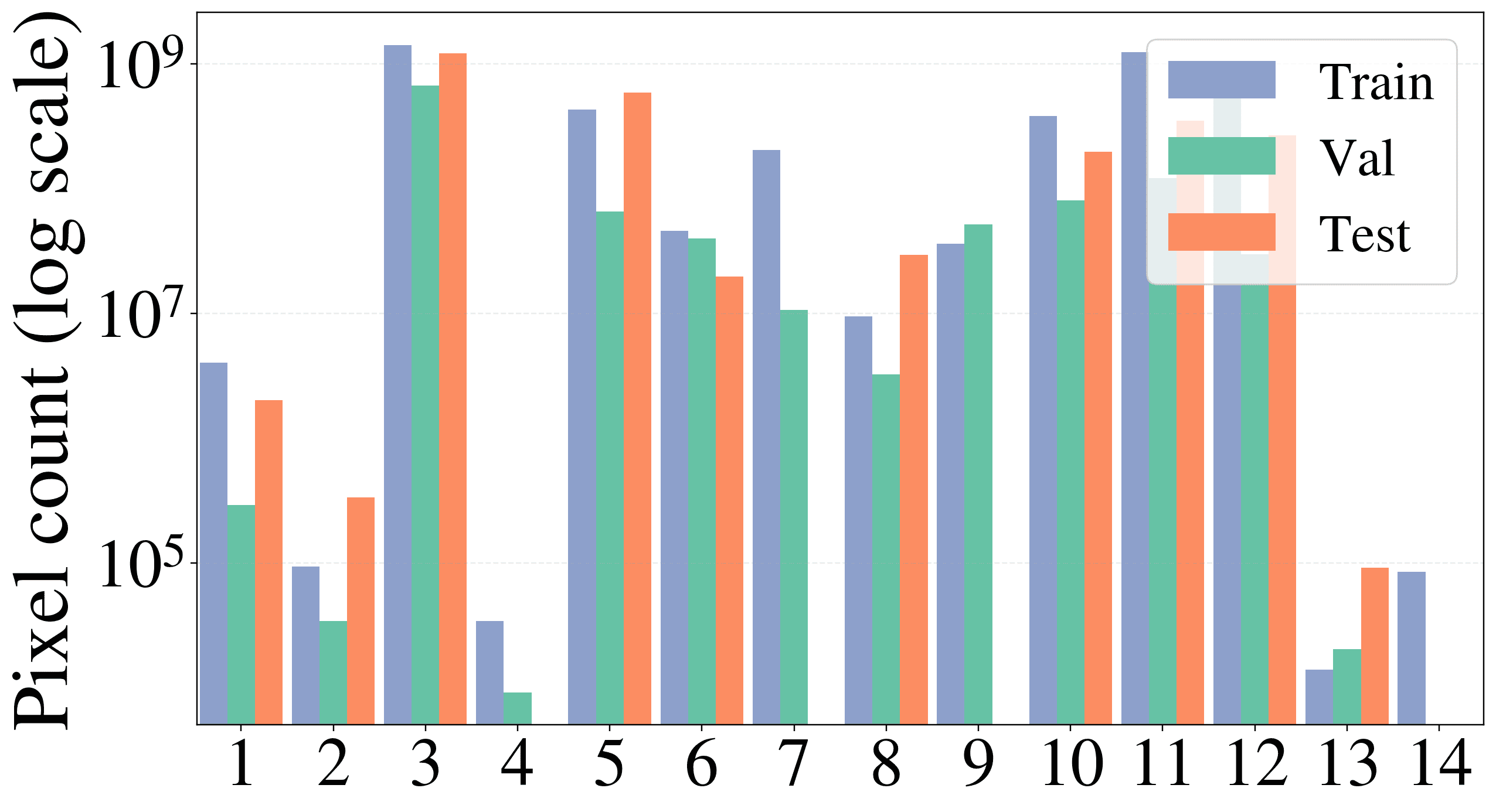}

\vspace{0.6em}

\scriptsize
\setlength{\tabcolsep}{3pt}
\renewcommand{\arraystretch}{1.05}
\arrayrulecolor{rulegray}
\begin{tabular}{lcc}
\toprule
\rowcolor{headergray}
\textbf{Split} & \textbf{Number of Sites} & \textbf{Patches} \\
\midrule
Train & 4 & 65,798 \\
Val   & 3 & 15,988 \\
Test  & 5 & 40,095 \\
\midrule
\rowcolor{totalgray}
\textbf{Total} & \textbf{12} & \textbf{121,881} \\
\bottomrule
\end{tabular}
\arrayrulecolor{black}

\end{minipage}

\caption{
Left: spatial distribution of the study sites in the MDD region.
Top right: class distribution across dataset splits.
Bottom right: patch statistics after preprocessing for each split.
}
\label{fig:dataset_overview}
\end{figure}

ELDOR covers more than 2,500 hectares and contains over 8.1 billion labeled pixels. The data exhibit strong class imbalance, see Figure~\ref{fig:dataset_overview} (top-right). Dominant classes such as primary forest and water bodies occupy a large portion of the area, while some mining-related classes (e.g., buildings, mining rafts, and sluices) appear only sparsely. These patterns reflect multiple stages of mining and recovery, resulting in fine-grained distinctions between visually similar categories; detailed site-level class distributions and class-wise RGB statistics are provided in Appendix~\ref{app:data} (Figures~\ref{fig:site_class_distribution} and~\ref{fig:class_rgb_distribution}). The dataset contains 12 sites and is split into training, validation, and test sets at the site level, with each site containing a distinct subset of semantic classes (Appendix~\ref{app:data}, Table~\ref{tab:site_cards}). The split avoids spatial overlap while ensuring that all semantic classes are represented in the training set. As shown in Figure~\ref{fig:dataset_overview}, all sites lie within the same MDD region but remain spatially disjoint. All 14 classes are present in the training set, while the validation set contains 13 classes, with vehicles not observed, and the test set contains 10 classes, with vehicles, heavy machinery, compact mounds, and grass absent. These missing categories are mobile or site-dependent, and their absence reflects natural variability of mining activity across locations rather than annotation bias. This makes the test set more focused on ambiguous and practically relevant categories.

To support model development, the orthomosaics are divided into fixed-size patches using a sliding-window scheme with a size of $512 \times 512$ pixels and a stride of 256. Patches with more than 80\% no-image background pixels are discarded to reduce trivial samples. The remaining patches are used to construct both semantic segmentation and multi-label recognition datasets. For segmentation, each patch is paired with a dense pixel-level label map. For recognition, labels are derived from the presence of semantic classes within each patch. This formulation allows the dataset to support both fine-grained spatial understanding and holistic semantic recognition within a unified framework. After preprocessing, the dataset contains 65,798 training patches, 15,988 validation patches, and 40,095 test patches, as shown in Figure~\ref{fig:dataset_overview} (bottom-right). These properties introduce challenges including class imbalance, fine-grained visual ambiguity, small object detection, and cross-site generalization. Appendix analysis further shows that difficulty arises not only from long-tailed class frequencies, but also from overlap in appearance among ecologically and operationally related categories.

\section{Benchmark Tasks}

ELDOR supports multiple levels of semantic understanding derived from a unified set of annotations. These tasks span dense spatial prediction, image-level recognition, and language-conditioned recognition, which enables direct comparison across different supervision levels and output formats. 
Segmentation outputs can also be converted into class-presence vectors, allowing us to examine whether accurate spatial modeling leads to reliable semantic recognition. All tasks are integrated into an interactive explorer that supports data exploration and model inference in a unified environment.

\paragraph{Task 1: Semantic Segmentation}

Semantic segmentation is formulated as a dense pixel-wise prediction over the predefined taxonomy. The task requires models to capture both small objects, such as buildings and mining rafts, and broader land-cover patterns such as forest, water, and regeneration. We evaluate performance using mean Intersection over Union (mIoU) over all classes and over classes present in each image. We also report overall accuracy (OA) and macro F1 score. To account for practical deployment, we report model efficiency, including parameter count, GFLOPs, latency, and peak VRAM usage. Metric definitions and evaluation details are provided in Appendix~\ref{app:seg_protocol}.

\paragraph{Task 2: Segmentation-derived Recognition}

Segmentation-derived recognition converts predicted masks into image-level class-presence vectors using a predefined criterion. This makes the outputs directly comparable to multi-label classification and allows us to test whether dense supervision improves image-level recognition. We evaluate performance using class-wise precision (CP), recall (CR), and F1 (CF1), overall precision (OP), recall (OR), and F1 (OF1), mean average precision (mAP), and aggregated metrics including macro, micro, and sample-wise F1. Details of the presence definition, score computation, evaluation protocol, and per-class results are given in Appendix~\ref{app:seg_rec_protocol}.

\paragraph{Task 3: Direct Multi-label Classification}

Direct multi-label classification predicts a binary vector of class presence for each image patch without pixel-level supervision. This formulation focuses on semantic content and provides a more efficient alternative to segmentation for large-scale analysis. In contrast to segmentation-derived recognition, this task relies on sparse image-level supervision, which allows us to assess the role of dense annotations in improving recognition performance. We evaluate performance using the same metrics as in segmentation-derived recognition. When available, we also report model efficiency to complement accuracy results. Details of the training setup, loss functions, and evaluation protocol are provided in Appendix~\ref{app:mlc_protocol}.

\paragraph{Task 4: VLM-based Recognition}

VLM-based recognition evaluates whether vision-language models can infer image-level class presence through prompt-based queries under the same closed label space. In contrast to direct multi-label classification, this setting does not use task-specific training, allowing us to assess zero-shot remote-sensing foundation models on mining-related semantics. We consider both prompt-based image-level recognition and segmentation-capable VLMs under a unified protocol. Performance is measured using the same metrics as in the other recognition tasks. Figures~\ref{fig:vlm_prompt_recognition} and~\ref{fig:vlm_prompt_segmentation} summarize the protocols, and Appendix~\ref{app:vlm_protocol} provides details.

\section{Experiments}
\label{sec:experiments}

\subsection{Implementation Details}
We follow the site-level split and patch construction protocol described in Section~\ref{sec:dataset}. For semantic segmentation, all models are trained on $512 \times 512$ patches. We use AdamW with a base learning rate of $2\times10^{-4}$, weight decay 0.05, and a warmup followed by cosine decay schedule. Unless otherwise required by a specific framework, models are trained for 80 epochs with automatic mixed precision. Training uses the same augmentation policy across methods. For methods that support direct loss comparison, we evaluate multiple loss settings; for methods with native set-prediction or framework-specific objectives, we keep their original training loss. All experiments are run on NVIDIA L40S GPUs. Full training and evaluation protocols, augmentation details, checkpoint selection rules, and loss definitions are provided in Appendix~\ref{app:protocol}.

\begin{figure*}[t]
\centering

\begin{minipage}[t]{0.5\textwidth}
\centering
\captionsetup{type=table}
\scriptsize
\caption{Selected segmentation and derived classification results. The first four metric columns report segmentation and the last three the corresponding classification results. Shaded blocks indicate general, remote-sensing-specific, and VFM-based methods; darker rows are compared in the right panel; $^{*}$ denotes an auxiliary loss and $^{\dagger}$ pretrained models with post-processing only.}
\label{tab:main_selected_combined}

\setlength{\tabcolsep}{2.5pt}
\renewcommand{\arraystretch}{1.05}

\definecolor{generalbg}{RGB}{247,249,252}
\definecolor{rsbg}{RGB}{245,250,245}
\definecolor{vfmbg}{RGB}{252,247,242}

\definecolor{generalhi}{RGB}{234,240,248}
\definecolor{rshi}{RGB}{232,243,232}
\definecolor{vfmhi}{RGB}{248,237,227}
\arrayrulecolor{rulegray}
\resizebox{\linewidth}{!}{%
\begin{tabular}{@{}clccccccc@{}}
\toprule
\multirow{2}{*}{\textbf{Model}} & \multirow{2}{*}{\textbf{Loss}} 
& \multicolumn{4}{c}{\textbf{Segmentation}} 
& \multicolumn{3}{c}{\textbf{Classification}} \\
\cmidrule(lr){3-6}\cmidrule(l){7-9}
& & \textbf{mIoU$_p$} & \textbf{mIoU} & \textbf{Macro-F1} & \textbf{OA} 
& \textbf{CF1} & \textbf{OF1} & \textbf{mAP} \\
\midrule

\cellcolor{generalbg} & \cellcolor{generalbg} CE+Dice
& \cellcolor{generalbg} 0.322 & \cellcolor{generalbg} 0.230 & \cellcolor{generalbg} 0.459 & \cellcolor{generalbg} 0.648
& \cellcolor{generalbg} 0.575 & \cellcolor{generalbg} 0.707 & \cellcolor{generalbg} 0.538 \\
\cellcolor{generalbg} & \cellcolor{generalbg} Focal+Dice
& \cellcolor{generalbg} 0.333 & \cellcolor{generalbg} 0.238 & \cellcolor{generalbg} 0.476 & \cellcolor{generalbg} 0.639
& \cellcolor{generalbg} 0.601 & \cellcolor{generalbg} 0.705 & \cellcolor{generalbg} 0.586 \\
\cellcolor{generalbg}\multirow[c]{-3}{*}{SegFormer} & \cellcolor{generalhi} WCE+Dice
& \cellcolor{generalhi} \textbf{0.401} & \cellcolor{generalhi} \textbf{0.308} & \cellcolor{generalhi} \textbf{0.530} & \cellcolor{generalhi} 0.716
& \cellcolor{generalhi} 0.618 & \cellcolor{generalhi} \textbf{0.754} & \cellcolor{generalhi} \textbf{0.658} \\

\cellcolor{generalbg} & \cellcolor{generalbg} CE+Dice
& \cellcolor{generalbg} 0.380 & \cellcolor{generalbg} 0.271 & \cellcolor{generalbg} 0.507 & \cellcolor{generalbg} \textbf{0.726}
& \cellcolor{generalbg} 0.595 & \cellcolor{generalbg} 0.747 & \cellcolor{generalbg} 0.600 \\
\cellcolor{generalbg} & \cellcolor{generalbg} Focal+Dice
& \cellcolor{generalbg} 0.369 & \cellcolor{generalbg} 0.264 & \cellcolor{generalbg} 0.509 & \cellcolor{generalbg} 0.678
& \cellcolor{generalbg} 0.625 & \cellcolor{generalbg} 0.732 & \cellcolor{generalbg} 0.610 \\
\cellcolor{generalbg}\multirow[c]{-3}{*}{EfficientViT} & \cellcolor{generalhi} WCE+Dice
& \cellcolor{generalhi} 0.379 & \cellcolor{generalhi} 0.271 & \cellcolor{generalhi} 0.518 & \cellcolor{generalhi} 0.715
& \cellcolor{generalhi} \textbf{0.632} & \cellcolor{generalhi} 0.749 & \cellcolor{generalhi} 0.623 \\

\midrule

\cellcolor{rsbg} & \cellcolor{rshi} CE+Dice$^{*}$
& \cellcolor{rshi} 0.394 & \cellcolor{rshi} \textbf{0.328} & \cellcolor{rshi} 0.515 & \cellcolor{rshi} 0.728
& \cellcolor{rshi} 0.575 & \cellcolor{rshi} \textbf{0.763} & \cellcolor{rshi} 0.591 \\
\cellcolor{rsbg} & \cellcolor{rsbg} Focal+Dice$^{*}$
& \cellcolor{rsbg} 0.377 & \cellcolor{rsbg} 0.314 & \cellcolor{rsbg} 0.499 & \cellcolor{rsbg} 0.718
& \cellcolor{rsbg} 0.547 & \cellcolor{rsbg} 0.728 & \cellcolor{rsbg} 0.581 \\
\cellcolor{rsbg}\multirow[c]{-3}{*}{UNetFormer} & \cellcolor{rsbg} WCE+Dice$^{*}$
& \cellcolor{rsbg} 0.357 & \cellcolor{rsbg} 0.274 & \cellcolor{rsbg} 0.493 & \cellcolor{rsbg} 0.681
& \cellcolor{rsbg} 0.557 & \cellcolor{rsbg} 0.711 & \cellcolor{rsbg} 0.612 \\

\cellcolor{rsbg} & \cellcolor{rsbg} CE+Dice
& \cellcolor{rsbg} 0.371 & \cellcolor{rsbg} 0.309 & \cellcolor{rsbg} 0.492 & \cellcolor{rsbg} 0.686
& \cellcolor{rsbg} 0.534 & \cellcolor{rsbg} 0.719 & \cellcolor{rsbg} 0.556 \\
\cellcolor{rsbg} & \cellcolor{rshi} Focal+Dice
& \cellcolor{rshi} 0.400 & \cellcolor{rshi} 0.286 & \cellcolor{rshi} 0.543 & \cellcolor{rshi} 0.685
& \cellcolor{rshi} 0.577 & \cellcolor{rshi} 0.701 & \cellcolor{rshi} 0.617 \\
\cellcolor{rsbg}\multirow[c]{-3}{*}{MANet} & \cellcolor{rsbg} WCE+Dice
& \cellcolor{rsbg} 0.383 & \cellcolor{rsbg} 0.273 & \cellcolor{rsbg} 0.523 & \cellcolor{rsbg} 0.676
& \cellcolor{rsbg} 0.584 & \cellcolor{rsbg} 0.703 & \cellcolor{rsbg} 0.682 \\

\cellcolor{rsbg} & \cellcolor{rsbg} CE+Dice
& \cellcolor{rsbg} 0.399 & \cellcolor{rsbg} 0.285 & \cellcolor{rsbg} 0.536 & \cellcolor{rsbg} 0.683
& \cellcolor{rsbg} 0.616 & \cellcolor{rsbg} 0.734 & \cellcolor{rsbg} 0.630 \\
\cellcolor{rsbg} & \cellcolor{rsbg} Focal+Dice
& \cellcolor{rsbg} 0.396 & \cellcolor{rsbg} 0.283 & \cellcolor{rsbg} 0.530 & \cellcolor{rsbg} 0.670
& \cellcolor{rsbg} 0.570 & \cellcolor{rsbg} 0.721 & \cellcolor{rsbg} 0.636 \\
\cellcolor{rsbg}\multirow[c]{-3}{*}{PyramidMamba} & \cellcolor{rshi} WCE+Dice
& \cellcolor{rshi} \textbf{0.441} & \cellcolor{rshi} 0.315 & \cellcolor{rshi} \textbf{0.586} & \cellcolor{rshi} 0.697
& \cellcolor{rshi} \textbf{0.651} & \cellcolor{rshi} 0.720 & \cellcolor{rshi} \textbf{0.702} \\

\midrule

\cellcolor{vfmbg} & \cellcolor{vfmbg} CE+Dice
& \cellcolor{vfmbg} 0.326 & \cellcolor{vfmbg} 0.251 & \cellcolor{vfmbg} 0.447 & \cellcolor{vfmbg} 0.698
& \cellcolor{vfmbg} 0.512 & \cellcolor{vfmbg} 0.716 & \cellcolor{vfmbg} 0.535 \\
\cellcolor{vfmbg} & \cellcolor{vfmhi} Focal+Dice
& \cellcolor{vfmhi} 0.345 & \cellcolor{vfmhi} 0.265 & \cellcolor{vfmhi} 0.464 & \cellcolor{vfmhi} \textbf{0.743}
& \cellcolor{vfmhi} 0.590 & \cellcolor{vfmhi} 0.759 & \cellcolor{vfmhi} 0.570 \\
\cellcolor{vfmbg}\multirow[c]{-3}{*}{RSAM-Seg} & \cellcolor{vfmbg} WCE+Dice
& \cellcolor{vfmbg} 0.370 & \cellcolor{vfmbg} 0.264 & \cellcolor{vfmbg} 0.509 & \cellcolor{vfmbg} 0.698
& \cellcolor{vfmbg} 0.592 & \cellcolor{vfmbg} 0.726 & \cellcolor{vfmbg} 0.621 \\

\cellcolor{vfmbg} & \cellcolor{vfmbg} CE+Dice
& \cellcolor{vfmbg} 0.360 & \cellcolor{vfmbg} 0.300 & \cellcolor{vfmbg} 0.482 & \cellcolor{vfmbg} 0.678
& \cellcolor{vfmbg} 0.527 & \cellcolor{vfmbg} 0.720 & \cellcolor{vfmbg} 0.556 \\
\cellcolor{vfmbg} & \cellcolor{vfmhi} Focal+Dice
& \cellcolor{vfmhi} 0.403 & \cellcolor{vfmhi} 0.288 & \cellcolor{vfmhi} 0.547 & \cellcolor{vfmhi} 0.685
& \cellcolor{vfmhi} 0.577 & \cellcolor{vfmhi} 0.701 & \cellcolor{vfmhi} 0.617 \\
\cellcolor{vfmbg}\multirow[c]{-3}{*}{\makecell[c]{SESSRS$^{\dagger}$\\(MANet)}} & \cellcolor{vfmbg} WCE+Dice
& \cellcolor{vfmbg} 0.384 & \cellcolor{vfmbg} 0.274 & \cellcolor{vfmbg} 0.524 & \cellcolor{vfmbg} 0.676
& \cellcolor{vfmbg} 0.584 & \cellcolor{vfmbg} 0.703 & \cellcolor{vfmbg} \textbf{0.682} \\

\cellcolor{vfmbg} & \cellcolor{vfmhi} CE+Dice$^{*}$
& \cellcolor{vfmhi} 0.396 & \cellcolor{vfmhi} \textbf{0.330} & \cellcolor{vfmhi} 0.517 & \cellcolor{vfmhi} 0.728
& \cellcolor{vfmhi} 0.575 & \cellcolor{vfmhi} \textbf{0.763} & \cellcolor{vfmhi} 0.591 \\
\cellcolor{vfmbg} & \cellcolor{vfmbg} Focal+Dice$^{*}$
& \cellcolor{vfmbg} 0.387 & \cellcolor{vfmbg} 0.323 & \cellcolor{vfmbg} 0.509 & \cellcolor{vfmbg} 0.720
& \cellcolor{vfmbg} 0.547 & \cellcolor{vfmbg} 0.728 & \cellcolor{vfmbg} 0.581 \\
\cellcolor{vfmbg}\multirow[c]{-3}{*}{\makecell[c]{SESSRS$^{\dagger}$\\(UNetFormer)}} & \cellcolor{vfmbg} WCE+Dice$^{*}$
& \cellcolor{vfmbg} 0.358 & \cellcolor{vfmbg} 0.275 & \cellcolor{vfmbg} 0.494 & \cellcolor{vfmbg} 0.682
& \cellcolor{vfmbg} 0.557 & \cellcolor{vfmbg} 0.711 & \cellcolor{vfmbg} 0.612 \\

\bottomrule
\end{tabular}%
}
\arrayrulecolor{black}
\end{minipage}
\hfill
\begin{minipage}[t]{0.48\textwidth}
\centering
\captionsetup{type=figure}
\includegraphics[width=\linewidth]{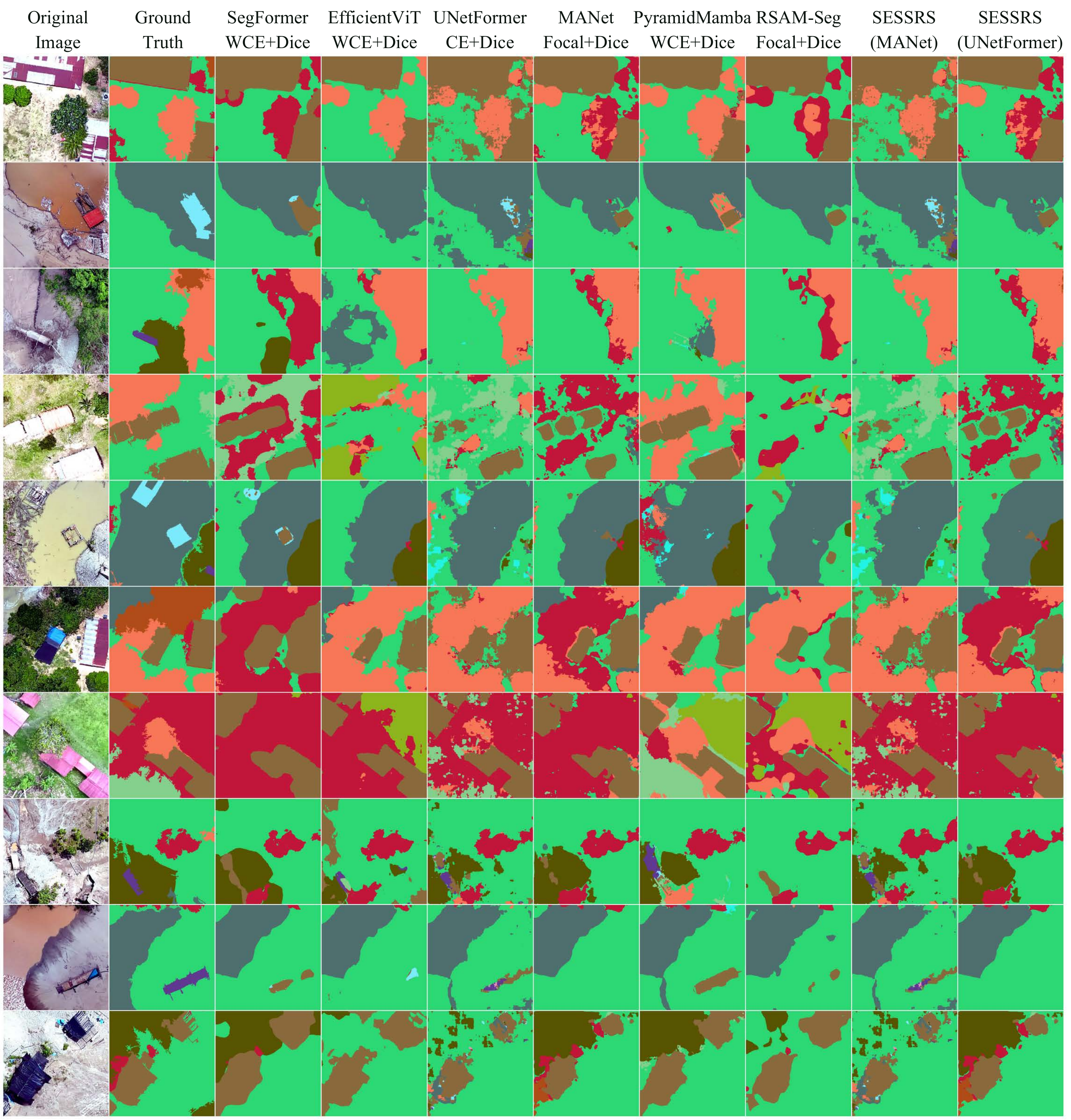}
\caption{Qualitative comparison of representative UAV patches for the highlighted configurations in Table~\ref{tab:main_selected_combined}. From left to right are the original image, ground truth, and predictions from the selected methods. While major classes are often recovered reasonably well, small or rare classes remain difficult across all methods.}
\label{fig:qualitative_selected}
\end{minipage}

\end{figure*}

\subsection{Results on Segmentation and Derived Recognition}

We evaluate semantic segmentation across three method families: general segmentation models, remote-sensing-specific models, and methods related to recent VFMs. We further assess the same models under a segmentation-derived recognition protocol, where predicted masks are converted into image-level class-presence vectors and evaluated under the same label space and metrics as direct multi-label classification (Appendix~\ref{app:seg_rec_protocol}). This setting tests whether accurate dense prediction can also support reliable semantic recognition. For models that support standard supervised training, we consider three loss settings: cross-entropy with Dice loss (CE+Dice), focal loss with Dice loss (Focal+Dice), and weighted cross-entropy with Dice loss (WCE+Dice), while retaining native objectives for methods with framework-specific designs (Appendix~\ref{app:seg_protocol}).

\textbf{General segmentation methods} include DeepLabV3+~\cite{chen2018encoder}, UPerNet~\cite{xiao2018unified}, OCRNet~\cite{yuan2020object}, BiSeNetv2~\cite{yu2021bisenet}, SegFormer~\cite{xie2021segformer}, STDC2~\cite{fan2021rethinking}, Mask2Former~\cite{cheng2022masked}, SegNeXt~\cite{guo2022segnext}, DDRNet~\cite{pan2022deep}, Afformer~\cite{dong2023head}, EfficientViT~\cite{cai2023efficientvit}, SeaFormer~\cite{wan2023seaformer}, PIDNet~\cite{xu2023pidnet}, CGRSeg~\cite{ni2024context}, PEM~\cite{cavagnero2024pem}, and VMamba~\cite{liu2024vmamba}. \textbf{Remote-sensing-specific methods} include FarSeg~\cite{zheng2020foreground}, BANet~\cite{wang2021transformer}, ABCNet~\cite{li2021abcnet}, MANet~\cite{li2021multiattention}, UNetFormer~\cite{wang2022unetformer}, DC-Swin~\cite{wang2022novel}, A2-FPN~\cite{li2022a2}, LoGCAN~\cite{ma2023log}, FarSeg++~\cite{zheng2023farseg++}, SACANet~\cite{ma2023sacanet}, DOCNet~\cite{ma2024docnet}, PPMambaSeg~\cite{mu2024ppmamba}, RS3Mamba~\cite{ma2024rs}, PyramidMamba~\cite{wang2025pyramidmamba}, LoGCAN++~\cite{ma2025logcan++}, MF-Mamba~\cite{xiao2025mf}, and MCPNet~\cite{zhang2025asymmetric}. \textbf{Methods related to VFMs} include HQ-SAM~\cite{ke2023segment}, SAM\_RS~\cite{ma2024sam}, SAM2~\cite{ravi2024sam}, RSAM-Seg~\cite{zhang2025rsam}, and SESSRS~\cite{qiao2025sam}.

\begin{figure*}[t]
\centering

\begin{minipage}[t]{0.7\textwidth}
\centering
\includegraphics[width=\linewidth]{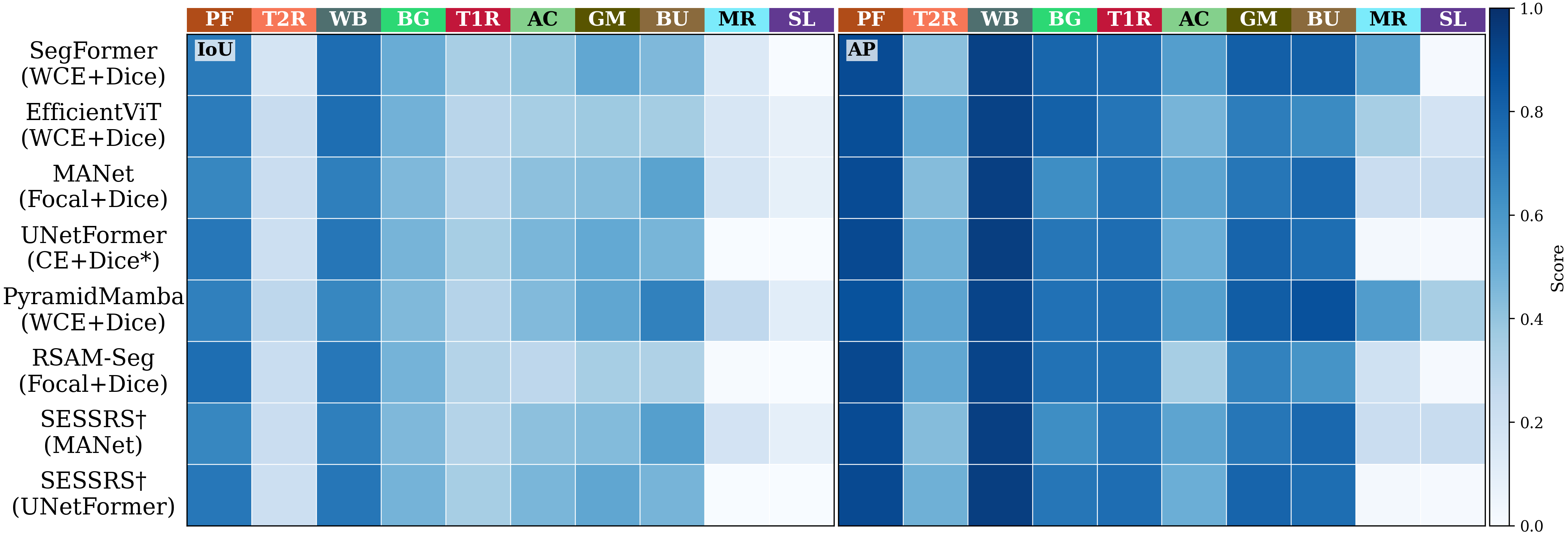}
\end{minipage}
\hfill
\begin{minipage}[t]{0.278\textwidth}
\centering
\includegraphics[width=\linewidth]{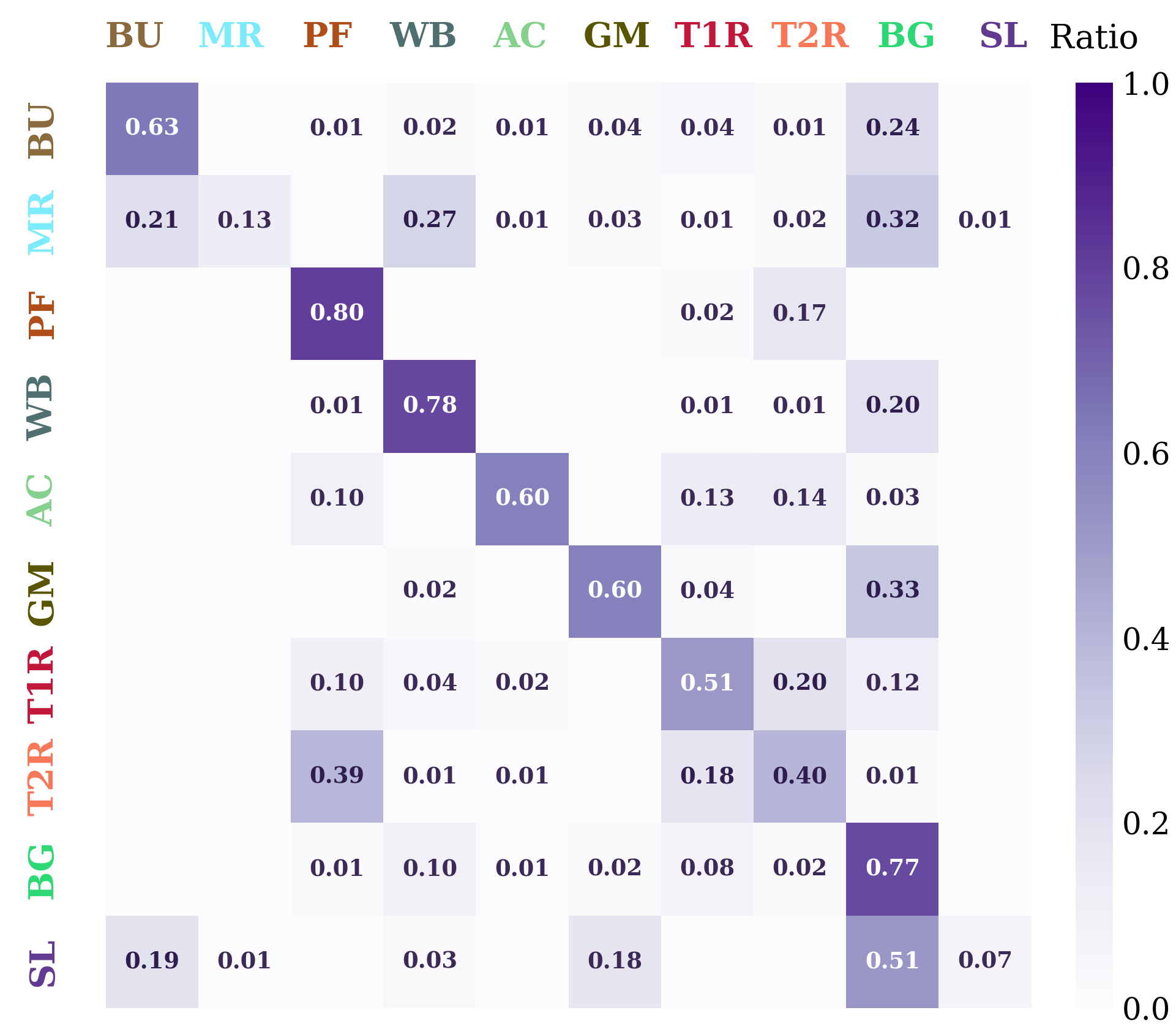}
\end{minipage}
\caption{
Left: per-class test-set IoU and AP of eight representative methods on the 10 foreground classes in the test split. Right: average row-normalized pixel-level confusion matrix of the same eight methods, with rows as ground truth and columns as predictions. Dominant classes are generally easier, while disturbance-related and rare mining-related classes remain more challenging.
}
\label{fig:perclass_main}
\end{figure*}

Detailed results for all methods are provided in the Appendix~\ref{app:seg}--\ref{app:seg_reg}. In the main text, Table~\ref{tab:main_selected_combined} highlights representative strong configurations from each method family, Figure~\ref{fig:qualitative_selected} provides qualitative showcase examples, and Figure~\ref{fig:perclass_main} summarizes per-class IoU, AP, and average confusion patterns for selected methods. More detailed analysis is further provided in Appendix~\ref{app:error_seg}: Appendix~\ref{app:perclass_selected_seg} presents the full per-class trends across method variants, Appendix~\ref{app:confusion_selected_seg} analyzes shared confusion patterns, and Appendix~\ref{app:visual_selected_seg} provides class-wise qualitative visualizations and discussion in Figures~\ref{fig:qual_bu}--\ref{fig:qual_t2r}. Overall, remote-sensing-specific methods are generally more competitive than generic baselines, especially on mining-related structures and disturbed land-cover categories. Methods related to VFMs show mixed behavior: direct SAM-style adaptation is not consistently advantageous, while stronger results more often come from refinement or integration with downstream segmenters. The effect of the training loss is also architecture-dependent, although class-imbalance-aware objectives are often helpful for difficult foreground categories, especially small mining-related targets. Similar trends appear in segmentation-derived recognition, where stronger dense prediction usually supports better image-level recognition without fully determining it.

Figure~\ref{fig:perclass_main} and the additional analysis in Appendix~\ref{app:error_seg} further clarify these trends. Segmentation and segmentation-derived recognition follow broadly similar class-level patterns: dominant categories such as primary forest and water bodies are more stable, while rare mining-related categories such as mining rafts and sluices remain the most difficult. However, frequency is not the whole story. Appendix~\ref{app:perclass_selected_seg} shows that Type~1 and Type~2 regeneration remain challenging despite their larger support because they are repeatedly confused with primary forest, bare ground, and each other. Appendix~\ref{app:confusion_selected_seg} further shows that many error modes are shared across methods, including confusions among buildings, mining rafts, and sluices, between gravel mounds and bare ground, and between water and bare ground. Appendix~\ref{app:visual_selected_seg} then expands these patterns into class-wise qualitative evidence in Figures~\ref{fig:qual_bu}--\ref{fig:qual_t2r}, showing that small targets are often only partially recovered, while even major classes remain sensitive to boundary ambiguity and contextual overlap. These results highlight that ELDOR is challenging not only because of long-tailed class frequency, but also because of strong visual and contextual overlap among mining, disturbance, and recovery categories.

\subsection{Results on Multi-label Classification}

\begin{figure*}[t]
\centering

\begin{minipage}[t]{0.34\textwidth}
\vspace{0pt}
\centering
\captionsetup{type=table}
\scriptsize
\caption{Selected results of direct multi-label classification methods. Shaded blocks indicate general and remote-sensing-specific methods. Best values are in bold within each group. Darker rows correspond to the representative methods highlighted for comparison.}
\label{tab:main_mlc_selected}
\setlength{\tabcolsep}{3.0pt}
\renewcommand{\arraystretch}{1.05}

\definecolor{generalbg}{RGB}{247,249,252}
\definecolor{rsbg}{RGB}{245,250,245}
\definecolor{generalhi}{RGB}{234,240,248}
\definecolor{rshi}{RGB}{232,243,232}
\definecolor{rulegray}{RGB}{190,190,190}

\arrayrulecolor{rulegray}
\resizebox{\linewidth}{!}{%
\begin{tabular}{@{}lccccc@{}}
\toprule
\textbf{Model} & \textbf{mAP} & \textbf{CF1} & \textbf{OF1} & \textbf{Macro-F1} & \textbf{Sample-F1} \\
\midrule

\rowcolor{generalbg}
ML-GCN     & 0.6499 & 0.5591 & 0.7495 & 0.5743 & 0.7670 \\
\rowcolor{generalbg}
C-Tran     & 0.6947 & 0.5672 & 0.7412 & \textbf{0.7019} & 0.7408 \\
\rowcolor{generalhi}
TDRG       & \textbf{0.6983} & \textbf{0.6737} & \textbf{0.7640} & 0.6161 & \textbf{0.7712} \\
\rowcolor{generalhi}
CPCL       & 0.6121 & 0.5833 & 0.7314 & 0.6335 & 0.7302 \\
\rowcolor{generalbg}
ML-Decoder & 0.6140 & 0.6051 & 0.7369 & 0.5862 & 0.7502 \\
\rowcolor{generalbg}
DDA-MLIC   & 0.6086 & 0.5986 & 0.7041 & 0.5859 & 0.7095 \\
\rowcolor{generalhi}
SpliceMix  & 0.6602 & 0.6246 & 0.7470 & 0.6003 & 0.7504 \\

\midrule

\rowcolor{rshi}
RelationNet & \textbf{0.6142} & 0.6029 & 0.7188 & 0.5961 & 0.7216 \\
\rowcolor{rshi}
GRN         & 0.6011 & 0.6010 & 0.7187 & 0.5859 & 0.7185 \\
\rowcolor{rsbg}
RSMLC       & 0.5926 & 0.5305 & 0.7254 & 0.5794 & \textbf{0.7476} \\
\rowcolor{rshi}
SIGNA       & 0.6112 & \textbf{0.6062} & \textbf{0.7301} & \textbf{0.6006} & 0.7339 \\

\bottomrule
\end{tabular}%
}
\arrayrulecolor{black}
\end{minipage}
\hfill
\begin{minipage}[t]{0.64\textwidth}
\vspace{0pt}
\centering
\captionsetup{type=figure,skip=3pt}
\includegraphics[width=\linewidth]{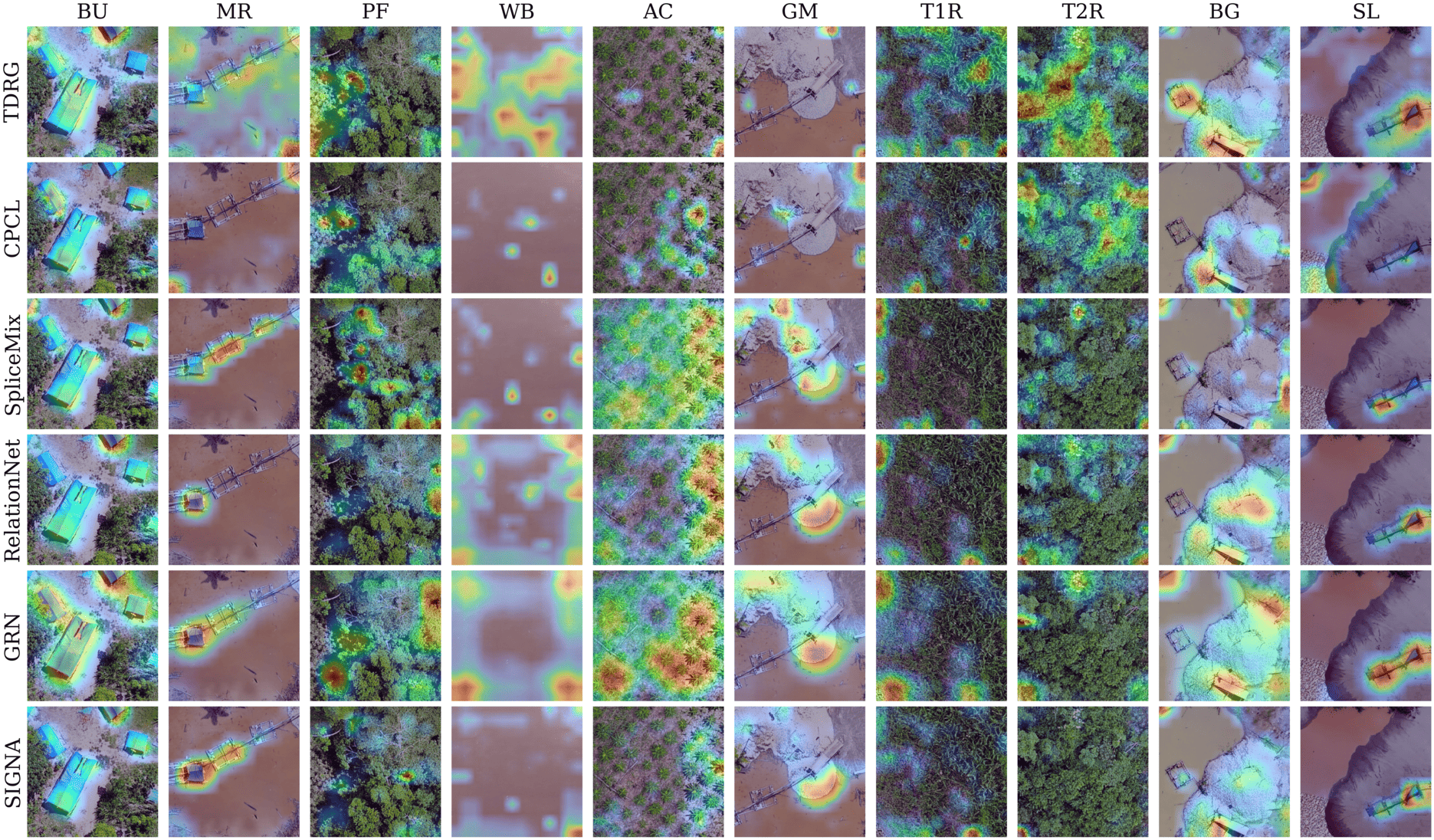}
\caption{Grad-CAM visualizations for representative direct multi-label classification methods. Warmer regions indicate stronger contributions to the predicted class presence.}
\label{fig:mlc_gradcam}
\end{minipage}

\end{figure*}

We evaluate direct multi-label classification on the same ELDOR split used for semantic segmentation. This setting complements segmentation-derived recognition by testing whether image-level semantic recognition can be learned directly from weak supervision. Unless restricted by native training recipes, models are trained under a unified direct multi-label classification protocol (Appendix~\ref{app:mlc_protocol}).

\textbf{General multi-label classification methods} include ML-GCN~\cite{chen2019multi}, ADD-GCN~\cite{ye2020attention}, C-Tran~\cite{lanchantin2021general}, CSRA~\cite{zhu2021residual}, TDRG~\cite{zhao2021transformer}, Q2L~\cite{liu2021query2label}, CPCL~\cite{zhou2021multi}, DualCoOp~\cite{sun2022dualcoop}, SALGL~\cite{zhu2023scene}, HSVLT~\cite{ouyang2023hsvlt}, ML-Decoder~\cite{ridnik2023ml}, SGRE~\cite{zhu2024semantic}, DDA-MLIC~\cite{singh2024discriminator}, DRL~\cite{lin2024distributionally}, and SpliceMix~\cite{wang2025splicemix}. \textbf{Remote-sensing-specific methods} include RelationNet~\cite{hua2020relation}, GRN~\cite{kang2020graph}, RSMLC~\cite{stoimchev2023deep}, SIGNA~\cite{liu2024semantic}, and IDMN~\cite{kim2024instance}.

Detailed results for all methods are provided in Appendix~\ref{app:mlc}. In the main text, Table~\ref{tab:main_mlc_selected} highlights representative direct multi-label classification methods, and Figure~\ref{fig:mlc_gradcam} shows Grad-CAM examples for the selected configurations. Appendix~\ref{app:mlc_overall}--\ref{app:mlc_gradcam} further provides overall and class-wise quantitative comparisons together with class-wise Grad-CAM visualizations, including the qualitative examples in Figures~\ref{fig:gradcam_buildings}--\ref{fig:gradcam_type2_regeneration}. Overall, direct multi-label classification on ELDOR remains challenging, with the main difficulties concentrated in rare mining-related classes, ambiguous recovery categories, and large variation in the spatial evidence used by different methods.

Figure~\ref{fig:mlc_gradcam} and Appendix~\ref{app:mlc_gradcam} show that many direct multi-label classification methods can still learn meaningful spatial information even though they are trained only with coarse image-level labels. In other words, coarse supervision is already sufficient for the model to learn not only which categories are present, but also where to look. This spatial evidence is not distributed uniformly across classes. For major classes, the models often rely on the most decision-relevant local regions rather than on the whole semantic area, such as canopy gaps and edge transitions for primary forest, shorelines and nearby context for water bodies, or mound boundaries for gravel mounds. For smaller mining-related classes, the responses more often need to cover the target structure itself, and method-to-method differences become more visible. A related quantitative comparison in Appendix~\ref{app:mlc_overall}--\ref{app:mlc_perclass} and Tables~\ref{tab:app_general_multilabel}--\ref{tab:app_vfm_multilabel} also shows that direct multi-label classification already reaches comparable mAP, higher CF1, and nearly identical OF1, compared with the strongest segmentation-derived recognition models. This suggests that coarse labels are already effective for image-level category understanding on ELDOR. By contrast, stronger segmentation-derived models remain better in Macro-F1 and Sample-F1, which indicates that pixel-level supervision is more helpful for balanced recognition across classes and for recovering complete label sets in ambiguous mixed patches.

\begin{table*}[t]
\centering
\scriptsize
\caption{Selected VLM-based recognition results across splits. Overall, generative VLMs provide the strongest image-level performance, while SAM3 also remains competitive. Here, SAM3$^\dagger$ denotes the version using detailed class-description prompts. See Appendix~\ref{app:vlm_protocol} for the protocol definitions.}
\label{tab:main_vqa_selected}
\setlength{\tabcolsep}{2.6pt}
\renewcommand{\arraystretch}{1.05}

\definecolor{vfmbg}{RGB}{252,247,242}
\definecolor{vfmhi}{RGB}{248,237,227}
\definecolor{rulegray}{RGB}{190,190,190}

\arrayrulecolor{rulegray}
\resizebox{\textwidth}{!}{%
\begin{tabular}{@{}llccccccccccccccc@{}}
\toprule
\multirow{2}{*}{\textbf{Model}} & \multirow{2}{*}{\textbf{Protocol}}
& \multicolumn{5}{c}{\textbf{Train}}
& \multicolumn{5}{c}{\textbf{Validation}}
& \multicolumn{5}{c}{\textbf{Test}} \\
\cmidrule(lr){3-7}\cmidrule(lr){8-12}\cmidrule(lr){13-17}
& & \textbf{mAP} & \textbf{CF1} & \textbf{OF1} & \textbf{Macro-F1} & \textbf{Sample-F1}
  & \textbf{mAP} & \textbf{CF1} & \textbf{OF1} & \textbf{Macro-F1} & \textbf{Sample-F1}
  & \textbf{mAP} & \textbf{CF1} & \textbf{OF1} & \textbf{Macro-F1} & \textbf{Sample-F1} \\
\midrule
\rowcolor{vfmbg}
RemoteCLIP & A2 & 0.2064 & 0.3611 & 0.4429 & 0.2651 & 0.4256 & 0.1983 & 0.2922 & 0.4001 & 0.2668 & 0.3813 & 0.3064 & 0.3632 & 0.4183 & 0.3354 & 0.3751 \\
\rowcolor{vfmbg}
GeoRSCLIP & A2 & 0.2818 & 0.3980 & 0.4091 & 0.2366 & 0.3973 & 0.2967 & 0.3551 & 0.4422 & 0.2722 & 0.4298 & 0.3544 & 0.5030 & 0.4151 & 0.2893 & 0.3704 \\
\rowcolor{vfmbg}
RS-LLaVA & A1 & 0.3607 & \cellcolor{vfmhi}\textbf{0.4294} & \cellcolor{vfmhi}\textbf{0.5113} & \cellcolor{vfmhi}\textbf{0.3643} & \cellcolor{vfmhi}\textbf{0.5023} & 0.4214 & \cellcolor{vfmhi}\textbf{0.4868} & 0.5484 & \cellcolor{vfmhi}\textbf{0.4326} & 0.5291 & 0.4848 & \cellcolor{vfmhi}\textbf{0.5362} & 0.5362 & 0.4368 & 0.4861 \\
\rowcolor{vfmbg}
VHM & A1 & \cellcolor{vfmhi}\textbf{0.4214} & 0.3495 & 0.3285 & 0.2890 & 0.3479 & \cellcolor{vfmhi}\textbf{0.4506} & 0.3669 & 0.5038 & 0.3495 & 0.5359 & \cellcolor{vfmhi}\textbf{0.5159} & 0.4272 & 0.4426 & \cellcolor{vfmhi}\textbf{0.4630} & 0.4343 \\
\rowcolor{vfmbg}
RemoteSAM & A4 & 0.1968 & 0.2894 & 0.3136 & 0.2511 & 0.3015 & 0.1526 & 0.2496 & 0.2992 & 0.2604 & 0.2889 & 0.2580 & 0.3616 & 0.3168 & 0.3119 & 0.3052 \\
\rowcolor{vfmbg}
SAM3$^\dagger$ & A5 & 0.3292 & 0.3755 & 0.4805 & 0.3330 & 0.4960 & 0.3324 & 0.4012 & \cellcolor{vfmhi}\textbf{0.6341} & 0.3675 & \cellcolor{vfmhi}\textbf{0.6946} & 0.3786 & 0.3845 & \cellcolor{vfmhi}\textbf{0.5636} & 0.3536 & \cellcolor{vfmhi}\textbf{0.5461} \\
\bottomrule
\end{tabular}%
}
\arrayrulecolor{black}
\end{table*}

\subsection{Results on VLM-based Recognition}


We evaluate VLM-based recognition on ELDOR across the train, validation, and test splits using RemoteCLIP~\cite{liu2024remoteclip}, GeoRSCLIP~\cite{zhang2024rs5m}, DOFA-CLIP~\cite{xiong2025dofa}, RS-LLaVA~\cite{bazi2024rs}, GeoChat~\cite{kuckreja2024geochat}, VHM~\cite{pang2025vhm}, RemoteSAM~\cite{yao2025remotesam}, and SAM3~\cite{carion2025sam}. See detailed protocols and results in Appendix~\ref{app:vlm_protocol} and~\ref{app:vqa}.

Table~\ref{tab:main_vqa_selected} highlights representative image-level results. Overall, generative models are the strongest zero-shot baselines, while SAM3-based recognition is also competitive. The full overall results in Tables~\ref{tab:app_vqa_overall_recognition}--\ref{tab:app_vqa_overall_segmentation} further show that both protocol design and prompt design matter: for CLIP-style models, positive--negative comparison gives much more meaningful predictions than positive-only scoring, and for SAM3, using class descriptions improves over class-name prompts in both recognition and prompted segmentation. The class-wise results in Tables~\ref{tab:app_vqa_classwise_apf1}--\ref{tab:app_vqa_seg_classwise} show a similar pattern to the previous tasks, with dominant categories being more stable and sparse mining-related categories remaining more difficult. Overall, zero-shot VLM-based recognition gives reasonable results, but still remains clearly below the strongest fine-tuned segmentation or multi-label classification methods.

\subsection{ELDOR Demonstration with Interactive Explorer}

\begin{figure}[t]
    \centering
    \includegraphics[width=\linewidth]{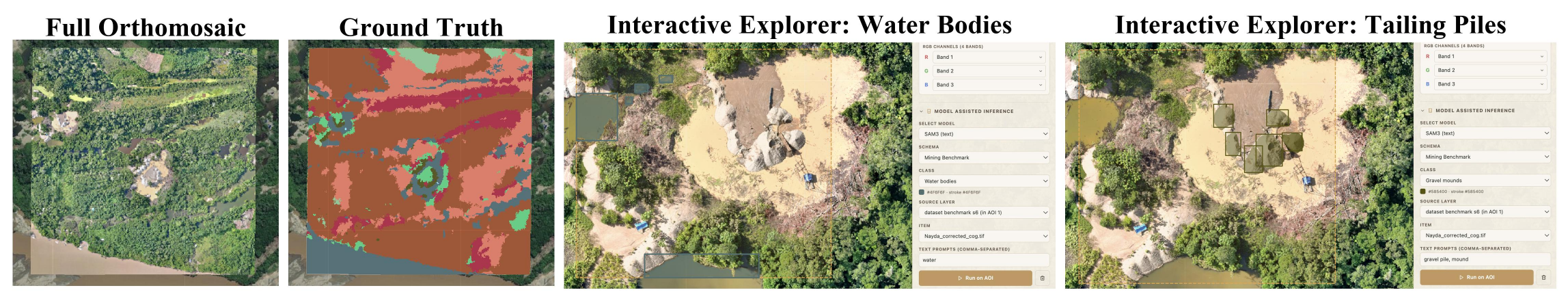}
    \caption{ELDOR demonstration in explorer. The left two panels show efficient loading and visualization of geospatial TIFF imagery with a global site view and surrounding contextual satellite layers provided through plugins. The right two panels show two example uses of the system, with SAM3 loaded as one supported model for interactive segmentation of water bodies and tailing piles.}
    \label{fig:interactive-explorer}
\end{figure}

We further implement an interactive explorer to make ELDOR dataset and pretrained models more actionable for ecologists in practice. As shown in Figure~\ref{fig:interactive-explorer}, the explorer supports efficient visualization of high-resolution imagery, which allows users to move from site-level overview to local inspection within the same interface. The left two panels show the orthomosaic and corresponding ground truth, while additional contextual layers, such as surrounding satellite imagery, provide broader environmental context. The same interface also supports interactive model use: users can define a region of interest, run a selected pretrained model, and inspect the predictions directly as overlay layers, as illustrated in the right two panels. This is enabled by a geospatial-native backend built on FastAPI~\cite{fastapi}, PostgreSQL \& PostGIS~\cite{postgis}, pgSTAC~\cite{pgstac}, TiTiler-pgSTAC~\cite{titiler}, MinIO~\cite{minio}, and Celery~\cite{celery}. More details are provided in Appendix~\ref{app:system}.

\section{Conclusions and Visions}
\label{sec:conclusion}

We introduced ELDOR, a large-scale UAV dataset and benchmark for monitoring environmental and landscape disturbance from illegal gold mining in the Amazon rainforest. Built from manually annotated high-resolution orthomosaics, ELDOR supports four complementary tasks: semantic segmentation, segmentation-derived recognition, direct multi-label classification, and VLM-based recognition. We also developed an interactive explorer that allows ecologists and domain experts to inspect orthomosaics, visualize annotations, and run model inference through an accessible interface. Across these tasks, existing models still face substantial challenges from fine-grained visual ambiguity, severe class imbalance, small mining-related objects, and cross-site variation, making ELDOR a difficult testbed for robust remote sensing perception under realistic ecological conditions.

The results reveal a useful contrast between dense and weak supervision: coarse labels can already support competitive image-level recognition, while dense annotations remain important for precise localization and balanced class coverage. This is especially important because object-level mining structures characterize not only whether mining has occurred, but also the modality, intensity, and likely consequences of land-use change. Features such as machinery, sluices, rafts, and mound-like deposits can indicate how the landscape is being modified, how much material may have been processed, and what forms of disturbance, remediation need, or recovery may follow. Future evaluation may therefore consider practical detection value in addition to complete mask accuracy, especially when partial localization supports rapid screening, field prioritization, restoration planning, and environmental risk assessment. Looking forward, spatial context, multimodal signals, temporal observations, uncertainty estimation, and human-in-the-loop refinement could connect UAV-scale interpretation with scalable monitoring of mining disturbance and recovery. Broader visions, practical uses, and limitations are discussed in Appendix~\ref{app:impact_limitations}.

\bibliographystyle{unsrtnat}
\bibliography{references}







\appendix
\section{Detailed Information of ELDOR Dataset}
\label{app:data}

This section provides detailed information about the ELDOR dataset to complement the summary in the main text. We present (i) class-level semantic definitions with visual examples (Table~\ref{tab:class_cards}), (ii) site-level metadata and semantic coverage (Table~\ref{tab:site_cards}), and (iii) per-site class distributions (Figure~\ref{fig:site_class_distribution}). These materials clarify the semantic structure of the dataset and the variability across sites.

\subsection{Class-level definitions.}
Table~\ref{tab:class_cards} provides detailed semantic descriptions and representative visual examples for all categories. Compared to the summary in Table~\ref{tab:class_mapping} in the main text, this table focuses on visual characteristics that are critical for interpretation in UAV orthomosaics, together with additional attributes such as source and generalizability. Here, source indicates whether a class arises from natural processes (e.g., forest, water) or mining-related activities (e.g., mining rafts, sluices), while generalizability reflects whether the visual pattern is expected to transfer beyond the Amazonian mining context or is specific to this region.

Several categories exhibit strong visual similarity, such as compact mounds and gravel mounds, or early-stage regeneration and bare ground. These ambiguities arise from mining processes and recovery stages, where texture, color, and spatial context often overlap. As a result, reliable recognition typically requires combining local appearance with surrounding context rather than relying on isolated pixel patterns.

\newcommand{\ClassCard}[4]{%
\multicolumn{1}{@{}p{\textwidth}@{}}{%
\begin{minipage}{\linewidth}
\textbf{#1}

\vspace{0.4ex}
\hrule height \lightrulewidth
\vspace{0.8ex}

\begin{tabular}{@{}p{0.23\textwidth}p{0.34\textwidth}p{0.35\textwidth}@{}}
#2 & #3 & #4
\end{tabular}
\end{minipage}
}\\
}

{\scriptsize
\begin{longtable}{@{}p{\textwidth}@{}}

\caption{Detailed semantic definitions, attributes, and visual examples for all categories in the ELDOR dataset. Each category is described with its source, temporal characteristics, and generalizability, along with representative image examples to illustrate its visual appearance.}
\label{tab:class_cards} \\

\toprule
\endfirsthead

\multicolumn{1}{@{}l@{}}{\small\itshape Table~\ref{tab:class_cards} continued from previous page.} \\
\toprule
\endhead

\bottomrule
\multicolumn{1}{r}{\small\itshape Continued on next page.} \\
\endfoot

\bottomrule
\endlastfoot

\ClassCard{Buildings}{
\begin{minipage}[c]{\linewidth}
\vspace{0pt}
\raggedright
\textbf{Source:} Anthropogenic

\textbf{Duration:} Persistent

\textbf{Generalizability:} Global
\end{minipage}
}{
\begin{minipage}[c]{\linewidth}
\centering
\includegraphics[width=0.48\linewidth]{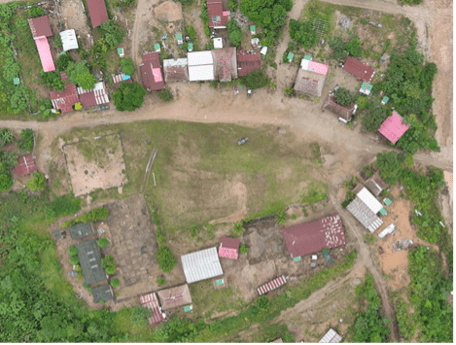}
\hfill
\includegraphics[width=0.48\linewidth]{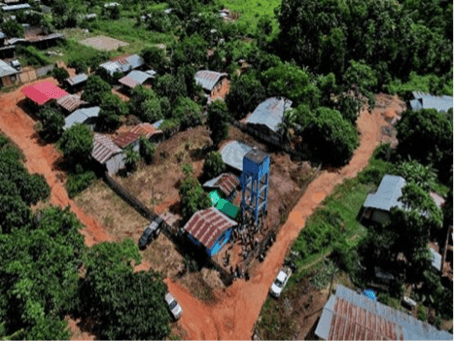}
\end{minipage}
}{
\begin{minipage}[c]{\linewidth}
\vspace{0pt}
\raggedright
Infrastructure built for human use, generally grouped and organized by blocks, streets, parks, etc.. Roofs of houses or dwellings are generally made of zinc-type calamine, painted in different colors such as red, blue, or silver. Generally interconnected by carriage roads, rivers or navigable streams.
\end{minipage}
}
\midrule

\ClassCard{Mining rafts}{
\begin{minipage}[c]{\linewidth}
\vspace{0pt}
\raggedright
\textbf{Source:} Anthropogenic

\textbf{Duration:} Mobile/Seasonal

\textbf{Generalizability:} Global
\end{minipage}
}{
\begin{minipage}[c]{\linewidth}
\centering
\includegraphics[width=0.48\linewidth]{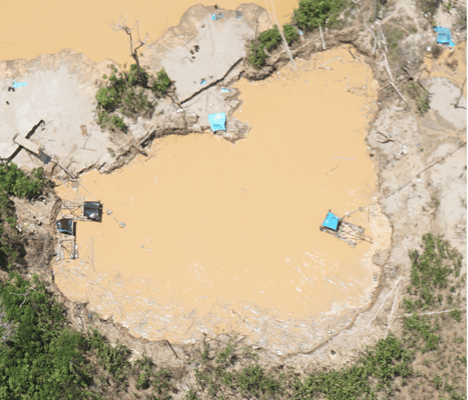}
\hfill
\includegraphics[width=0.48\linewidth]{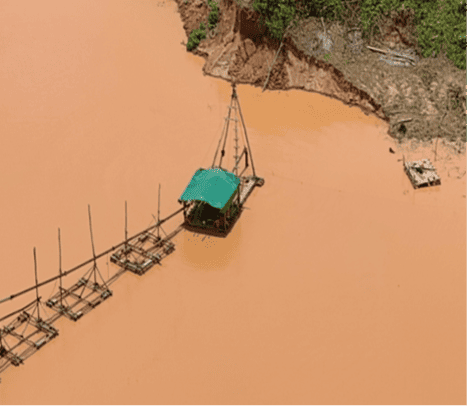}
\end{minipage}
}{
\begin{minipage}[c]{\linewidth}
\vspace{0pt}
\raggedright
Rectangular or square structures, generally with a blue or black plastic roof, supported by cylindrical metal floats. Found in rivers, streams, lakes and mining ponds. Often have one tube that connects to a sluice for washing the gold-bearing material and another tube at the front that sucks the gold-bearing material from the bottom of the water body. This system is powered by a motor attached to the raft and tubes, which is operated by a minimum of two people.
\end{minipage}
}
\midrule

\ClassCard{Primary Forest}{
\begin{minipage}[c]{\linewidth}
\vspace{0pt}
\raggedright
\textbf{Source:} Natural

\textbf{Duration:} Persistent

\textbf{Generalizability:} Amazonia
\end{minipage}
}{
\begin{minipage}[c]{\linewidth}
\centering
\includegraphics[width=0.48\linewidth]{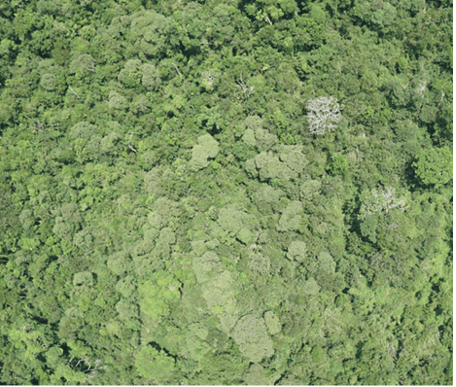}
\hfill
\includegraphics[width=0.48\linewidth]{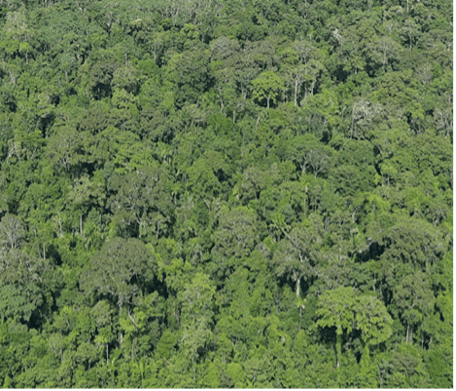}
\end{minipage}
}{
\begin{minipage}[c]{\linewidth}
\vspace{0pt}
\raggedright
Areas with arboreal vegetation that have not been intervened or significantly logged. They represent old-growth forests with little or no human intervention and high biological diversity. Some may have undergone selective logging, mainly for mining activities. Secondary forests (previously logged and recovering arboreal vegetation) may be mistaken for primary forests in orthomosaic imagery.
\end{minipage}
}
\midrule

\ClassCard{Heavy machinery}{
\begin{minipage}[c]{\linewidth}
\vspace{0pt}
\raggedright
\textbf{Source:} Anthropogenic

\textbf{Duration:} Mobile

\textbf{Generalizability:} Global
\end{minipage}
}{
\begin{minipage}[c]{\linewidth}
\centering
\includegraphics[width=0.35\linewidth]{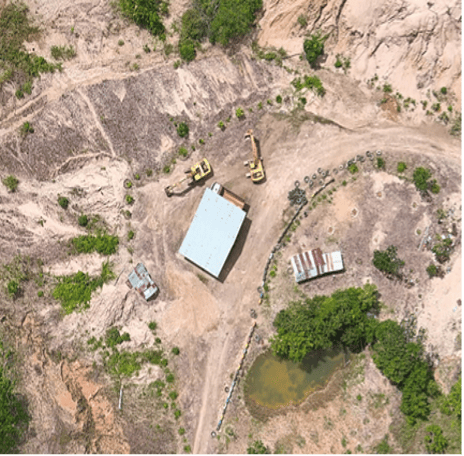}
\hfill
\includegraphics[width=0.60\linewidth]{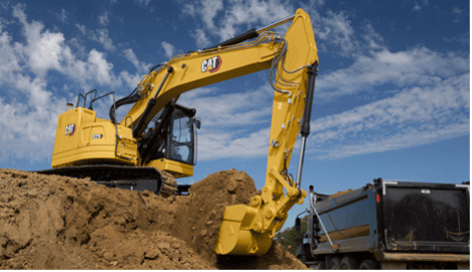}
\end{minipage}
}{
\begin{minipage}[c]{\linewidth}
\vspace{0pt}
\raggedright
Heavy machinery that cannot be determined among front loaders, excavators and dump trucks.
\end{minipage}
}
\midrule

\ClassCard{Water bodies}{
\begin{minipage}[c]{\linewidth}
\vspace{0pt}
\raggedright
\textbf{Source:} Anthropogenic + Natural

\textbf{Duration:} Persistent

\textbf{Generalizability:} Global
\end{minipage}
}{
\begin{minipage}[c]{\linewidth}
\centering
\includegraphics[width=0.48\linewidth]{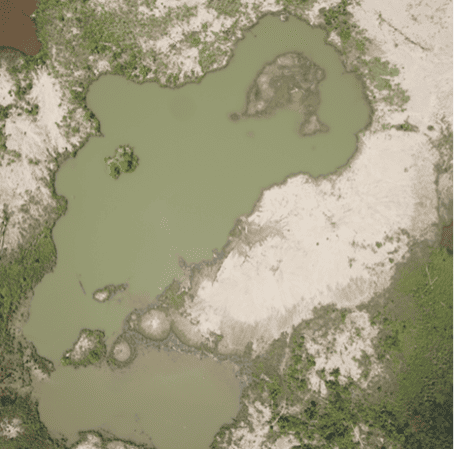}
\hfill
\includegraphics[width=0.48\linewidth]{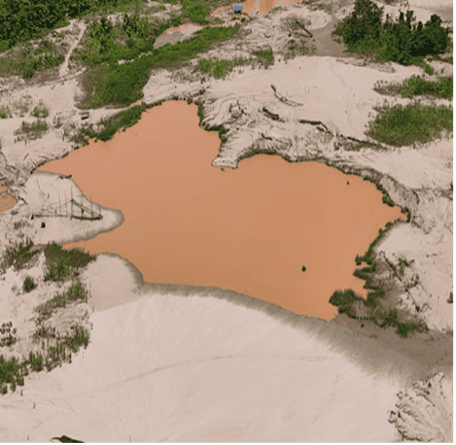}
\end{minipage}
}{
\begin{minipage}[c]{\linewidth}
\vspace{0pt}
\raggedright
Depressions or hollows in the landscape filled with water, generated by and for gold mining operations. Identified by their circular and irregular shape, the water is generally cloudy, light or dark in color depending on the activity or abandonment of the pond. Can be found with higher density and generally maintain the circular shape in mining operations with suction pumps. Usually interconnected by small channels where water flows during the rainy season. Less common in mining operations with heavy machinery in foothills, but occur with crescent-shaped grooves, are usually interconnected, and can have very large dimensions.
\end{minipage}
}
\midrule

\ClassCard{Agricultural crops}{
\begin{minipage}[c]{\linewidth}
\vspace{0pt}
\raggedright
\textbf{Source:} Anthropogenic

\textbf{Duration:} Persistent

\textbf{Generalizability:} Global
\end{minipage}
}{
\begin{minipage}[c]{\linewidth}
\centering
\includegraphics[width=0.42\linewidth]{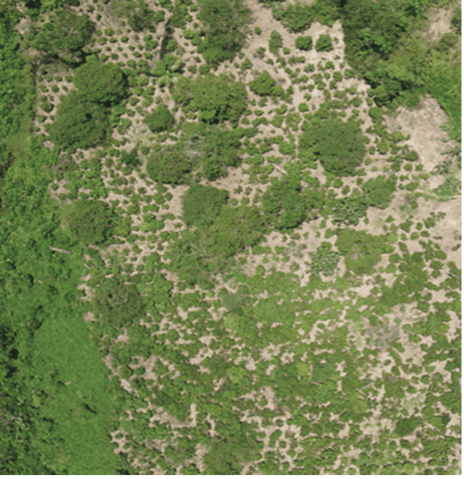}
\hfill
\includegraphics[width=0.54\linewidth]{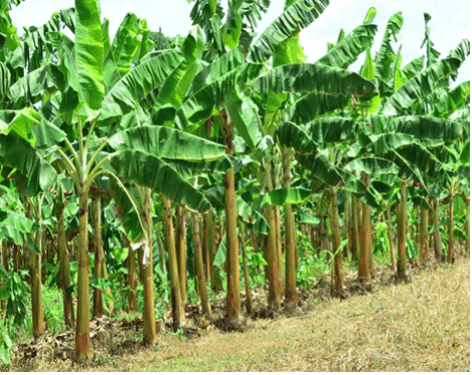}
\end{minipage}
}{
\begin{minipage}[c]{\linewidth}
\vspace{0pt}
\raggedright
Areas used to produce staple foods, identified by patterns of order and symmetrical distance visible in orthomosaic imagery. Generally banana, corn, cassava or fruit trees such as citrus, copoazú and guava.
\end{minipage}
}
\midrule

\ClassCard{Compact mounds}{
\begin{minipage}[c]{\linewidth}
\vspace{0pt}
\raggedright
\textbf{Source:} Anthropogenic

\textbf{Duration:} Persistent

\textbf{Generalizability:} Global
\end{minipage}
}{
\begin{minipage}[c]{\linewidth}
\centering
\includegraphics[width=0.48\linewidth]{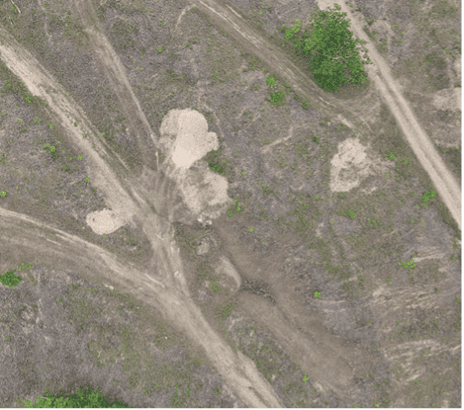}
\hfill
\includegraphics[width=0.48\linewidth]{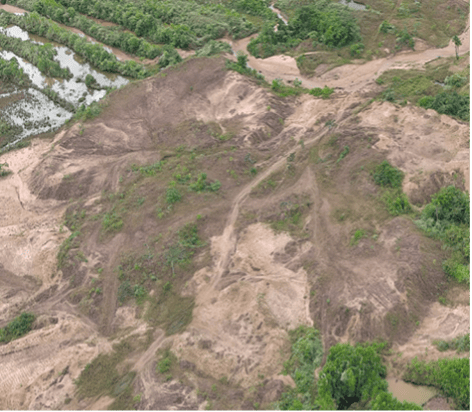}
\end{minipage}
}{
\begin{minipage}[c]{\linewidth}
\vspace{0pt}
\raggedright
Mounds of gravel, sand, and stone amassed by heavy machinery to be washed. Identified by the shape of a truncated cone, flattened at the top, often corresponding to places where the old mining chutes were located. Generally found in areas mined with heavy machinery, larger than rubble mounds, around 5--15 m high, and appear dark brown in orthomosaics with light-brown slopes resulting from landslides.
\end{minipage}
}
\midrule

\ClassCard{Gravel mounds}{
\begin{minipage}[c]{\linewidth}
\vspace{0pt}
\raggedright
\textbf{Source:} Anthropogenic

\textbf{Duration:} Persistent

\textbf{Generalizability:} Global
\end{minipage}
}{
\begin{minipage}[c]{\linewidth}
\centering
\includegraphics[width=0.48\linewidth]{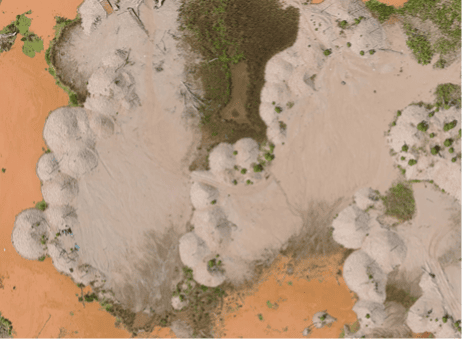}
\hfill
\includegraphics[width=0.48\linewidth]{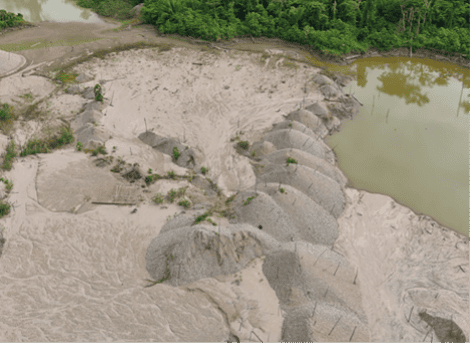}
\end{minipage}
}{
\begin{minipage}[c]{\linewidth}
\vspace{0pt}
\raggedright
Mounds gravel, sand and loose stone deposited by suction pumps as the gold-bearing material is washed, separating stones from fine material. Identified by a cone shape, generally found in areas mined with suction pumps, smaller than compact mounds, around 2--5 m high, and appear dark colors in orthomosaics.
\end{minipage}
}
\midrule

\ClassCard{Grass}{
\begin{minipage}[c]{\linewidth}
\vspace{0pt}
\raggedright
\textbf{Source:} Anthropogenic + Natural

\textbf{Duration:} Persistent

\textbf{Generalizability:} Amazonia
\end{minipage}
}{
\begin{minipage}[c]{\linewidth}
\centering
\includegraphics[width=0.48\linewidth]{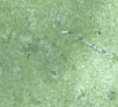}
\hfill
\includegraphics[width=0.48\linewidth]{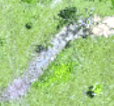}
\end{minipage}
}{
\begin{minipage}[c]{\linewidth}
\vspace{0pt}
\raggedright
Pasture used to feed and house cattle.
\end{minipage}
}
\midrule

\ClassCard{Type 1 disturbed vegetation}{
\begin{minipage}[c]{\linewidth}
\vspace{0pt}
\raggedright
\textbf{Source:} Natural

\textbf{Duration:} Multi-year

\textbf{Generalizability:} Amazonia
\end{minipage}
}{
\begin{minipage}[c]{\linewidth}
\centering
\includegraphics[width=0.48\linewidth]{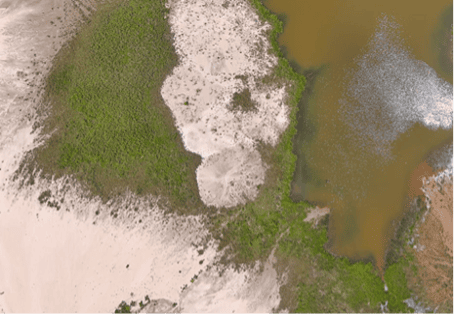}
\hfill
\includegraphics[width=0.48\linewidth]{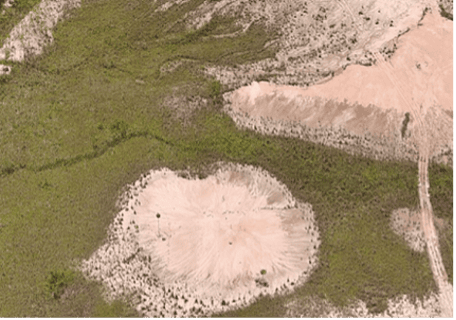}
\end{minipage}
}{
\begin{minipage}[c]{\linewidth}
\vspace{0pt}
\raggedright
Secondary forests dominated by non-woody vegetation such as grasses and lianas. Generally less than three meters high, commonly occurs near mining ponds and in wetlands, and appear dark brown in orthomosaics. More often reflects mined areas that have been clear-cut and had soil removed.
\end{minipage}
}
\midrule

\ClassCard{Type 2 disturbed vegetation}{
\begin{minipage}[c]{\linewidth}
\vspace{0pt}
\raggedright
\textbf{Source:} Natural

\textbf{Duration:} Multi-year

\textbf{Generalizability:} Amazonia
\end{minipage}
}{
\begin{minipage}[c]{\linewidth}
\centering
\includegraphics[width=0.48\linewidth]{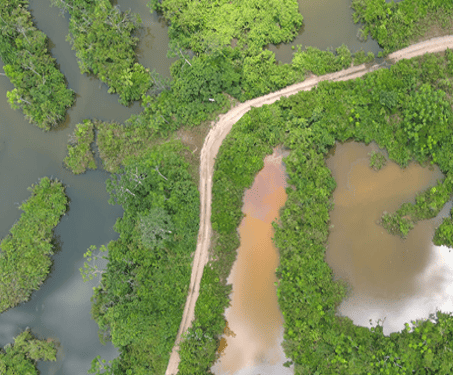}
\hfill
\includegraphics[width=0.48\linewidth]{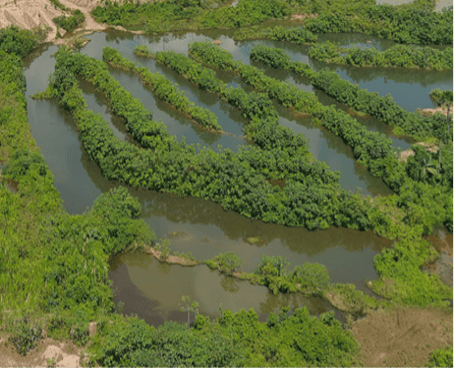}
\end{minipage}
}{
\begin{minipage}[c]{\linewidth}
\vspace{0pt}
\raggedright
Secondary forests dominated by woody vegetation such as shrubs, trees and lianas. Generally around 5--15 meters high, commonly occurs on rubble mounds and areas with decomposing organic matter. Often includes regions never mined, but forested or thinned, with intact soil.
\end{minipage}
}
\midrule

\ClassCard{Bare ground}{
\begin{minipage}[c]{\linewidth}
\vspace{0pt}
\raggedright
\textbf{Source:} Anthropogenic + Natural

\textbf{Duration:} Persistent / Multi-year

\textbf{Generalizability:} Global
\end{minipage}
}{
\begin{minipage}[c]{\linewidth}
\centering
\includegraphics[width=0.48\linewidth]{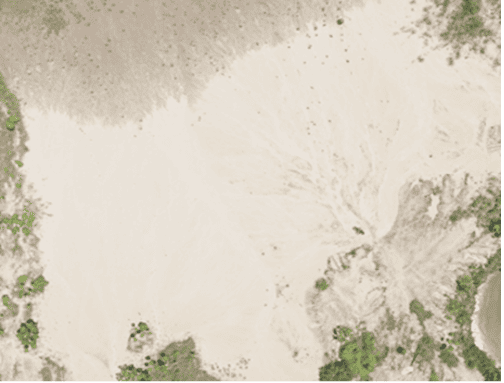}
\hfill
\includegraphics[width=0.48\linewidth]{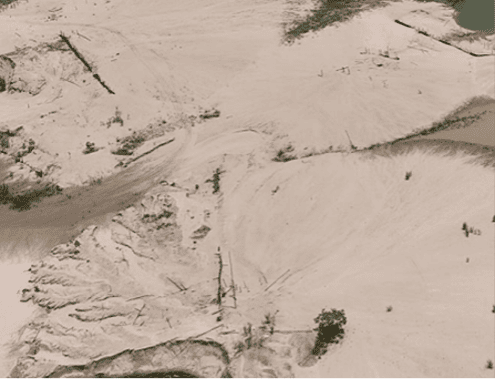}
\end{minipage}
}{
\begin{minipage}[c]{\linewidth}
\vspace{0pt}
\raggedright
Areas without vegetation. May occur alongside natural regeneration type 1 or 2, depending on the soil’s initial conditions and time of abandonment since mining. Appears as different shades of gray in orthomosaics.
\end{minipage}
}
\midrule

\ClassCard{Sluices}{
\begin{minipage}[c]{\linewidth}
\vspace{0pt}
\raggedright
\textbf{Source:} Anthropogenic

\textbf{Duration:} Seasonal

\textbf{Generalizability:} Global
\end{minipage}
}{
\begin{minipage}[c]{\linewidth}
\centering
\includegraphics[width=0.48\linewidth]{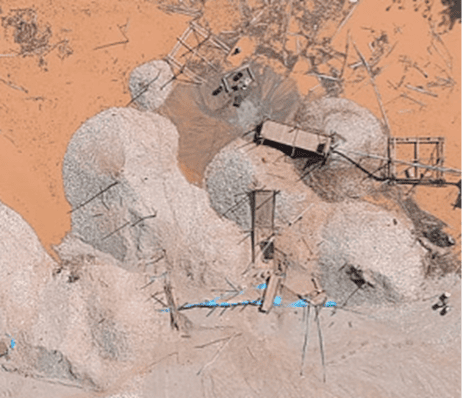}
\hfill
\includegraphics[width=0.48\linewidth]{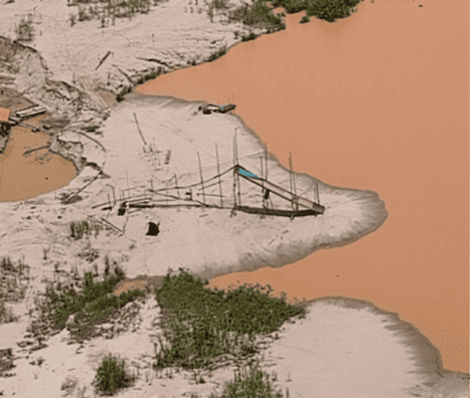}
\end{minipage}
}{
\begin{minipage}[c]{\linewidth}
\vspace{0pt}
\raggedright
Wooden infrastructure identified by its rectangular shape, located at the edge of gold mining ponds. Includes a ramp on a rubble mound or wooden infrastructure, with a pipe at the top, connected to the mining pond, which transports the gold-bearing material.
\end{minipage}
}
\midrule

\ClassCard{Vehicles}{
\begin{minipage}[c]{\linewidth}
\vspace{0pt}
\raggedright
\textbf{Source:} Anthropogenic

\textbf{Duration:} Mobile

\textbf{Generalizability:} Global
\end{minipage}
}{
\begin{minipage}[c]{\linewidth}
\centering
\includegraphics[width=0.48\linewidth]{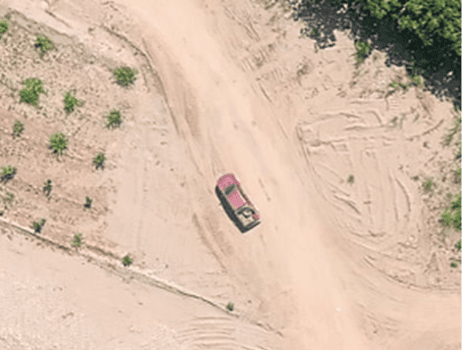}
\hfill
\includegraphics[width=0.48\linewidth]{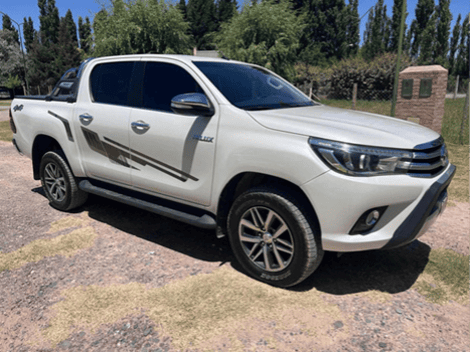}
\end{minipage}
}{
\begin{minipage}[c]{\linewidth}
\vspace{0pt}
\raggedright
Transportation vehicles such as pick-up trucks or cars, generally located in or near mining camps, used to transport supplies, equipment and people.
\end{minipage}
}

\end{longtable}
}

\subsection{Site-level metadata and coverage.}
Table~\ref{tab:site_cards} summarizes the properties of each site, including spatial extent, resolution, and the set of semantic classes present. Each site is assigned to a single split (train, validation, or test), and no spatial overlap exists between sites. The table shows substantial variability across sites in both scale and semantic composition. Some sites contain diverse mining-related structures, while others are dominated by natural or partially recovered landscapes. In addition, certain classes are absent in specific splits, particularly in the test set. These missing categories are often mobile or site-dependent, such as vehicles or heavy machinery, and their absence reflects real-world variability rather than annotation bias. This design introduces a cross-site generalization challenge, where models must transfer knowledge across sites with different semantic distributions and visual characteristics.

\newcommand{\SiteCardHeader}{%
\begin{tabular}{@{}p{0.32\textwidth}p{0.18\textwidth}p{0.18\textwidth}p{0.22\textwidth}@{}}
\textbf{Site Metadata} & \textbf{Orthomosaic} & \textbf{Label} & \textbf{Classes Present}
\end{tabular}
}

\newcommand{\SiteCard}[5]{%
\multicolumn{1}{@{}p{\textwidth}@{}}{%
\begin{minipage}{\linewidth}
\textbf{#1}

\vspace{0.4ex}
\hrule height \lightrulewidth
\vspace{0.8ex}

\begin{tabular}{@{}p{0.32\textwidth}p{0.18\textwidth}p{0.18\textwidth}p{0.22\textwidth}@{}}
#2 & #3 & #4 & #5
\end{tabular}
\end{minipage}
}\\
}

{\scriptsize
\begin{longtable}{@{}p{\textwidth}@{}}
\caption{Site-level metadata, visual examples, and semantic coverage in ELDOR. For each site, we report split assignment, spatial properties, geographic extent, representative orthomosaic and label visualizations, and the set of semantic classes present.}
\label{tab:site_cards} \\

\toprule
\SiteCardHeader \\
\midrule
\endfirsthead

\multicolumn{1}{@{}l@{}}{\small\itshape Table~\ref{tab:site_cards} continued from previous page.} \\
\toprule
\SiteCardHeader \\
\midrule
\endhead

\bottomrule
\multicolumn{1}{r}{\small\itshape Continued on next page.} \\
\endfoot

\bottomrule
\endlastfoot

\SiteCard{AcumulacionAaron2B}{
\begin{minipage}[c]{\linewidth}
\vspace{0pt}
\raggedright
\textbf{Split:} train

\textbf{Output size:} 38122$\times$17643

\textbf{Resolution:} 0.0563 m/px

\textbf{Area:} 212.0501 ha

\textbf{Longitude range:} 
[-70.419597, -70.399788]

\textbf{Latitude range:} 
[-12.627643, -12.618558]
\end{minipage}
}{
\begin{minipage}[c]{\linewidth}
\centering
\includegraphics[width=0.95\linewidth]{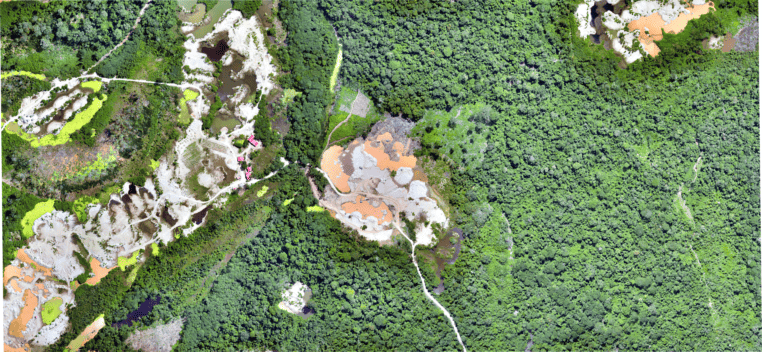}
\end{minipage}
}{
\begin{minipage}[c]{\linewidth}
\centering
\includegraphics[width=0.95\linewidth]{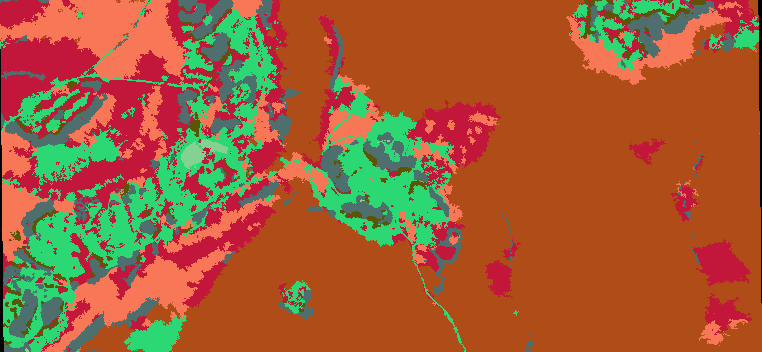}
\end{minipage}
}{
\begin{minipage}[c]{\linewidth}
\vspace{0pt}
\raggedright
Building; Mining raft; Primary Forest; Heavy machinery; Water bodies; Agricultural crop; Gravel mounds; Type 1 natural regeneration; Type 2 natural regeneration; Bare ground; Sluice; Vehicles
\end{minipage}
}
\midrule

\SiteCard{Anel}{
\begin{minipage}[c]{\linewidth}
\vspace{0pt}
\raggedright
\textbf{Split:} test

\textbf{Output size:} 18274$\times$18420

\textbf{Resolution:} 0.0566 m/px

\textbf{Area:} 102.7175 ha

\textbf{Longitude range:} 
[-69.714044, -69.699976]

\textbf{Latitude range:} 
[-12.713767, -12.701071]
\end{minipage}
}{
\begin{minipage}[c]{\linewidth}
\centering
\includegraphics[width=0.95\linewidth]{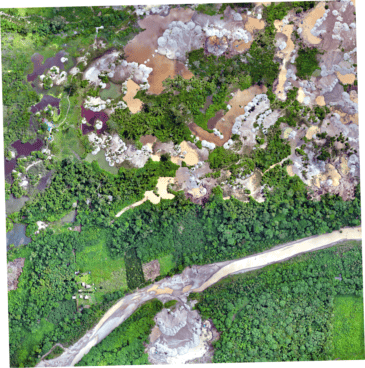}
\end{minipage}
}{
\begin{minipage}[c]{\linewidth}
\centering
\includegraphics[width=0.95\linewidth]{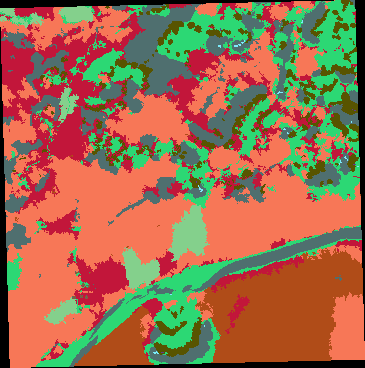}
\end{minipage}
}{
\begin{minipage}[c]{\linewidth}
\vspace{0pt}
\raggedright
Building; Mining raft; Primary Forest; Water bodies; Agricultural crop; Gravel mounds; Type 1 natural regeneration; Type 2 natural regeneration; Bare ground; Sluice
\end{minipage}
}
\midrule

\SiteCard{Clavelito}{
\begin{minipage}[c]{\linewidth}
\vspace{0pt}
\raggedright
\textbf{Split:} val

\textbf{Output size:} 19556$\times$40304

\textbf{Resolution:} 0.0756 m/px

\textbf{Area:} 344.5330 ha

\textbf{Longitude range:} 
[-70.605945, -70.591929]

\textbf{Latitude range:} 
[-12.961391, -12.933761]
\end{minipage}
}{
\begin{minipage}[c]{\linewidth}
\centering
\includegraphics[width=0.95\linewidth]{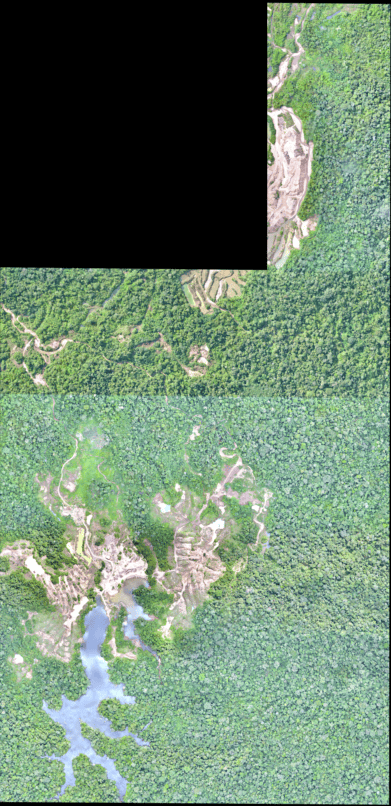}
\end{minipage}
}{
\begin{minipage}[c]{\linewidth}
\centering
\includegraphics[width=0.95\linewidth]{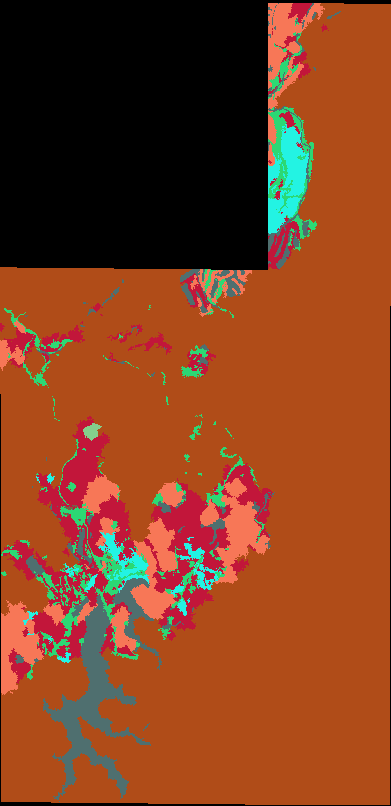}
\end{minipage}
}{
\begin{minipage}[c]{\linewidth}
\vspace{0pt}
\raggedright
Building; Primary Forest; Heavy machinery; Water bodies; Agricultural crop; Compact mounds; Type 1 natural regeneration; Type 2 natural regeneration; Bare ground
\end{minipage}
}
\midrule

\SiteCard{ElEngano}{
\begin{minipage}[c]{\linewidth}
\vspace{0pt}
\raggedright
\textbf{Split:} test

\textbf{Output size:} 17575$\times$17770

\textbf{Resolution:} 0.0572 m/px

\textbf{Area:} 101.2404 ha

\textbf{Longitude range:} 
[-69.962478, -69.949966]

\textbf{Latitude range:} 
[-13.020161, -13.007312]
\end{minipage}
}{
\begin{minipage}[c]{\linewidth}
\centering
\includegraphics[width=0.95\linewidth]{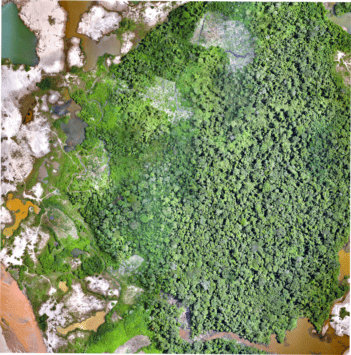}
\end{minipage}
}{
\begin{minipage}[c]{\linewidth}
\centering
\includegraphics[width=0.95\linewidth]{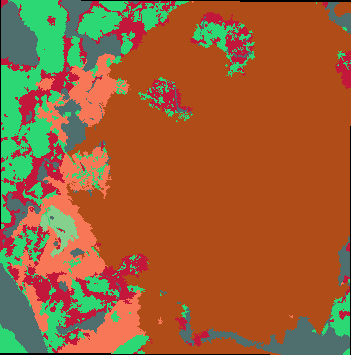}
\end{minipage}
}{
\begin{minipage}[c]{\linewidth}
\vspace{0pt}
\raggedright
Primary Forest; Water bodies; Agricultural crop; Type 1 natural regeneration; Type 2 natural regeneration; Bare ground
\end{minipage}
}
\midrule

\SiteCard{Kotsimba}{
\begin{minipage}[c]{\linewidth}
\vspace{0pt}
\raggedright
\textbf{Split:} train

\textbf{Output size:} 57328$\times$53959

\textbf{Resolution:} 0.0728 m/px

\textbf{Area:} 693.7270 ha

\textbf{Longitude range:} 
[-70.283655, -70.244998]

\textbf{Latitude range:} 
[-13.136635, -13.100951]
\end{minipage}
}{
\begin{minipage}[c]{\linewidth}
\centering
\includegraphics[width=0.95\linewidth]{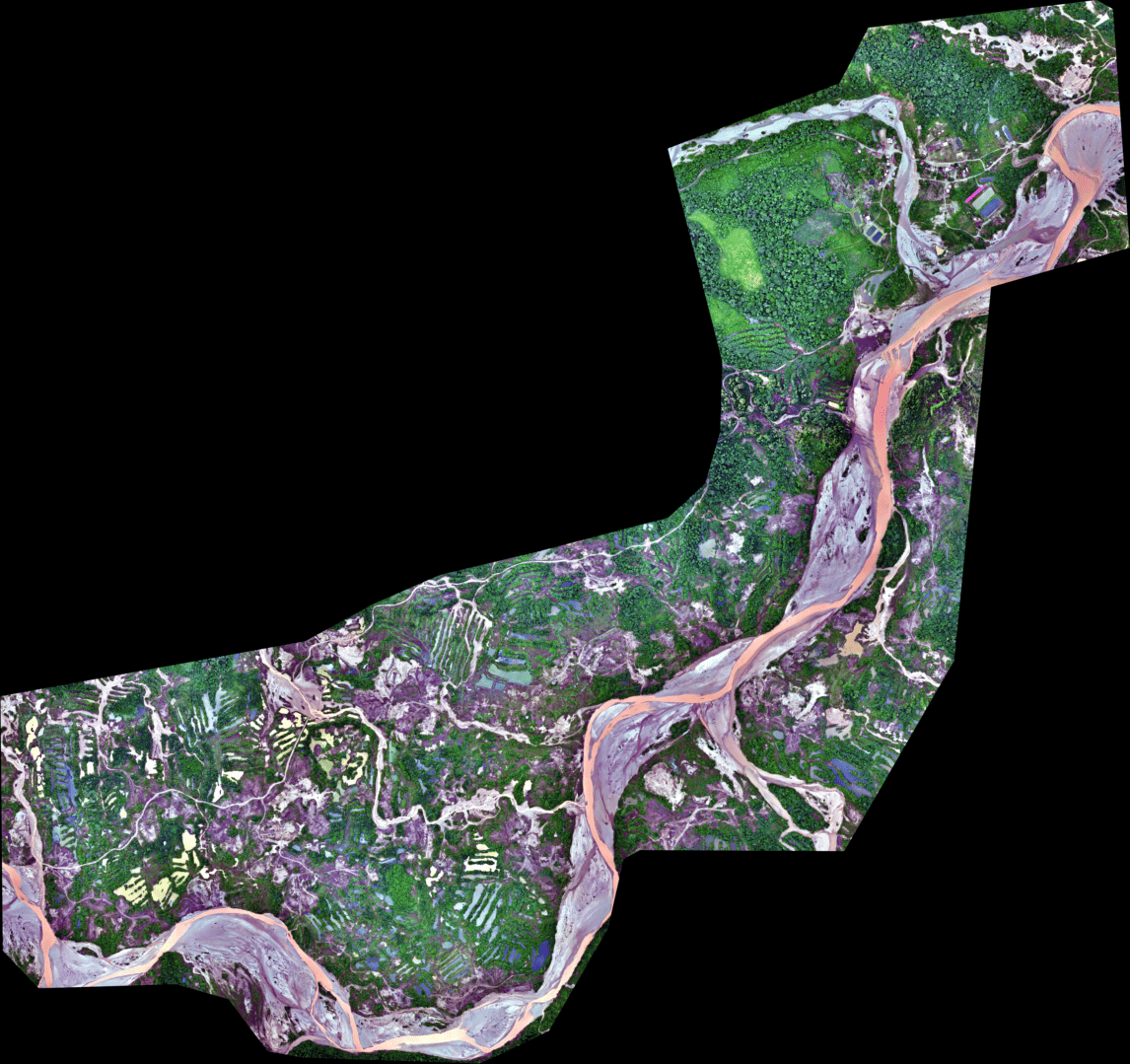}
\end{minipage}
}{
\begin{minipage}[c]{\linewidth}
\centering
\includegraphics[width=0.95\linewidth]{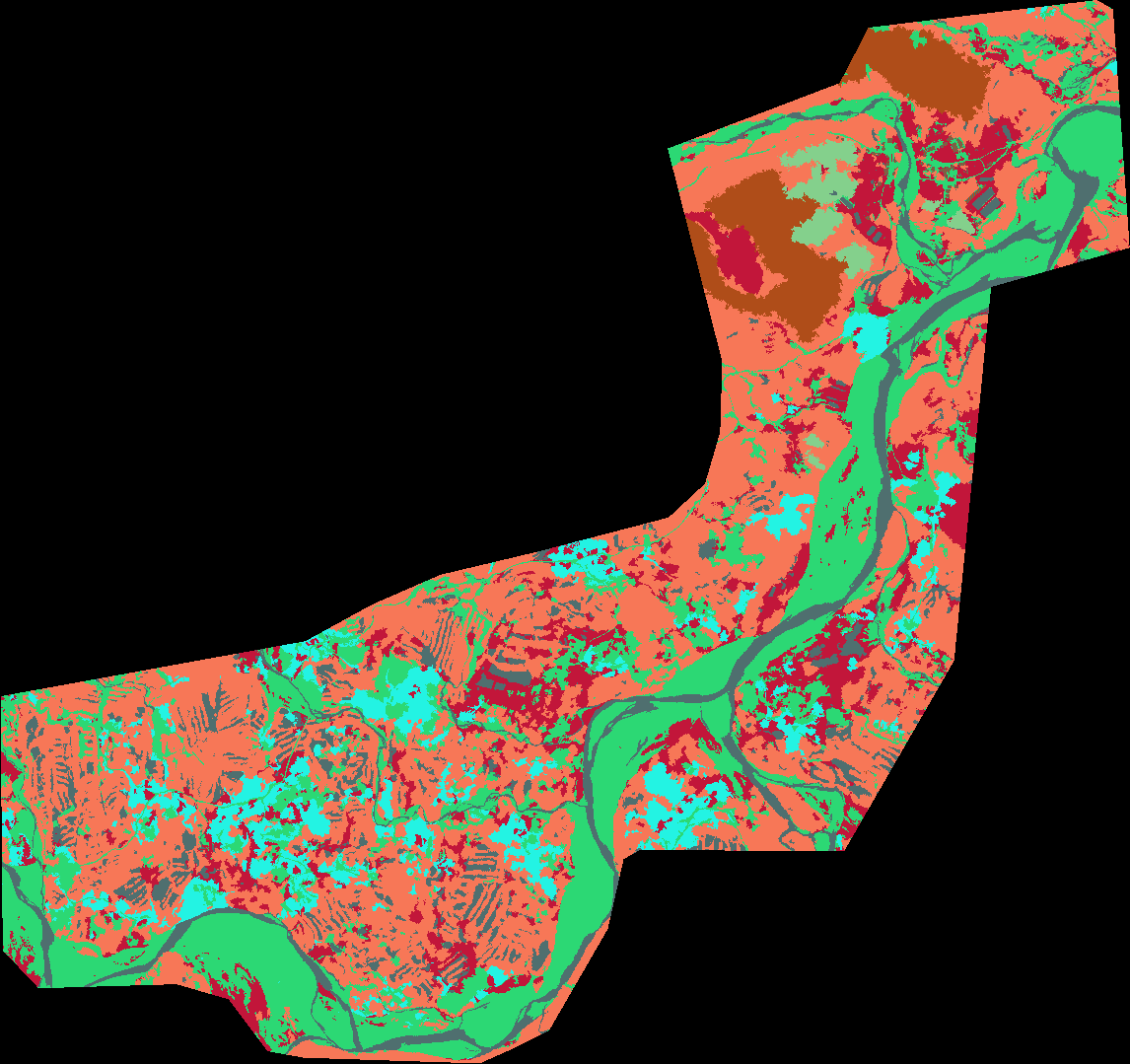}
\end{minipage}
}{
\begin{minipage}[c]{\linewidth}
\vspace{0pt}
\raggedright
Building; Primary Forest; Water bodies; Agricultural crop; Compact mounds; Type 1 natural regeneration; Type 2 natural regeneration; Bare ground
\end{minipage}
}
\midrule

\SiteCard{Linda}{
\begin{minipage}[c]{\linewidth}
\vspace{0pt}
\raggedright
\textbf{Split:} test

\textbf{Output size:} 17852$\times$17888

\textbf{Resolution:} 0.0576 m/px

\textbf{Area:} 98.8187 ha

\textbf{Longitude range:} 
[-69.953378, -69.940789]

\textbf{Latitude range:} 
[-13.020082, -13.007688]
\end{minipage}
}{
\begin{minipage}[c]{\linewidth}
\centering
\includegraphics[width=0.95\linewidth]{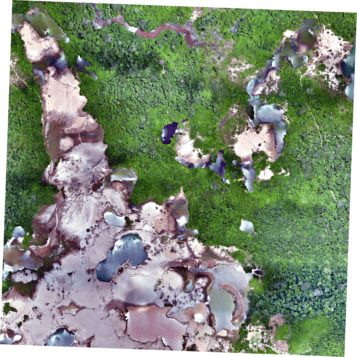}
\end{minipage}
}{
\begin{minipage}[c]{\linewidth}
\centering
\includegraphics[width=0.95\linewidth]{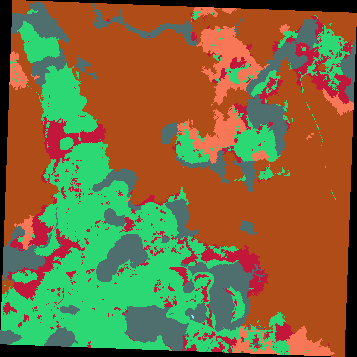}
\end{minipage}
}{
\begin{minipage}[c]{\linewidth}
\vspace{0pt}
\raggedright
Building; Mining raft; Primary Forest; Water bodies; Type 1 natural regeneration; Type 2 natural regeneration; Bare ground; Sluice
\end{minipage}
}
\midrule

\SiteCard{Los5Rebeldes}{
\begin{minipage}[c]{\linewidth}
\vspace{0pt}
\raggedright
\textbf{Split:} train

\textbf{Output size:} 33490$\times$66672

\textbf{Resolution:} 0.0300 m/px

\textbf{Area:} 199.9257 ha

\textbf{Longitude range:} 
[-70.672354, -70.660084]

\textbf{Latitude range:} 
[-13.034336, -13.013573]
\end{minipage}
}{
\begin{minipage}[c]{\linewidth}
\centering
\includegraphics[width=0.95\linewidth]{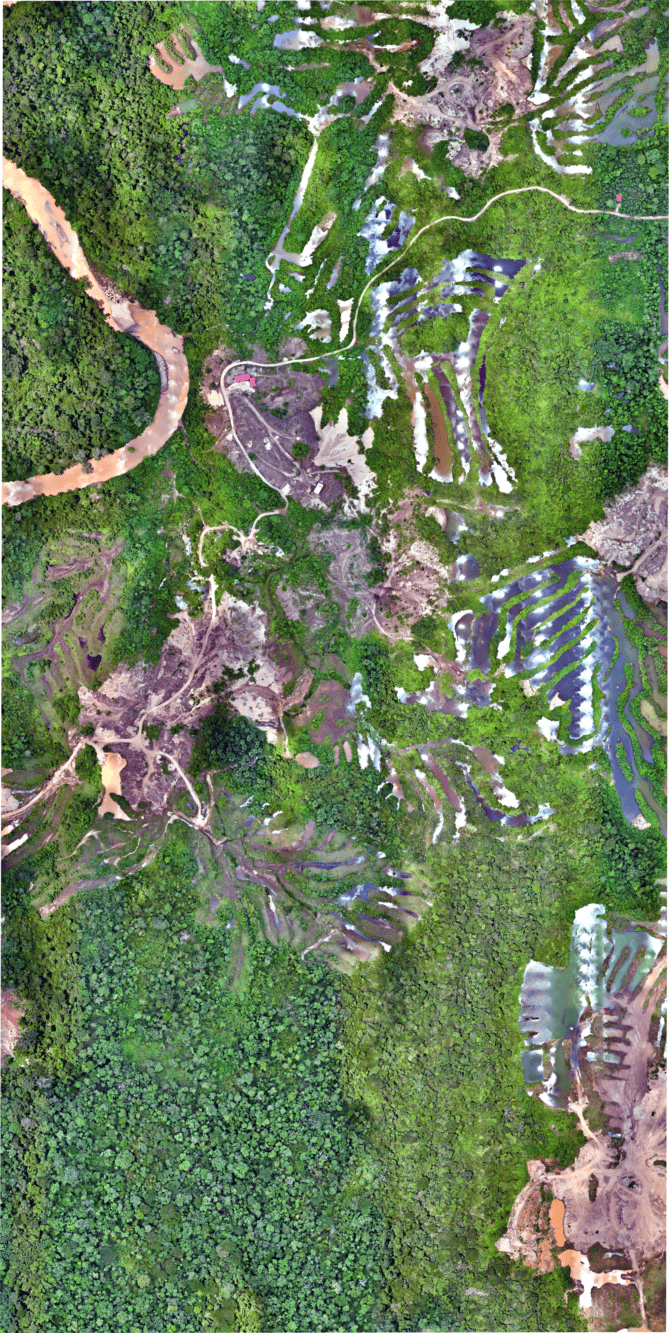}
\end{minipage}
}{
\begin{minipage}[c]{\linewidth}
\centering
\includegraphics[width=0.95\linewidth]{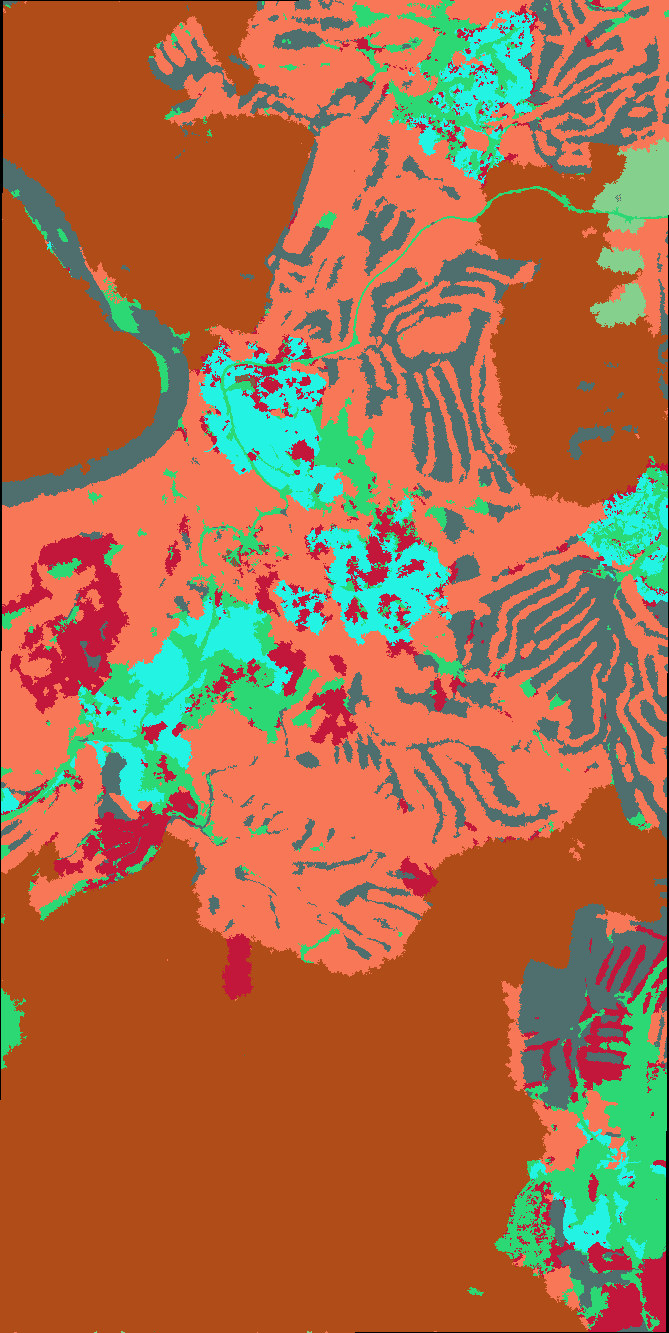}
\end{minipage}
}{
\begin{minipage}[c]{\linewidth}
\vspace{0pt}
\raggedright
Building; Primary Forest; Water bodies; Agricultural crop; Compact mounds; Type 1 natural regeneration; Type 2 natural regeneration; Bare ground; Vehicles
\end{minipage}
}
\midrule

\SiteCard{Nayda}{
\begin{minipage}[c]{\linewidth}
\vspace{0pt}
\raggedright
\textbf{Split:} val

\textbf{Output size:} 17954$\times$17908

\textbf{Resolution:} 0.0566 m/px

\textbf{Area:} 101.9933 ha

\textbf{Longitude range:} 
[-69.721603, -69.709746]

\textbf{Latitude range:} 
[-12.713363, -12.701313]
\end{minipage}
}{
\begin{minipage}[c]{\linewidth}
\centering
\includegraphics[width=0.95\linewidth]{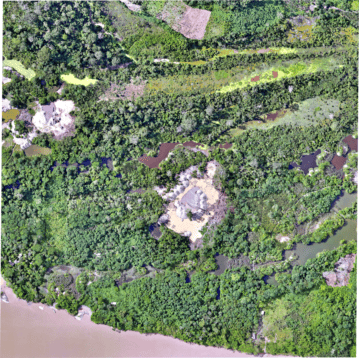}
\end{minipage}
}{
\begin{minipage}[c]{\linewidth}
\centering
\includegraphics[width=0.95\linewidth]{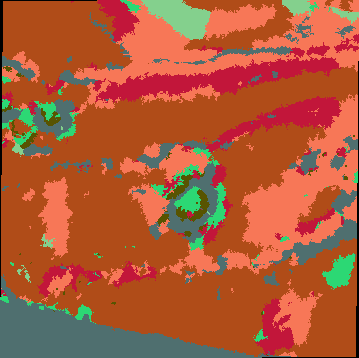}
\end{minipage}
}{
\begin{minipage}[c]{\linewidth}
\vspace{0pt}
\raggedright
Building; Mining raft; Primary Forest; Water bodies; Agricultural crop; Gravel mounds; Type 1 natural regeneration; Type 2 natural regeneration; Bare ground; Sluice
\end{minipage}
}
\midrule

\SiteCard{Paolita}{
\begin{minipage}[c]{\linewidth}
\vspace{0pt}
\raggedright
\textbf{Split:} test

\textbf{Output size:} 53943$\times$27054

\textbf{Resolution:} 0.0554 m/px

\textbf{Area:} 445.5226 ha

\textbf{Longitude range:} 
[-69.630457, -69.600462]

\textbf{Latitude range:} 
[-12.689889, -12.673012]
\end{minipage}
}{
\begin{minipage}[c]{\linewidth}
\centering
\includegraphics[width=0.95\linewidth]{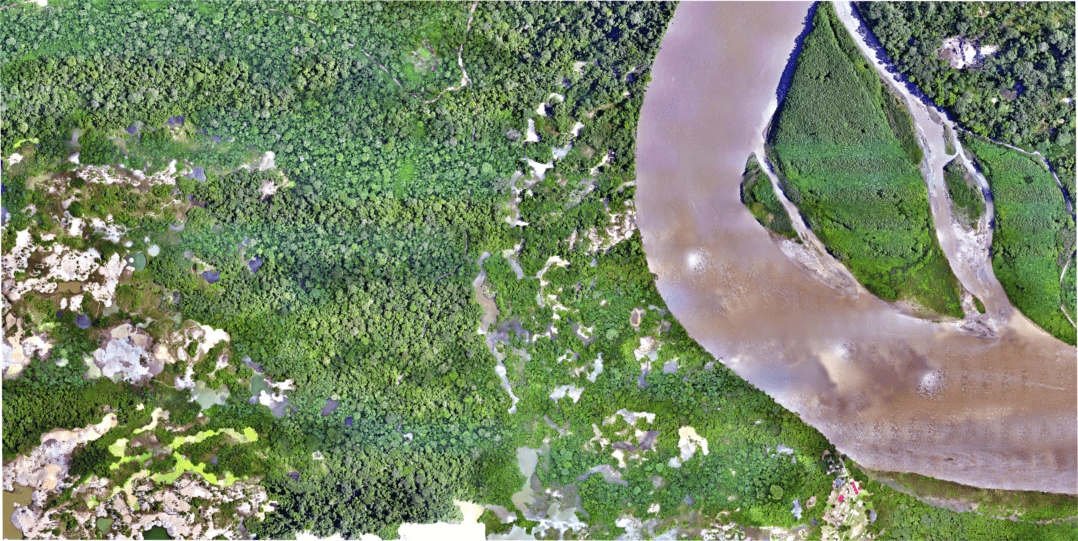}
\end{minipage}
}{
\begin{minipage}[c]{\linewidth}
\centering
\includegraphics[width=0.95\linewidth]{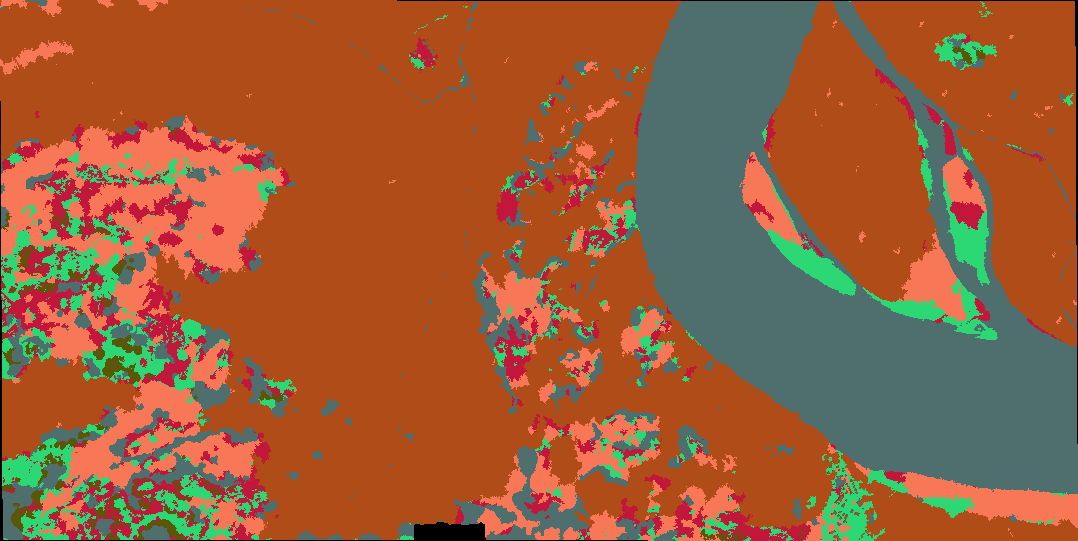}
\end{minipage}
}{
\begin{minipage}[c]{\linewidth}
\vspace{0pt}
\raggedright
Building; Mining raft; Primary Forest; Water bodies; Agricultural crop; Gravel mounds; Type 1 natural regeneration; Type 2 natural regeneration; Bare ground; Sluice
\end{minipage}
}
\midrule

\SiteCard{PlayaMirador1}{
\begin{minipage}[c]{\linewidth}
\vspace{0pt}
\raggedright
\textbf{Split:} train

\textbf{Output size:} 17624$\times$8883

\textbf{Resolution:} 0.0568 m/px

\textbf{Area:} 50.1027 ha

\textbf{Longitude range:} 
[-70.377536, -70.365474]

\textbf{Latitude range:} 
[-13.063390, -13.051829]
\end{minipage}
}{
\begin{minipage}[c]{\linewidth}
\centering
\includegraphics[width=0.95\linewidth]{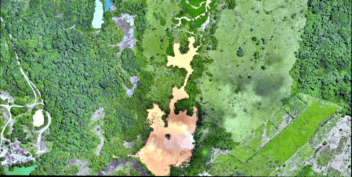}
\end{minipage}
}{
\begin{minipage}[c]{\linewidth}
\centering
\includegraphics[width=0.95\linewidth]{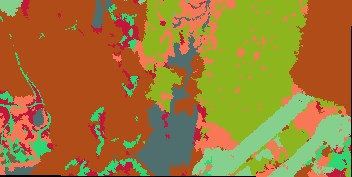}
\end{minipage}
}{
\begin{minipage}[c]{\linewidth}
\vspace{0pt}
\raggedright
Building; Primary Forest; Water bodies; Agricultural crop; Grass; Type 1 natural regeneration; Type 2 natural regeneration; Bare ground
\end{minipage}
}
\midrule

\SiteCard{PlayaMirador2}{
\begin{minipage}[c]{\linewidth}
\vspace{0pt}
\raggedright
\textbf{Split:} val

\textbf{Output size:} 17624$\times$8883

\textbf{Resolution:} 0.0568 m/px

\textbf{Area:} 50.1021 ha

\textbf{Longitude range:} 
[-70.377536, -70.365474]

\textbf{Latitude range:} 
[-13.057446, -13.052861]
\end{minipage}
}{
\begin{minipage}[c]{\linewidth}
\centering
\includegraphics[width=0.95\linewidth]{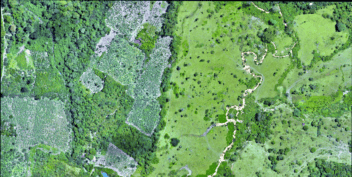}
\end{minipage}
}{
\begin{minipage}[c]{\linewidth}
\centering
\includegraphics[width=0.95\linewidth]{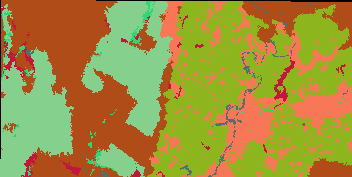}
\end{minipage}
}{
\begin{minipage}[c]{\linewidth}
\vspace{0pt}
\raggedright
Building; Primary Forest; Water bodies; Agricultural crop; Grass; Type 1 natural regeneration; Type 2 natural regeneration; Bare ground
\end{minipage}
}
\midrule

\SiteCard{SantaInesDosMil}{
\begin{minipage}[c]{\linewidth}
\vspace{0pt}
\raggedright
\textbf{Split:} test

\textbf{Output size:} 17387$\times$17267

\textbf{Resolution:} 0.0582 m/px

\textbf{Area:} 100.9625 ha

\textbf{Longitude range:} 
[-70.386657, -70.374307]

\textbf{Latitude range:} 
[-13.063436, -13.050589]
\end{minipage}
}{
\begin{minipage}[c]{\linewidth}
\centering
\includegraphics[width=0.95\linewidth]{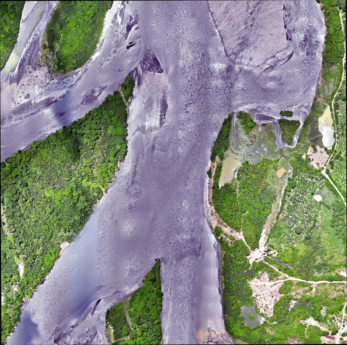}
\end{minipage}
}{
\begin{minipage}[c]{\linewidth}
\centering
\includegraphics[width=0.95\linewidth]{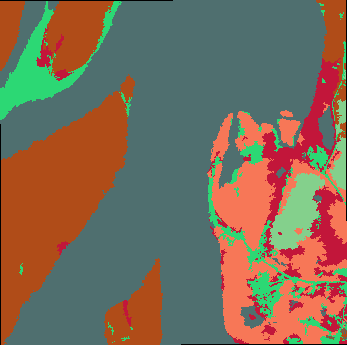}
\end{minipage}
}{
\begin{minipage}[c]{\linewidth}
\vspace{0pt}
\raggedright
Building; Primary Forest; Water bodies; Agricultural crop; Type 1 natural regeneration; Type 2 natural regeneration; Bare ground
\end{minipage}
}

\end{longtable}
}

\subsection{Site-level Composition and Class-wise Appearance Statistics.}

\begin{figure*}[t]
\centering
\includegraphics[width=\textwidth]{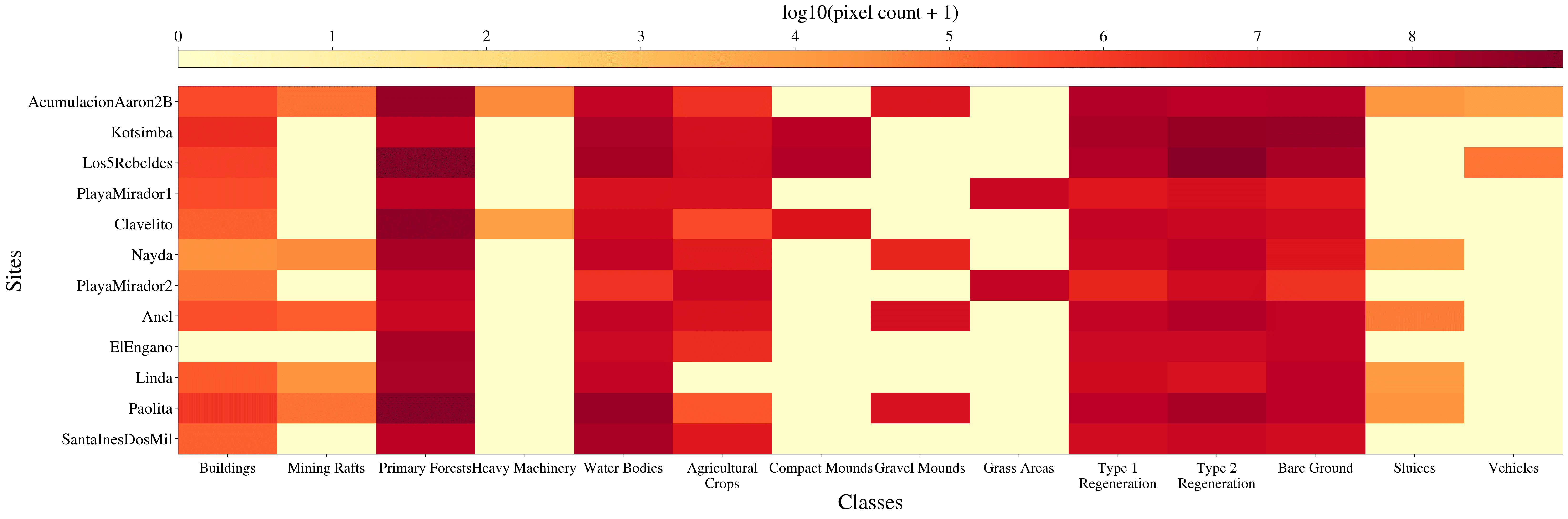}
\caption{Site-level class distributions in ELDOR. The figure shows the semantic composition of each site and provides a detailed view of class imbalance and split-specific variation across the dataset.}
\label{fig:site_class_distribution}
\end{figure*}

\begin{figure*}[t]
\centering
\includegraphics[width=\textwidth]{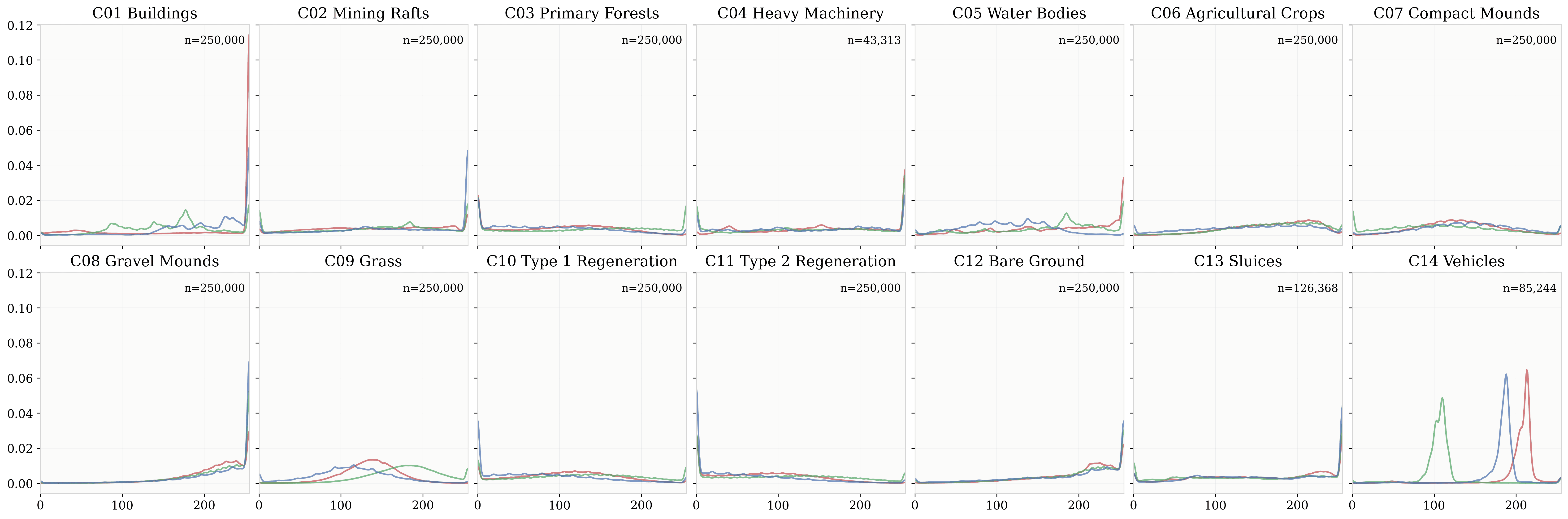}
\caption{Class-wise RGB intensity distributions in ELDOR. For each class, labeled pixels are aggregated from the full orthomosaics across all splits before patch cropping, with up to 250{,}000 pixels sampled per class. Red, green, and blue channel values are histogrammed over the 0--255 range, normalized to probability distributions, and lightly smoothed for visualization. All subplots share the same axes to support direct comparison across classes.}
\label{fig:class_rgb_distribution}
\end{figure*}

Figures~\ref{fig:site_class_distribution} and~\ref{fig:class_rgb_distribution} provide two complementary views of dataset variability in ELDOR. Figure~\ref{fig:site_class_distribution} shows the semantic composition of individual sites and reveals substantial variation in class balance across locations and splits. The distribution is highly imbalanced at both the global and local levels. Dominant classes such as primary forest and water bodies appear across most sites, whereas mining-specific classes are sparse and unevenly distributed. Some categories are concentrated in only a few sites, which increases the difficulty of generalization under the site-level split. These patterns suggest that performance cannot be explained by global class frequency alone. Site-specific composition and spatial context also play an important role in model behavior.

Figure~\ref{fig:class_rgb_distribution} summarizes the RGB intensity distributions of all 14 semantic classes. For each class, labeled pixels are aggregated from the full orthomosaics across the training, validation, and test splits before patch cropping. To keep the computation tractable while preserving the overall appearance statistics, we randomly sample up to 250{,}000 pixels per class from all sites where that class is present. For each class, the red, green, and blue channel values are histogrammed over the 0--255 range, normalized into probability distributions, and lightly smoothed for visualization.

These statistics show that ELDOR is challenging not only because of class imbalance, but also because of visual similarity across several important categories. Major classes often benefit from much larger pixel counts, which makes them easier to learn. Some of them also have relatively more distinctive RGB profiles. This is especially true for water bodies, which are visually more separable than many disturbed land-cover categories. However, the pattern is not uniform across all frequent classes. Primary forest still overlaps with type 1 and type 2 regeneration in its RGB distribution, which helps explain why these ecologically related classes remain difficult to separate in some cases. A similar issue appears among bare ground, gravel mounds, and compact mounds, whose color distributions are also close. These results indicate that color information alone is often insufficient for reliable discrimination. Effective modeling in ELDOR therefore requires the use of spatial context, structure, and scale in addition to local appearance information.

\section{Training and Evaluation Protocols}

\label{app:protocol}

\subsection{Semantic Segmentation Protocol}
\label{app:seg_protocol}

Semantic segmentation is performed on the ELDOR dataset with 14 foreground classes, while the original background label is excluded from training loss computation. For optimization, foreground labels are remapped from $1,\dots,14$ to $0,\dots,13$, and the original background label is assigned \texttt{ignore\_index}=255. We follow the site-level split described in the main text. The training set contains AcumulacionAaron2B, Kotsimba, Los5Rebeldes, and PlayaMirador1; the validation set contains Clavelito, Nayda, and PlayaMirador2; and the test set contains Anel, ElEngano, Linda, Paolita, and SantaInesDosMil.

All orthomosaics are divided into $512 \times 512$ patches with stride 256. Patches with more than 80\% background are removed. This produces 65{,}798 training patches, 15{,}988 validation patches, and 40{,}095 test patches. Unless constrained by a framework's native training recipe, we use AdamW and train for 80 epochs. We use batch size 8 as the unified target setting; where frameworks use different internal execution modes, schedules are adjusted to preserve the same effective training budget. Learning-rate scheduling is framework-aligned: most custom PyTorch baselines use a short warmup followed by cosine decay, while some framework-native implementations use polynomial decay. For iteration-based frameworks, we use the equivalent schedule with \texttt{MAX\_ITERS}=658000 and batch size 8. Automatic mixed precision is enabled.

We adopt a fixed augmentation policy for all segmentation models during training. This policy includes random scale augmentation around the patch size in the range $0.8\times$ to $1.2\times$, together with aligned label transforms, random brightness, contrast, and color jitter, random Gaussian blur with low probability, random horizontal and vertical flips, and random 90-degree rotations. No augmentation is applied at validation or test time. Validation is performed every epoch, and the final test result is reported using the checkpoint that achieves the best validation mIoU. For our unified training scripts, only the best and last checkpoints are retained; some third-party frameworks may additionally produce native bookkeeping artifacts.

For models that support standard supervised training, we evaluate three loss settings: cross-entropy loss combined with Dice loss (CE+Dice), weighted cross-entropy combined with Dice loss (WCE+Dice), and focal loss combined with Dice loss (Focal+Dice) with $\gamma=2$. These loss combinations are applied across multiple architectures when compatible with the training framework. Class weights in the weighted setting are computed from foreground frequencies in the training set using a power-law scheme. For methods with native objectives, such as set-prediction frameworks, we retain their original loss formulations, which may include architecture-specific classification, mask, Dice, or auxiliary terms.

The individual loss components are defined as follows. The standard cross-entropy loss is
\begin{equation}
\mathcal{L}_{\mathrm{CE}} = - \frac{1}{N} \sum_{i=1}^{N} \sum_{c=1}^{C} y_{i,c} \log p_{i,c},
\end{equation}
where $p_{i,c}$ is the predicted probability for class $c$ at pixel $i$, and $y_{i,c}$ is the corresponding one-hot label. The weighted cross-entropy loss is
\begin{equation}
\mathcal{L}_{\mathrm{WCE}} = - \frac{1}{N} \sum_{i=1}^{N} \sum_{c=1}^{C} w_c \, y_{i,c} \log p_{i,c},
\end{equation}
where $w_c$ denotes the class weight for class $c$. We compute these weights from foreground class frequencies in the training set using a power-law scheme, so that rarer classes receive larger weights. We use a multiclass focal loss over foreground classes with ignore masking for background pixels:
\begin{equation}
\mathcal{L}_{\mathrm{Focal}} = - \frac{1}{N} \sum_{i=1}^{N} \sum_{c=1}^{C} y_{i,c}(1-p_{i,c})^{\gamma}\log p_{i,c}.
\end{equation}
Dice loss is implemented as multiclass soft Dice averaged over foreground classes only:
\begin{equation}
\mathcal{L}_{\mathrm{Dice}} =
1 - \frac{1}{C} \sum_{c=1}^{C}
\frac{2 \sum_i p_{i,c} y_{i,c} + \epsilon}
{\sum_i p_{i,c} + \sum_i y_{i,c} + \epsilon},
\end{equation}
where $\epsilon$ is a small constant for numerical stability. For standard supervised baselines, the total loss is formed as the weighted sum of its constituent terms; auxiliary terms are retained where required by the native architecture.

All segmentation metrics are computed from confusion matrices over foreground classes only. Background pixels are ignored during training via \texttt{ignore\_index}=255, and the evaluation confusion matrix is accumulated only over the 14 foreground classes. We report mIoU over all foreground classes, mIoU over classes present in the evaluation split, macro F1, macro F1 over present classes, foreground overall accuracy, and per-class IoU, F1, and ground-truth pixel counts for class-wise analysis. Because some classes are absent in a given split, we report both all-class and present-class variants, and use present-class metrics as an important auxiliary view of performance. Definitions of all metrics are provided in Table~\ref{tab:metrics_definition}.

Efficiency statistics are measured with an input of size $1\times3\times512\times512$. We report the number of parameters in millions, GFLOPs, average inference latency in milliseconds, and peak VRAM usage in GB. Latency is measured on NVIDIA L40S GPUs using 10 warmup iterations followed by 30 timed iterations. The same hardware is used for the reported inference measurements.

\begin{table*}[t]
\centering
\scriptsize
\caption{Summary of evaluation metrics used in ELDOR.}
\label{tab:metrics_definition}
\setlength{\tabcolsep}{4pt}
\renewcommand{\arraystretch}{1.15}

\begin{tabular}{llcl}
\toprule
Category & Metric & Formula & Description \\
\midrule

\multicolumn{4}{l}{\textbf{Semantic Segmentation}} \\
\midrule
 & IoU & $\frac{TP}{TP + FP + FN}$ & Intersection over Union for each class \\
 & mIoU & $\frac{1}{C} \sum_{c=1}^{C} IoU_c$ & Mean IoU over all classes \\
 & mIoU$_p$ & $\frac{1}{|\mathcal{C}_x|} \sum_{c \in \mathcal{C}_x} IoU_c$ & Mean IoU over classes present in the image \\
 & OA & $\frac{TP + TN}{TP + TN + FP + FN}$ & Overall pixel accuracy \\
 & Macro F1 & $\frac{1}{C} \sum_{c=1}^{C} F1_c$ & Average F1 across classes \\
\midrule

\multicolumn{4}{l}{\textbf{Multi-label Classification}} \\
\midrule
 & CP & $\frac{1}{C} \sum_{c=1}^{C} \frac{TP_c}{TP_c + FP_c}$ & Class-wise precision (macro precision) \\
 & CR & $\frac{1}{C} \sum_{c=1}^{C} \frac{TP_c}{TP_c + FN_c}$ & Class-wise recall (macro recall) \\
 & CF1 & $\frac{2 \cdot CP \cdot CR}{CP + CR}$ & Class-wise F1 computed from CP and CR \\
 & OP & $\frac{\sum_c TP_c}{\sum_c (TP_c + FP_c)}$ & Overall precision \\
 & OR & $\frac{\sum_c TP_c}{\sum_c (TP_c + FN_c)}$ & Overall recall \\
 & OF1 & $\frac{2 \cdot OP \cdot OR}{OP + OR}$ & Overall F1 score \\
 & mAP & $\frac{1}{C} \sum_{c=1}^{C} AP_c$ & Mean average precision over classes \\
 & Macro F1 & $\frac{1}{C} \sum_{c=1}^{C} F1_c$ & Mean of class-wise F1 scores \\
 & Sample F1 & $\frac{1}{N} \sum_{i=1}^{N} F1_i$ & F1 computed per sample and averaged \\
\midrule

\multicolumn{4}{l}{\textbf{Efficiency Metrics}} \\
\midrule
 & Params & -- & Number of model parameters \\
 & GFLOPs & -- & Floating point operations (in billions) \\
 & Latency & -- & Inference time per image \\
 & Peak VRAM & -- & Maximum GPU memory usage during inference \\
\bottomrule
\end{tabular}

\end{table*}

\subsection{Segmentation-derived Recognition Protocol}
\label{app:seg_rec_protocol}

Segmentation-derived recognition converts final segmentation outputs into image-level multi-label predictions, so that dense prediction models can be evaluated under a classification-based formulation. Each image is mapped to a 14-dimensional binary vector over foreground classes.

For each image $i$ and foreground class $c$, the ground-truth label is defined as
\begin{equation}
y_{i,c} = 1 \;\; \text{if class } c \text{ appears in the ground-truth segmentation map, and } 0 \text{ otherwise}.
\end{equation}
The predicted label is defined analogously from the final predicted segmentation map: it is set to 1 if class $c$ appears at least once in the prediction, and 0 otherwise. Ignored pixels are excluded from all computations.

Performance is evaluated using multi-label classification metrics, including class-wise precision (CP), recall (CR), and F1 score (CF1), overall precision (OP), recall (OR), and F1 score (OF1), as well as macro F1 and sample-wise F1. We also compute mean average precision (mAP). For each class, average precision (AP) is computed from the predicted scores $s_{i,c}$ and binary ground-truth labels $y_{i,c}$ across all images.

For dense segmentation models with calibrated per-pixel logits, the score $s_{i,c}$ is defined as the maximum softmax confidence for class $c$ over valid pixels whose final predicted label is class $c$:
\begin{equation}
s_{i,c} =
\begin{cases}
\max\limits_{p \in \Omega_i:\,\hat{z}_i(p)=c} P_{i,c}(p), & \text{if class } c \text{ is predicted in image } i, \\
0, & \text{otherwise},
\end{cases}
\end{equation}
where $\Omega_i$ denotes the set of valid pixels, $\hat{z}_i(p)$ is the final predicted class at pixel $p$, and $P_{i,c}(p)$ is the softmax probability of class $c$. This confidence definition ties the ranking score to the final hard prediction while avoiding systematic preference for spatially large classes, which is important for small mining-related categories such as mining rafts and sluices.

For query-based segmentation models, we use the final semantic class score map as a proxy confidence and apply the same maximum-over-final-predicted-pixels rule. For SESSRS, whose final post-processed masks do not have calibrated logits, we use a proxy score obtained by taking the maximum base-model softmax confidence within the final SESSRS mask for class $c$. The score is set to 0 when class $c$ is absent from the final mask.

The final mAP is obtained by averaging class-wise AP values over foreground classes that appear in the evaluation split. Definitions of all metrics are provided in Table~\ref{tab:metrics_definition}. We also report per-class precision, recall, F1, and AP for detailed analysis.

\subsection{Direct Multi-label Classification Protocol}
\label{app:mlc_protocol}

Direct multi-label classification is performed on the same split used for semantic segmentation. We use the same $512 \times 512$ cropped RGB patches described in the main text and Appendix~\ref{app:seg_protocol}. Each patch is associated with a 14-dimensional binary vector over the foreground classes. The image-level labels are constructed in the same way as in Appendix~\ref{app:seg_rec_protocol}, except that the classifier is trained directly on these binary targets rather than on dense segmentation masks.

Unless constrained by a framework's native training recipe, we use AdamW and train for 80 epochs with learning rate $2\times10^{-4}$, weight decay $0.05$, and batch size 8. We use 5 epochs of linear warmup followed by cosine decay. Automatic mixed precision is enabled, gradients are clipped with \texttt{max\_norm}=1.0, and the final test result is reported using the checkpoint that achieves the best validation mAP. We adopt a unified augmentation policy for direct multi-label classification. During training, this includes random flips, random rotation, color jitter, padding if needed, random cropping to $512 \times 512$, and ImageNet normalization. At validation and test time, we apply the same normalization without stochastic augmentation.

Unlike semantic segmentation, direct multi-label classification models are trained using their original loss functions or native training objectives as proposed in the respective papers. No cross-model loss standardization is applied. The main training objectives in this benchmark are binary cross-entropy (BCE), asymmetric loss (ASL)~\cite{ben2022multi}, resample loss~\cite{lin2024distributionally}, BCE combined with a triplet term~\cite{zhou2021multi}, scalable neighbor discriminative loss with BCE (SNDL-BCE)~\cite{kang2020graph}, and self-adaptive training (SAT)~\cite{huang2020self}.

The standard BCE objective is
\begin{equation}
\mathcal{L}_{\mathrm{BCE}} =
-\frac{1}{C}\sum_{c=1}^{C}
\left[
y_c \log \sigma(z_c) + (1-y_c)\log\left(1-\sigma(z_c)\right)
\right],
\end{equation}
where $z_c$ is the logit for class $c$, $\sigma(\cdot)$ is the sigmoid function, and $y_c \in \{0,1\}$ is the ground-truth label. Some implementations use multi-label soft margin loss, which is equivalent in this setting.

For ASL-based methods, we use the asymmetric loss~\cite{ben2022multi}
\begin{equation}
\mathcal{L}_{\mathrm{ASL}} =
-\frac{1}{C}\sum_{c=1}^{C}
\begin{cases}
(1-p_c)^{\gamma^+}\log(p_c), & y_c = 1, \\[4pt]
(p_{m,c})^{\gamma^-}\log(1-p_{m,c}), & y_c = 0,
\end{cases}
\end{equation}
where $p_c=\sigma(z_c)$, $p_{m,c}=\max(p_c-\delta,0)$, $\gamma^+=0$, $\gamma^-=4$, and $\delta=0.05$.

For methods using resample loss, we retain the native long-tail rebalancing formulation~\cite{lin2024distributionally}:
\begin{equation}
\mathcal{L}_{\mathrm{DRL}} =
-\frac{1}{C}\sum_{c=1}^{C}
w_c \, f_c \, \ell_{\mathrm{BCE}}(z_c, y_c),
\end{equation}
where $w_c$ is a class-frequency-based reweighting term and $f_c$ is a focal-style modulation term. Class frequencies are computed from the ELDOR training split.

For methods that combine classification and metric learning, we use~\cite{zhou2021multi}
\begin{equation}
\mathcal{L}_{\mathrm{CPCL}} =
\mathcal{L}_{\mathrm{BCE}} + \beta \mathcal{L}_{\mathrm{triplet}},
\end{equation}
where $\mathcal{L}_{\mathrm{triplet}}$ is a margin-based triplet term that encourages class-aware separation in the learned feature space. We follow the original setting with triplet margin $m=20$ and loss weight $\beta=0.5$.

For GRN, we retain the original SNDL-BCE objective~\cite{kang2020graph}, which combines BCE with a neighborhood-discriminative term to encourage class-aware structure in the learned embedding space. For methods using self-adaptive training (SAT)~\cite{huang2020self}, we retain the original adaptive training scheme.

Performance is evaluated using the same multi-label metrics defined in Table~\ref{tab:metrics_definition}, including CP, CR, CF1, OP, OR, OF1, mAP, macro F1, and sample-wise F1. We also report per-class average precision for detailed analysis. For threshold-based metrics, sigmoid outputs are thresholded at 0.5. The main model-selection metric is validation mAP, while mAP itself is computed in a threshold-free manner by integrating the precision--recall curve for each class and then averaging over all 14 classes.

\begin{figure*}[t]
\centering
\begin{tcolorbox}[
  width=0.96\textwidth,
  colback=white,
  colframe=black,
  colbacktitle=black!12,
  coltitle=black,
  title=\textbf{Protocol A: VLM-based Multi-label Presence Prediction},
  boxrule=0.8pt,
  arc=1.2mm,
  left=1.4mm,right=1.4mm,top=1.1mm,bottom=1.1mm
]
\small
\textbf{A1. Binary QA (GeoChat, RS-LLaVA, VHM).}
\textbf{Prompt template:}
\begin{quote}\ttfamily\small
Is there [CLASS] in this image? Answer yes or no.
\end{quote}
\normalfont
\textbf{Examples:}
\begin{quote}\ttfamily\small
Is there bare ground in this image? Answer yes or no.\\
Is there mining raft in this image? Answer yes or no.
\end{quote}

\textbf{A2. Contrastive scoring (RemoteCLIP, GeoRSCLIP, DOFA-CLIP).}
\textbf{Prompt pair per class:}
\begin{quote}\ttfamily\small
There is [CLASS] in this image.\\
There is no [CLASS] in this image.
\end{quote}
\normalfont
\textbf{Example pair:}
\begin{quote}\ttfamily\small
There is bare ground in this image.\\
There is no bare ground in this image.
\end{quote}
For RemoteCLIP and GeoRSCLIP, we apply paired softmax scoring. For DOFA-CLIP, we use SigLIP-style sigmoid scoring with independent thresholding.

\textbf{A3. Positive-threshold scoring (RemoteCLIP$^+$, GeoRSCLIP$^+$).}
For RemoteCLIP$^+$ and GeoRSCLIP$^+$, we use the same positive prompt as in A2 but define the image-level confidence using only the positive prompt score. Specifically, the confidence is computed from the positive prompt alone and thresholded at $0.5$ to determine binary presence.

\textbf{A4. Model-native multi-label classification (RemoteSAM).}
For RemoteSAM, we evaluate the model's official native multi-label classification head. For each class, the image-level score is computed as
$0.5 \cdot \max p_{\mathrm{fg}} + 0.5 \cdot \operatorname{avg} p_{\mathrm{fg}}$,
where $\max p_{\mathrm{fg}}$ and $\operatorname{avg} p_{\mathrm{fg}}$ denote the maximum and mean foreground probabilities for that class. A class is treated as present if the resulting score exceeds $0.5$.

\textbf{A5. Segmentation-derived recognition (RemoteSAM, SAM3).}
For both RemoteSAM and SAM3, we additionally derive image-level recognition from the segmentation masks produced by Protocol~B. For each class, the image-level confidence is defined as the maximum mask confidence returned for that class in the image. Binary presence is then obtained from the aggregated segmentation map by treating a class as present if any pixel is assigned to that class. For RemoteSAM, this segmentation-derived recognition is evaluated separately from its native image-level classification in A4 and is not equivalent to it.
\end{tcolorbox}
\caption{\textbf{Prompt templates for Protocol~A.} Protocol~A covers image-level multi-label presence prediction through binary QA for generative models (A1), contrastive scoring for CLIP-style models (A2), positive-threshold scoring for RemoteCLIP$^+$ and GeoRSCLIP$^+$ (A3), model-native multi-label classification for RemoteSAM (A4), and segmentation-derived recognition for RemoteSAM and SAM3 (A5).}
\label{fig:vlm_prompt_recognition}
\end{figure*}

\begin{figure*}[t]
\centering
\begin{tcolorbox}[
  width=0.96\textwidth,
  colback=white,
  colframe=black,
  colbacktitle=black!12,
  coltitle=black,
  title=\textbf{Protocol B: Prompted Segmentation},
  boxrule=0.8pt,
  arc=1.2mm,
  left=1.4mm,
  right=1.4mm,
  top=1.1mm,
  bottom=1.1mm
]
\small

\textbf{Prompting strategy.}
For RemoteSAM, we use its model-native default prompting interface, where each class is queried by its class name using the default RemoteSAM template. For SAM3, we use fixed class-specific descriptive prompts.

\textbf{Inference.}
For RemoteSAM, we query each image--class pair using the model-native default class query and aggregate the resulting binary masks into a final segmentation map using confidence-weighted assignment. For SAM3, we use the prompted mask-generation interface with the descriptive prompt listed below, combine all returned masks for a class using confidence-weighted max pooling, and assign each pixel to the class with the highest confidence score.

\vspace{0.3em}
\scriptsize
\renewcommand{\arraystretch}{1.06}
\setlength{\tabcolsep}{5pt}

\begin{tabular}{p{0.28\textwidth}p{0.60\textwidth}}
\toprule
\textbf{Class} & \textbf{SAM3 Description Prompt} \\
\midrule
Building & \texttt{building with metal roof} \\
Mining raft & \texttt{rectangular floating mining platform} \\
Primary Forest & \texttt{dense forest tree cover} \\
Heavy machinery & \texttt{large mining excavator machinery} \\
Water bodies & \texttt{water filled pond or pool} \\
Agricultural crop & \texttt{agricultural crop field} \\
Compact mounds & \texttt{large truncated cone gravel mound} \\
Gravel mounds & \texttt{small cone gravel pile} \\
Grass & \texttt{grass pasture} \\
Type 1 natural regeneration & \texttt{low vegetation regeneration} \\
Type 2 natural regeneration & \texttt{woody shrub tree regeneration} \\
Bare ground & \texttt{bare exposed ground} \\
Sluice & \texttt{wooden sluice mining structure} \\
Vehicles & \texttt{pickup truck or car vehicle} \\
\bottomrule
\end{tabular}

\vspace{0.35em}
\scriptsize
RemoteSAM uses its default class-name query for each class, e.g., \texttt{[CLASS\_NAME] in the image}.

\end{tcolorbox}
\caption{\textbf{Prompt templates for Protocol~B.} Protocol~B evaluates prompted segmentation using model-specific prompting: RemoteSAM uses its model-native default class-name query, while SAM3 uses fixed descriptive prompts.}
\label{fig:vlm_prompt_segmentation}
\end{figure*}

\subsection{VLM-based Recognition Protocol}
\label{app:vlm_protocol}

We evaluate vision-language models (VLMs) on ELDOR in a fully zero-shot setting, without any fine-tuning or training on ELDOR data. The goal is to assess whether pretrained remote-sensing VLMs can recognize the presence of the 14 foreground classes under a fixed closed label space, rather than to benchmark open-ended captioning or unconstrained visual question answering. We evaluate eight models in total: three generative or chat-style models (GeoChat, RS-LLaVA, and VHM), three CLIP-style contrastive models (RemoteCLIP, GeoRSCLIP, and DOFA-CLIP), and two models with spatial grounding or segmentation capability (RemoteSAM and SAM~3). All models are tested on the same $512\times512$ patches.

Since no model fitting is performed, we evaluate VLMs separately on the train, validation, and test splits in the same zero-shot setting. Image-level ground-truth labels are derived from the segmentation masks as described above, with background and ignored pixels excluded.

For image-level recognition, we term the task VLM-based multi-label presence prediction. This is the primary VLM evaluation setting and is implemented through five variants under a unified Protocol~A. For generative models, namely GeoChat~\cite{kuckreja2024geochat}, RS-LLaVA~\cite{bazi2024rs}, and VHM~\cite{pang2025vhm}, we use binary QA and query each class independently. For contrastive models, namely RemoteCLIP~\cite{liu2024remoteclip}, GeoRSCLIP~\cite{zhang2024rs5m}, and DOFA-CLIP~\cite{xiong2025dofa}, we replace binary QA with paired positive and negative textual statements and determine the binary decision from their image-text scores. RemoteCLIP and GeoRSCLIP use paired softmax scoring, whereas DOFA-CLIP uses SigLIP-style sigmoid scoring with independent thresholding. We additionally evaluate a positive-threshold variant for RemoteCLIP and GeoRSCLIP, denoted as A3, where the image-level confidence is computed from the positive prompt alone and thresholded at $0.5$ for binary prediction. For RemoteSAM~\cite{yao2025remotesam}, we evaluate two image-level recognition variants. Under A4, we use its official native multi-label classification head, where the image-level score for each class is computed as $0.5 \cdot \max p_{\mathrm{fg}} + 0.5 \cdot \operatorname{avg} p_{\mathrm{fg}}$, and the class is predicted as present if this score exceeds $0.5$. Under A5, we additionally derive image-level recognition from the segmentation masks produced by Protocol~B. For SAM3~\cite{carion2025sam}, image-level recognition is evaluated only under this segmentation-derived setting, since it does not provide a dedicated native image-level classification head. In A5, for both RemoteSAM and SAM3, the image-level confidence for each class is defined as the maximum mask confidence returned for that class in the image, and binary presence is obtained from the aggregated segmentation map by treating a class as present if any pixel is assigned to that class. We report the same multi-label metrics as in the previous recognition protocols, including CP, CR, CF1, OP, OR, OF1, macro F1, micro F1, and sample F1, together with per-class results. We note that RemoteCLIP and GeoRSCLIP exhibit the known all-positive thresholding behavior induced by paired softmax scoring, which makes threshold-based binary metrics less reliable. This issue does not apply to DOFA-CLIP.

For models that support spatial output, we additionally evaluate a unified prompted segmentation setting under Protocol~B. Specifically, RemoteSAM and SAM3 are queried independently for each foreground class and return class-specific masks. For RemoteSAM, we use its native text-guided segmentation interface with the model-native default class-name query. For SAM3, we use the prompted mask-generation interface with fixed class-specific descriptive prompts and a fixed confidence threshold. The class-specific masks are aggregated into a final semantic prediction map using confidence-based assignment. For RemoteSAM, the binary masks returned for each class are assigned according to their associated confidence scores. For SAM3, multiple masks returned for the same class are first combined using confidence-weighted max pooling, and each pixel is then assigned to the class with the highest confidence. The resulting segmentation maps are evaluated using the same dense prediction metrics as the supervised benchmark, including mIoU, image-level present-class mIoU, macro F1, foreground overall accuracy, and per-class IoU and F1. Figures~\ref{fig:vlm_prompt_recognition} and~\ref{fig:vlm_prompt_segmentation} summarize the prompt templates used in Protocols~A and~B.

\FloatBarrier

\section{Detailed Experimental Results}
\label{app:experiments}

\subsection{Semantic Segmentation}
\label{app:seg}

This appendix provides complete semantic segmentation results for three groups of methods: general segmentation models, remote-sensing-specific methods, and methods related to vision foundation models. We report both aggregate metrics and class-wise results in order to provide a more comprehensive view of segmentation quality. The overall comparison demonstrates differences in accuracy and efficiency, while the per-class comparison reveals how different methods behave on dominant versus rare categories.

\begin{table*}[t]
\centering
\scriptsize
\caption{Results of general segmentation models.}
\label{tab:app_seg_general}
\setlength{\tabcolsep}{2.2pt}
\renewcommand{\arraystretch}{1.05}
\begin{tabular*}{\textwidth}{@{\extracolsep{\fill}}llcccccccc}
\toprule
Model & Backbone & Loss 
& mIoU$_p$ & mIoU & Macro-F1 & OA 
& Params & GFLOPs & Latency \\
\midrule

\multirow{6}{*}{DeepLabV3+}
& ConvNeXt-Tiny & CE+Dice    & 0.3895 & 0.2782 & 0.5260 & 0.7189 & 29.31 & 75.91 & 4.47 \\
& ConvNeXt-Tiny & Focal+Dice & 0.3735 & 0.2668 & 0.5068 & 0.7048 & 29.31 & 75.91 & 4.47 \\
& ConvNeXt-Tiny & WCE+Dice   & 0.3766 & 0.2690 & 0.5193 & 0.6840 & 29.31 & 75.91 & 4.47 \\
& ResNet-50     & CE+Dice    & 0.3662 & 0.2817 & 0.4897 & 0.7092 & 26.68 & 73.59 & 3.00 \\
& ResNet-50     & Focal+Dice & 0.3515 & 0.2929 & 0.4819 & 0.7000 & 26.68 & 73.59 & 3.00 \\
& ResNet-50     & WCE+Dice   & 0.3438 & 0.2644 & 0.4755 & 0.6951 & 26.68 & 73.59 & 3.00 \\

\midrule
UPerNet
& Swin-Tiny     & CE+Dice    & 0.3371 & 0.2593 & 0.4651 & 0.6729 & 59.84 & 472.12 & 22.63 \\

\midrule
OCRNet
& HRNet-W48     & CE+Dice    & 0.2722 & 0.2268 & 0.3954 & 0.5735 & 70.37 & 325.35 & 61.49 \\

\midrule
\multirow{3}{*}{BiSeNetv2}
& Custom Bilateral & CE+Dice    & 0.3486 & 0.2905 & 0.4673 & \cellcolor{blue!10}\textbf{0.7330} & 5.22 & 35.86 & 4.02 \\
& Custom Bilateral & Focal+Dice & 0.3576 & 0.2980 & 0.4847 & 0.6872 & 5.22 & 35.86 & 4.02 \\
& Custom Bilateral & WCE+Dice   & 0.3475 & 0.2895 & 0.4685 & 0.6873 & 5.22 & 35.86 & 4.02 \\

\midrule
\multirow{3}{*}{SegFormer}
& MiT-B2 & CE+Dice    & 0.3222 & 0.2301 & 0.4594 & 0.6484 & 27.36 & 121.93 & 8.28 \\
& MiT-B2 & Focal+Dice & 0.3332 & 0.2380 & 0.4760 & 0.6385 & 27.36 & 121.93 & 8.28 \\
& MiT-B2 & WCE+Dice   & \cellcolor{blue!10}\textbf{0.4010} & \cellcolor{blue!10}\textbf{0.3084} & \cellcolor{blue!10}\textbf{0.5297} & 0.7163 & 27.36 & 121.93 & 8.28 \\

\midrule
\multirow{3}{*}{STDC2}
& STDCNet   & CE+Dice    & 0.3438 & 0.2865 & 0.4617 & 0.7186 & 16.08 & 44.43 & 6.33 \\
& STDCNet   & Focal+Dice & 0.3286 & 0.2738 & 0.4493 & 0.6979 & 16.08 & 44.43 & 6.33 \\
& STDCNet   & WCE+Dice   & 0.3371 & 0.2408 & 0.4659 & 0.6781 & 16.08 & 44.43 & 6.33 \\

\midrule
Mask2Former
& ResNet-50 & Set CE+Mask+Dice & 0.2985 & 0.2985 & 0.4285 & 0.6561 & 44.01 & 133.29 & 17.46 \\

\midrule
SegNeXt
& MSCAN-Tiny & CE+Dice & 0.2682 & 0.1916 & 0.3884 & 0.6065 & 4.23 & 12.64 & 9.26 \\

\midrule
DDRNet
& DDRNet-23-slim & Focal+Dice & 0.2869 & 0.2391 & 0.4089 & 0.6188 & 5.70 & 9.13 & \cellcolor{blue!10}\textbf{3.41} \\

\midrule
Afformer
& AFFormer-Base & CE+Dice & 0.3047 & 0.2176 & 0.4362 & 0.6389 & \cellcolor{blue!10}\textbf{2.97} & 8.57 & 7.47 \\

\midrule
\multirow{3}{*}{EfficientViT}
& EfficientViT-B2 & CE+Dice    & 0.3799 & 0.2713 & 0.5065 & 0.7258 & 15.28 & 18.32 & 6.42 \\
& EfficientViT-B2 & Focal+Dice & 0.3693 & 0.2638 & 0.5086 & 0.6781 & 15.28 & 18.32 & 6.42 \\
& EfficientViT-B2 & WCE+Dice   & 0.3790 & 0.2707 & 0.5181 & 0.7154 & 15.28 & 18.32 & 6.42 \\

\midrule
SeaFormer
& SeaFormer-Base & CE+Dice & 0.3117 & 0.2398 & 0.4408 & 0.6392 & 8.58 & \cellcolor{blue!10}\textbf{3.47} & 12.47 \\

\midrule
\multirow{3}{*}{PIDNet}
& PIDNet-M & CE+Dice    & 0.3437 & 0.2865 & 0.4675 & 0.6834 & 28.54 & 44.57 & 6.21 \\
& PIDNet-M & Focal+Dice & 0.3431 & 0.2859 & 0.4671 & 0.7098 & 28.54 & 44.57 & 6.21 \\
& PIDNet-M & WCE+Dice   & 0.3401 & 0.2430 & 0.4727 & 0.6792 & 28.54 & 44.57 & 6.21 \\

\midrule
CGRSeg
& EfficientFormerV2-B & CE+Dice & 0.2679 & 0.1913 & 0.3961 & 0.5844 & 19.08 & 7.50 & 14.46 \\

\midrule
PEM
& ResNet-50 & Set CE+Mask+Dice & 0.2789 & 0.2789 & 0.4011 & 0.6502 & 35.53 & 60.60 & 11.52 \\

\midrule
VMamba
& VMamba-Tiny & CE+Dice & 0.2829 & 0.2176 & 0.4040 & 0.6312 & 48.73 & 450.41 & 31.45 \\

\bottomrule
\end{tabular*}
\end{table*}

\subsubsection{Overall Segmentation Performance}

\begin{table*}[t]
\centering
\scriptsize
\caption{Results of remote-sensing-specific methods. An asterisk ($^{*}$) denotes the use of an auxiliary loss.}
\label{tab:app_seg_rs}
\setlength{\tabcolsep}{2.2pt}
\renewcommand{\arraystretch}{1.05}
\begin{tabular*}{\textwidth}{@{\extracolsep{\fill}}llcccccccc}
\toprule
Model & Backbone & Loss
& mIoU$_p$ & mIoU & Macro-F1 & OA
& Params & GFLOPs & Latency \\
\midrule

\multirow{4}{*}{FarSeg}
& ResNet-50 & CE  & 0.3564 & 0.2970 & 0.4726 & 0.7130 & 31.37 & 94.12 & 3.92 \\
& ResNet-50 & CE+Dice      & 0.3717 & 0.3098 & 0.4934 & 0.6893 & 31.37 & 94.12 & 3.92 \\
& ResNet-50 & Focal+Dice   & 0.3744 & 0.3120 & 0.4899 & 0.7342 & 31.37 & 94.12 & 3.92 \\
& ResNet-50 & WCE+Dice     & 0.3642 & 0.2601 & 0.5066 & 0.6636 & 31.37 & 94.12 & 3.92 \\

\midrule
BANet
& ResT-Lite & CE+Dice & 0.2926 & 0.2250 & 0.4147 & 0.6535 & 12.86 & 31.38 & 4.81 \\

\midrule
ABCNet
& ResNet-18 & CE+Dice$^{*}$ & 0.3145 & 0.2621 & 0.4302 & 0.6831 & 13.96 & 32.39 & 2.81 \\

\midrule
\multirow{3}{*}{MANet}
& ResNet-50 & CE+Dice    & 0.3711 & 0.3093 & 0.4922 & 0.6862 & 35.86 & 109.62 & 4.79 \\
& ResNet-50 & Focal+Dice & 0.3999 & 0.2856 & 0.5431 & 0.6848 & 35.86 & 109.62 & 4.79 \\
& ResNet-50 & WCE+Dice   & 0.3828 & 0.2734 & 0.5228 & 0.6759 & 35.86 & 109.62 & 4.79 \\

\midrule
\multirow{3}{*}{UNetFormer}
& ResNet-18 & CE+Dice$^{*}$    & 0.3941 & \cellcolor{green!10}\textbf{0.3284} & 0.5152 & 0.7276 & 11.73 & \cellcolor{green!10}\textbf{23.55} & 3.40 \\
& ResNet-18 & Focal+Dice$^{*}$ & 0.3765 & 0.3137 & 0.4989 & 0.7182 & 11.73 & 23.55 & 3.40 \\
& ResNet-18 & WCE+Dice$^{*}$   & 0.3566 & 0.2743 & 0.4931 & 0.6811 & 11.73 & 23.55 & 3.40 \\

\midrule
DC-Swin
& Swin-Small & CE+Dice & 0.2971 & 0.2476 & 0.4173 & 0.6584 & 66.95 & 144.39 & 12.38 \\

\midrule
\multirow{3}{*}{A2-FPN}
& ResNet-18 & CE+Dice    & 0.3688 & 0.3073 & 0.4834 & 0.7335 & 12.16 & 27.14 & \cellcolor{green!10}\textbf{2.12} \\
& ResNet-18 & Focal+Dice & 0.3363 & 0.2802 & 0.4602 & 0.6659 & 12.16 & 27.14 & 2.12 \\
& ResNet-18 & WCE+Dice   & 0.3720 & 0.2657 & 0.5107 & 0.7094 & 12.16 & 27.14 & 2.12 \\

\midrule
LoGCAN
& ResNet-50 & CE$^{*}$ & 0.3108 & 0.2590 & 0.4081 & \cellcolor{green!10}\textbf{0.7474} & 30.92 & 99.23 & 6.05 \\

\midrule
FarSeg++
& MiT-B2 & CE & 0.3062 & 0.2187 & 0.4358 & 0.6669 & 32.56 & 95.08 & 8.77 \\

\midrule
SACANet
& HRNet-W32 & CE$^{*}$ & 0.3294 & 0.2534 & 0.4557 & 0.6573 & 30.27 & 115.90 & 19.02 \\

\midrule
DOCNet
& HRNet-W32 & CE$^{*}$ & 0.3147 & 0.2421 & 0.4398 & 0.6785 & 39.13 & 395.32 & 20.64 \\

\midrule
\multirow{3}{*}{PPMambaSeg}
& SWSL & CE+Dice     & 0.3520 & 0.2934 & 0.4780 & 0.6683 & 21.70 & 45.99 & 11.28 \\
& SWSL & Focal+Dice  & 0.3897 & 0.3248 & 0.5100 & 0.7103 & 21.70 & 45.99 & 11.28 \\
& SWSL & WCE+Dice    & 0.3854 & 0.2753 & 0.5298 & 0.6816 & 21.70 & 45.99 & 11.28 \\

\midrule
\multirow{3}{*}{RS3Mamba}
& VMamba-Tiny & CE+Dice    & 0.2385 & 0.1987 & 0.3080 & 0.7257 & 43.33 & 78.59 & 11.60 \\
& VMamba-Tiny & Focal+Dice & 0.2399 & 0.1999 & 0.3125 & 0.7251 & 43.33 & 78.59 & 11.60 \\
& VMamba-Tiny & WCE+Dice   & 0.3068 & 0.2556 & 0.4280 & 0.6519 & 43.33 & 78.59 & 11.60 \\

\midrule
\multirow{3}{*}{PyramidMamba}
& Swin-Base & CE+Dice    & 0.3985 & 0.2847 & 0.5360 & 0.6833 & 125.11 & 217.75 & 13.76 \\
& Swin-Base & Focal+Dice & 0.3961 & 0.2830 & 0.5304 & 0.6699 & 125.11 & 217.75 & 13.76 \\
& Swin-Base & WCE+Dice   & \cellcolor{green!10}\textbf{0.4414} & 0.3153 & \cellcolor{green!10}\textbf{0.5864} & 0.6967 & 125.11 & 217.75 & 13.76 \\

\midrule
LoGCAN++
& RepViT-M2.3 & CE$^{*}$ & 0.2264 & 0.2264 & 0.3066 & 0.6353 & 25.19 & 74.37 & 17.19 \\

\midrule
MF-Mamba
& HRNet-W18 & CE+Dice & 0.3001 & 0.2501 & 0.4242 & 0.6376 & \cellcolor{green!10}\textbf{11.27} & 38.94 & 20.54 \\

\midrule
\multirow{3}{*}{MCPNet}
& ResNet-50 & CE+Dice    & 0.3056 & 0.2183 & 0.4267 & 0.6680 & 45.15 & 110.99 & 6.88 \\
& ResNet-50 & Focal+Dice & 0.3233 & 0.2487 & 0.4448 & 0.6898 & 45.15 & 110.99 & 6.88 \\
& ResNet-50 & WCE+Dice   & 0.3193 & 0.2281 & 0.4552 & 0.6405 & 45.15 & 110.99 & 6.88 \\

\bottomrule
\end{tabular*}
\end{table*}

\begin{table*}[t]
\centering
\scriptsize
\caption{Results of remote-sensing segmentation methods with VFM. An asterisk ($^{*}$) denotes the use of an auxiliary loss.}
\label{tab:app_seg_vfm}
\setlength{\tabcolsep}{2.2pt}
\renewcommand{\arraystretch}{1.05}
\begin{tabular*}{\textwidth}{@{\extracolsep{\fill}}llcccccccc}
\toprule
Model & Backbone & Loss
& mIoU$_p$ & mIoU & Macro-F1 & OA
& Params & GFLOPs & Latency \\
\midrule

\multirow{3}{*}{HQ-SAM}
& ViT-B + HQ Decoder & CE+Dice    & 0.2485 & 0.1775 & 0.3558 & 0.6390 & 97.83 & 983.13 & 74.34 \\
& ViT-B + HQ Decoder & Focal+Dice & 0.2503 & 0.1925 & 0.3561 & 0.6539 & 97.83 & 983.13 & 74.34 \\
& ViT-B + HQ Decoder & WCE+Dice   & 0.2538 & 0.1813 & 0.3711 & 0.6150 & 97.83 & 983.13 & 74.34 \\

\midrule
\multirow{4}{*}{SAM\_RS}
& ABCNet + SAM Priors       & Seg+Bdy+Obj & 0.2964 & 0.2470 & 0.4098 & 0.6573 & 13.96 & 32.35 & \cellcolor{red!10}\textbf{2.57} \\
& CMTFNet + SAM Priors      & Seg+Bdy+Obj & 0.2916 & 0.2243 & 0.4084 & 0.6598 & 30.07 & 66.14 & 6.22 \\
& FTUNetFormer + SAM Priors & Seg+Bdy+Obj & 0.2922 & 0.2435 & 0.4094 & 0.6871 & 96.14 & 51.04 & 14.66 \\
& UNetFormer + SAM Priors   & Seg+Bdy+Obj & 0.3241 & 0.2700 & 0.4452 & 0.6839 & \cellcolor{red!10}\textbf{11.69} & 23.52 & 3.32 \\

\midrule
\multirow{3}{*}{RSAM-Seg}
& SAM-ViT-B (Frozen Encoder) & CE+Dice    & 0.3263 & 0.2510 & 0.4472 & 0.6978 & 98.59 & 247.05 & 15.36 \\
& SAM-ViT-B (Frozen Encoder) & Focal+Dice & 0.3450 & 0.2654 & 0.4642 & \cellcolor{red!10}\textbf{0.7430} & 98.59 & 247.05 & 15.36 \\
& SAM-ViT-B (Frozen Encoder) & WCE+Dice   & 0.3696 & 0.2640 & 0.5085 & 0.6978 & 98.59 & 247.05 & 15.36 \\

\midrule
\multirow{6}{*}{SAM2.1}
& Hiera-B+ (Frozen, MSFPN)  & CE+Dice    & 0.2422 & 0.1863 & 0.3510 & 0.6193 & 83.90 & 191.82 & 10.49 \\
& Hiera-B+ (Frozen, MSFPN)  & Focal+Dice & 0.2351 & 0.1809 & 0.3438 & 0.6158 & 83.90 & 191.82 & 10.49 \\
& Hiera-B+ (Frozen, MSFPN)  & WCE+Dice   & 0.2207 & 0.1577 & 0.3235 & 0.5938 & 83.90 & 191.82 & 10.49 \\
& Hiera-B+ (Full FT, MSFPN) & CE+Dice    & 0.2885 & 0.2405 & 0.4089 & 0.6562 & 83.90 & 191.82 & 10.49 \\
& Hiera-B+ (Full FT, MSFPN) & Focal+Dice & 0.2980 & 0.2483 & 0.4155 & 0.6870 & 83.90 & 191.82 & 10.49 \\
& Hiera-B+ (Full FT, MSFPN) & WCE+Dice   & 0.2875 & 0.2054 & 0.4058 & 0.6769 & 83.90 & 191.82 & 10.49 \\

\midrule
\multirow{11}{*}{SESSRS}
& A2-FPN (CE+Dice)         & Postprocess & 0.3702 & 0.3085 & 0.4848 & 0.7338 & 12.16 & 27.14 & 3.20 \\
& A2-FPN (Focal+Dice)      & Postprocess & 0.3374 & 0.2812 & 0.4613 & 0.6663 & 12.16 & 27.14 & 3.32 \\
& A2-FPN (WCE+Dice)        & Postprocess & 0.3745 & 0.2675 & 0.5139 & 0.7098 & 12.16 & 27.14 & 3.24 \\
& ABCNet (CE+Dice)$^{*}$   & Postprocess & 0.3154 & 0.2629 & 0.4311 & 0.6835 & 13.96 & 32.39 & 20.95 \\
& BANet (CE+Dice)          & Postprocess & 0.2937 & 0.2259 & 0.4161 & 0.6536 & 12.86 & 31.38 & 7.71 \\
& MANet (CE+Dice)          & Postprocess & 0.3604 & 0.3004 & 0.4820 & 0.6775 & 35.86 & 109.62 & 7.76 \\
& MANet (Focal+Dice)       & Postprocess & \cellcolor{red!10}\textbf{0.4032} & 0.2880 & \cellcolor{red!10}\textbf{0.5467} & 0.6849 & 35.86 & 109.62 & 7.07 \\
& MANet (WCE+Dice)         & Postprocess & 0.3839 & 0.2742 & 0.5238 & 0.6763 & 35.86 & 109.62 & 8.62 \\
& UNetFormer (CE+Dice)     & Postprocess & 0.3958 & \cellcolor{red!10}\textbf{0.3298} & 0.5167 & 0.7279 & 11.73 & \cellcolor{red!10}\textbf{23.55} & 4.93 \\
& UNetFormer (Focal+Dice)  & Postprocess & 0.3873 & 0.3228 & 0.5091 & 0.7195 & 11.73 & 23.55 & 4.78 \\
& UNetFormer (WCE+Dice)    & Postprocess & 0.3578 & 0.2752 & 0.4943 & 0.6816 & 11.73 & 23.55 & 5.06 \\

\bottomrule
\end{tabular*}
\end{table*}

Tables~\ref{tab:app_seg_general}, \ref{tab:app_seg_rs}, and \ref{tab:app_seg_vfm} report the complete overall comparison for general segmentation models, remote-sensing-specific methods, and methods related to vision foundation models, respectively. Several consistent patterns can be observed across these three groups.

Table~\ref{tab:app_seg_general} shows that, among general segmentation models, SegFormer with WCE+Dice provides the strongest overall performance, achieving the best mIoU$_p$ (0.4010), mIoU (0.3084), and Macro-F1 (0.5297) within this group. This indicates that a well-trained generic transformer baseline already offers a strong reference point on this benchmark. At the same time, lightweight models such as Afformer, SeaFormer, and DDRNet remain attractive from an efficiency perspective, achieving the lowest parameter count, GFLOPs, or latency in the group, although their segmentation quality is clearly below that of the strongest baselines. In addition, BiSeNetv2 attains the highest OA (0.7330), showing that high OA does not necessarily coincide with the strongest foreground-oriented metrics due to the class imbalance.

Table~\ref{tab:app_seg_rs} shows that remote-sensing-specific methods are generally more competitive and better balanced than generic segmentation models. In particular, PyramidMamba with WCE+Dice achieves the best mIoU$_p$ (0.4414) and Macro-F1 (0.5864) in this group, indicating the strongest overall segmentation quality among the remote-sensing-specific methods evaluated here. UNetFormer with auxiliary supervision achieves the best mIoU (0.3284) while remaining relatively efficient, with the lowest GFLOPs (23.55) in this group. This suggests that architectures tailored to aerial or remote-sensing imagery continue to provide a meaningful advantage, especially under large appearance variation, small foreground structures, and severe class imbalance. In addition, some specialized methods show strengths on specific metrics; for example, LoGCAN achieves the highest OA (0.7474), A2-FPN has the lowest latency (2.12), and MF-Mamba uses the fewest parameters (11.27).

Table~\ref{tab:app_seg_vfm} presents the results of methods related to vision foundation models and shows a mixed overall picture. Direct adaptation of SAM-style models does not consistently outperform strong task-specific baselines, especially when the backbone is frozen or only lightly adapted. Stronger results in this group are instead obtained when foundation-model priors are combined with downstream segmenters or explicit refinement, as in SESSRS. Notably, no single method dominates all metrics. SESSRS with MANet and Focal+Dice achieves the best mIoU$_p$ (0.4032) and Macro-F1 (0.5467), while SESSRS with UNetFormer and CE+Dice yields the best mIoU (0.3298). In contrast, RSAM-Seg with a frozen SAM-ViT-B encoder and Focal+Dice attains the highest OA (0.7430). These results suggest that, at the current stage, vision foundation models are more effective when used as complementary components within downstream segmentation pipelines rather than as direct replacements for strong segmentation architectures.

In summary, Tables~\ref{tab:app_seg_general}--\ref{tab:app_seg_vfm} reveal a clear trade-off between segmentation accuracy and computational efficiency. They also show that OA alone can be misleading, since some methods remain competitive in OA while lagging behind in mIoU$_p$ or Macro-F1. For this reason, the main text emphasizes mIoU$_p$ and Macro-F1 when discussing model quality on foreground and rare mining-related categories.

\subsubsection{Per-class Segmentation Performance}

\begin{table*}[t]
\centering
\scriptsize
\caption{Per-class IoU results of general segmentation methods.}
\label{tab:app_seg_class_general}
\setlength{\tabcolsep}{3pt}
\renewcommand{\arraystretch}{1.1}
\resizebox{\textwidth}{!}{
\begin{tabular}{llccccccccccc}
\toprule
\toprule
\multirow{2}{*}{Model} &
\multirow{2}{*}{Backbone} &
\multirow{2}{*}{Loss} &
\multirow{2}{*}{Buildings} &
Mining & Primary & Water & Agricultural & Gravel & Type 1 & Type 2 & Bare & \multirow{2}{*}{Sluices} \\
& & & &
rafts & forests & bodies & crops & mounds & regeneration & regeneration & ground & \\
\midrule

\multirow{6}{*}{DeepLabV3+}
& ConvNeXt-Tiny & CE+Dice    & 0.3304 & 0.1906 & 0.7089 & 0.7395 & 0.4176 & 0.4109 & 0.2980 & 0.2636 & \cellcolor{blue!10}\textbf{0.5200} & 0.0158 \\
& ConvNeXt-Tiny & Focal+Dice & 0.4205 & 0.1563 & 0.6750 & 0.7615 & 0.3090 & 0.3516 & 0.2940 & 0.2595 & 0.5048 & 0.0025 \\
& ConvNeXt-Tiny & WCE+Dice   & 0.4039 & 0.1888 & 0.6503 & 0.7353 & 0.2867 & 0.3817 & 0.3098 & 0.2395 & 0.4885 & \cellcolor{blue!10}\textbf{0.0814} \\
& ResNet-50     & CE+Dice    & 0.4126 & \cellcolor{blue!10}\textbf{0.1955} & 0.7100 & 0.7005 & 0.3608 & 0.4825 & 0.2906 & 0.2546 & 0.4506 & 0.0000 \\
& ResNet-50     & Focal+Dice & 0.3971 & 0.1066 & 0.6884 & 0.7172 & 0.2859 & 0.2962 & 0.2972 & 0.2549 & 0.4715 & 0.0000 \\
& ResNet-50     & WCE+Dice   & 0.3376 & 0.1888 & 0.6927 & 0.6909 & 0.1500 & 0.3743 & 0.3336 & 0.2400 & 0.4230 & 0.0000 \\

\midrule
UPerNet
& Swin-Tiny & CE+Dice & 0.3163 & 0.0689 & 0.6443 & 0.7172 & 0.3436 & 0.3114 & 0.3008 & 0.2055 & 0.4629 & 0.0000 \\

\midrule
OCRNet
& HRNet-W48 & CE+Dice & 0.2982 & 0.0000 & 0.5517 & 0.4619 & 0.1134 & 0.3948 & 0.2974 & 0.2469 & 0.3575 & 0.0000 \\

\midrule
\multirow{3}{*}{BiSeNetv2}
& Custom Bilateral & CE+Dice    & 0.1960 & 0.0000 & 0.7412 & 0.7308 & 0.3244 & 0.4212 & 0.3418 & \cellcolor{blue!10}\textbf{0.2735} & 0.4575 & 0.0000 \\
& Custom Bilateral & WCE+Dice   & 0.2838 & 0.0000 & 0.6529 & 0.7809 & 0.3576 & 0.3988 & 0.3172 & 0.2293 & 0.4540 & 0.0000 \\
& Custom Bilateral & Focal+Dice & 0.3798 & 0.0000 & 0.7020 & 0.6074 & 0.3807 & 0.4860 & \cellcolor{blue!10}\textbf{0.3513} & 0.2515 & 0.4169 & 0.0000 \\

\midrule
\multirow{3}{*}{SegFormer}
& MiT-B2 & CE+Dice    & 0.3190 & 0.1321 & 0.6619 & 0.5524 & 0.2314 & 0.3474 & 0.3223 & 0.2548 & 0.3886 & 0.0119 \\
& MiT-B2 & Focal+Dice & 0.3588 & 0.1843 & 0.6842 & 0.4357 & 0.3276 & 0.3999 & 0.3138 & 0.2634 & 0.3570 & 0.0078 \\
& MiT-B2 & WCE+Dice   & \cellcolor{blue!10}\textbf{0.4500} & 0.1358 & 0.7137 & 0.7636 & 0.3985 & \cellcolor{blue!10}\textbf{0.5275} & 0.3404 & 0.1760 & 0.5043 & 0.0000 \\

\midrule
\multirow{3}{*}{STDC2}
& STDCNet & CE+Dice    & 0.3258 & 0.0000 & 0.7213 & 0.7558 & 0.3652 & 0.2950 & 0.2713 & 0.2133 & 0.4902 & 0.0000 \\
& STDCNet & Focal+Dice & 0.3029 & 0.0000 & 0.7156 & 0.6860 & 0.2838 & 0.3458 & 0.3073 & 0.2115 & 0.4331 & 0.0000 \\
& STDCNet & WCE+Dice   & 0.1923 & 0.0961 & 0.6456 & 0.7188 & 0.3701 & 0.3238 & 0.3154 & 0.2565 & 0.4522 & 0.0000 \\

\midrule
Mask2Former
& ResNet-50 & Set CE+Mask+Dice & 0.2218 & 0.1155 & 0.7008 & 0.5283 & 0.2786 & 0.2635 & 0.2846 & 0.2232 & 0.3682 & 0.0000 \\

\midrule
SegNeXt
& MSCAN-Tiny & CE+Dice & 0.3313 & 0.0385 & 0.6085 & 0.4889 & 0.0846 & 0.2239 & 0.3168 & 0.2280 & 0.3804 & 0.0000 \\

\midrule
DDRNet
& DDRNet-23-slim & Focal+Dice & 0.1707 & 0.0000 & 0.6028 & 0.5551 & 0.2109 & 0.4222 & 0.2866 & 0.2637 & 0.3572 & 0.0000 \\

\midrule
Afformer
& AFFormer-Base & CE+Dice & 0.3622 & 0.0633 & 0.6749 & 0.4837 & 0.2671 & 0.2531 & 0.3398 & 0.2436 & 0.3587 & 0.0004 \\

\midrule
\multirow{3}{*}{EfficientViT}
& EfficientViT-B2 & CE+Dice    & 0.3306 & 0.0949 & 0.7048 & \cellcolor{blue!10}\textbf{0.7956} & 0.3462 & 0.4750 & 0.3115 & 0.2275 & 0.5126 & 0.0000 \\
& EfficientViT-B2 & Focal+Dice & 0.3134 & 0.1549 & 0.6499 & 0.6826 & 0.3583 & 0.4868 & 0.3036 & 0.2498 & 0.4628 & 0.0311 \\
& EfficientViT-B2 & WCE+Dice   & 0.3526 & 0.1570 & 0.7041 & 0.7616 & 0.3469 & 0.3744 & 0.2903 & 0.2405 & 0.4831 & 0.0799 \\

\midrule
SeaFormer
& SeaFormer-Base & CE+Dice & 0.3221 & 0.0141 & 0.6851 & 0.4595 & 0.4053 & 0.3342 & 0.3298 & 0.2221 & 0.3451 & 0.0000 \\

\midrule
\multirow{3}{*}{PIDNet}
& PIDNet-M & CE+Dice    & 0.3190 & 0.0000 & 0.6561 & 0.7084 & \cellcolor{blue!10}\textbf{0.4202} & 0.2948 & 0.3391 & 0.2373 & 0.4623 & 0.0000 \\
& PIDNet-M & Focal+Dice & 0.3088 & 0.0000 & \cellcolor{blue!10}\textbf{0.7545} & 0.6233 & 0.3319 & 0.4304 & 0.3315 & 0.2533 & 0.3973 & 0.0000 \\
& PIDNet-M & WCE+Dice   & 0.3057 & 0.1590 & 0.6573 & 0.7137 & 0.3158 & 0.2556 & 0.2724 & 0.2553 & 0.4667 & 0.0000 \\

\midrule
CGRSeg
& EfficientFormerV2-B & CE+Dice & 0.3415 & 0.0863 & 0.6122 & 0.4065 & 0.1962 & 0.1772 & 0.3054 & 0.2107 & 0.3427 & 0.0000 \\

\midrule
PEM
& ResNet-50 & Set CE+Mask+Dice & 0.1083 & 0.0880 & 0.7338 & 0.4404 & 0.3529 & 0.2254 & 0.2835 & 0.2227 & 0.3336 & 0.0000 \\

\midrule
VMamba
& VMamba-Tiny & CE+Dice & 0.2919 & 0.0000 & 0.6920 & 0.4388 & 0.1667 & 0.4053 & 0.2869 & 0.2086 & 0.3390 & 0.0000 \\

\bottomrule
\end{tabular}}
\end{table*}

\begin{table*}[t]
\centering
\scriptsize
\caption{Per-class IoU results of remote-sensing-specific methods.}
\label{tab:app_seg_class_rs}
\setlength{\tabcolsep}{3pt}
\renewcommand{\arraystretch}{1.1}
\resizebox{\textwidth}{!}{
\begin{tabular}{llccccccccccc}
\toprule
\multirow{2}{*}{Model} &
\multirow{2}{*}{Backbone} &
\multirow{2}{*}{Loss} &
\multirow{2}{*}{Buildings} &
Mining & Primary & Water & Agricultural & Gravel & Type 1 & Type 2 & Bare & \multirow{2}{*}{Sluices} \\
& & & &
rafts & forests & bodies & crops & mounds & regeneration & regeneration & ground & \\
\midrule

\multirow{4}{*}{FarSeg}
& ResNet-50 & CE  & 0.2996 & 0.0000 & 0.6518 & 0.8273 & 0.2050 & 0.4567 & 0.3081 & 0.2718 & 0.5442 & 0.0000 \\
& ResNet-50 & CE+Dice    & 0.3666 & 0.0000 & 0.6105 & 0.8184 & 0.3664 & 0.4747 & 0.3118 & 0.2522 & 0.5167 & 0.0000 \\
& ResNet-50 & Focal+Dice & 0.3855 & 0.0000 & 0.6992 & \cellcolor{green!10}\textbf{0.8519} & 0.2370 & 0.4306 & 0.3358 & 0.2478 & \cellcolor{green!10}\textbf{0.5560} & 0.0000 \\
& ResNet-50 & WCE+Dice   & 0.3047 & 0.1328 & 0.6141 & 0.7264 & 0.3078 & 0.4693 & 0.2636 & 0.2445 & 0.4610 & 0.1179 \\

\midrule
BANet
& ResT-Lite & CE+Dice & 0.2564 & 0.0000 & 0.6911 & 0.5580 & 0.2844 & 0.2729 & 0.2906 & 0.2283 & 0.3438 & 0.0000 \\

\midrule
ABCNet
& ResNet-18 & CE+Dice$^{*}$ & 0.2693 & 0.0000 & 0.6888 & 0.7210 & 0.1825 & 0.3820 & 0.2787 & 0.1956 & 0.4275 & 0.0000 \\

\midrule
\multirow{3}{*}{MANet}
& ResNet-50 & CE+Dice    & 0.5667 & 0.0000 & 0.6558 & 0.7265 & 0.3444 & 0.4506 & 0.2938 & 0.2029 & 0.4700 & 0.0000 \\
& ResNet-50 & Focal+Dice & 0.5479 & 0.1745 & 0.6646 & 0.6925 & 0.4170 & 0.4341 & 0.3045 & 0.2271 & 0.4509 & 0.0848 \\
& ResNet-50 & WCE+Dice   & 0.4332 & 0.2108 & 0.6441 & 0.7291 & 0.3572 & 0.4760 & 0.2907 & 0.1884 & 0.4533 & 0.0440 \\

\midrule
\multirow{3}{*}{UNetFormer}
& ResNet-18 & CE+Dice$^{*}$    & 0.4657 & 0.0000 & 0.7259 & 0.7274 & 0.4635 & 0.5224 & \cellcolor{green!10}\textbf{0.3449} & 0.2181 & 0.4720 & 0.0000 \\
& ResNet-18 & Focal+Dice$^{*}$ & 0.4652 & 0.0000 & 0.7098 & 0.7641 & 0.3846 & 0.4806 & 0.3143 & 0.2627 & 0.4828 & 0.0000 \\
& ResNet-18 & WCE+Dice$^{*}$   & 0.3318 & 0.1946 & 0.6817 & 0.6710 & 0.2861 & 0.4238 & 0.3321 & 0.2116 & 0.4327 & 0.0000 \\

\midrule
DC-Swin
& Swin-Small & CE+Dice & 0.2815 & 0.0000 & 0.6661 & 0.6159 & 0.1668 & 0.3052 & 0.3172 & 0.2400 & 0.3787 & 0.0000 \\

\midrule
\multirow{3}{*}{A2-FPN}
& ResNet-18 & CE+Dice    & 0.3454 & 0.0000 & 0.7085 & 0.8334 & 0.2368 & 0.4817 & 0.2882 & 0.2428 & 0.5512 & 0.0000 \\
& ResNet-18 & Focal+Dice & 0.3992 & 0.0000 & 0.6585 & 0.6340 & 0.2202 & 0.4728 & 0.3150 & 0.2345 & 0.4285 & 0.0000 \\
& ResNet-18 & WCE+Dice   & 0.3624 & 0.1482 & 0.7198 & 0.7322 & 0.3032 & 0.4070 & 0.3043 & 0.2122 & 0.4455 & 0.0856 \\

\midrule
LoGCAN
& ResNet-50 & CE$^{*}$ & 0.2460 & 0.1024 & 0.7280 & 0.7215 & 0.2772 & 0.0170 & 0.2716 & 0.2706 & 0.4524 & 0.0000 \\

\midrule
FarSeg++
& MiT-B2 & CE & 0.3413 & 0.1459 & 0.7048 & 0.5551 & 0.1487 & 0.2526 & 0.2895 & 0.2436 & 0.3800 & 0.0000 \\

\midrule
SACANet
& HRNet-W32 & CE$^{*}$ & 0.3359 & 0.0018 & 0.6371 & 0.6208 & 0.4380 & 0.2665 & 0.3319 & 0.2333 & 0.4282 & 0.0000 \\

\midrule
DOCNet
& HRNet-W32 & CE$^{*}$ & 0.2662 & 0.0731 & 0.6593 & 0.6889 & 0.2644 & 0.2455 & 0.2725 & 0.2254 & 0.4512 & 0.0000 \\

\midrule
\multirow{3}{*}{PPMambaSeg}
& SWSL & CE+Dice    & 0.4223 & 0.0000 & 0.6743 & 0.6150 & 0.3353 & 0.5308 & 0.2833 & 0.2537 & 0.4057 & 0.0000 \\
& SWSL & Focal+Dice & 0.4594 & 0.0000 & 0.7028 & 0.7118 & \cellcolor{green!10}\textbf{0.4937} & \cellcolor{green!10}\textbf{0.5388} & 0.3357 & 0.1800 & 0.4750 & 0.0000 \\
& SWSL & WCE+Dice   & 0.4859 & 0.1289 & 0.6944 & 0.6466 & 0.3336 & 0.4880 & 0.2873 & 0.2228 & 0.4248 & \cellcolor{green!10}\textbf{0.1412} \\

\midrule
\multirow{3}{*}{RS3Mamba}
& VMamba-Tiny & CE+Dice    & 0.0000 & 0.0000 & \cellcolor{green!10}\textbf{0.7796} & 0.6756 & 0.0000 & 0.0000 & 0.2615 & 0.2298 & 0.4381 & 0.0000 \\
& VMamba-Tiny & Focal+Dice & 0.0000 & 0.0000 & 0.7723 & 0.6926 & 0.0147 & 0.0416 & 0.2654 & 0.1766 & 0.4352 & 0.0000 \\
& VMamba-Tiny & WCE+Dice   & 0.1480 & 0.0000 & 0.6587 & 0.5998 & 0.4101 & 0.3283 & 0.2975 & 0.2396 & 0.3856 & 0.0000 \\

\midrule
\multirow{3}{*}{PyramidMamba}
& Swin-Base & CE+Dice    & 0.6295 & 0.2154 & 0.6818 & 0.6364 & 0.3555 & 0.4653 & 0.3036 & 0.2522 & 0.4457 & 0.0000 \\
& Swin-Base & Focal+Dice & \cellcolor{green!10}\textbf{0.7059} & 0.1549 & 0.6602 & 0.6184 & 0.3811 & 0.4395 & 0.3284 & 0.2406 & 0.4324 & 0.0000 \\
& Swin-Base & WCE+Dice   & 0.6875 & \cellcolor{green!10}\textbf{0.2659} & 0.6891 & 0.6642 & 0.4437 & 0.5345 & 0.3030 & \cellcolor{green!10}\textbf{0.2743} & 0.4459 & 0.1060 \\

\midrule
LoGCAN++
& RepViT-M2.3 & CE$^{*}$ & 0.1737 & 0.0165 & 0.6562 & 0.5126 & 0.1640 & 0.1577 & 0.2803 & 0.2214 & 0.3635 & 0.0000 \\

\midrule
MF-Mamba
& HRNet-W18 & CE+Dice & 0.2849 & 0.0000 & 0.6570 & 0.5295 & 0.2176 & 0.4245 & 0.2855 & 0.2247 & 0.3776 & 0.0000 \\

\midrule
\multirow{3}{*}{MCPNet}
& ResNet-50 & CE+Dice    & 0.1656 & 0.0000 & 0.6890 & 0.5591 & 0.2642 & 0.4480 & 0.2901 & 0.2606 & 0.3795 & 0.0000 \\
& ResNet-50 & Focal+Dice & 0.1958 & 0.0000 & 0.7028 & 0.6344 & 0.4144 & 0.3116 & 0.3241 & 0.2410 & 0.4088 & 0.0000 \\
& ResNet-50 & WCE+Dice   & 0.1766 & 0.1451 & 0.6789 & 0.4895 & 0.3775 & 0.3701 & 0.3182 & 0.2658 & 0.3710 & 0.0000 \\

\bottomrule
\end{tabular}}
\end{table*}

\begin{table*}[t]
\centering
\scriptsize
\caption{Per-class IoU results of methods related to vision foundation models.}
\label{tab:app_seg_class_vfm}
\setlength{\tabcolsep}{3pt}
\renewcommand{\arraystretch}{1.1}
\resizebox{\textwidth}{!}{
\begin{tabular}{llccccccccccc}
\toprule
\multirow{2}{*}{Model} &
\multirow{2}{*}{Backbone} &
\multirow{2}{*}{Loss} &
\multirow{2}{*}{Buildings} &
Mining & Primary & Water & Agricultural & Gravel & Type 1 & Type 2 & Bare & \multirow{2}{*}{Sluices} \\
& & & &
rafts & forests & bodies & crops & mounds & regeneration & regeneration & ground & \\
\midrule

\multirow{3}{*}{HQ-SAM}
& ViT-B + HQ Decoder & CE+Dice    & 0.2233 & 0.0211 & 0.6664 & 0.5210 & 0.0429 & 0.1300 & 0.2979 & 0.2299 & 0.3530 & 0.0000 \\
& ViT-B + HQ Decoder & Focal+Dice & 0.2408 & 0.0521 & 0.7055 & 0.5072 & 0.0366 & 0.0794 & 0.2916 & 0.2327 & 0.3569 & 0.0000 \\
& ViT-B + HQ Decoder & WCE+Dice   & 0.1961 & 0.0877 & 0.6578 & 0.4813 & 0.0741 & 0.1949 & 0.3270 & 0.2207 & 0.2989 & 0.0000 \\

\midrule
\multirow{4}{*}{SAM\_{RS}}
& ABCNet + SAM Priors       & Seg+Bdy+Obj & 0.0894 & 0.0000 & 0.6257 & 0.6902 & 0.2007 & 0.3997 & 0.3006 & 0.2397 & 0.4174 & 0.0000 \\
& CMTFNet + SAM Priors      & Seg+Bdy+Obj & 0.2048 & 0.0000 & 0.6620 & 0.6011 & 0.1270 & 0.3726 & 0.2936 & 0.2437 & 0.4108 & 0.0000 \\
& FTUNetFormer + SAM Priors & Seg+Bdy+Obj & 0.2637 & 0.0000 & 0.7463 & 0.5706 & 0.1557 & 0.3117 & 0.2710 & 0.2614 & 0.3418 & 0.0000 \\
& UNetFormer + SAM Priors   & Seg+Bdy+Obj & 0.2783 & 0.0000 & 0.6849 & 0.6421 & 0.1958 & 0.4495 & 0.2846 & \cellcolor{red!10}\textbf{0.2670} & 0.4384 & 0.0000 \\

\midrule
\multirow{6}{*}{SAM2.1}
& Hiera-B+ (Frozen, MSFPN)  & CE+Dice    & 0.2552 & 0.0441 & 0.6777 & 0.4286 & 0.0149 & 0.2028 & 0.3060 & 0.1635 & 0.3288 & 0.0000 \\
& Hiera-B+ (Frozen, MSFPN)  & Focal+Dice & 0.2074 & 0.0479 & 0.6960 & 0.3906 & 0.0482 & 0.1869 & 0.2984 & 0.1556 & 0.3204 & 0.0000 \\
& Hiera-B+ (Frozen, MSFPN)  & WCE+Dice   & 0.0778 & 0.0403 & 0.6850 & 0.3653 & 0.0532 & 0.2506 & 0.2871 & 0.1351 & 0.3102 & 0.0027 \\
& Hiera-B+ (Full FT, MSFPN) & CE+Dice    & 0.2944 & 0.0000 & 0.6846 & 0.5456 & 0.1633 & 0.2935 & 0.2949 & 0.2355 & 0.3737 & 0.0000 \\
& Hiera-B+ (Full FT, MSFPN) & Focal+Dice & 0.2670 & 0.0000 & 0.7298 & 0.6004 & 0.1825 & 0.3277 & 0.3039 & 0.1975 & 0.3712 & 0.0000 \\
& Hiera-B+ (Full FT, MSFPN) & WCE+Dice   & 0.1230 & 0.0774 & 0.7163 & 0.6069 & 0.1829 & 0.2916 & 0.2798 & 0.2168 & 0.3805 & 0.0000 \\

\midrule
\multirow{3}{*}{RSAM-Seg}
& SAM-ViT-B (Frozen Encoder) & CE+Dice    & 0.3254 & 0.0000 & 0.7030 & 0.6784 & 0.2550 & 0.3121 & 0.3022 & 0.2285 & 0.4579 & 0.0000 \\
& SAM-ViT-B (Frozen Encoder) & Focal+Dice & 0.3239 & 0.0068 & \cellcolor{red!10}\textbf{0.7604} & 0.7255 & 0.2763 & 0.3453 & 0.3057 & 0.2320 & 0.4741 & 0.0000 \\
& SAM-ViT-B (Frozen Encoder) & WCE+Dice   & 0.3694 & 0.1755 & 0.7118 & 0.6970 & 0.3144 & 0.4291 & 0.3081 & 0.1970 & 0.4374 & 0.0563 \\

\midrule
\multirow{11}{*}{SESSRS}
& A2-FPN (CE+Dice)        & Postprocess & 0.3508 & 0.0000 & 0.7085 & \cellcolor{red!10}\textbf{0.8338} & 0.2368 & 0.4890 & 0.2883 & 0.2430 & \cellcolor{red!10}\textbf{0.5519} & 0.0000 \\
& A2-FPN (Focal+Dice)     & Postprocess & 0.4018 & 0.0000 & 0.6584 & 0.6353 & 0.2203 & 0.4785 & 0.3156 & 0.2347 & 0.4292 & 0.0000 \\
& A2-FPN (WCE+Dice)       & Postprocess & 0.3669 & 0.1513 & 0.7198 & 0.7334 & 0.3035 & 0.4118 & 0.3048 & 0.2127 & 0.4463 & \cellcolor{red!10}\textbf{0.0943} \\
& ABCNet (CE+Dice)$^{*}$  & Postprocess & 0.2720 & 0.0000 & 0.6887 & 0.7224 & 0.1825 & 0.3858 & 0.2794 & 0.1958 & 0.4277 & 0.0000 \\
& BANet (CE+Dice)         & Postprocess & 0.2585 & 0.0000 & 0.6911 & 0.5584 & 0.2844 & 0.2812 & 0.2907 & 0.2286 & 0.3438 & 0.0000 \\
& MANet (CE+Dice)         & Postprocess & 0.5642 & 0.0000 & 0.6667 & 0.6537 & 0.2792 & 0.4624 & 0.3063 & 0.1918 & 0.4800 & 0.0000 \\
& MANet (Focal+Dice)      & Postprocess & \cellcolor{red!10}\textbf{0.5654} & 0.1813 & 0.6645 & 0.6930 & 0.4172 & 0.4393 & 0.3048 & 0.2272 & 0.4512 & 0.0883 \\
& MANet (WCE+Dice)        & Postprocess & 0.4388 & \cellcolor{red!10}\textbf{0.2110} & 0.6440 & 0.7298 & 0.3573 & 0.4806 & 0.2910 & 0.1887 & 0.4537 & 0.0440 \\
& UNetFormer (CE+Dice)    & Postprocess & 0.4708 & 0.0000 & 0.7259 & 0.7281 & \cellcolor{red!10}\textbf{0.4633} & \cellcolor{red!10}\textbf{0.5327} & \cellcolor{red!10}\textbf{0.3453} & 0.2184 & 0.4735 & 0.0000 \\
& UNetFormer (Focal+Dice) & Postprocess & 0.4662 & 0.0000 & 0.7098 & 0.7649 & 0.3850 & 0.4858 & 0.3149 & 0.2631 & 0.4838 & 0.0000 \\
& UNetFormer (WCE+Dice)   & Postprocess & 0.3334 & 0.1955 & 0.6817 & 0.6728 & 0.2858 & 0.4299 & 0.3330 & 0.2123 & 0.4335 & 0.0000 \\

\bottomrule
\end{tabular}}
\end{table*}

Tables~\ref{tab:app_seg_class_general}, \ref{tab:app_seg_class_rs}, and \ref{tab:app_seg_class_vfm} report class-wise IoU results for general segmentation models, remote-sensing-specific methods, and methods related to vision foundation models, respectively. Compared with the aggregate metrics in the main text and the overall tables in the appendix, these class-wise results provide a more detailed view of where different methods succeed or fail. In particular, they help distinguish improvements on dominant land-cover categories from improvements on rare and operationally important mining-related classes.

A consistent pattern across all three tables is that performance varies substantially by class. Large and visually consistent categories, especially primary forests and water bodies, are easier for most methods. In contrast, rare and spatially sparse categories such as mining rafts and sluices are much harder. This difference is large in every method group. It also explains why methods with similar overall scores can still behave very differently on the small mining-related classes.

Table~\ref{tab:app_seg_class_general} shows that the best results in the general-model group are spread across several architectures. SegFormer with WCE+Dice gives the best score on buildings (0.4500) and gravel mounds (0.5275). PIDNet with Focal+Dice gives the best result on primary forests (0.7545). EfficientViT with CE+Dice performs best on water bodies (0.7956). PIDNet with CE+Dice is strongest on agricultural crops (0.4202). BiSeNetv2 gives the best results on type 1 regeneration (0.3513) and type 2 regeneration (0.2735). DeepLabV3+ performs best on bare ground (0.5200) and sluices (0.0814). These results show that strong general segmentation models remain competitive, but their strengths are not consistent across all classes.

Table~\ref{tab:app_seg_class_rs} shows that remote-sensing-specific methods are stronger on several difficult classes, although no single model is best everywhere. PyramidMamba gives the best results on buildings (0.7059), mining rafts (0.2659), and type 2 regeneration (0.2743). RS3Mamba achieves the highest score on primary forests (0.7796). FarSeg with Focal+Dice performs best on water bodies (0.8519) and bare ground (0.5560). PPMambaSeg with Focal+Dice gives the best results on agricultural crops (0.4937) and gravel mounds (0.5388). UNetFormer with auxiliary supervision achieves the best score on type 1 regeneration (0.3449), while PPMambaSeg with WCE+Dice gives the strongest sluices result (0.1412). Overall, this table shows that remote-sensing-specific methods fit the dataset better, especially on the small and imbalanced disturbance-related classes.

Table~\ref{tab:app_seg_class_vfm} shows a mixed picture for methods related to vision foundation models. Direct SAM-style adaptation does not consistently improve the more difficult classes, and the strongest results in this group mostly come from integration or refinement strategies rather than from direct replacement of the segmentation backbone. No single method dominates all categories. RSAM-Seg with Focal+Dice gives the best primary forests score in this group (0.7604), and SAM\_RS with UNetFormer gives the best type 2 regeneration result (0.2670). Most of the strongest class-wise results, however, come from SESSRS. SESSRS with MANet and Focal+Dice gives the best buildings score (0.5654), while SESSRS with MANet and WCE+Dice performs best on mining rafts (0.2110). SESSRS with UNetFormer and CE+Dice gives the best results on agricultural crops (0.4633) and gravel mounds (0.5327), and also achieves the best type 1 regeneration score (0.3453). SESSRS with A2-FPN and CE+Dice gives the best water bodies (0.8338) and bare ground (0.5519) scores, while SESSRS with A2-FPN and WCE+Dice gives the best sluices result (0.0943). Overall, these results again suggest that foundation-model priors are currently more useful when they are combined with a strong downstream segmenter and an explicit refinement stage, rather than being used alone.

Looking across Tables~\ref{tab:app_seg_class_general}--\ref{tab:app_seg_class_vfm}, three points stand out. First, this benchmark is still difficult, especially for rare mining-related categories. Second, no method family is best on every class. Third, remote-sensing-specific methods and refinement-based VFM pipelines are more reliable on the hard foreground categories than direct general-purpose baselines. For this reason, the class-wise tables are important for understanding model behavior beyond the aggregate scores.

\subsection{Segmentation-derived Recognition}

\label{app:seg_reg}

This appendix also reports the full results for segmentation-derived recognition. We organize the comparison into the same three method families used in semantic segmentation: general segmentation models, remote-sensing-specific methods, and methods related to vision foundation models. Since recognition is obtained from segmentation outputs, these results complement the dense prediction analysis by showing how well each method recovers image-level semantic presence. The overall comparison highlights differences in label-level precision, recall, and balance, while the per-class comparison shows how different methods behave on dominant categories versus rare mining-related classes.

\subsubsection{Overall Segmentation-derived Recognition Performance}

Tables~\ref{tab:app_general_multilabel}, \ref{tab:app_rs_multilabel}, and \ref{tab:app_vfm_multilabel} summarize the overall recognition results for the three method groups. Although these results are derived from the same segmentation outputs analyzed earlier, the ranking is not identical, which indicates that accurate spatial delineation and reliable image-level presence prediction are related but not interchangeable.

\begin{table*}[t]
\centering
\scriptsize
\caption{Results of general segmentation models on segmentation-derived multi-label classification.}
\label{tab:app_general_multilabel}
\setlength{\tabcolsep}{2.2pt}
\renewcommand{\arraystretch}{1.05}
\resizebox{\textwidth}{!}{%
\begin{tabular}{llcccccccccc}
\toprule
Model & Backbone & Loss
& CP & CR & CF1 & OP & OR & OF1 & mAP & Macro-F1 & Sample-F1 \\
\midrule

\multirow{6}{*}{DeepLabV3+}
& ConvNeXt-Tiny & CE+Dice & 0.5586 & 0.6951 & 0.6194 & 0.6850 & 0.8133 & 0.7437 & 0.6230 & 0.6135 & 0.7574 \\
& ConvNeXt-Tiny & Focal+Dice & 0.5536 & 0.6728 & 0.6074 & 0.6665 & 0.8223 & 0.7362 & 0.6179 & 0.5954 & 0.7458 \\
& ConvNeXt-Tiny & WCE+Dice & 0.5500 & 0.6924 & 0.6131 & 0.6732 & 0.8044 & 0.7330 & 0.6107 & 0.6015 & 0.7395 \\
& ResNet-50 & CE+Dice & 0.4948 & 0.6274 & 0.5533 & 0.6781 & 0.8083 & 0.7375 & 0.5776 & \cellcolor{blue!10}\textbf{0.6833} & 0.7561 \\
& ResNet-50 & Focal+Dice & 0.4881 & \cellcolor{blue!10}\textbf{0.7140} & 0.5798 & 0.6325 & 0.8604 & 0.7291 & 0.6120 & 0.6257 & 0.7424 \\
& ResNet-50 & WCE+Dice & 0.5404 & 0.6461 & 0.5886 & 0.6803 & 0.7885 & 0.7304 & 0.5664 & 0.6417 & 0.7447 \\

\midrule
UPerNet
& Swin-Tiny & CE+Dice & 0.5441 & 0.5653 & 0.5545 & 0.7040 & 0.7645 & 0.7330 & 0.5252 & 0.6095 & 0.7422 \\

\midrule
OCRNet
& HRNet-W48 & CE+Dice & 0.5009 & 0.5395 & 0.5195 & 0.6579 & 0.7061 & 0.6811 & 0.4867 & 0.6200 & 0.6589 \\

\midrule
\multirow{3}{*}{BiSeNetv2}
& Custom Bilateral & CE+Dice & 0.3733 & 0.6991 & 0.4867 & 0.5099 & \cellcolor{blue!10}\textbf{0.8911} & 0.6486 & 0.5377 & 0.5546 & 0.6718 \\
& Custom Bilateral & Focal+Dice & 0.4648 & 0.6253 & 0.5332 & 0.6632 & 0.8427 & 0.7423 & 0.5263 & 0.6306 & 0.7478 \\
& Custom Bilateral & WCE+Dice & 0.3646 & 0.6953 & 0.4783 & 0.5039 & 0.8836 & 0.6418 & 0.5369 & 0.5484 & 0.6739 \\

\midrule
\multirow{3}{*}{SegFormer}
& MiT-B2 & CE+Dice & 0.5309 & 0.6277 & 0.5753 & 0.6929 & 0.7207 & 0.7065 & 0.5382 & 0.5597 & 0.7045 \\
& MiT-B2 & Focal+Dice & 0.5288 & 0.6958 & 0.6009 & 0.6582 & 0.7594 & 0.7052 & 0.5857 & 0.5735 & 0.6986 \\
& MiT-B2 & WCE+Dice & 0.5665 & 0.6796 & 0.6179 & 0.7340 & 0.7750 & \cellcolor{blue!10}\textbf{0.7539} & \cellcolor{blue!10}\textbf{0.6576} & 0.6747 & \cellcolor{blue!10}\textbf{0.7790} \\

\midrule
\multirow{3}{*}{STDC2}
& STDCNet & CE+Dice & 0.5032 & 0.5483 & 0.5248 & 0.7110 & 0.7875 & 0.7473 & 0.5016 & 0.6086 & 0.7705 \\
& STDCNet & Focal+Dice & 0.4859 & 0.5673 & 0.5235 & 0.7121 & 0.7753 & 0.7424 & 0.5010 & 0.6241 & 0.7586 \\
& STDCNet & WCE+Dice & 0.6086 & 0.6020 & 0.6052 & 0.7034 & 0.7688 & 0.7347 & 0.5756 & 0.6550 & 0.7464 \\

\midrule
Mask2Former
& ResNet-50 & Set CE+Mask+Dice & \cellcolor{blue!10}\textbf{0.6584} & 0.5126 & 0.5764 & 0.7630 & 0.6625 & 0.7092 & 0.5297 & 0.6317 & 0.7125 \\

\midrule
SegNeXt
& MSCAN-Tiny & CE+Dice & 0.5306 & 0.5334 & 0.5320 & 0.6659 & 0.7484 & 0.7048 & 0.4756 & 0.5636 & 0.6888 \\

\midrule
DDRNet
& DDRNet-23-slim & Focal+Dice & 0.4979 & 0.5512 & 0.5232 & 0.6634 & 0.7621 & 0.7093 & 0.5081 & 0.6349 & 0.7061 \\

\midrule
Afformer
& AFFormer-Base & CE+Dice & 0.5220 & 0.6169 & 0.5655 & 0.6572 & 0.7639 & 0.7065 & 0.5297 & 0.5497 & 0.6983 \\

\midrule
\multirow{3}{*}{EfficientViT}
& EfficientViT-B2 & CE+Dice & 0.5583 & 0.6376 & 0.5953 & 0.7063 & 0.7937 & 0.7474 & 0.6002 & 0.6240 & 0.7711 \\
& EfficientViT-B2 & Focal+Dice & 0.5822 & 0.6753 & 0.6253 & 0.6840 & 0.7881 & 0.7324 & 0.6100 & 0.6056 & 0.7375 \\
& EfficientViT-B2 & WCE+Dice & 0.5896 & 0.6812 & \cellcolor{blue!10}\textbf{0.6321} & 0.7031 & 0.8023 & 0.7494 & 0.6228 & 0.6113 & 0.7683 \\

\midrule
SeaFormer
& SeaFormer-Base & CE+Dice & 0.5405 & 0.5770 & 0.5582 & 0.6775 & 0.7554 & 0.7143 & 0.5171 & 0.5941 & 0.7064 \\

\midrule
\multirow{3}{*}{PIDNet}
& PIDNet-M & CE+Dice & 0.5287 & 0.5554 & 0.5417 & 0.7152 & 0.7664 & 0.7399 & 0.5254 & 0.6547 & 0.7440 \\
& PIDNet-M & Focal+Dice & 0.3792 & 0.5382 & 0.4449 & 0.5830 & 0.8004 & 0.6746 & 0.4298 & 0.5229 & 0.6882 \\
& PIDNet-M & WCE+Dice & 0.5655 & 0.5914 & 0.5782 & 0.6922 & 0.7603 & 0.7247 & 0.5254 & 0.5977 & 0.7335 \\

\midrule
CGRSeg
& EfficientFormerV2-B & CE+Dice & 0.5860 & 0.5601 & 0.5728 & 0.6472 & 0.7446 & 0.6925 & 0.4990 & 0.5247 & 0.6736 \\

\midrule
PEM
& ResNet-50 & Set CE+Mask+Dice & 0.6373 & 0.4949 & 0.5571 & \cellcolor{blue!10}\textbf{0.7678} & 0.6664 & 0.7135 & 0.5069 & 0.6062 & 0.7110 \\

\midrule
VMamba
& VMamba-Tiny & CE+Dice & 0.5231 & 0.4907 & 0.5064 & 0.7046 & 0.7232 & 0.7138 & 0.4648 & 0.6237 & 0.6990 \\

\bottomrule
\end{tabular}%
}
\end{table*}

\begin{table*}[t]
\centering
\scriptsize
\caption{Results of remote-sensing-specific methods on segmentation-derived multi-label classification. An asterisk ($^{*}$) denotes the use of an auxiliary loss.}
\label{tab:app_rs_multilabel}
\setlength{\tabcolsep}{2.2pt}
\renewcommand{\arraystretch}{1.05}
\resizebox{\textwidth}{!}{%
\begin{tabular}{llcccccccccc}
\toprule
Model & Backbone & Loss
& CP & CR & CF1 & OP & OR & OF1 & mAP & Macro-F1 & Sample-F1 \\
\midrule

\multirow{4}{*}{FarSeg}
& ResNet-50 & CE & 0.4880 & 0.6234 & 0.5475 & 0.7013 & 0.8159 & 0.7543 & 0.5576 & 0.6695 & 0.7658 \\
& ResNet-50 & CE+Dice & 0.4617 & 0.6430 & 0.5375 & 0.6716 & 0.8039 & 0.7318 & 0.5752 & 0.6499 & 0.7420 \\
& ResNet-50 & Focal+Dice & 0.4884 & 0.6471 & 0.5567 & 0.6889 & 0.8537 & 0.7625 & 0.5773 & 0.6822 & \cellcolor{green!10}\textbf{0.7887} \\
& ResNet-50 & WCE+Dice & 0.4766 & 0.7369 & 0.5788 & 0.6265 & 0.7857 & 0.6971 & 0.6159 & 0.5455 & 0.7050 \\

\midrule
BANet
& ResT-Lite & CE+Dice & 0.4616 & 0.5574 & 0.5050 & 0.6664 & 0.7606 & 0.7104 & 0.4863 & 0.6238 & 0.7155 \\

\midrule
ABCNet
& ResNet-18 & CE+Dice$^{*}$ & 0.5194 & 0.5399 & 0.5294 & 0.7339 & 0.7366 & 0.7353 & 0.5150 & 0.6593 & 0.7460 \\

\midrule
\multirow{3}{*}{MANet}
& ResNet-50 & CE+Dice & 0.4338 & 0.6937 & 0.5338 & 0.6062 & 0.8818 & 0.7185 & 0.5556 & 0.6545 & 0.7376 \\
& ResNet-50 & Focal+Dice & 0.4344 & \cellcolor{green!10}\textbf{0.8610} & 0.5774 & 0.5777 & 0.8897 & 0.7005 & 0.6170 & 0.5444 & 0.7203 \\
& ResNet-50 & WCE+Dice & 0.4458 & 0.8443 & 0.5835 & 0.5860 & 0.8794 & 0.7034 & 0.6817 & 0.5535 & 0.7236 \\

\midrule
\multirow{3}{*}{UNetFormer}
& ResNet-18 & CE+Dice$^{*}$ & 0.5221 & 0.6390 & 0.5747 & 0.6912 & 0.8521 & \cellcolor{green!10}\textbf{0.7633} & 0.5913 & 0.7105 & 0.7839 \\
& ResNet-18 & Focal+Dice$^{*}$ & 0.4633 & 0.6671 & 0.5469 & 0.6344 & 0.8535 & 0.7278 & 0.5807 & 0.6704 & 0.7533 \\
& ResNet-18 & WCE+Dice$^{*}$ & 0.4444 & 0.7464 & 0.5571 & 0.6076 & 0.8561 & 0.7107 & 0.6119 & 0.5893 & 0.7314 \\

\midrule
DC-Swin
& Swin-Small & CE+Dice & 0.4535 & 0.5773 & 0.5080 & 0.6489 & 0.8161 & 0.7230 & 0.4953 & 0.6255 & 0.7345 \\

\midrule
\multirow{3}{*}{A2-FPN}
& ResNet-18 & CE+Dice & 0.5133 & 0.5665 & 0.5386 & \cellcolor{green!10}\textbf{0.7381} & 0.7526 & 0.7453 & 0.5592 & 0.6678 & 0.7725 \\
& ResNet-18 & Focal+Dice & 0.4574 & 0.6230 & 0.5275 & 0.6502 & 0.7846 & 0.7111 & 0.5414 & 0.6394 & 0.7161 \\
& ResNet-18 & WCE+Dice & 0.5660 & 0.6809 & 0.6182 & 0.7081 & 0.7544 & 0.7305 & 0.6176 & 0.6064 & 0.7554 \\

\midrule
LoGCAN
& ResNet-50 & CE$^{*}$ & 0.5961 & 0.5941 & 0.5951 & 0.6678 & 0.8413 & 0.7446 & 0.5499 & 0.6306 & 0.7664 \\

\midrule
FarSeg++
& MiT-B2 & CE & 0.5400 & 0.5397 & 0.5399 & 0.7078 & 0.7409 & 0.7240 & 0.4957 & 0.5932 & 0.7260 \\

\midrule
SACANet
& HRNet-W32 & CE$^{*}$ & \cellcolor{green!10}\textbf{0.6200} & 0.5916 & 0.6055 & 0.6800 & 0.7801 & 0.7266 & 0.5525 & 0.6192 & 0.7213 \\

\midrule
DOCNet
& HRNet-W32 & CE$^{*}$ & 0.6101 & 0.5911 & 0.6004 & 0.7001 & 0.7815 & 0.7385 & 0.5317 & 0.6207 & 0.7455 \\

\midrule
\multirow{3}{*}{PPMambaSeg}
& SWSL & CE+Dice & 0.4914 & 0.6305 & 0.5523 & 0.6670 & 0.8063 & 0.7301 & 0.5615 & 0.6806 & 0.7332 \\
& SWSL & Focal+Dice & 0.4956 & 0.6459 & 0.5609 & 0.6726 & 0.8394 & 0.7468 & 0.5828 & 0.6895 & 0.7636 \\
& SWSL & WCE+Dice & 0.5786 & 0.7293 & 0.6453 & 0.6509 & 0.8175 & 0.7247 & 0.6408 & 0.6273 & 0.7318 \\

\midrule
\multirow{3}{*}{RS3Mamba}
& VMamba-Tiny & CE+Dice & 0.3354 & 0.4376 & 0.3797 & 0.6214 & 0.8425 & 0.7153 & 0.4235 & \cellcolor{green!10}\textbf{0.7530} & 0.7490 \\
& VMamba-Tiny & Focal+Dice & 0.4559 & 0.5104 & 0.4816 & 0.6167 & 0.8720 & 0.7225 & 0.4709 & 0.6587 & 0.7541 \\
& VMamba-Tiny & WCE+Dice & 0.3574 & 0.6920 & 0.4714 & 0.5223 & \cellcolor{green!10}\textbf{0.8905} & 0.6584 & 0.5263 & 0.5523 & 0.6762 \\

\midrule
\multirow{3}{*}{PyramidMamba}
& Swin-Base & CE+Dice & 0.5523 & 0.6965 & 0.6161 & 0.6568 & 0.8324 & 0.7343 & 0.6301 & 0.6680 & 0.7337 \\
& Swin-Base & Focal+Dice & 0.4768 & 0.7094 & 0.5703 & 0.6385 & 0.8289 & 0.7213 & 0.6365 & 0.6183 & 0.7184 \\
& Swin-Base & WCE+Dice & 0.5602 & 0.7758 & \cellcolor{green!10}\textbf{0.6506} & 0.6263 & 0.8459 & 0.7197 & \cellcolor{green!10}\textbf{0.7024} & 0.6181 & 0.7357 \\

\midrule
LoGCAN++
& RepViT-M2.3 & CE$^{*}$ & 0.5133 & 0.5051 & 0.5091 & 0.6588 & 0.7460 & 0.6997 & 0.4524 & 0.5142 & 0.6945 \\

\midrule
MF-Mamba
& HRNet-W18 & CE+Dice & 0.4403 & 0.6114 & 0.5119 & 0.6508 & 0.7890 & 0.7133 & 0.5137 & 0.6195 & 0.7104 \\

\midrule
\multirow{3}{*}{MCPNet}
& ResNet-50 & CE+Dice & 0.4849 & 0.5565 & 0.5182 & 0.6956 & 0.7545 & 0.7238 & 0.5063 & 0.6357 & 0.7242 \\
& ResNet-50 & Focal+Dice & 0.4945 & 0.5715 & 0.5302 & 0.6993 & 0.7839 & 0.7392 & 0.5267 & 0.5524 & 0.7538 \\
& ResNet-50 & WCE+Dice & 0.5195 & 0.6184 & 0.5646 & 0.6749 & 0.7615 & 0.7156 & 0.5545 & 0.6160 & 0.7130 \\

\bottomrule
\end{tabular}%
}
\end{table*}

\begin{table*}[t]
\centering
\scriptsize
\caption{Results of methods related to vision foundation models on segmentation-derived multi-label classification. An asterisk ($^{*}$) denotes the use of an auxiliary loss.}
\label{tab:app_vfm_multilabel}
\setlength{\tabcolsep}{2.2pt}
\renewcommand{\arraystretch}{1.05}
\resizebox{\textwidth}{!}{%
\begin{tabular}{llcccccccccc}
\toprule
Model & Backbone & Loss
& CP & CR & CF1 & OP & OR & OF1 & mAP & Macro-F1 & Sample-F1 \\
\midrule

\multirow{3}{*}{HQ-SAM}
& ViT-B + HQ Decoder & CE+Dice & 0.3647 & 0.6824 & 0.4754 & 0.4954 & \cellcolor{red!10}\textbf{0.9257} & 0.6455 & 0.4697 & 0.5098 & 0.6488 \\
& ViT-B + HQ Decoder & Focal+Dice & 0.3846 & 0.6560 & 0.4849 & 0.5066 & 0.9178 & 0.6529 & 0.4710 & 0.5258 & 0.6623 \\
& ViT-B + HQ Decoder & WCE+Dice & 0.3280 & 0.7370 & 0.4540 & 0.4455 & 0.9243 & 0.6012 & 0.4693 & 0.4226 & 0.5995 \\

\midrule
\multirow{4}{*}{SAM\_RS}
& ABCNet + SAM Priors & Seg+Bdy+Obj & 0.5300 & 0.5205 & 0.5252 & 0.7355 & 0.7214 & 0.7284 & 0.4938 & 0.6442 & 0.7352 \\
& CMTFNet + SAM Priors & Seg+Bdy+Obj & 0.5355 & 0.5312 & 0.5334 & 0.7243 & 0.7374 & 0.7308 & 0.4867 & 0.6571 & 0.7269 \\
& FTUNetFormer + SAM Priors & Seg+Bdy+Obj & 0.4921 & 0.5071 & 0.4995 & 0.7213 & 0.7331 & 0.7271 & 0.4757 & 0.6172 & 0.7409 \\
& UNetFormer + SAM Priors & Seg+Bdy+Obj & 0.4799 & 0.5933 & 0.5306 & 0.6829 & 0.7843 & 0.7301 & 0.5091 & 0.6442 & 0.7384 \\

\midrule
\multirow{6}{*}{SAM2.1}
& Hiera-B+ (Frozen, MSFPN) & CE+Dice & 0.3745 & 0.6554 & 0.4767 & 0.5165 & 0.9060 & 0.6579 & 0.4668 & 0.5117 & 0.6603 \\
& Hiera-B+ (Frozen, MSFPN) & Focal+Dice & 0.3508 & 0.6742 & 0.4615 & 0.4933 & 0.9156 & 0.6412 & 0.4648 & 0.4904 & 0.6411 \\
& Hiera-B+ (Frozen, MSFPN) & WCE+Dice & 0.3157 & 0.7515 & 0.4446 & 0.4238 & 0.9081 & 0.5779 & 0.4665 & 0.4031 & 0.5986 \\
& Hiera-B+ (Full FT, MSFPN) & CE+Dice & 0.4654 & 0.5394 & 0.4996 & 0.6785 & 0.7464 & 0.7108 & 0.4929 & 0.6116 & 0.7148 \\
& Hiera-B+ (Full FT, MSFPN) & Focal+Dice & 0.4484 & 0.5770 & 0.5046 & 0.6897 & 0.7728 & 0.7289 & 0.5124 & 0.6143 & 0.7440 \\
& Hiera-B+ (Full FT, MSFPN) & WCE+Dice & 0.4270 & 0.6527 & 0.5163 & 0.6359 & 0.7738 & 0.6981 & 0.5113 & 0.5313 & 0.7258 \\

\midrule
\multirow{3}{*}{RSAM-Seg}
& SAM-ViT-B (Frozen Encoder) & CE+Dice & 0.4322 & 0.6280 & 0.5121 & 0.6393 & 0.8147 & 0.7164 & 0.5349 & 0.6171 & 0.7400 \\
& SAM-ViT-B (Frozen Encoder) & Focal+Dice & 0.5499 & 0.6354 & 0.5896 & 0.7046 & 0.8230 & 0.7592 & 0.5705 & 0.6147 & \cellcolor{red!10}\textbf{0.7843} \\
& SAM-ViT-B (Frozen Encoder) & WCE+Dice & 0.4842 & 0.7614 & 0.5920 & 0.6559 & 0.8140 & 0.7264 & 0.6206 & 0.5666 & 0.7512 \\

\midrule
\multirow{11}{*}{SESSRS}
& A2-FPN (CE+Dice) & Postprocess & 0.5133 & 0.5665 & 0.5386 & \cellcolor{red!10}\textbf{0.7381} & 0.7526 & 0.7453 & 0.5592 & 0.6678 & 0.7725 \\
& A2-FPN (Focal+Dice) & Postprocess & 0.4574 & 0.6230 & 0.5275 & 0.6502 & 0.7846 & 0.7111 & 0.5414 & 0.6394 & 0.7161 \\
& A2-FPN (WCE+Dice) & Postprocess & \cellcolor{red!10}\textbf{0.5660} & 0.6809 & \cellcolor{red!10}\textbf{0.6181} & 0.7081 & 0.7544 & 0.7305 & 0.6176 & 0.6064 & 0.7553 \\
& ABCNet (CE+Dice)$^{*}$ & Postprocess & 0.5193 & 0.5399 & 0.5294 & 0.7338 & 0.7367 & 0.7353 & 0.5150 & 0.6593 & 0.7460 \\
& BANet (CE+Dice) & Postprocess & 0.4616 & 0.5574 & 0.5050 & 0.6664 & 0.7606 & 0.7104 & 0.4863 & 0.6238 & 0.7155 \\
& MANet (CE+Dice) & Postprocess & 0.4261 & 0.6917 & 0.5273 & 0.6085 & 0.8824 & 0.7203 & 0.5556 & 0.6430 & 0.7340 \\
& MANet (Focal+Dice) & Postprocess & 0.4344 & \cellcolor{red!10}\textbf{0.8610} & 0.5774 & 0.5777 & 0.8897 & 0.7005 & 0.6170 & 0.5444 & 0.7203 \\
& MANet (WCE+Dice) & Postprocess & 0.4458 & 0.8443 & 0.5835 & 0.5860 & 0.8794 & 0.7033 & \cellcolor{red!10}\textbf{0.6817} & 0.5535 & 0.7236 \\
& UNetFormer (CE+Dice) & Postprocess & 0.5221 & 0.6390 & 0.5747 & 0.6912 & 0.8521 & \cellcolor{red!10}\textbf{0.7633} & 0.5913 & \cellcolor{red!10}\textbf{0.7105} & 0.7838 \\
& UNetFormer (Focal+Dice) & Postprocess & 0.4633 & 0.6671 & 0.5468 & 0.6343 & 0.8535 & 0.7278 & 0.5807 & 0.6704 & 0.7533 \\
& UNetFormer (WCE+Dice) & Postprocess & 0.4444 & 0.7464 & 0.5571 & 0.6075 & 0.8561 & 0.7107 & 0.6120 & 0.5893 & 0.7314 \\

\bottomrule
\end{tabular}%
}
\end{table*}

Among general segmentation models, the strongest overall performance is obtained by a small set of methods rather than a single uniformly dominant architecture. As shown in Table~\ref{tab:app_general_multilabel}, SegFormer with WCE+Dice gives the best OF1 (0.7539), mAP (0.6576), and Sample-F1 (0.7790), making it the most balanced general model for segmentation-derived recognition. EfficientViT with WCE+Dice achieves the highest CF1 (0.6321), while DeepLabV3+ with ResNet-50 and CE+Dice gives the best Macro-F1 (0.6833). Other models peak on more isolated metrics, such as Mask2Former on CP (0.6584), PEM on OP (0.7678), and BiSeNetv2 on OR (0.8911). These results suggest that general-purpose segmentation backbones can produce competitive recognition signals, but their strengths remain scattered across different evaluation criteria.

The remote-sensing-specific models in Table~\ref{tab:app_rs_multilabel} are stronger overall and more stable across metrics. PyramidMamba with WCE+Dice reaches the highest CF1 (0.6506) and mAP (0.7024), indicating the strongest class-balanced recognition quality in this group. UNetFormer with auxiliary supervision attains the best OF1 (0.7633), while RS3Mamba with CE+Dice yields the highest Macro-F1 (0.7530). A2-FPN with CE+Dice gives the best OP (0.7381), and FarSeg with Focal+Dice achieves the best Sample-F1 (0.7887). Other peak values are distributed across SACANet for CP (0.6200), MANet with Focal+Dice for CR (0.8610), and RS3Mamba with WCE+Dice for OR (0.8905). Compared with the general baselines, these methods convert segmentation outputs into recognition labels more effectively, consistent with their stronger adaptation to remote-sensing imagery and imbalanced foreground categories.

The methods related to vision foundation models show a less uniform pattern. Table~\ref{tab:app_vfm_multilabel} indicates that direct use of SAM-style backbones is not sufficient to guarantee uniformly strong recognition performance. SESSRS-based post-processing improves several thresholded recognition metrics and also gives the strongest ranking-based result in this group: SESSRS with A2-FPN and WCE+Dice gives the best CP (0.5660) and CF1 (0.6181), SESSRS with A2-FPN and CE+Dice gives the best OP (0.7381), SESSRS with UNetFormer and CE+Dice gives the best OF1 (0.7633) and Macro-F1 (0.7105), and SESSRS with MANet and WCE+Dice achieves the highest mAP (0.6817). HQ-SAM mainly stands out on OR (0.9257), while RSAM-Seg with Focal+Dice gives the highest Sample-F1 (0.7843). Overall, these results suggest that foundation-model-related methods can be useful components in recognition-oriented segmentation pipelines, but their benefits depend strongly on how the segmentation output and confidence scores are produced.

Overall, Tables~\ref{tab:app_general_multilabel}--\ref{tab:app_vfm_multilabel} show that segmentation-derived recognition should be judged from several angles at once. Some methods favor recall-heavy behavior, others are more precise, and the models with the strongest thresholded multi-label scores are not always the ones with the best ranking-based mAP. For that reason, the main text focuses on CF1, mAP, OF1, and Macro-F1 when discussing recognition quality derived from segmentation outputs.

\subsubsection{Per-class Segmentation-derived Recognition Performance}

Tables~\ref{tab:app_general_perclass_f1_ap}, \ref{tab:app_rs_perclass_f1_ap}, and \ref{tab:app_vfm_perclass_f1_ap} further break down segmentation-derived recognition into class-wise F1 and AP results. These tables make it possible to see whether an improvement in aggregate recognition is driven by large and visually stable classes or by the smaller disturbance-related categories that are more important for mining analysis.

\begin{table*}[t]
\centering
\scriptsize
\caption{Per-class F1 and AP results of general segmentation models on segmentation-derived multi-label classification. The upper block reports F1, and the lower block reports AP.}
\label{tab:app_general_perclass_f1_ap}
\setlength{\tabcolsep}{2.0pt}
\renewcommand{\arraystretch}{0.95}
\resizebox{\textwidth}{!}{%
\begin{tabular}{lllcccccccccc}
\toprule
\multirow{2}{*}{Model} &
\multirow{2}{*}{Backbone} &
\multirow{2}{*}{Loss} &
\multirow{2}{*}{Buildings} &
Mining & Primary & Water & Agricultural & Gravel & Type 1 & Type 2 & Bare & \multirow{2}{*}{Sluices} \\
& & &
& rafts & forests & bodies & crops & mounds & regeneration & regeneration & ground & \\
\midrule
\rowcolor{gray!15}\multicolumn{13}{c}{\textbf{F1}} \\
\midrule

\multirow{6}{*}{DeepLabV3+}
& ConvNeXt-Tiny & CE+Dice    & 0.5939 & 0.4916 & 0.8251 & 0.8661 & 0.4102 & 0.6344 & 0.7037 & 0.6014 & 0.7709 & 0.2374 \\
& ConvNeXt-Tiny & Focal+Dice & 0.5936 & 0.4536 & 0.8083 & 0.8651 & 0.4324 & 0.6431 & 0.7069 & 0.5856 & 0.7795 & 0.0862 \\
& ConvNeXt-Tiny & WCE+Dice   & 0.5728 & 0.4542 & 0.7956 & 0.8732 & 0.3479 & 0.5901 & 0.7149 & 0.5873 & 0.7669 & 0.3125 \\
& ResNet-50     & CE+Dice    & 0.6523 & 0.0000 & 0.8194 & 0.8725 & 0.4295 & 0.6443 & 0.7047 & 0.5917 & 0.7518 & 0.0000 \\
& ResNet-50     & Focal+Dice & 0.4694 & 0.4967 & 0.8352 & 0.8690 & 0.2771 & 0.6573 & 0.7059 & 0.5817 & 0.7390 & 0.0000 \\
& ResNet-50     & WCE+Dice   & \cellcolor{blue!10}\textbf{0.6729} & 0.4865 & 0.8168 & 0.8486 & 0.2468 & 0.6664 & 0.7241 & 0.5609 & 0.7522 & 0.0000 \\

\midrule
UPerNet
& Swin-Tiny & CE+Dice & 0.5324 & 0.3091 & 0.8082 & 0.8656 & 0.3937 & 0.5697 & 0.7051 & 0.5376 & 0.7639 & 0.0000 \\

\midrule
OCRNet
& HRNet-W48 & CE+Dice & 0.5709 & 0.0000 & 0.6941 & 0.8088 & 0.2351 & 0.6486 & 0.7148 & 0.5756 & 0.7125 & 0.0000 \\

\midrule
\multirow{3}{*}{BiSeNetv2}
& Custom Bilateral & CE+Dice    & 0.0425 & 0.0000 & \cellcolor{blue!10}\textbf{0.8591} & 0.8662 & 0.1901 & 0.4760 & 0.7044 & 0.5929 & 0.7054 & 0.0000 \\
& Custom Bilateral & Focal+Dice & 0.2432 & 0.0000 & 0.8442 & 0.8630 & 0.3338 & \cellcolor{blue!10}\textbf{0.6986} & 0.7328 & \cellcolor{blue!10}\textbf{0.6139} & 0.7156 & 0.0000 \\
& Custom Bilateral & WCE+Dice   & 0.0777 & 0.0000 & 0.8336 & 0.8197 & 0.2670 & 0.3548 & 0.6854 & 0.5847 & 0.7646 & 0.0000 \\

\midrule
\multirow{3}{*}{SegFormer}
& MiT-B2 & CE+Dice    & 0.3756 & 0.3594 & 0.7784 & 0.8180 & 0.3902 & 0.6545 & 0.7001 & 0.5718 & 0.7311 & 0.2180 \\
& MiT-B2 & Focal+Dice & 0.2960 & 0.4968 & 0.8100 & 0.7926 & 0.4639 & 0.6335 & 0.7095 & 0.6035 & 0.7159 & 0.2131 \\
& MiT-B2 & WCE+Dice   & 0.6226 & \cellcolor{blue!10}\textbf{0.5798} & 0.8372 & \cellcolor{blue!10}\textbf{0.8914} & 0.5355 & 0.5991 & \cellcolor{blue!10}\textbf{0.7573} & 0.4656 & 0.7842 & 0.0000 \\

\midrule
\multirow{3}{*}{STDC2}
& STDCNet & CE+Dice    & 0.1669 & 0.0000 & 0.8276 & 0.8780 & 0.4406 & 0.5003 & 0.7136 & 0.5765 & 0.7652 & 0.0000 \\
& STDCNet & Focal+Dice & 0.1763 & 0.0000 & 0.8313 & 0.8641 & 0.3832 & 0.6956 & 0.7289 & 0.5628 & 0.7504 & 0.0000 \\
& STDCNet & WCE+Dice   & 0.6539 & 0.3725 & 0.7831 & 0.8685 & 0.4738 & 0.6863 & 0.7210 & 0.5847 & 0.7511 & 0.0000 \\

\midrule
Mask2Former
& ResNet-50 & Set CE+Mask+Dice & 0.6077 & 0.4348 & 0.8194 & 0.7269 & 0.4487 & 0.6810 & 0.7000 & 0.5520 & 0.7151 & 0.0000 \\

\midrule
SegNeXt
& MSCAN-Tiny & CE+Dice & 0.5026 & 0.2286 & 0.7631 & 0.8023 & 0.1934 & 0.5572 & 0.7192 & 0.5817 & 0.7245 & 0.0000 \\

\midrule
DDRNet
& DDRNet-23-slim & Focal+Dice & 0.5458 & 0.0000 & 0.7718 & 0.8277 & 0.3037 & 0.6373 & 0.7111 & 0.5798 & 0.7021 & 0.0000 \\

\midrule
Afformer
& AFFormer-Base & CE+Dice & 0.4366 & 0.3792 & 0.7892 & 0.7787 & 0.3474 & 0.5931 & 0.7267 & 0.5951 & 0.7155 & 0.1353 \\

\midrule
\multirow{3}{*}{EfficientViT}
& EfficientViT-B2 & CE+Dice    & 0.5413 & 0.3279 & 0.8307 & 0.8786 & 0.4021 & 0.5579 & 0.7151 & 0.5642 & \cellcolor{blue!10}\textbf{0.7987} & 0.0000 \\
& EfficientViT-B2 & Focal+Dice & 0.5509 & 0.4940 & 0.8022 & 0.8616 & 0.4443 & 0.6182 & 0.7155 & 0.5610 & 0.7682 & 0.2400 \\
& EfficientViT-B2 & WCE+Dice   & 0.4946 & 0.4667 & 0.8248 & 0.8706 & 0.3823 & 0.6608 & 0.7165 & 0.5913 & 0.7861 & \cellcolor{blue!10}\textbf{0.3194} \\

\midrule
SeaFormer
& SeaFormer-Base & CE+Dice & 0.4777 & 0.1818 & 0.8078 & 0.7835 & 0.5055 & 0.5808 & 0.7362 & 0.5645 & 0.7089 & 0.0000 \\

\midrule
\multirow{3}{*}{PIDNet}
& PIDNet-M & CE+Dice    & 0.4570 & 0.0000 & 0.7880 & 0.8546 & 0.4785 & 0.5682 & 0.7470 & 0.5756 & 0.7685 & 0.0000 \\
& PIDNet-M & Focal+Dice & 0.0445 & 0.0000 & 0.8197 & 0.8325 & 0.1134 & 0.3915 & 0.7227 & 0.5646 & 0.6942 & 0.0000 \\
& PIDNet-M & WCE+Dice   & 0.2486 & 0.4327 & 0.7848 & 0.8634 & 0.4181 & 0.5586 & 0.6990 & 0.6021 & 0.7724 & 0.0000 \\

\midrule
MP-Former
& Swin-L & Focal+Dice & 0.0000 & 0.0000 & 0.7564 & 0.4384 & 0.0000 & 0.0008 & 0.4132 & 0.3935 & 0.4655 & 0.0000 \\

\midrule
CGRSeg
& EfficientFormerV2-B & CE+Dice & 0.4978 & 0.3277 & 0.7787 & 0.7721 & 0.3091 & 0.5336 & 0.7085 & 0.5563 & 0.7049 & 0.0583 \\

\midrule
PEM
& ResNet-50 & Set CE+Mask+Dice & 0.5202 & 0.2412 & 0.8336 & 0.7024 & \cellcolor{blue!10}\textbf{0.5593} & 0.6283 & 0.7163 & 0.5592 & 0.6949 & 0.0000 \\

\midrule
VMamba
& VMamba-Tiny & CE+Dice & 0.5333 & 0.0000 & 0.8168 & 0.7620 & 0.2646 & 0.6642 & 0.6994 & 0.5386 & 0.7109 & 0.0000 \\

\midrule
\rowcolor{gray!15}\multicolumn{13}{c}{\textbf{AP}} \\
\midrule

\multirow{6}{*}{DeepLabV3+}
& ConvNeXt-Tiny & CE+Dice & 0.5605 & 0.4701 & 0.8847 & 0.9299 & 0.4738 & 0.7446 & 0.7435 & 0.5157 & 0.8092 & 0.0980 \\
& ConvNeXt-Tiny & Focal+Dice & 0.6140 & 0.4814 & 0.8796 & 0.9305 & 0.4805 & 0.6994 & 0.7469 & 0.5092 & 0.8130 & 0.0247 \\
& ConvNeXt-Tiny & WCE+Dice & 0.6399 & 0.3737 & 0.8786 & 0.9266 & 0.3730 & 0.7248 & 0.7488 & 0.4562 & 0.7829 & \cellcolor{blue!10}\textbf{0.2023} \\
& ResNet-50 & CE+Dice & 0.7742 & 0.0189 & 0.8792 & 0.9309 & 0.4392 & 0.7893 & 0.7381 & 0.4932 & 0.7031 & 0.0100 \\
& ResNet-50 & Focal+Dice & 0.6631 & 0.3782 & 0.8915 & 0.9342 & 0.4640 & 0.7316 & 0.7507 & 0.5020 & 0.7952 & 0.0100 \\
& ResNet-50 & WCE+Dice & 0.7206 & 0.3840 & 0.8781 & 0.9187 & 0.1457 & 0.7374 & 0.7482 & 0.4850 & 0.6364 & 0.0100 \\

\midrule
UPerNet
& Swin-Tiny & CE+Dice & 0.4622 & 0.1962 & 0.8641 & 0.9184 & 0.3475 & 0.5842 & 0.7145 & 0.4247 & 0.7305 & 0.0100 \\

\midrule
OCRNet
& HRNet-W48 & CE+Dice & 0.4815 & 0.0189 & 0.8225 & 0.8599 & 0.3158 & 0.6986 & 0.7299 & 0.4226 & 0.5076 & 0.0100 \\

\midrule
\multirow{3}{*}{BiSeNetv2}
& Custom Bilateral & CE+Dice & 0.4594 & 0.0189 & 0.8778 & 0.9257 & 0.4250 & 0.7560 & \cellcolor{blue!10}\textbf{0.7736} & 0.4815 & 0.6496 & 0.0100 \\
& Custom Bilateral & Focal+Dice & 0.3515 & 0.0189 & \cellcolor{blue!10}\textbf{0.9016} & 0.9179 & 0.4481 & 0.6983 & 0.7525 & 0.5067 & 0.6578 & 0.0100 \\
& Custom Bilateral & WCE+Dice & 0.3610 & 0.0189 & 0.8482 & \cellcolor{blue!10}\textbf{0.9442} & 0.4472 & 0.7794 & 0.7533 & 0.4421 & 0.7650 & 0.0100 \\

\midrule
\multirow{3}{*}{SegFormer}
& MiT-B2 & CE+Dice & 0.5163 & 0.3023 & 0.8575 & 0.8678 & 0.2783 & 0.7244 & 0.7231 & 0.4819 & 0.5599 & 0.0701 \\
& MiT-B2 & Focal+Dice & 0.5922 & 0.4603 & 0.8903 & 0.8403 & 0.5325 & 0.7161 & 0.7384 & 0.4723 & 0.5539 & 0.0605 \\
& MiT-B2 & WCE+Dice & \cellcolor{blue!10}\textbf{0.8137} & \cellcolor{blue!10}\textbf{0.5563} & 0.8925 & 0.9333 & \cellcolor{blue!10}\textbf{0.5681} & \cellcolor{blue!10}\textbf{0.8198} & 0.7717 & 0.4194 & 0.7908 & 0.0100 \\

\midrule
\multirow{3}{*}{STDC2}
& STDCNet & CE+Dice & 0.2907 & 0.0189 & 0.8943 & 0.9225 & 0.3793 & 0.5378 & 0.6640 & 0.4966 & 0.8024 & 0.0100 \\
& STDCNet & Focal+Dice & 0.2553 & 0.0189 & 0.8924 & 0.9105 & 0.3562 & 0.6966 & 0.7334 & 0.4529 & 0.6834 & 0.0100 \\
& STDCNet & WCE+Dice & 0.5716 & 0.2593 & 0.8718 & 0.9157 & 0.4984 & 0.6962 & 0.7486 & 0.4659 & 0.7181 & 0.0100 \\

\midrule
Mask2Former
& ResNet-50 & Set CE+Mask+Dice & 0.4482 & 0.3164 & 0.7934 & 0.7863 & 0.3046 & 0.6250 & 0.6906 & 0.5706 & 0.7516 & 0.0100 \\

\midrule
SegNeXt
& MSCAN-Tiny & CE+Dice & 0.4830 & 0.1331 & 0.8519 & 0.8595 & 0.1045 & 0.5783 & 0.7317 & 0.4265 & 0.5776 & 0.0100 \\

\midrule
DDRNet
& DDRNet-23-slim & Focal+Dice & 0.4808 & 0.0189 & 0.8657 & 0.8715 & 0.3883 & 0.6356 & 0.7235 & 0.4485 & 0.6381 & 0.0100 \\

\midrule
Afformer
& AFFormer-Base & CE+Dice & 0.5744 & 0.2345 & 0.8592 & 0.8407 & 0.3401 & 0.6210 & 0.7575 & 0.4570 & 0.5805 & 0.0319 \\

\midrule
\multirow{3}{*}{EfficientViT}
& EfficientViT-B2 & CE+Dice & 0.6018 & 0.2333 & 0.8910 & 0.9257 & 0.5477 & 0.7617 & 0.7343 & 0.4750 & \cellcolor{blue!10}\textbf{0.8210} & 0.0100 \\
& EfficientViT-B2 & Focal+Dice & 0.5371 & 0.4208 & 0.8769 & 0.9121 & 0.4961 & 0.7526 & 0.7441 & 0.4659 & 0.7921 & 0.1021 \\
& EfficientViT-B2 & WCE+Dice & 0.6490 & 0.3473 & 0.8843 & 0.9280 & 0.4724 & 0.7022 & 0.7340 & 0.5165 & 0.8111 & 0.1827 \\

\midrule
SeaFormer
& SeaFormer-Base & CE+Dice & 0.5452 & 0.1100 & 0.8709 & 0.8387 & 0.4316 & 0.6091 & 0.7631 & 0.4593 & 0.5326 & 0.0100 \\

\midrule
\multirow{3}{*}{PIDNet}
& PIDNet-M & CE+Dice & 0.4811 & 0.0189 & 0.8766 & 0.9101 & 0.4882 & 0.5863 & 0.7391 & 0.4270 & 0.7172 & 0.0100 \\
& PIDNet-M & Focal+Dice & 0.1607 & 0.0189 & 0.8815 & 0.8879 & 0.0578 & 0.3648 & 0.7354 & 0.4331 & 0.7475 & 0.0100 \\
& PIDNet-M & WCE+Dice & 0.3210 & 0.3226 & 0.8719 & 0.9172 & 0.3865 & 0.5997 & 0.6427 & 0.4775 & 0.7047 & 0.0100 \\

\midrule
CGRSeg
& EfficientFormerV2-B & CE+Dice & 0.5869 & 0.2319 & 0.8501 & 0.8265 & 0.2199 & 0.5555 & 0.7172 & 0.4265 & 0.5447 & 0.0307 \\

\midrule
PEM
& ResNet-50 & Set CE+Mask+Dice & 0.3560 & 0.1290 & 0.8110 & 0.7546 & 0.3815 & 0.5961 & 0.6932 & \cellcolor{blue!10}\textbf{0.5735} & 0.7643 & 0.0100 \\

\midrule
VMamba
& VMamba-Tiny & CE+Dice & 0.3841 & 0.0189 & 0.8790 & 0.8191 & 0.1858 & 0.6347 & 0.6951 & 0.4674 & 0.5540 & 0.0100 \\

\bottomrule
\end{tabular}%
}
\end{table*}

\begin{table*}[t]
\centering
\scriptsize
\caption{Per-class F1 and AP results of remote-sensing-specific methods on segmentation-derived multi-label classification. An asterisk ($^{*}$) denotes the use of an auxiliary loss. The upper block reports F1, and the lower block reports AP.}
\label{tab:app_rs_perclass_f1_ap}
\setlength{\tabcolsep}{1.8pt}
\renewcommand{\arraystretch}{0.6}
\resizebox{\textwidth}{!}{%
\begin{tabular}{lllcccccccccc}
\toprule
\multirow{2}{*}{Model} &
\multirow{2}{*}{Backbone} &
\multirow{2}{*}{Loss} &
\multirow{2}{*}{Buildings} &
Mining & Primary & Water & Agricultural & Gravel & Type 1 & Type 2 & Bare & \multirow{2}{*}{Sluices} \\
& & &
& rafts & forests & bodies & crops & mounds & regeneration & regeneration & ground & \\
\midrule
\rowcolor{gray!15}\multicolumn{13}{c}{\textbf{F1}} \\
\midrule

\multirow{4}{*}{FarSeg}
& ResNet-50 & CE         & 0.5925 & 0.0000 & 0.8051 & 0.8933 & 0.3569 & 0.5930 & 0.7230 & 0.5989 & 0.7932 & 0.0000 \\
& ResNet-50 & CE+Dice    & 0.5564 & 0.0000 & 0.7547 & 0.8888 & 0.3416 & 0.5353 & 0.7255 & 0.6158 & 0.7808 & 0.0000 \\
& ResNet-50 & Focal+Dice & 0.5394 & 0.0000 & 0.8470 & \cellcolor{green!10}\textbf{0.8959} & 0.3319 & \cellcolor{green!10}\textbf{0.7288} & 0.7281 & 0.6107 & 0.7758 & 0.0000 \\
& ResNet-50 & WCE+Dice   & 0.2652 & 0.3295 & 0.7696 & 0.8666 & 0.3615 & 0.5254 & 0.6504 & 0.5685 & 0.7568 & 0.3617 \\

\midrule
BANet
& ResT-Lite & CE+Dice & 0.4342 & 0.0000 & 0.8144 & 0.8050 & 0.3899 & 0.5463 & 0.6995 & 0.5767 & 0.7248 & 0.0000 \\

\midrule
ABCNet
& ResNet-18 & CE+Dice$^{*}$ & 0.5893 & 0.0000 & 0.8138 & 0.8642 & 0.3391 & 0.6776 & 0.7153 & 0.5203 & 0.7547 & 0.0000 \\

\midrule
\multirow{3}{*}{MANet}
& ResNet-50 & CE+Dice    & 0.5924 & 0.0000 & 0.7979 & 0.8628 & 0.3572 & 0.5627 & 0.7265 & 0.5965 & 0.7402 & 0.0000 \\
& ResNet-50 & Focal+Dice & 0.3888 & 0.2873 & 0.8260 & 0.8623 & 0.3103 & 0.5072 & 0.7063 & 0.5914 & 0.7202 & 0.2445 \\
& ResNet-50 & WCE+Dice   & 0.5355 & 0.2603 & 0.8073 & 0.8545 & 0.2753 & 0.4907 & 0.7216 & 0.5878 & 0.7526 & 0.2494 \\

\midrule
\multirow{3}{*}{UNetFormer}
& ResNet-18 & CE+Dice$^{*}$    & \cellcolor{green!10}\textbf{0.6641} & 0.0000 & 0.8310 & 0.8755 & 0.5662 & 0.6208 & \cellcolor{green!10}\textbf{0.7490} & 0.6188 & 0.7588 & 0.0000 \\
& ResNet-18 & Focal+Dice$^{*}$ & 0.5568 & 0.0000 & 0.8276 & 0.8852 & 0.4554 & 0.5489 & 0.7282 & \cellcolor{green!10}\textbf{0.6194} & 0.7417 & 0.0000 \\
& ResNet-18 & WCE+Dice$^{*}$   & 0.3654 & 0.4491 & 0.8347 & 0.8727 & 0.2961 & 0.4452 & 0.7184 & 0.5952 & 0.7271 & 0.0000 \\

\midrule
DC-Swin
& Swin-Small & CE+Dice & 0.4747 & 0.0000 & 0.8144 & 0.8602 & 0.2968 & 0.4934 & 0.7178 & 0.6054 & 0.7411 & 0.0000 \\

\midrule
\multirow{3}{*}{A2-FPN}
& ResNet-18 & CE+Dice    & 0.6086 & 0.0000 & 0.8133 & 0.8862 & 0.3868 & 0.6286 & 0.6758 & 0.5440 & \cellcolor{green!10}\textbf{0.7993} & 0.0000 \\
& ResNet-18 & Focal+Dice & 0.6231 & 0.0000 & 0.8057 & 0.8505 & 0.3343 & 0.5007 & 0.7184 & 0.5485 & 0.7344 & 0.0000 \\
& ResNet-18 & WCE+Dice   & 0.6170 & 0.4527 & 0.8267 & 0.8706 & 0.4241 & 0.5826 & 0.7208 & 0.4983 & 0.7593 & 0.3119 \\

\midrule
LoGCAN
& ResNet-50 & CE$^{*}$ & 0.6328 & 0.3789 & 0.8479 & 0.8577 & 0.5305 & 0.3573 & 0.7033 & 0.6089 & 0.7583 & 0.0000 \\

\midrule
FarSeg++
& MiT-B2 & CE & 0.4934 & 0.3983 & 0.8239 & 0.8195 & 0.2602 & 0.5459 & 0.6741 & 0.5960 & 0.7280 & 0.0000 \\

\midrule
SACANet
& HRNet-W32 & CE$^{*}$ & 0.5948 & 0.1307 & 0.7916 & 0.8419 & 0.5326 & 0.6411 & 0.7340 & 0.5719 & 0.7338 & 0.0000 \\

\midrule
DOCNet
& HRNet-W32 & CE$^{*}$ & 0.5722 & 0.3140 & 0.8037 & 0.8512 & 0.3907 & 0.6091 & 0.6987 & 0.5876 & 0.7593 & 0.0000 \\

\midrule
\multirow{3}{*}{PPMambaSeg}
& SWSL & CE+Dice    & 0.5998 & 0.0000 & 0.8231 & 0.8462 & 0.4891 & 0.6484 & 0.7110 & 0.5967 & 0.7305 & 0.0000 \\
& SWSL & Focal+Dice & 0.5963 & 0.0000 & 0.8311 & 0.8611 & \cellcolor{green!10}\textbf{0.5933} & 0.5763 & 0.7365 & 0.5527 & 0.7685 & 0.0000 \\
& SWSL & WCE+Dice   & 0.5586 & 0.5191 & 0.8322 & 0.8496 & 0.3887 & 0.6104 & 0.7078 & 0.5897 & 0.7266 & \cellcolor{green!10}\textbf{0.4906} \\

\midrule
\multirow{3}{*}{RS3Mamba}
& VMamba-Tiny & CE+Dice    & 0.0000 & 0.0000 & \cellcolor{green!10}\textbf{0.8569} & 0.8671 & 0.0000 & 0.0000 & 0.7340 & 0.5869 & 0.7202 & 0.0000 \\
& VMamba-Tiny & Focal+Dice & 0.0000 & 0.0000 & 0.8522 & 0.8708 & 0.3955 & 0.4499 & 0.7350 & 0.5832 & 0.7240 & 0.0000 \\
& VMamba-Tiny & WCE+Dice   & 0.1652 & 0.0000 & 0.8418 & 0.8342 & 0.2752 & 0.4392 & 0.6430 & 0.5909 & 0.6289 & 0.0000 \\

\midrule
\multirow{3}{*}{PyramidMamba}
& Swin-Base & CE+Dice    & 0.6051 & 0.5667 & 0.8208 & 0.8591 & 0.3988 & 0.7015 & 0.7309 & 0.5866 & 0.7424 & 0.0000 \\
& Swin-Base & Focal+Dice & 0.5441 & 0.3413 & 0.7982 & 0.8547 & 0.3971 & 0.6041 & 0.7315 & 0.5896 & 0.7038 & 0.0000 \\
& Swin-Base & WCE+Dice   & 0.4966 & \cellcolor{green!10}\textbf{0.6014} & 0.8046 & 0.8476 & 0.2091 & 0.6757 & 0.7289 & 0.6093 & 0.7507 & 0.4575 \\

\midrule
LoGCAN++
& RepViT-M2.3 & CE$^{*}$ & 0.3050 & 0.1039 & 0.7984 & 0.7759 & 0.2000 & 0.4497 & 0.6850 & 0.5850 & 0.7249 & 0.0000 \\

\midrule
MF-Mamba
& HRNet-W18 & CE+Dice & 0.5027 & 0.0000 & 0.8152 & 0.8093 & 0.2650 & 0.5562 & 0.7071 & 0.5781 & 0.7226 & 0.0000 \\

\midrule
\multirow{3}{*}{MCPNet}
& ResNet-50 & CE+Dice    & 0.4235 & 0.0000 & 0.8162 & 0.8111 & 0.3775 & 0.6619 & 0.6900 & 0.5767 & 0.7291 & 0.0000 \\
& ResNet-50 & Focal+Dice & 0.2488 & 0.0000 & 0.8287 & 0.8540 & 0.4191 & 0.5460 & 0.7167 & 0.5910 & 0.7465 & 0.0206 \\
& ResNet-50 & WCE+Dice   & 0.4634 & 0.3242 & 0.8147 & 0.7986 & 0.4993 & 0.5915 & 0.7239 & 0.6102 & 0.7180 & 0.0000 \\

\midrule
\rowcolor{gray!15}\multicolumn{13}{c}{\textbf{AP}} \\
\midrule

\multirow{4}{*}{FarSeg}
& ResNet-50 & CE & 0.5435 & 0.0189 & 0.8746 & 0.9459 & 0.3968 & 0.7467 & 0.7406 & 0.4529 & 0.8462 & 0.0100 \\
& ResNet-50 & CE+Dice & 0.5892 & 0.0189 & 0.8490 & 0.9422 & 0.5241 & 0.7795 & 0.7598 & 0.4413 & 0.8379 & 0.0100 \\
& ResNet-50 & Focal+Dice & 0.7438 & 0.0189 & \cellcolor{green!10}\textbf{0.9053} & \cellcolor{green!10}\textbf{0.9516} & 0.2878 & 0.7579 & 0.7630 & 0.4816 & \cellcolor{green!10}\textbf{0.8528} & 0.0100 \\
& ResNet-50 & WCE+Dice & 0.6762 & 0.3389 & 0.8606 & 0.9338 & 0.5183 & 0.7662 & 0.6375 & 0.4348 & 0.6965 & 0.2962 \\

\midrule
BANet
& ResT-Lite & CE+Dice & 0.5043 & 0.0185 & 0.8800 & 0.8655 & 0.2897 & 0.5945 & 0.6983 & 0.4473 & 0.5548 & 0.0100 \\

\midrule
ABCNet
& ResNet-18 & CE+Dice$^{*}$ & 0.5336 & 0.0189 & 0.8696 & 0.9077 & 0.2597 & 0.7413 & 0.6850 & 0.4197 & 0.7041 & 0.0100 \\

\midrule
\multirow{3}{*}{MANet}
& ResNet-50 & CE+Dice & 0.7770 & 0.0181 & 0.8558 & 0.9490 & 0.5034 & 0.5916 & 0.7336 & 0.4234 & 0.6941 & 0.0100 \\
& ResNet-50 & Focal+Dice & 0.7816 & 0.2292 & 0.8944 & 0.9397 & 0.5425 & 0.7294 & 0.7422 & 0.4348 & 0.6382 & 0.2377 \\
& ResNet-50 & WCE+Dice & \cellcolor{green!10}\textbf{0.8797} & \cellcolor{green!10}\textbf{0.7104} & 0.8894 & 0.9461 & 0.5387 & 0.8221 & 0.7289 & 0.4043 & 0.6655 & 0.2316 \\

\midrule
\multirow{3}{*}{UNetFormer}
& ResNet-18 & CE+Dice$^{*}$ & 0.7615 & 0.0189 & 0.9018 & 0.9415 & 0.4988 & 0.7970 & 0.7621 & 0.4917 & 0.7298 & 0.0100 \\
& ResNet-18 & Focal+Dice$^{*}$ & 0.7688 & 0.0189 & 0.8928 & 0.9451 & 0.4542 & 0.7449 & 0.7203 & 0.5087 & 0.7437 & 0.0100 \\
& ResNet-18 & WCE+Dice$^{*}$ & 0.8290 & 0.4190 & 0.8935 & 0.9273 & 0.4719 & 0.7618 & 0.7550 & 0.4596 & 0.5921 & 0.0100 \\

\midrule
DC-Swin
& Swin-Small & CE+Dice & 0.4915 & 0.0189 & 0.8828 & 0.9204 & 0.2186 & 0.6074 & 0.7343 & 0.4530 & 0.6165 & 0.0100 \\

\midrule
\multirow{3}{*}{A2-FPN}
& ResNet-18 & CE+Dice & 0.6169 & 0.0189 & 0.8789 & 0.9370 & 0.3888 & 0.7417 & 0.6752 & 0.4866 & 0.8384 & 0.0100 \\
& ResNet-18 & Focal+Dice & 0.7121 & 0.0189 & 0.8722 & 0.9017 & 0.3364 & 0.7490 & 0.7505 & 0.4352 & 0.6282 & 0.0100 \\
& ResNet-18 & WCE+Dice & 0.6652 & 0.3397 & 0.8789 & 0.9183 & 0.4605 & 0.7435 & 0.7266 & 0.4551 & 0.6861 & 0.3019 \\

\midrule
LoGCAN
& ResNet-50 & CE$^{*}$ & 0.5547 & 0.2676 & \cellcolor{green!10}\textbf{0.9053} & 0.9331 & 0.4860 & 0.4014 & 0.7218 & 0.5253 & 0.6934 & 0.0100 \\

\midrule
FarSeg++
& MiT-B2 & CE & 0.4085 & 0.3274 & 0.8789 & 0.8719 & 0.1357 & 0.5440 & 0.7032 & 0.4743 & 0.6031 & 0.0099 \\

\midrule
SACANet
& HRNet-W32 & CE$^{*}$ & 0.5654 & 0.0937 & 0.8755 & 0.8899 & 0.5717 & 0.6314 & 0.7664 & 0.4211 & 0.7002 & 0.0100 \\

\midrule
DOCNet
& HRNet-W32 & CE$^{*}$ & 0.5083 & 0.2296 & 0.8815 & 0.9130 & 0.3333 & 0.6102 & 0.7011 & 0.4370 & 0.6935 & 0.0100 \\

\midrule
\multirow{3}{*}{PPMambaSeg}
& SWSL & CE+Dice & 0.6559 & 0.0189 & 0.8838 & 0.8978 & 0.4776 & 0.8051 & 0.7090 & 0.5149 & 0.6416 & 0.0100 \\
& SWSL & Focal+Dice & 0.7406 & 0.0189 & 0.8977 & 0.9258 & 0.5313 & 0.8103 & 0.7524 & 0.4300 & 0.7112 & 0.0100 \\
& SWSL & WCE+Dice & 0.7717 & 0.3993 & 0.8787 & 0.9093 & 0.4862 & 0.7686 & 0.7121 & 0.4693 & 0.6474 & \cellcolor{green!10}\textbf{0.3652} \\

\midrule
\multirow{3}{*}{RS3Mamba}
& VMamba-Tiny & CE+Dice & 0.0150 & 0.0189 & 0.9011 & 0.9291 & 0.0453 & 0.2438 & 0.7405 & \cellcolor{green!10}\textbf{0.5423} & 0.7889 & 0.0100 \\
& VMamba-Tiny & Focal+Dice & 0.0150 & 0.0189 & 0.8981 & 0.9312 & 0.3472 & 0.4767 & 0.7502 & 0.5131 & 0.7485 & 0.0100 \\
& VMamba-Tiny & WCE+Dice & 0.4916 & 0.0189 & 0.8792 & 0.9219 & 0.5172 & 0.5694 & 0.7669 & 0.4904 & 0.5971 & 0.0100 \\

\midrule
\multirow{3}{*}{PyramidMamba}
& Swin-Base & CE+Dice & 0.7195 & 0.4744 & 0.8822 & 0.9089 & 0.4577 & 0.7697 & 0.7673 & 0.4868 & 0.8241 & 0.0100 \\
& Swin-Base & Focal+Dice & 0.8647 & 0.4124 & 0.8799 & 0.9025 & \cellcolor{green!10}\textbf{0.5748} & 0.7770 & \cellcolor{green!10}\textbf{0.7689} & 0.4686 & 0.7060 & 0.0100 \\
& Swin-Base & WCE+Dice & 0.8734 & 0.5746 & 0.8683 & 0.9185 & 0.5646 & \cellcolor{green!10}\textbf{0.8256} & 0.7681 & 0.5418 & 0.7480 & 0.3406 \\

\midrule
LoGCAN++
& RepViT-M2.3 & CE$^{*}$ & 0.3992 & 0.0829 & 0.8489 & 0.8472 & 0.1764 & 0.4763 & 0.6906 & 0.4567 & 0.5363 & 0.0100 \\

\midrule
MF-Mamba
& HRNet-W18 & CE+Dice & 0.5553 & 0.0189 & 0.8765 & 0.8671 & 0.3626 & 0.6620 & 0.7263 & 0.4751 & 0.5832 & 0.0100 \\

\midrule
\multirow{3}{*}{MCPNet}
& ResNet-50 & CE+Dice & 0.4156 & 0.0186 & 0.8820 & 0.8715 & 0.3612 & 0.6907 & 0.7297 & 0.5011 & 0.5824 & 0.0100 \\
& ResNet-50 & WCE+Dice & 0.3707 & 0.3980 & 0.8808 & 0.8533 & 0.5057 & 0.6440 & 0.7542 & 0.5084 & 0.6202 & 0.0100 \\
& ResNet-50 & Focal+Dice & 0.4573 & 0.0189 & 0.8901 & 0.9069 & 0.4804 & 0.5895 & 0.7350 & 0.5057 & 0.6699 & 0.0136 \\

\bottomrule
\end{tabular}%
}
\end{table*}

\begin{table*}[t]
\centering
\scriptsize
\caption{Per-class F1 and AP results of methods related to vision foundation models on segmentation-derived multi-label classification. An asterisk ($^{*}$) denotes the use of an auxiliary loss. The upper block reports F1, and the lower block reports AP.}
\label{tab:app_vfm_perclass_f1_ap}
\setlength{\tabcolsep}{1.8pt}
\renewcommand{\arraystretch}{0.95}
\resizebox{\textwidth}{!}{%
\begin{tabular}{lllcccccccccc}
\toprule
\multirow{2}{*}{Model} &
\multirow{2}{*}{Backbone} &
\multirow{2}{*}{Loss} &
\multirow{2}{*}{Buildings} &
Mining & Primary & Water & Agricultural & Gravel & Type 1 & Type 2 & Bare & \multirow{2}{*}{Sluices} \\
& & &
& rafts & forests & bodies & crops & mounds & regeneration & regeneration & ground & \\
\midrule
\rowcolor{gray!15}\multicolumn{13}{c}{\textbf{F1}} \\
\midrule

\multirow{3}{*}{HQ-SAM}
& ViT-B + HQ Decoder & CE+Dice    & 0.2650 & 0.2742 & 0.8317 & 0.8034 & 0.1554 & 0.4401 & 0.6075 & 0.5295 & 0.6816 & 0.0000 \\
& ViT-B + HQ Decoder & Focal+Dice & 0.3891 & 0.2571 & 0.8420 & 0.8230 & 0.1538 & 0.4547 & 0.6074 & 0.5350 & 0.6704 & 0.0000 \\
& ViT-B + HQ Decoder & WCE+Dice   & 0.1885 & 0.1317 & 0.8370 & 0.8043 & 0.0900 & 0.3779 & 0.5638 & 0.5392 & 0.6825 & 0.0114 \\

\midrule
\multirow{4}{*}{SAM\_RS}
& ABCNet + SAM Priors       & Seg+Bdy+Obj & 0.4706 & 0.0000 & 0.7746 & 0.8683 & 0.3770 & 0.6583 & 0.6995 & 0.5612 & 0.7438 & 0.0000 \\
& CMTFNet + SAM Priors      & Seg+Bdy+Obj & 0.5960 & 0.0000 & 0.7991 & 0.8265 & 0.3537 & 0.6539 & 0.7021 & 0.5735 & 0.7524 & 0.0000 \\
& FTUNetFormer + SAM Priors & Seg+Bdy+Obj & 0.5283 & 0.0000 & 0.8385 & 0.8197 & 0.2546 & 0.5327 & 0.6512 & 0.5877 & 0.7252 & 0.0000 \\
& UNetFormer + SAM Priors   & Seg+Bdy+Obj & 0.4683 & 0.0000 & 0.8209 & 0.8461 & 0.2873 & \cellcolor{red!10}\textbf{0.6799} & 0.7035 & 0.6103 & 0.7374 & 0.0000 \\

\midrule
\multirow{6}{*}{SAM2.1}
& Hiera-B+ (Frozen, MSFPN)  & CE+Dice    & 0.2931 & 0.2764 & 0.8286 & 0.7907 & 0.1616 & 0.3809 & 0.6310 & 0.5655 & 0.6772 & 0.0000 \\
& Hiera-B+ (Frozen, MSFPN)  & Focal+Dice & 0.1646 & 0.2939 & 0.8265 & 0.7839 & 0.1003 & 0.4316 & 0.5924 & 0.5522 & 0.6685 & 0.0000 \\
& Hiera-B+ (Frozen, MSFPN)  & WCE+Dice   & 0.0811 & 0.0733 & 0.8244 & 0.7743 & 0.0833 & 0.3363 & 0.5908 & 0.5555 & 0.6794 & 0.0327 \\
& Hiera-B+ (Full FT, MSFPN) & CE+Dice    & 0.4919 & 0.0000 & 0.8156 & 0.8086 & 0.3076 & 0.4769 & 0.6901 & 0.5747 & 0.7276 & 0.0000 \\
& Hiera-B+ (Full FT, MSFPN) & Focal+Dice & 0.4240 & 0.0000 & 0.8426 & 0.8277 & 0.3034 & 0.5269 & 0.7127 & 0.5422 & 0.7347 & 0.0000 \\
& Hiera-B+ (Full FT, MSFPN) & WCE+Dice   & 0.2929 & 0.1856 & 0.8314 & 0.8328 & 0.2342 & 0.4182 & 0.6931 & 0.5589 & 0.7348 & 0.0000 \\

\midrule
\multirow{3}{*}{RSAM-Seg}
& SAM-ViT-B (Frozen Encoder) & CE+Dice    & 0.4356 & 0.0000 & 0.8117 & 0.8568 & 0.3332 & 0.4318 & 0.7229 & 0.5892 & 0.7556 & 0.0000 \\
& SAM-ViT-B (Frozen Encoder) & Focal+Dice & 0.3829 & 0.2938 & \cellcolor{red!10}\textbf{0.8451} & 0.8635 & 0.4645 & 0.5876 & 0.7322 & 0.5965 & 0.7666 & 0.0000 \\
& SAM-ViT-B (Frozen Encoder) & WCE+Dice   & 0.5163 & 0.2919 & 0.8330 & 0.8503 & 0.3877 & 0.5205 & 0.7216 & 0.5643 & 0.7508 & 0.2297 \\

\midrule
\multirow{11}{*}{SESSRS}
& A2-FPN (CE+Dice)        & Postprocess & 0.6086 & 0.0000 & 0.8133 & \cellcolor{red!10}\textbf{0.8861} & 0.3868 & 0.6286 & 0.6758 & 0.5440 & \cellcolor{red!10}\textbf{0.7993} & 0.0000 \\
& A2-FPN (Focal+Dice)     & Postprocess & 0.6231 & 0.0000 & 0.8057 & 0.8503 & 0.3343 & 0.5007 & 0.7184 & 0.5485 & 0.7344 & 0.0000 \\
& A2-FPN (WCE+Dice)       & Postprocess & 0.6170 & \cellcolor{red!10}\textbf{0.4527} & 0.8267 & 0.8704 & 0.4241 & 0.5826 & 0.7208 & 0.4983 & 0.7593 & \cellcolor{red!10}\textbf{0.3119} \\
& ABCNet (CE+Dice)$^{*}$  & Postprocess & 0.5893 & 0.0000 & 0.8138 & 0.8641 & 0.3391 & 0.6776 & 0.7153 & 0.5203 & 0.7547 & 0.0000 \\
& BANet (CE+Dice)         & Postprocess & 0.4342 & 0.0000 & 0.8144 & 0.8050 & 0.3899 & 0.5463 & 0.6995 & 0.5767 & 0.7248 & 0.0000 \\
& MANet (CE+Dice)         & Postprocess & 0.5380 & 0.0000 & 0.8104 & 0.8524 & 0.3014 & 0.5640 & 0.7374 & 0.5885 & 0.7521 & 0.0000 \\
& MANet (Focal+Dice)      & Postprocess & 0.3888 & 0.2873 & 0.8260 & 0.8623 & 0.3103 & 0.5072 & 0.7063 & 0.5914 & 0.7202 & 0.2445 \\
& MANet (WCE+Dice)        & Postprocess & 0.5355 & 0.2603 & 0.8073 & 0.8544 & 0.2753 & 0.4907 & 0.7216 & 0.5878 & 0.7525 & 0.2494 \\
& UNetFormer (CE+Dice)    & Postprocess & \cellcolor{red!10}\textbf{0.6641} & 0.0000 & 0.8310 & 0.8753 & \cellcolor{red!10}\textbf{0.5662} & 0.6208 & \cellcolor{red!10}\textbf{0.7490} & 0.6188 & 0.7588 & 0.0000 \\
& UNetFormer (Focal+Dice) & Postprocess & 0.5568 & 0.0000 & 0.8276 & 0.8850 & 0.4554 & 0.5489 & 0.7282 & \cellcolor{red!10}\textbf{0.6194} & 0.7417 & 0.0000 \\
& UNetFormer (WCE+Dice)   & Postprocess & 0.3654 & 0.4491 & 0.8347 & 0.8724 & 0.2961 & 0.4452 & 0.7184 & 0.5952 & 0.7271 & 0.0000 \\

\midrule
\rowcolor{gray!15}\multicolumn{13}{c}{\textbf{AP}} \\
\midrule

\multirow{3}{*}{HQ-SAM}
& ViT-B + HQ Decoder & CE+Dice & 0.4272 & 0.0977 & 0.8905 & 0.9218 & 0.1728 & 0.5084 & 0.7092 & 0.4227 & 0.5372 & 0.0100 \\
& ViT-B + HQ Decoder & Focal+Dice & 0.5042 & 0.1114 & 0.8973 & 0.9182 & 0.1216 & 0.4533 & 0.7032 & 0.4449 & 0.5463 & 0.0100 \\
& ViT-B + HQ Decoder & WCE+Dice & 0.4210 & 0.1291 & 0.8922 & 0.9088 & 0.1921 & 0.5121 & 0.7033 & 0.4122 & 0.5130 & 0.0095 \\

\midrule
\multirow{4}{*}{SAM\_RS}
& ABCNet + SAM Priors & Seg+Bdy+Obj & 0.4019 & 0.0189 & 0.8645 & 0.9125 & 0.2399 & 0.6537 & 0.6969 & 0.4391 & 0.7011 & 0.0100 \\
& CMTFNet + SAM Priors & Seg+Bdy+Obj & 0.5065 & 0.0189 & 0.8661 & 0.8742 & 0.1383 & 0.6519 & 0.7126 & 0.4589 & 0.6297 & 0.0100 \\
& FTUNetFormer + SAM Priors & Seg+Bdy+Obj & 0.4173 & 0.0189 & 0.8821 & 0.8749 & 0.1570 & 0.6086 & 0.6996 & \cellcolor{red!10}\textbf{0.5340} & 0.5551 & 0.0100 \\
& UNetFormer + SAM Priors & Seg+Bdy+Obj & 0.4374 & 0.0189 & 0.8913 & 0.9069 & 0.2495 & 0.7081 & 0.7474 & 0.4871 & 0.6345 & 0.0100 \\

\midrule
\multirow{6}{*}{SAM2.1}
& Hiera-B+ (Frozen, MSFPN) & CE+Dice & 0.5237 & 0.1335 & 0.8854 & 0.8844 & 0.1010 & 0.5053 & 0.7059 & 0.4033 & 0.5151 & 0.0100 \\
& Hiera-B+ (Frozen, MSFPN) & Focal+Dice & 0.5176 & 0.1467 & 0.8940 & 0.8874 & 0.0643 & 0.5239 & 0.7078 & 0.3942 & 0.5023 & 0.0100 \\
& Hiera-B+ (Frozen, MSFPN) & WCE+Dice & 0.4531 & 0.2150 & 0.8880 & 0.8756 & 0.0649 & 0.5488 & 0.7134 & 0.3982 & 0.4952 & 0.0126 \\
& Hiera-B+ (Full FT, MSFPN) & CE+Dice & 0.5516 & 0.0189 & 0.8773 & 0.8599 & 0.2326 & 0.5910 & 0.7171 & 0.4537 & 0.6164 & 0.0100 \\
& Hiera-B+ (Full FT, MSFPN) & Focal+Dice & 0.6102 & 0.0189 & 0.8771 & 0.8750 & 0.2271 & 0.6144 & 0.7275 & 0.4858 & 0.6777 & 0.0100 \\
& Hiera-B+ (Full FT, MSFPN) & WCE+Dice & 0.3644 & 0.3615 & 0.8940 & 0.8916 & 0.2359 & 0.5707 & 0.7070 & 0.4720 & 0.6060 & 0.0100 \\

\midrule
\multirow{3}{*}{RSAM-Seg}
& SAM-ViT-B (Frozen Encoder) & CE+Dice & 0.5917 & 0.0189 & 0.8828 & 0.9084 & 0.3755 & 0.6160 & 0.7402 & 0.4752 & 0.7300 & 0.0100 \\
& SAM-ViT-B (Frozen Encoder) & Focal+Dice & 0.6099 & 0.2017 & \cellcolor{red!10}\textbf{0.9024} & 0.9202 & 0.3439 & 0.6827 & 0.7598 & 0.5294 & 0.7449 & 0.0100 \\
& SAM-ViT-B (Frozen Encoder) & WCE+Dice & 0.6679 & 0.4979 & 0.8896 & 0.9172 & 0.4805 & 0.7221 & 0.7459 & 0.4666 & 0.6769 & 0.1417 \\

\midrule
\multirow{11}{*}{SESSRS}
& A2-FPN (CE+Dice) & Postprocess & 0.6169 & 0.0189 & 0.8789 & 0.9370 & 0.3888 & 0.7417 & 0.6752 & 0.4866 & \cellcolor{red!10}\textbf{0.8384} & 0.0100 \\
& A2-FPN (Focal+Dice) & Postprocess & 0.7121 & 0.0189 & 0.8722 & 0.9017 & 0.3364 & 0.7490 & 0.7505 & 0.4353 & 0.6282 & 0.0100 \\
& A2-FPN (WCE+Dice) & Postprocess & 0.6652 & 0.3398 & 0.8789 & 0.9183 & 0.4605 & 0.7433 & 0.7266 & 0.4550 & 0.6861 & \cellcolor{red!10}\textbf{0.3019} \\
& ABCNet (CE+Dice)$^{*}$ & Postprocess & 0.5336 & 0.0189 & 0.8696 & 0.9078 & 0.2597 & 0.7413 & 0.6850 & 0.4196 & 0.7041 & 0.0100 \\
& BANet (CE+Dice) & Postprocess & 0.5043 & 0.0185 & 0.8800 & 0.8655 & 0.2898 & 0.5945 & 0.6983 & 0.4473 & 0.5548 & 0.0100 \\
& MANet (CE+Dice) & Postprocess & 0.7770 & 0.0181 & 0.8558 & \cellcolor{red!10}\textbf{0.9489} & 0.5034 & 0.5917 & 0.7334 & 0.4235 & 0.6941 & 0.0100 \\
& MANet (Focal+Dice) & Postprocess & 0.7816 & 0.2291 & 0.8943 & 0.9397 & \cellcolor{red!10}\textbf{0.5424} & 0.7294 & 0.7421 & 0.4348 & 0.6382 & 0.2379 \\
& MANet (WCE+Dice) & Postprocess & \cellcolor{red!10}\textbf{0.8797} & \cellcolor{red!10}\textbf{0.7103} & 0.8894 & 0.9463 & 0.5389 & \cellcolor{red!10}\textbf{0.8220} & 0.7288 & 0.4043 & 0.6655 & 0.2316 \\
& UNetFormer (CE+Dice) & Postprocess & 0.7615 & 0.0189 & 0.9018 & 0.9415 & 0.4988 & 0.7970 & \cellcolor{red!10}\textbf{0.7620} & 0.4917 & 0.7298 & 0.0100 \\
& UNetFormer (Focal+Dice) & Postprocess & 0.7688 & 0.0189 & 0.8928 & 0.9451 & 0.4542 & 0.7449 & 0.7202 & 0.5087 & 0.7437 & 0.0100 \\
& UNetFormer (WCE+Dice) & Postprocess & 0.8290 & 0.4201 & 0.8934 & 0.9273 & 0.4719 & 0.7618 & 0.7550 & 0.4596 & 0.5921 & 0.0100 \\

\bottomrule
\end{tabular}%
}
\end{table*}

A broad trend across all three groups is that image-level recognition remains much easier for dominant categories such as primary forests, water bodies, and bare ground than for rare categories such as mining rafts and sluices. Even when a method achieves strong overall OF1 or mAP, its class-wise scores on sparse classes can still remain weak. This gap is especially visible in F1, where missing a small class entirely immediately produces a zero score. The class-wise tables are therefore necessary for interpreting whether a model is truly useful for rare-category monitoring or mainly strong on high-coverage categories.

For the general-model group, Table~\ref{tab:app_general_perclass_f1_ap} shows that the best class-wise scores are distributed across several architectures. On F1, SegFormer with WCE+Dice performs best on mining rafts (0.5798), water bodies (0.8914), and type 1 regeneration (0.7573), while BiSeNetv2 variants lead on primary forests (0.8591), gravel mounds (0.6986), and type 2 regeneration (0.6139). DeepLabV3+ with ResNet-50 and WCE+Dice gives the highest building F1 (0.6729), PEM gives the best agricultural-crop F1 (0.5593), and EfficientViT gives the best bare-ground (0.7987) and sluice F1 (0.3194). On AP, SegFormer with WCE+Dice performs best on buildings (0.8137), mining rafts (0.5563), agricultural crops (0.5681), and gravel mounds (0.8198). BiSeNetv2 variants are strongest on primary forests (0.9016), water bodies (0.9442), and type 1 regeneration (0.7736), while PEM gives the best type 2 regeneration AP (0.5735), EfficientViT with CE+Dice gives the best bare-ground AP (0.8210), and DeepLabV3+ with ConvNeXt-Tiny and WCE+Dice gives the best sluice AP (0.2023). This pattern again shows that general models can be strong, but their class-wise behavior is uneven.

The remote-sensing-specific models in Table~\ref{tab:app_rs_perclass_f1_ap} are more convincing on the difficult mining-related categories. On F1, UNetFormer with auxiliary supervision gives the best results on buildings (0.6641) and type 1 regeneration (0.7490), PyramidMamba with WCE+Dice performs best on mining rafts (0.6014), RS3Mamba with CE+Dice is strongest on primary forests (0.8569), and FarSeg with Focal+Dice leads on water bodies (0.8959) and gravel mounds (0.7288). PPMambaSeg with Focal+Dice gives the highest agricultural-crop F1 (0.5933), UNetFormer with Focal+Dice gives the best type 2 regeneration F1 (0.6194), A2-FPN with CE+Dice gives the best bare-ground F1 (0.7993), and PPMambaSeg with WCE+Dice gives the highest sluice F1 (0.4906). On AP, MANet with WCE+Dice performs best on buildings (0.8797) and mining rafts (0.7104), while FarSeg with Focal+Dice leads on water bodies (0.9516) and bare ground (0.8528), and is tied after rounding with LoGCAN on primary forests (0.9053). PyramidMamba variants are strongest on agricultural crops (0.5748), gravel mounds (0.8256), and type 1 regeneration (0.7689), RS3Mamba with CE+Dice gives the best type 2 regeneration AP (0.5423), and PPMambaSeg with WCE+Dice achieves the highest sluice AP (0.3652). Compared with the general group, these methods improve not only aggregate recognition scores but also the coverage of rare disturbance-related categories.

Table~\ref{tab:app_vfm_perclass_f1_ap} shows that methods related to vision foundation models remain heterogeneous at the class level. Frozen or lightly adapted variants such as HQ-SAM and SAM2.1 are comparatively weak on several small categories, even when they remain competitive on primary forests and water bodies. The strongest F1 results mainly come from SESSRS variants, RSAM-Seg, and SAM\_RS with downstream segmenters. On F1, SESSRS variants achieve the best scores on buildings (0.6641), mining rafts (0.4527), water bodies (0.8861), agricultural crops (0.5662), type 1 regeneration (0.7490), type 2 regeneration (0.6194), bare ground (0.7993), and sluices (0.3119), while RSAM-Seg with Focal+Dice gives the best primary-forest F1 (0.8451) and SAM\_RS with UNetFormer gives the best gravel-mound F1 (0.6799). On AP, the strongest results are also dominated by SESSRS variants but are distributed across different base segmenters: SESSRS with MANet and WCE+Dice leads on buildings (0.8797), mining rafts (0.7103), and gravel mounds (0.8220); SESSRS with MANet and CE+Dice gives the best water-body AP (0.9489); and SESSRS with MANet and Focal+Dice gives the best agricultural-crop AP (0.5424). RSAM-Seg with Focal+Dice remains strongest on primary forests (0.9024), SESSRS with UNetFormer and CE+Dice leads on type 1 regeneration (0.7620), SAM\_RS with FTUNetFormer gives the best type 2 regeneration AP (0.5340), and SESSRS with A2-FPN gives the best bare-ground AP (0.8384 with CE+Dice) and sluice AP (0.3019 with WCE+Dice). The class-wise evidence therefore supports the same conclusion drawn from the overall tables: foundation-model-related approaches become most effective when coupled with task-specific refinement, but their gains are not uniform across classes or confidence-based AP.

In summary, Tables~\ref{tab:app_general_perclass_f1_ap}--\ref{tab:app_vfm_perclass_f1_ap} show that segmentation-derived recognition is highly class-dependent. Gains in aggregate metrics often reflect better recovery of large-area categories, but the most useful methods are those that also improve recognition on rare mining-related classes. The class-wise F1 and AP results are therefore important for separating broad semantic improvement from performance that is concentrated only on easier categories.

\FloatBarrier

\begin{figure*}[t]
\centering
\includegraphics[width=\textwidth]{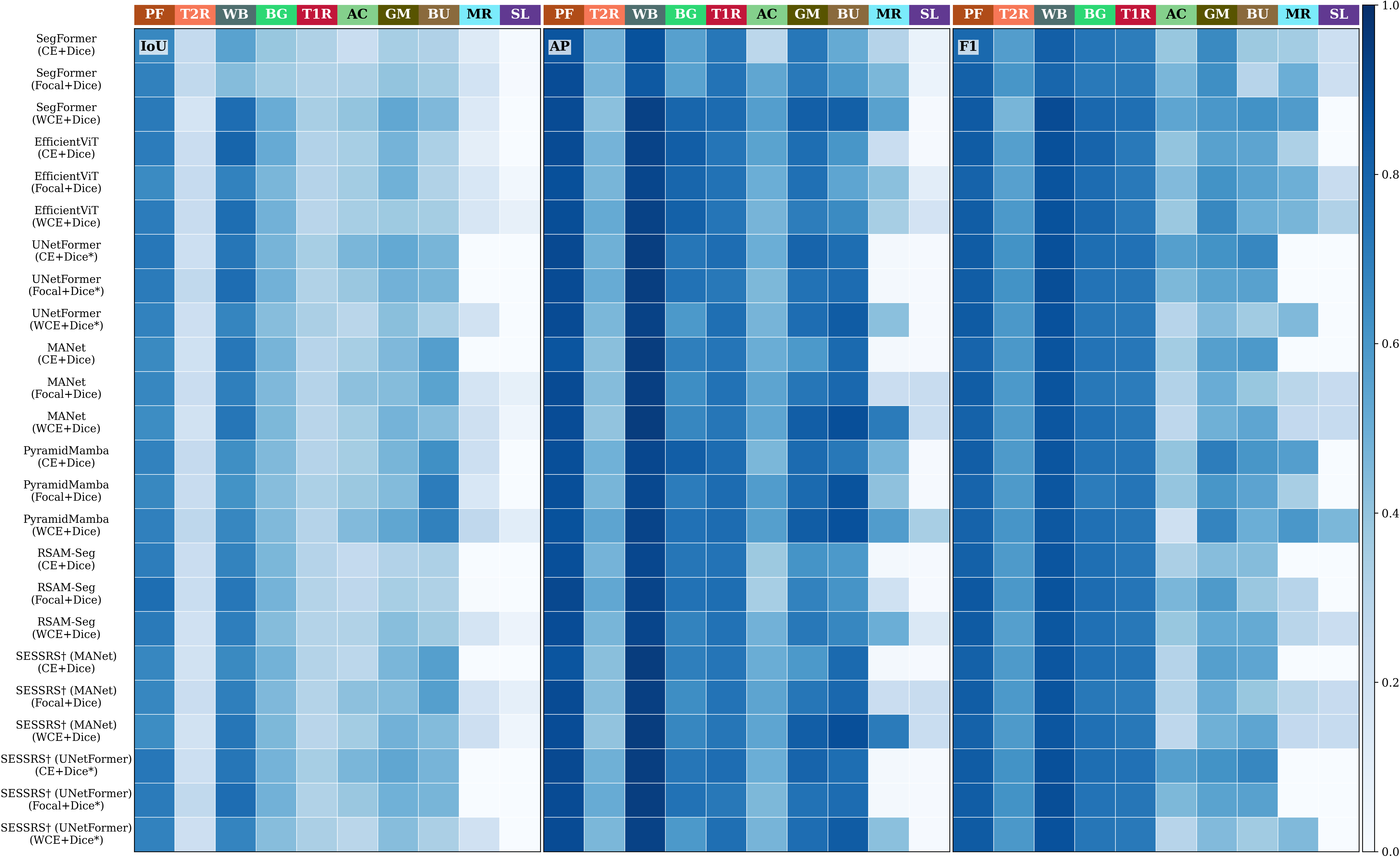}
\caption{Per-class test performance across the 24 model variants listed in the main comparison table. Rows follow the same method--loss ordering as the main comparison, and columns correspond to the 10 foreground classes present in the test split, ordered by decreasing pixel frequency. The three blocks report per-class segmentation IoU, segmentation-derived classification AP, and segmentation-derived classification F1. The figure shows that segmentation and segmentation-derived recognition are broadly consistent at the class level: dominant categories are generally easier, rare mining-related categories remain the most difficult, and visually ambiguous recovery and disturbance classes continue to challenge most methods.}
\label{fig:app_perclass_full_heatmap}
\end{figure*}

\subsection{Additional Per-class and Error Analysis}
\label{app:error_seg}

This section provides additional analysis beyond the detailed per-class tables reported above. We first summarize class-wise behavior across all representative method variants in a unified heatmap view, which makes it easier to compare segmentation and segmentation-derived recognition at the same time. We then examine error patterns of selected methods through confusion matrix analysis, in order to identify which categories remain difficult and which class pairs are most frequently confused. Finally, we provide qualitative visualizations of representative categories and model predictions, which offer a more direct view of how different methods behave on fine-grained mining-related structures and visually ambiguous recovery classes.

\subsubsection{Per-class Performance Across Method Variants}
\label{app:perclass_selected_seg}

The detailed per-class tables above provide complete results, but they make it harder to see the broader relationship between segmentation and segmentation-derived recognition across many model variants at once. Figure~\ref{fig:app_perclass_full_heatmap} addresses this by placing class-wise IoU, AP, and F1 in a single aligned view. This makes it possible to compare not only which classes are easy or difficult, but also how closely the ranking of classes agrees between dense prediction and image-level recognition.

A clear pattern is that the segmentation and classification blocks are broadly consistent. Classes that are strong in IoU are usually also strong in AP and F1, and classes that are weak in segmentation are usually weak in recognition as well. In other words, the relative difficulty of the classes is largely preserved across the two views. This is especially visible for dominant categories such as primary forest and water bodies, which remain among the easier classes for most methods, and for mining rafts and sluices, which remain consistently difficult across all three blocks.

Most classes also follow the expected relationship with their overall amount in the dataset. Classes with larger support tend to achieve stronger results, while sparse mining-related categories tend to remain much harder. However, the pattern is not purely explained by frequency. Type 1 regeneration and Type 2 regeneration remain relatively difficult even though they are not among the rarest categories, and their performance is clearly weaker than that of other high-frequency classes. This suggests that the challenge comes not only from class imbalance, but also from their visual similarity to neighboring ecological states such as primary forest, bare ground, and other disturbed or recovering regions. Agricultural crops show another notable pattern: their segmentation results are often reasonable, but their AP and F1 are comparatively weaker. This suggests that pixel-level delineation can be easier than reliable image-level presence prediction for this category.

Overall, the figure shows that class frequency explains an important part of the benchmark difficulty, but not all of it. The remaining difficulty is concentrated on visually ambiguous disturbance and recovery categories, where segmentation and recognition are related but still not fully interchangeable.

\begin{figure*}[t]
\centering
\includegraphics[width=\textwidth]{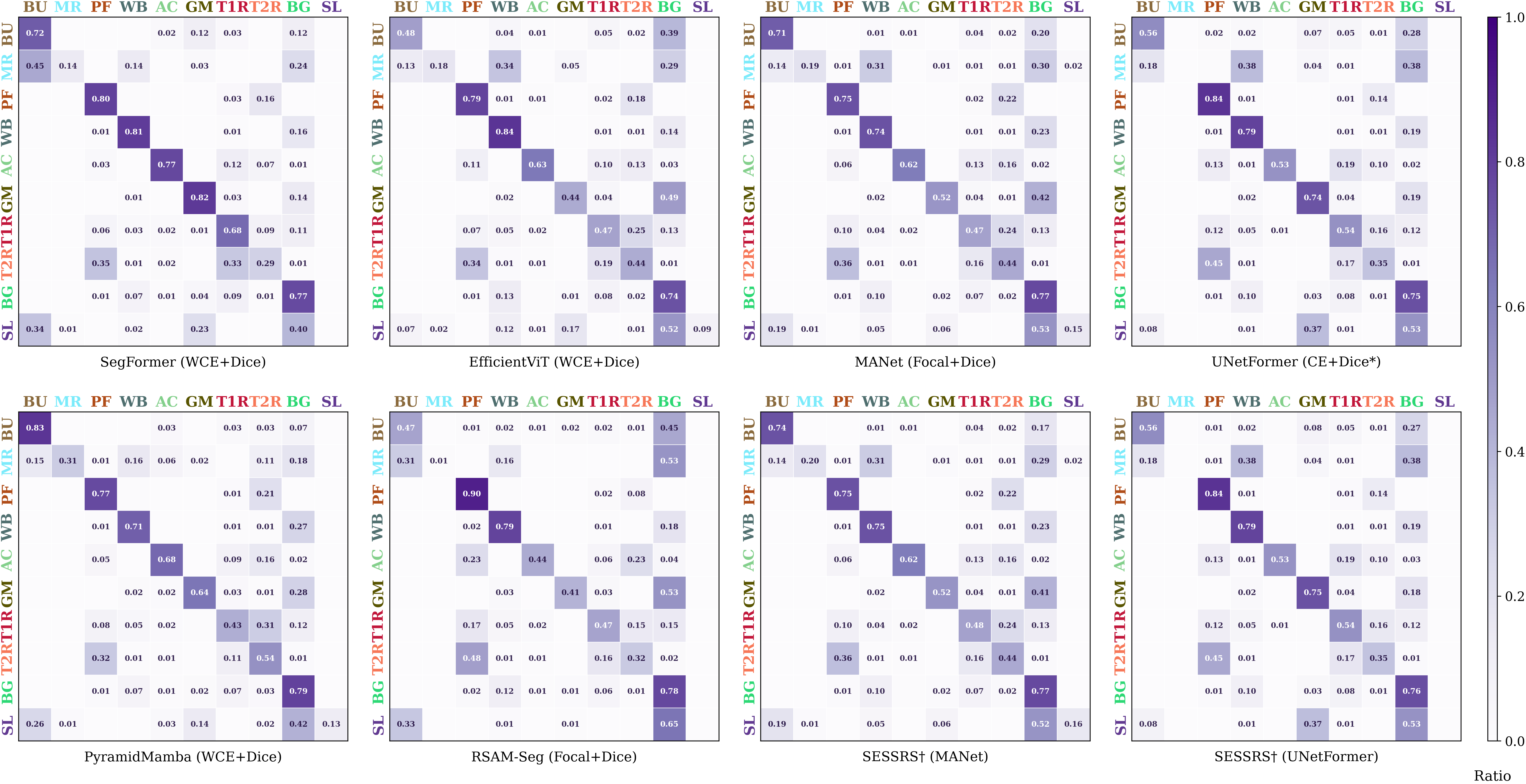}
\caption{Row-normalized pixel-level confusion matrices of eight selected methods on the ELDOR test split. Each subplot corresponds to one method and is computed over the 10 foreground classes present in the test set. Rows denote ground-truth classes and columns denote predicted classes. Darker diagonal cells indicate stronger class-specific segmentation performance, while darker off-diagonal cells highlight systematic confusion patterns. The class abbreviations are shown along the top and left using the same class colors as the other visualizations.}
\label{fig:app_selected_confusions}
\end{figure*}

\subsubsection{Confusion Matrix Analysis of Selected Methods}
\label{app:confusion_selected_seg}

Figure~\ref{fig:app_selected_confusions} provides a more direct view of the error structure of the eight selected methods. Although these methods differ in backbone, training loss, and model family, their confusion patterns are broadly similar. In most cases, the same groups of classes remain difficult, and the same pairs of categories are repeatedly confused. This suggests that the main errors are driven less by isolated model failures and more by the intrinsic visual and contextual ambiguity of the benchmark.

Several of these confusions are also semantically reasonable. Mining rafts and sluices are both small anthropogenic structures and are often confused with buildings. Mining rafts may additionally be confused with water bodies when they lie on ponds or rivers, and in some cases with bare ground or nearby background regions when they appear on land or at low spatial support. Sluices are often confused with bare ground, gravel mounds, or nearby disturbed regions, which is consistent with the fact that they typically occur along the edges of mining areas and are visually embedded in these surroundings.

The disturbance and recovery classes show another consistent source of error. Bare ground is frequently confused with water bodies and gravel mounds, which is understandable because these categories often share similar yellowish or light-brown tones in RGB imagery. Gravel mounds also tend to be confused with bare ground for the same reason. Type 1 regeneration is commonly confused with bare ground and Type 2 regeneration, reflecting its intermediate visual status between sparsely vegetated disturbed land and denser recovery. Type 2 regeneration is often confused with primary forest and Type 1 regeneration, which is also expected because it corresponds to a later recovery stage with taller and denser vegetation. Agricultural crops show a related pattern: they are often confused with Type 1 regeneration and, in some cases, Type 2 regeneration, since both contain low or moderately dense vegetation, while crops differ mainly through their stronger local structure and planting regularity.

The more dominant ecological classes are more stable, but their remaining errors also follow intuitive patterns. Water bodies are usually recognized well, yet some pixels are still confused with bare ground when turbid water and exposed mining surfaces have similar color. Primary forest is also relatively stable, but its main confusion is with Type 2 regeneration and, to a lesser extent, Type 1 regeneration, which reflects the gradual visual transition between intact forest and recovering vegetation. Overall, these matrices reinforce the main conclusion from the per-class analysis: most remaining errors arise from visually similar recovery and disturbance categories and from rare mining-related structures with limited spatial support.

\subsubsection{Qualitative Visualizations of Selected Methods}
\label{app:visual_selected_seg}

\begin{figure*}[t]
\centering
\includegraphics[width=\textwidth]{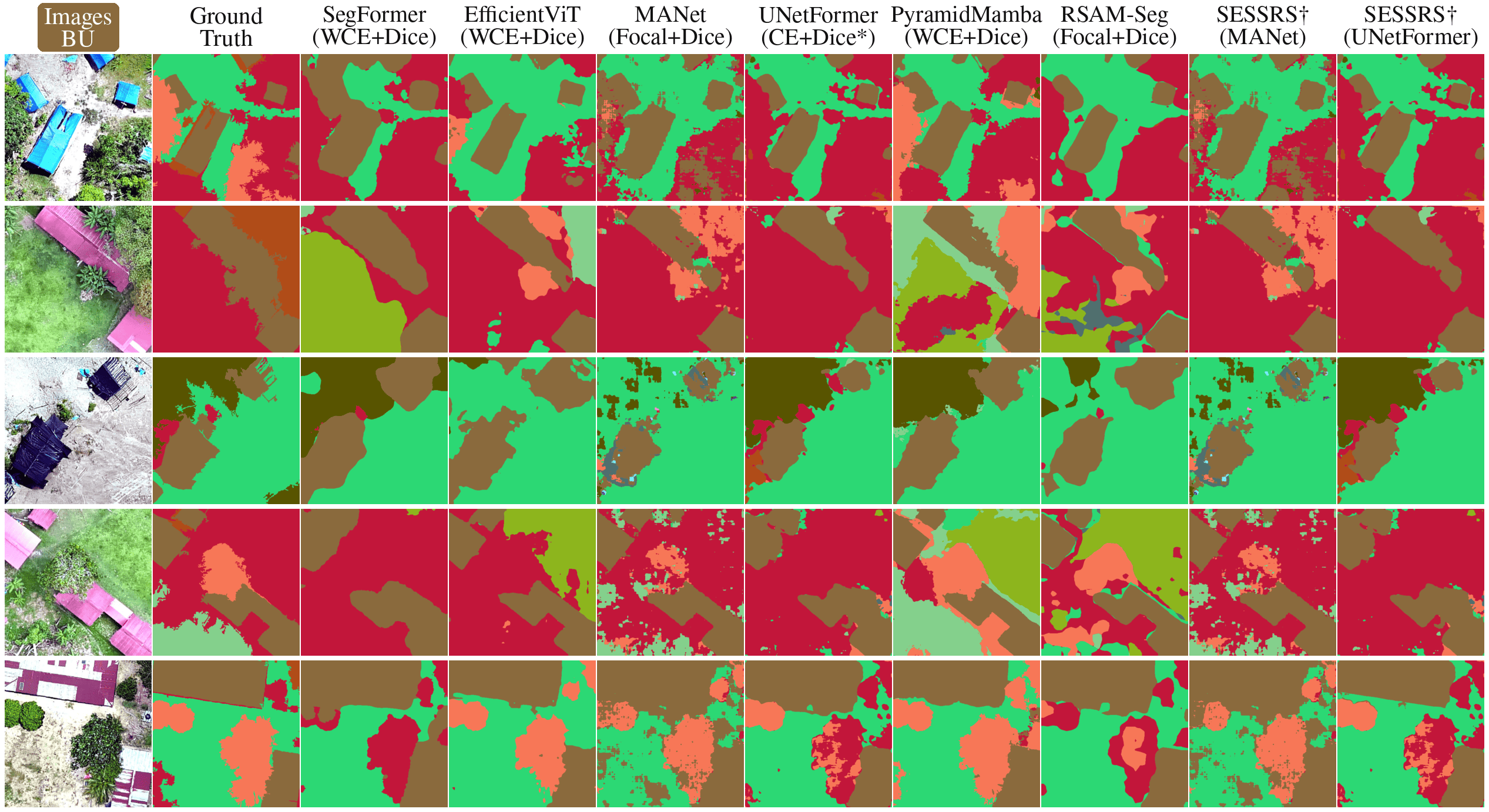}
\caption{Class-wise qualitative visualization for \textbf{BU} (Buildings) on the ELDOR test set. Each row shows one representative patch, and columns correspond to the original image, the ground truth, and predictions from the eight selected methods.}
\label{fig:qual_bu}
\end{figure*}

\begin{figure*}[t]
\centering
\includegraphics[width=\textwidth]{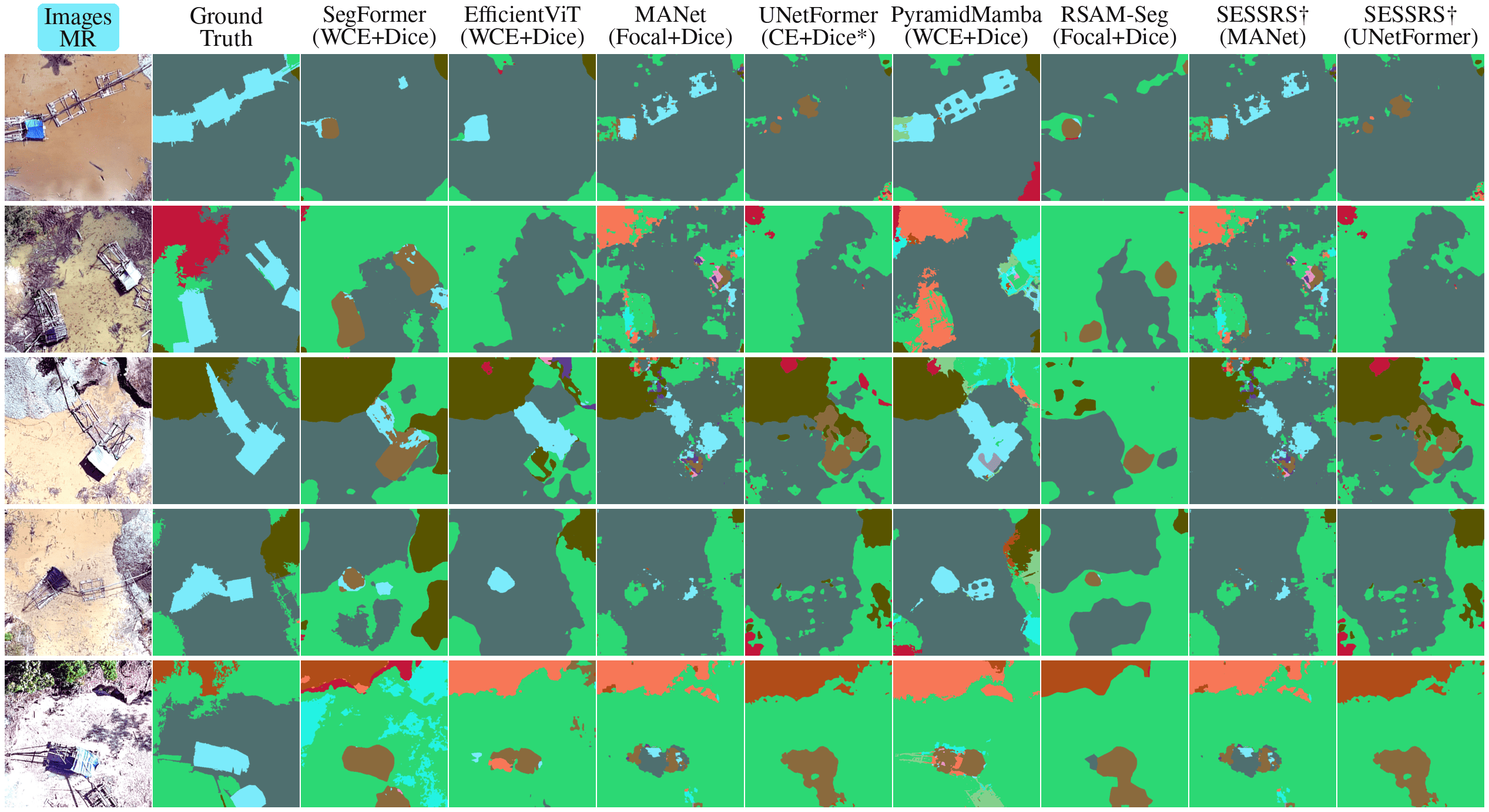}
\caption{Class-wise qualitative visualization for \textbf{MR} (Mining rafts) on the ELDOR test set. Each row shows one representative patch, and columns correspond to the original image, the ground truth, and predictions from the eight selected methods.}
\label{fig:qual_mr}
\end{figure*}

\begin{figure*}[t]
\centering
\includegraphics[width=\textwidth]{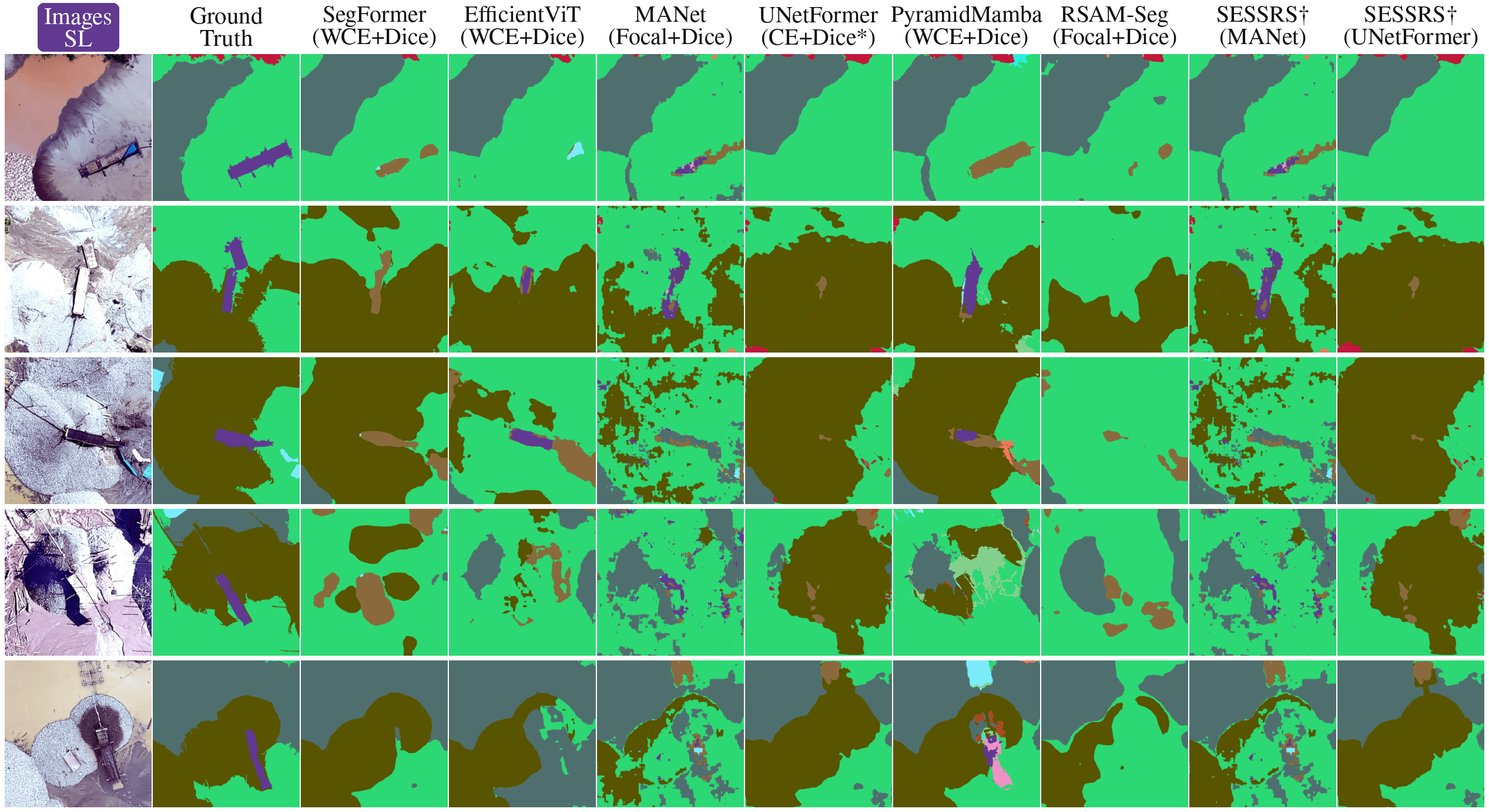}
\caption{Class-wise qualitative visualization for \textbf{SL} (Sluices) on the ELDOR test set. Each row shows one representative patch, and columns correspond to the original image, the ground truth, and predictions from the eight selected methods.}
\label{fig:qual_sl}
\end{figure*}

\begin{figure*}[t]
\centering
\includegraphics[width=\textwidth]{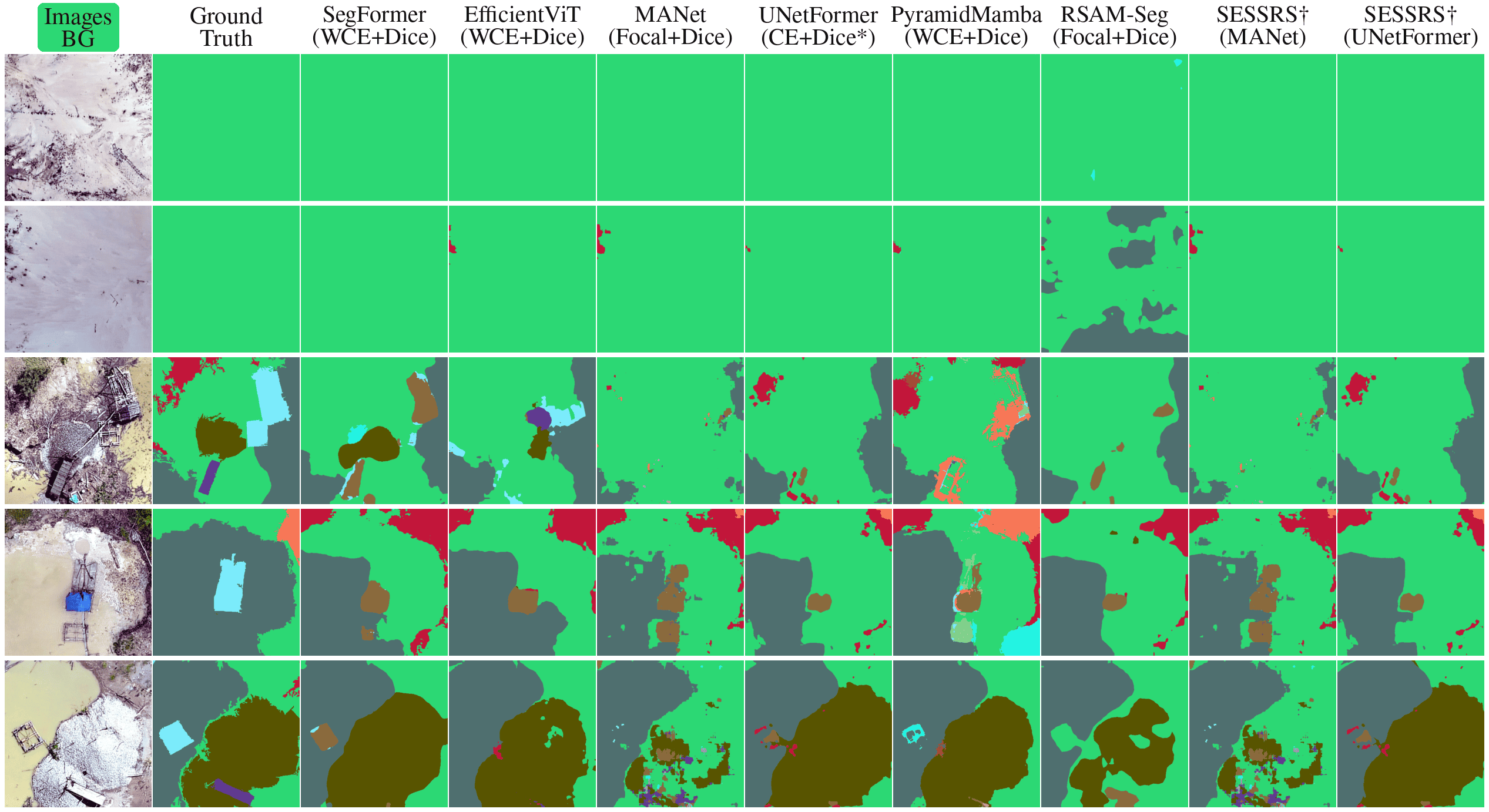}
\caption{Class-wise qualitative visualization for \textbf{BG} (Bare ground) on the ELDOR test set. Each row shows one representative patch, and columns correspond to the original image, the ground truth, and predictions from the eight selected methods.}
\label{fig:qual_bg}
\end{figure*}

\begin{figure*}[t]
\centering
\includegraphics[width=\textwidth]{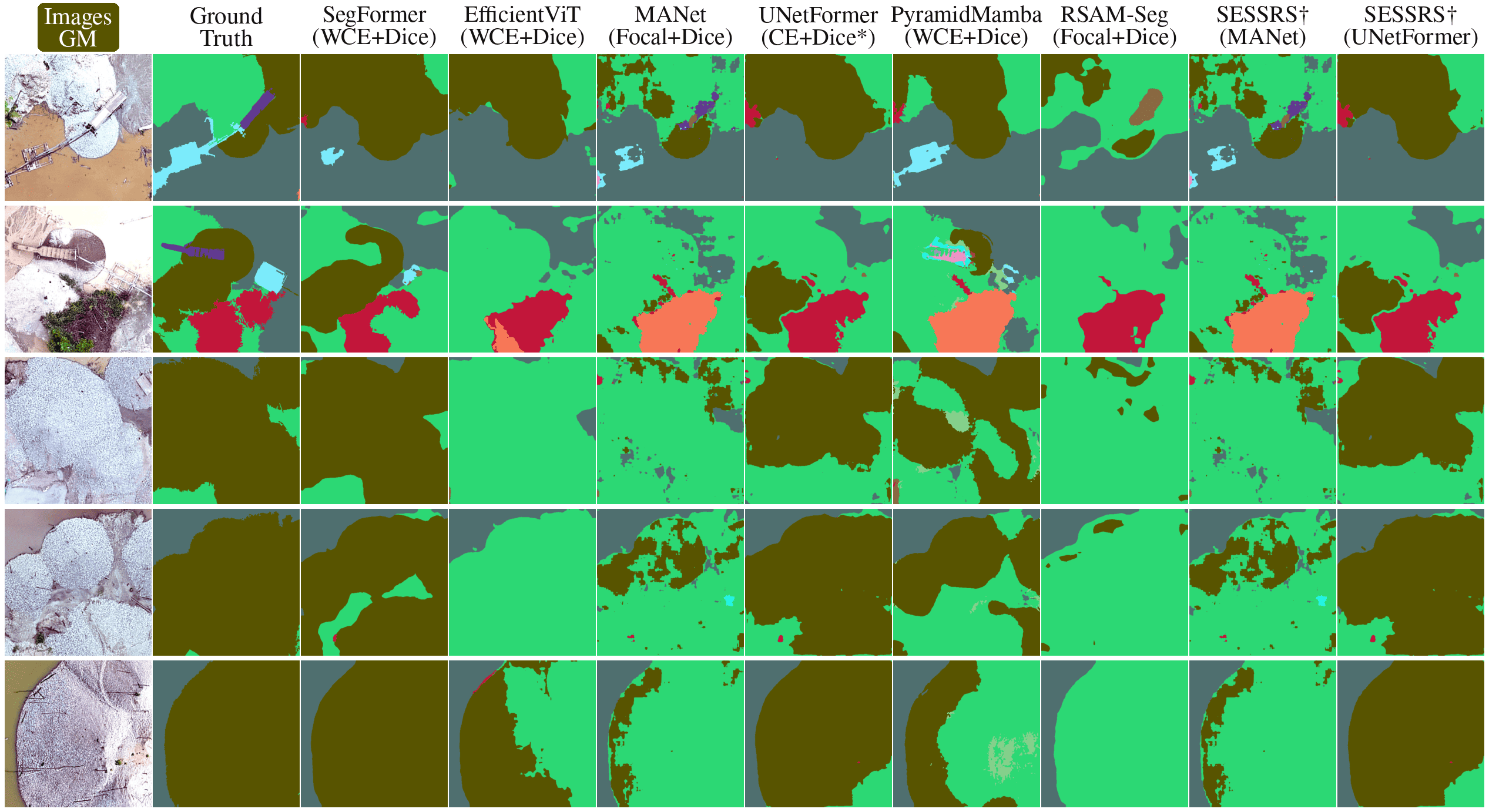}
\caption{Class-wise qualitative visualization for \textbf{GM} (Gravel mounds) on the ELDOR test set. Each row shows one representative patch, and columns correspond to the original image, the ground truth, and predictions from the eight selected methods.}
\label{fig:qual_gm}
\end{figure*}

\begin{figure*}[t]
\centering
\includegraphics[width=\textwidth]{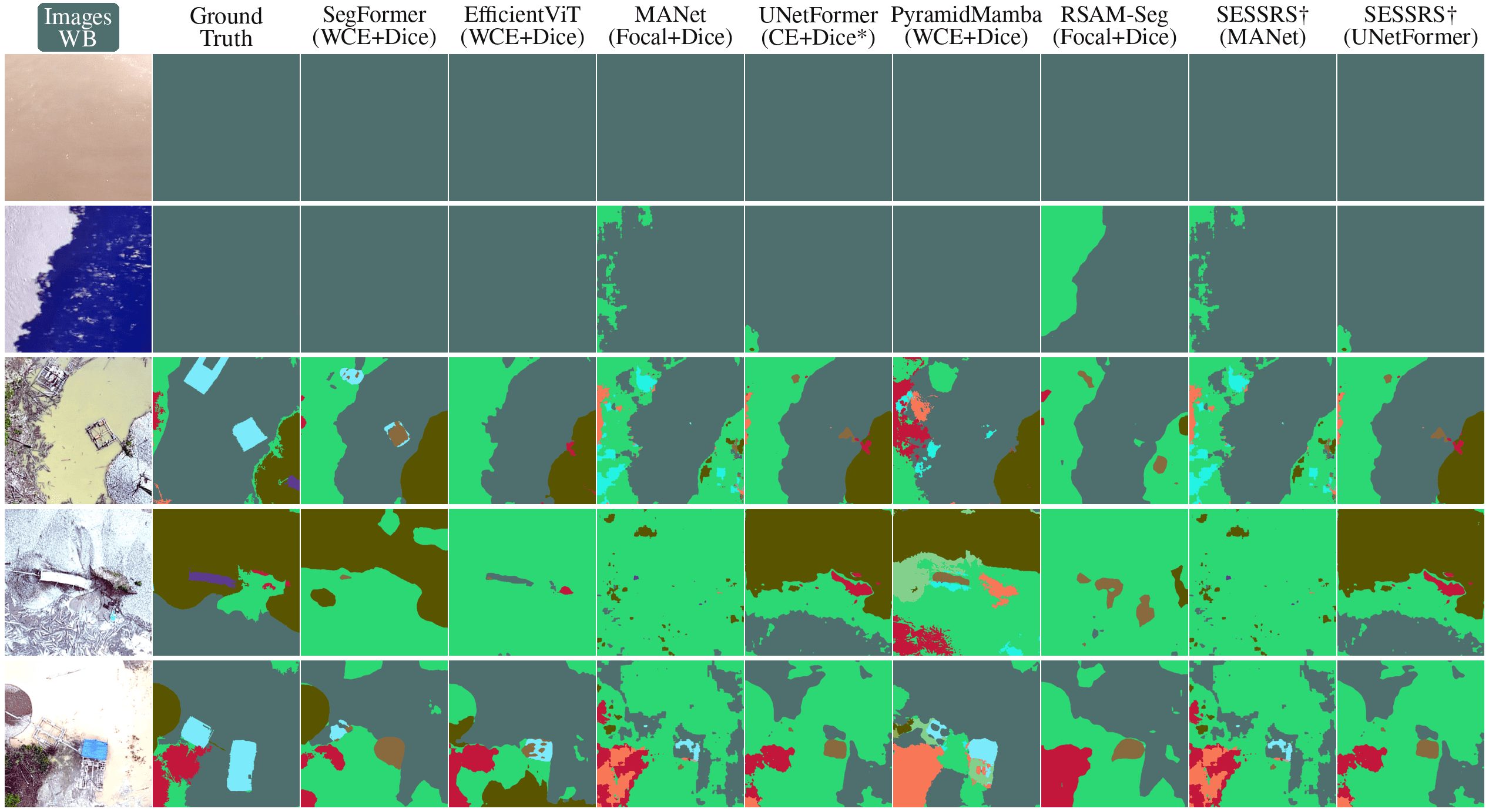}
\caption{Class-wise qualitative visualization for \textbf{WB} (Water bodies) on the ELDOR test set. Each row shows one representative patch, and columns correspond to the original image, the ground truth, and predictions from the eight selected methods.}
\label{fig:qual_wb}
\end{figure*}

\begin{figure*}[t]
\centering
\includegraphics[width=\textwidth]{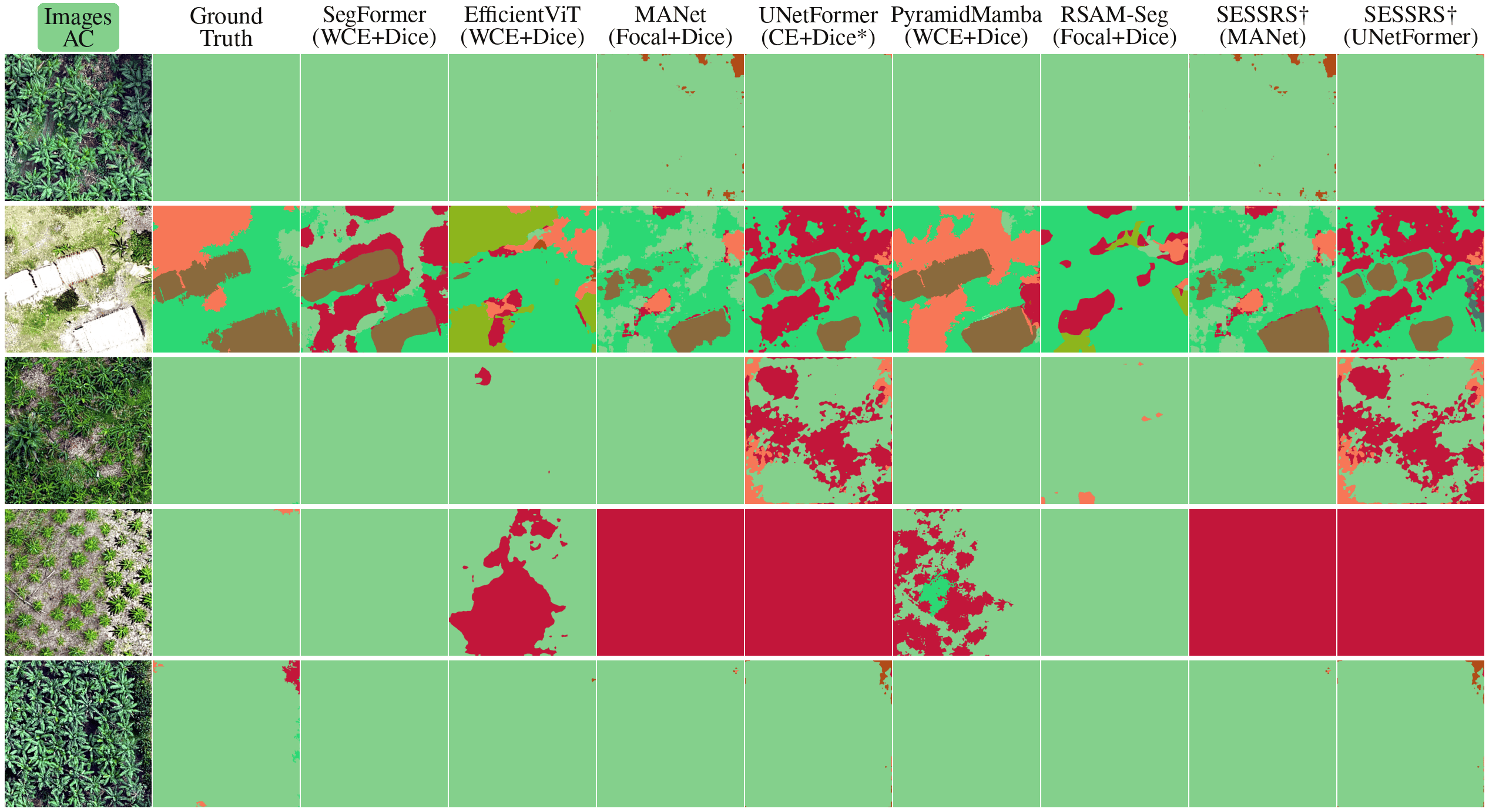}
\caption{Class-wise qualitative visualization for \textbf{AC} (Agricultural crops) on the ELDOR test set. Each row shows one representative patch, and columns correspond to the original image, the ground truth, and predictions from the eight selected methods.}
\label{fig:qual_ac}
\end{figure*}

\begin{figure*}[t]
\centering
\includegraphics[width=\textwidth]{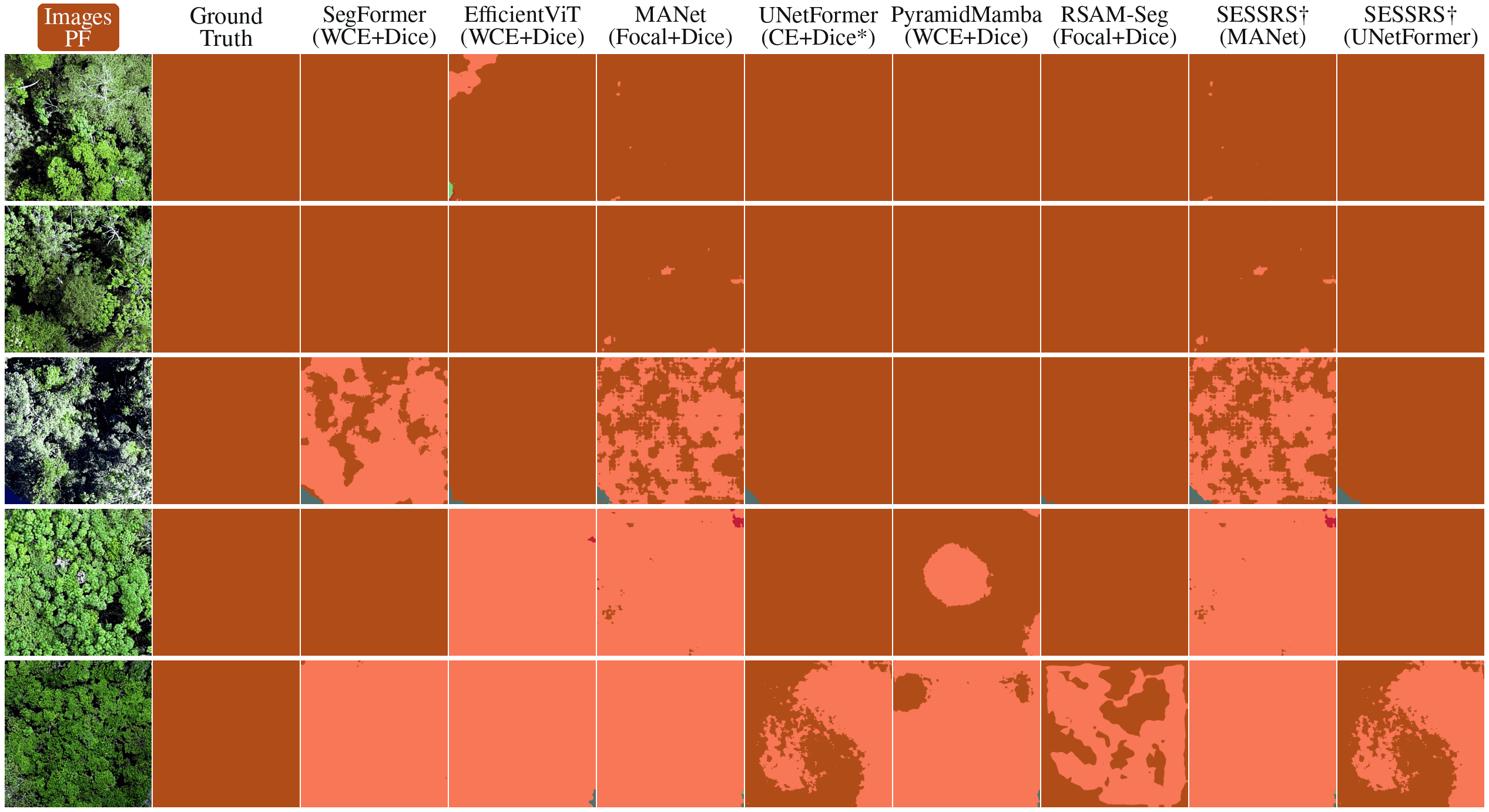}
\caption{Class-wise qualitative visualization for \textbf{PF} (Primary forests) on the ELDOR test set. Each row shows one representative patch, and columns correspond to the original image, the ground truth, and predictions from the eight selected methods.}
\label{fig:qual_pf}
\end{figure*}

\begin{figure*}[t]
\centering
\includegraphics[width=\textwidth]{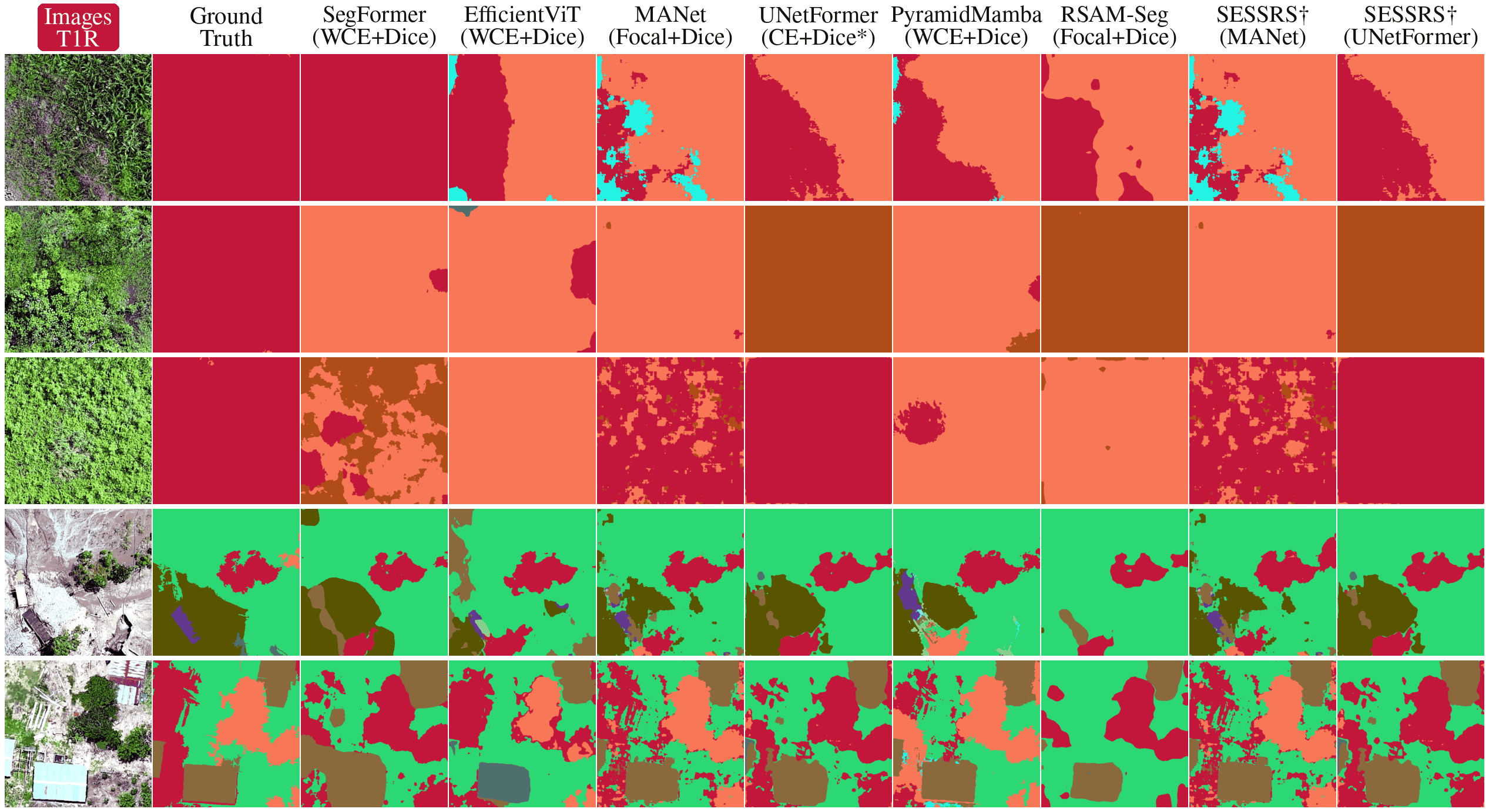}
\caption{Class-wise qualitative visualization for \textbf{T1R} (Type 1 regeneration) on the ELDOR test set. Each row shows one representative patch, and columns correspond to the original image, the ground truth, and predictions from the eight selected methods.}
\label{fig:qual_t1r}
\end{figure*}

\begin{figure*}[t]
\centering
\includegraphics[width=\textwidth]{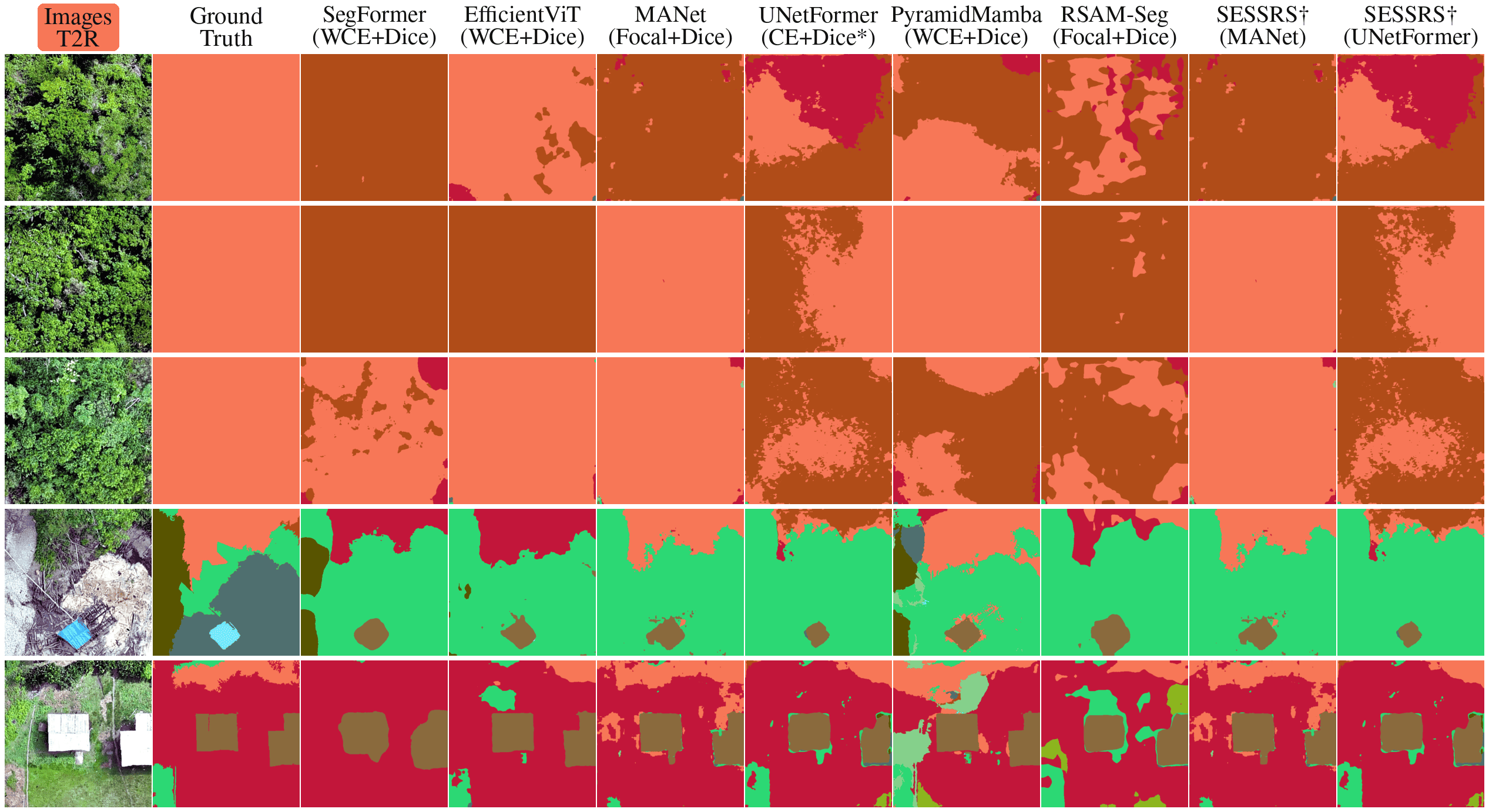}
\caption{Class-wise qualitative visualization for \textbf{T2R} (Type 2 regeneration) on the ELDOR test set. Each row shows one representative patch, and columns correspond to the original image, the ground truth, and predictions from the eight selected methods.}
\label{fig:qual_t2r}
\end{figure*}

Figures~\ref{fig:qual_bu}--\ref{fig:qual_t2r} provide class-wise qualitative comparisons of the eight selected methods on the ELDOR test set. Compared with the per-class heatmaps and confusion matrices above, these visual examples give a more direct view of how different methods behave on individual categories under varied contextual compositions. In particular, they make it easier to inspect whether a method recovers the full structure of a target, segments only part of it, or confuses it with visually similar surrounding classes.

Buildings in Figure~\ref{fig:qual_bu} provide a relatively stable example. Although this is a minor class, its visual structure is often more distinctive than that of many disturbance-related categories, so most methods can usually recover the main target, with the remaining errors appearing more often as boundary deviations than as complete misses. As also reflected in the confusion matrices in Figure~\ref{fig:app_selected_confusions}, true building pixels are most often predicted as bare ground. This is likely because buildings are small and their boundaries are easily mixed with the surrounding exposed surface, especially when only a limited number of pixels is available for learning. Overall, buildings remain more recoverable than other small mining-related targets, and they also serve as an important indicator of mining activity, since such man-made structures would not normally appear in otherwise intact forested areas.

Mining rafts in Figure~\ref{fig:qual_mr} are substantially more difficult. This is already clear from the confusion matrices in Figure~\ref{fig:app_selected_confusions} and from the per-class quantitative results in Figure~\ref{fig:app_perclass_full_heatmap}. True mining raft pixels are most often predicted as bare ground, water bodies, or buildings. The confusion with water bodies is visible in Figure~\ref{fig:qual_mr} (rows 1 and 4), which is expected because rafts are often located directly on water. The confusion with buildings can be seen especially in Figure~\ref{fig:qual_mr} (row 5) and reflects their similar man-made structure. The confusion with bare ground is also common because the local color and texture can be similar when the target is small. The qualitative comparison further suggests that WCE- and focal-loss-based models are better at recovering at least part of the raft, whereas CE-based models (e.g., UNetFormer and SESSRS$^\dagger$ (UNetFormer)) often fail to detect it.

Sluices in Figure~\ref{fig:qual_sl} are even more difficult than mining rafts. Their weakness is already evident in the earlier segmentation IoU and classification AP/F1 results in Figure~\ref{fig:app_perclass_full_heatmap}, and is further supported by the per-class AP and F1 values in Tables~\ref{tab:app_general_perclass_f1_ap} and~\ref{tab:app_vfm_perclass_f1_ap}, where this class remains one of the weakest and is zero for several methods. Consistent with Figure~\ref{fig:app_selected_confusions}, true sluice pixels are most often predicted as bare ground, buildings, or gravel mounds. The first confusion is especially clear in Figure~\ref{fig:qual_sl} (rows 4 and 5), where the class is extremely small and visually weak against the surrounding bare ground. The confusion with buildings can be seen in Figure~\ref{fig:qual_sl} (rows 2 and 3) and again follows from their similar man-made appearance. The confusion with gravel mounds is visible in Figure~\ref{fig:qual_sl} (row 2); this is also understandable because sluices are often built directly on or next to gravelly deposits generated by the washing process itself. In that example, the two classes also share very similar color, and apart from some boundary information, the distinction is difficult to recover reliably. Overall, this category appears to be so underrepresented in the training data that even WCE or focal loss provides only limited help, and several methods still miss it entirely.

Bare ground in Figure~\ref{fig:qual_bg} shows a different pattern because it is one of the major classes and is generally segmented more stably, as also suggested by Figure~\ref{fig:app_perclass_full_heatmap}. Even so, some ground-truth bare-ground pixels are still predicted as water bodies or Type 1 regeneration, and a smaller portion is predicted as Type 2 regeneration. The confusion with water bodies is visible in Figure~\ref{fig:qual_bg} (rows 2 and 4) and is mainly due to similar color and texture in turbid or shallow regions, where local appearance alone is often insufficient. The confusion with Type 1 regeneration is visible in Figure~\ref{fig:qual_bg} (row 4) and reflects the fact that this class corresponds to an early recovery stage, where sparse vegetation still lies visually close to exposed soil. A smaller amount of Type 2 regeneration is also absorbed into bare ground for a similar reason, since the regenerated vegetation is still not fully dense in some areas.

Gravel mounds in Figure~\ref{fig:qual_gm} show another characteristic type of confusion. As also reflected in Figure~\ref{fig:app_selected_confusions}, a large fraction of ground-truth gravel-mound pixels, nearly one third on average, is predicted as bare ground. Smaller portions are predicted as Type 1 regeneration and, in a few cases, water bodies. The first confusion is expected, since gravel mounds are usually embedded in exposed mining surfaces and often share similar color and texture with surrounding bare ground. This pattern is especially visible in Figure~\ref{fig:qual_gm} (row 5), where the patch is largely composed of gravel mounds, yet many methods still absorb substantial regions into bare ground, particularly on the right side where the local appearance is less distinctive. The confusion with Type 1 regeneration follows a related logic, since sparse vegetation can sometimes grow on or around these deposits; this can be observed in Figure~\ref{fig:qual_gm} (row 2), where part of the mound area is visually mixed with low vegetation. The main difficulty is not rarity but visual similarity, since gravel mounds often blend into the surrounding bare-ground region and are not always clearly separable from it.

Water bodies in Figure~\ref{fig:qual_wb} are one of the more stable major classes, and most methods segment them reasonably well. As reflected in Figure~\ref{fig:app_selected_confusions}, the main confusion is with bare ground. In the qualitative examples, this is visible in Figure~\ref{fig:qual_wb} (row 2), where the water region on the right shows a clear appearance change caused by surrounding tree shadows. Under this strong local contrast, and without sufficient global context, some methods tend to absorb part of the region into bare ground. Aside from such cases, the remaining errors are more concentrated around the exact boundary between water and bare ground, which appears to be the more persistent challenge for this class.

Agricultural crops in Figure~\ref{fig:qual_ac} are characterized less by local color than by their more regular global structure. This pattern is clear in Figure~\ref{fig:qual_ac} (rows 1, 4, and 5), where the crop regions appear relatively organized. Once that structure is weakened, the class becomes much easier to confuse with surrounding vegetation. In Figure~\ref{fig:qual_ac} (rows 3 and 4), for example, substantial bare-ground gaps appear inside the crop region, and methods that fail to capture the overall layout tend to push these areas toward Type 1 regeneration or bare ground. In Figure~\ref{fig:qual_ac} (row 2), a small region near the upper-right crop boundary is predicted as Type 2 regeneration because the visible area is too limited to reveal the larger agricultural pattern. Confusion with primary forest also appears when the vegetation is especially dense and the crop-specific arrangement is not fully captured.

Primary forests in Figure~\ref{fig:qual_pf} are generally segmented well, but their main confusion is with Type 2 regeneration, with only a smaller amount absorbed into Type 1 regeneration. This pattern is especially visible in Figure~\ref{fig:qual_pf} (rows 3, 4, and 5), where different methods assign substantial forest regions to Type 2 regeneration. The reason is that Type 2 regeneration has already reached a relatively advanced recovery stage and can appear visually very close to primary forest. In many cases, the distinction depends less on the local appearance of the trees themselves than on broader contextual cues, such as whether the surrounding area shows signs of past mining disturbance. This can also be seen in the darker regions of Figure~\ref{fig:qual_pf} (rows 3 and 5), where some methods tend to push parts of the forest toward Type 2 regeneration. From the image alone, these darker areas may appear less mature or less uniform, even though such appearance can also be caused by shadows, terrain variation, or other environmental factors rather than actual recovery status. Overall, this class shows that even a major category can remain difficult to separate from nearby recovery classes when the semantic distinction depends partly on ecological history rather than only on immediate visual evidence.

Type 1 regeneration in Figure~\ref{fig:qual_t1r} is mainly confused with Type 2 regeneration, bare ground, and primary forest. This pattern follows the role of the class itself, since it lies visually between exposed disturbed land and more developed vegetation. Confusion with Type 2 regeneration appears when the vegetation is interpreted as slightly denser or taller than it actually is, while confusion with bare ground occurs when the cover is too sparse and the exposed soil becomes more visually dominant. Confusion with primary forest is less frequent but still present, as seen in Figure~\ref{fig:qual_t1r} (row 3), where some regions are assigned to primary forest once the vegetation appears more continuous or mature. A small amount of confusion also occurs with water bodies and agricultural crops. The former is mostly associated with darker local appearance, while the latter comes from a similar mixture of vegetation and exposed soil, despite the fact that agricultural crops are typically more regular in structure.

Type 2 regeneration in Figure~\ref{fig:qual_t2r} is most often confused with primary forest, followed by Type 1 regeneration. The first confusion is the more dominant one and is already consistent with the discussion above for primary forest. Once recovery reaches this stage, the vegetation is already relatively tall and dense, so its local appearance can become very close to intact forest. This is especially visible in Figure~\ref{fig:qual_t2r} (rows 2 and 3), where different methods assign substantial Type 2 regions to primary forest. Confusion with Type 1 regeneration is also common, as shown in Figure~\ref{fig:qual_t2r} (rows 1, 3, and 4), and mainly reflects the gradual transition between the two recovery stages. In practice, the difference is often one of vegetation maturity and density rather than a sharp visual boundary, so this category naturally lies between earlier regeneration and fully developed forest.

%


\subsection{Multi-label Classification}
\label{app:mlc}

This appendix reports the complete results of direct multi-label classification for both general and remote-sensing-specific methods. We first present the overall comparison in terms of recognition accuracy and efficiency, and then provide a more detailed per-class analysis in the following subsection. Compared with the main-text summary, these appendix results offer a fuller view of how different architectures trade off balanced label prediction, overall recognition quality, and computational cost.

\subsubsection{Overall Multi-label Classification Performance}
\label{app:mlc_overall}

\begin{table*}[t]
\centering
\scriptsize
\caption{Overall results of direct multi-label classification methods. General methods are highlighted in blue and remote-sensing-specific methods in green. All general methods use ResNet-101 backbones unless otherwise noted; HSVLT uses ConvNeXt. All remote-sensing-specific methods use ResNet-50 backbones unless otherwise noted; GRN uses ResNet-18. Best values are boldfaced; lower is better for Params, GFLOPs, and Latency.}
\label{tab:app_mlc_overall}
\setlength{\tabcolsep}{2.2pt}
\renewcommand{\arraystretch}{1.05}
\resizebox{\textwidth}{!}{%
\begin{tabular}{llcccccccccccc}
\toprule
Model & Loss & mAP & CP & CR & CF1 & OP & OR & OF1 & Macro-F1 & Sample-F1 & Params & GFLOPs & Latency \\
\midrule

\rowcolor{blue!10}
\multicolumn{14}{c}{\textbf{General methods}} \\

ML-GCN     & BCE          & 0.6499 & 0.7464 & 0.4469 & 0.5591 & 0.8143 & 0.6942 & 0.7495 & 0.5743 & 0.7670 & \cellcolor{blue!10}\textbf{42.50}  & 82.17  & 6.44 \\
ADD-GCN    & BCE          & 0.6328 & 0.6591 & 0.5470 & 0.5978 & 0.7762 & 0.7363 & 0.7557 & 0.6479 & 0.7612 & 47.82  & 83.34  & 6.83 \\
CSRA       & BCE          & 0.6392 & 0.7760 & 0.4863 & 0.5979 & 0.7769 & 0.6830 & 0.7269 & 0.5489 & 0.7244 & 42.61  & 82.22  & 5.71 \\
C-Tran     & BCE          & 0.6947 & 0.5879 & 0.5480 & 0.5672 & 0.7642 & 0.7195 & 0.7412 & \cellcolor{blue!10}\textbf{0.7019} & 0.7408 & 120.18 & 62.22  & 7.38 \\
TDRG       & BCE          & \cellcolor{blue!10}\textbf{0.6983} & \cellcolor{blue!10}\textbf{0.7997} & 0.5820 & \cellcolor{blue!10}\textbf{0.6737} & 0.7733 & 0.7550 & \cellcolor{blue!10}\textbf{0.7640} & 0.6161 & \cellcolor{blue!10}\textbf{0.7712} & 107.14 & 348.60 & 11.15 \\
Q2L        & BCE          & 0.6730 & 0.7061 & 0.6207 & 0.6607 & 0.7486 & 0.7336 & 0.7410 & 0.6526 & 0.7470 & 193.53 & 100.56 & 12.90 \\
CPCL       & BCE+Triplet  & 0.6121 & 0.6122 & 0.5570 & 0.5833 & 0.7418 & 0.7213 & 0.7314 & 0.6335 & 0.7302 & 50.86  & \cellcolor{blue!10}\textbf{40.92}  & 5.58 \\
DualCoOp   & CE           & 0.6016 & 0.4792 & 0.6655 & 0.5572 & 0.5206 & 0.8697 & 0.6514 & 0.5036 & 0.6679 & 94.41  & 204.13 & 30.52 \\
SALGL      & ASL          & 0.5681 & 0.5209 & 0.6052 & 0.5599 & 0.6271 & 0.8064 & 0.7055 & 0.5479 & 0.7157 & 119.15 & 53.14  & 7.14 \\
HSVLT      & ASL          & 0.4902 & 0.4557 & 0.5505 & 0.4986 & 0.5849 & 0.8678 & 0.6988 & 0.5308 & 0.7160 & 154.33 & 57.32  & 37.36 \\
ML-Decoder & ASL          & 0.6140 & 0.5431 & 0.6830 & 0.6051 & 0.6305 & \cellcolor{blue!10}\textbf{0.8864} & 0.7369 & 0.5862 & 0.7502 & 49.61  & 83.71  & 6.08 \\
SGRE       & ASL          & 0.6279 & 0.4983 & 0.7178 & 0.5882 & 0.5873 & 0.8574 & 0.6971 & 0.5755 & 0.7051 & 42.53  & 82.16  & 5.29 \\
DDA-MLIC   & ASL          & 0.6086 & 0.5009 & \cellcolor{blue!10}\textbf{0.7437} & 0.5986 & 0.5948 & 0.8626 & 0.7041 & 0.5859 & 0.7095 & 43.02  & 82.16  & 5.27 \\
DRL        & ResampleLoss & 0.5971 & 0.7145 & 0.3255 & 0.4473 & \cellcolor{blue!10}\textbf{0.8783} & 0.4037 & 0.5531 & 0.4302 & 0.5306 & 42.53  & 82.16  & 5.29 \\
SpliceMix  & BCE          & 0.6602 & 0.7394 & 0.5406 & 0.6246 & 0.7707 & 0.7247 & 0.7470 & 0.6003 & 0.7504 & 42.53  & 82.16  & \cellcolor{blue!10}\textbf{5.25} \\

\midrule
\rowcolor{green!10}
\multicolumn{14}{c}{\textbf{Remote-sensing-specific methods}} \\

RelationNet & BCE & \cellcolor{green!10}\textbf{0.6142} & 0.6102 & 0.5958 & 0.6029 & 0.7400 & 0.6988 & 0.7188 & 0.5961 & 0.7216 & 25.59 & 86.21 & 2.67 \\
GRN         & BCE & 0.6011 & 0.5647 & \cellcolor{green!10}\textbf{0.6423} & 0.6010 & 0.7231 & 0.7144 & 0.7187 & 0.5859 & 0.7185 & \cellcolor{green!10}\textbf{11.24} & \cellcolor{green!10}\textbf{19.07} & \cellcolor{green!10}\textbf{1.15} \\
RSMLC       & BCE & 0.5926 & 0.5640 & 0.5007 & 0.5305 & \cellcolor{green!10}\textbf{0.7886} & 0.6716 & 0.7254 & 0.5794 & \cellcolor{green!10}\textbf{0.7476} & 23.54 & 43.16 & 2.70 \\
SIGNA       & BCE & 0.6112 & 0.6198 & 0.5931 & \cellcolor{green!10}\textbf{0.6062} & 0.7260 & \cellcolor{green!10}\textbf{0.7343} & \cellcolor{green!10}\textbf{0.7301} & \cellcolor{green!10}\textbf{0.6006} & 0.7339 & 43.30 & 43.21 & 8.67 \\
IDMN        & SAT & 0.5366 & \cellcolor{green!10}\textbf{0.6400} & 0.3900 & 0.4847 & 0.7765 & 0.6368 & 0.6997 & 0.5638 & 0.7123 & 23.54 & 43.17 & 3.46 \\

\bottomrule
\end{tabular}%
}
\end{table*}

Tables~\ref{tab:app_mlc_overall} reports the complete overall comparison for direct multi-label classification methods across both general and remote-sensing-specific groups. Several consistent patterns can be observed.

Within the general-method group, TDRG gives the strongest overall recognition performance, achieving the best mAP (0.6983), CP (0.7997), CF1 (0.6737), OF1 (0.7640), and Sample-F1 (0.7712). This suggests that transformer-based label modeling with explicit relational structure is particularly effective on this benchmark. Other methods peak on different metrics: C-Tran achieves the highest Macro-F1 (0.7019), DDA-MLIC gives the highest CR (0.7437), ML-Decoder attains the highest OR (0.8864), and DRL yields the highest OP (0.8783). From the efficiency perspective, ML-GCN uses the fewest parameters (42.50M), CPCL has the lowest GFLOPs (40.92), and SpliceMix gives the lowest latency (5.25 ms) in this group.

Within the remote-sensing-specific group, RelationNet achieves the best mAP (0.6142), while SIGNA gives the strongest balanced recognition results, including the best CF1 (0.6062), OR (0.7343), OF1 (0.7301), and Macro-F1 (0.6006). GRN achieves the highest CR (0.6423) and is also the most efficient model in this group, with the fewest parameters (11.24M), lowest GFLOPs (19.07), and lowest latency (1.15 ms). RSMLC attains the highest OP (0.7886) and Sample-F1 (0.7476), whereas IDMN achieves the highest CP (0.6400) but remains weaker on recall-sensitive metrics.

We also evaluated additional backbone-matched variants for the missing settings, including ResNet-50 and ResNet-101 counterparts where applicable, although these results are omitted here for brevity. In general, changing the backbone size did not substantially alter the overall performance trends or the relative ranking of methods. This suggests that the main performance differences on ELDOR are driven more by the recognition architecture itself than by backbone scale alone. Overall, direct multi-label classification on ELDOR is currently led by strong general recognition architectures in terms of absolute recognition quality, while remote-sensing-specific methods remain attractive when efficiency is an important consideration. This also suggests that image-level recognition on this benchmark benefits strongly from general-purpose multi-label architectures with stronger label-correlation modeling, even though remote-sensing-specific methods still offer competitive trade-offs in speed and model size.

\subsubsection{Per-class Multi-label Classification Performance}
\label{app:mlc_perclass}

\begin{table*}[t]
\centering
\scriptsize
\caption{Per-class results of direct multi-label classification methods. The upper block reports AP and the lower block reports F1. General methods are highlighted in blue and remote-sensing-specific methods in green. Only the 10 classes present in the test split are shown. Best values are boldfaced separately within each method family for each metric block.}
\label{tab:app_mlc_classwise}
\setlength{\tabcolsep}{2.8pt}
\renewcommand{\arraystretch}{1.08}
\resizebox{\textwidth}{!}{%
\begin{tabular}{llcccccccccc}
\toprule
\multirow{2}{*}{Model} &
\multirow{2}{*}{Loss} &
\multirow{2}{*}{Buildings} &
Mining & Primary & Water & Agricultural & Gravel & Type 1 & Type 2 & Bare & \multirow{2}{*}{Sluices} \\
& &
& rafts & forests & bodies & crops & mounds & regeneration & regeneration & ground & \\
\midrule

\rowcolor{gray!15}
\multicolumn{12}{c}{\textbf{Per-class AP}} \\

\midrule

\rowcolor{blue!10}
\multicolumn{12}{c}{\textbf{General methods}} \\

ML-GCN     & BCE          & 0.6518 & 0.3972 & \cellcolor{blue!10}\textbf{0.9157} & \cellcolor{blue!10}\textbf{0.9526} & 0.5271 & 0.7156 & 0.7577 & 0.5868 & 0.8629 & 0.1316 \\
ADD-GCN    & BCE          & 0.6438 & 0.2401 & 0.9060 & 0.9417 & 0.5182 & 0.7649 & 0.7710 & 0.5303 & 0.8334 & 0.1784 \\
CSRA       & BCE          & 0.5840 & 0.4032 & 0.9043 & 0.9363 & 0.4587 & 0.7259 & 0.7305 & \cellcolor{blue!10}\textbf{0.6097} & 0.8003 & 0.2388 \\
C-Tran     & BCE          & 0.7665 & \cellcolor{blue!10}\textbf{0.5402} & 0.9139 & 0.9433 & 0.4036 & 0.8055 & \cellcolor{blue!10}\textbf{0.7870} & 0.5527 & \cellcolor{blue!10}\textbf{0.8963} & 0.3380 \\
TDRG       & BCE          & \cellcolor{blue!10}\textbf{0.8081} & 0.4405 & 0.9037 & 0.9493 & 0.5551 & \cellcolor{blue!10}\textbf{0.8251} & 0.7727 & 0.5499 & 0.8894 & 0.2893 \\
Q2L        & BCE          & 0.7621 & 0.5230 & 0.8776 & 0.9233 & 0.4184 & 0.7527 & 0.7568 & 0.4954 & 0.8821 & \cellcolor{blue!10}\textbf{0.3383} \\
CPCL       & BCE+Triplet  & 0.6537 & 0.2176 & 0.9020 & 0.9332 & 0.5316 & 0.7638 & 0.7375 & 0.4870 & 0.8117 & 0.0826 \\
DualCoOp   & CE           & 0.6967 & 0.2636 & 0.8920 & 0.8792 & 0.4722 & 0.6753 & 0.7148 & 0.5369 & 0.8022 & 0.0828 \\
SALGL      & ASL          & 0.4442 & 0.1354 & 0.8955 & 0.8973 & 0.4352 & 0.7386 & 0.7185 & 0.5311 & 0.7865 & 0.0987 \\
HSVLT      & ASL          & 0.4690 & 0.1751 & 0.9007 & 0.8995 & 0.0951 & 0.3635 & 0.7527 & 0.5352 & 0.6881 & 0.0231 \\
ML-Decoder & ASL          & 0.5391 & 0.1563 & 0.9141 & 0.9414 & 0.4882 & 0.7998 & 0.7718 & 0.5492 & 0.8342 & 0.1461 \\
SGRE       & ASL          & 0.6641 & 0.2706 & 0.8978 & 0.9342 & 0.5061 & 0.7647 & 0.6874 & 0.5040 & 0.8628 & 0.1870 \\
DDA-MLIC   & ASL          & 0.6677 & 0.2398 & 0.8787 & 0.9386 & \cellcolor{blue!10}\textbf{0.5563} & 0.7723 & 0.5994 & 0.4344 & 0.8470 & 0.1515 \\
DRL        & ResampleLoss & 0.6060 & 0.1834 & 0.8819 & 0.9274 & 0.5408 & 0.7281 & 0.6609 & 0.5431 & 0.7832 & 0.1168 \\
SpliceMix  & BCE          & 0.6486 & 0.4316 & 0.8966 & 0.9388 & 0.4820 & 0.7897 & 0.7623 & 0.5611 & 0.8612 & 0.2303 \\

\rowcolor{green!10}
\multicolumn{12}{c}{\textbf{Remote-sensing-specific methods}} \\

RelationNet & BCE & 0.6186 & 0.1844 & 0.8844 & 0.9288 & \cellcolor{green!10}\textbf{0.5612} & 0.7305 & 0.7175 & 0.5382 & 0.8166 & \cellcolor{green!10}\textbf{0.1622} \\
GRN         & BCE & 0.5957 & \cellcolor{green!10}\textbf{0.2532} & 0.8947 & 0.9154 & 0.4889 & 0.7529 & 0.7010 & 0.5328 & 0.7525 & 0.1238 \\
RSMLC       & BCE & 0.5502 & 0.2087 & 0.9129 & \cellcolor{green!10}\textbf{0.9416} & 0.4526 & 0.6331 & \cellcolor{green!10}\textbf{0.7803} & 0.5125 & \cellcolor{green!10}\textbf{0.8368} & 0.0977 \\
SIGNA       & BCE & \cellcolor{green!10}\textbf{0.6552} & 0.2055 & 0.9067 & 0.9188 & 0.5357 & \cellcolor{green!10}\textbf{0.7596} & 0.7159 & 0.5598 & 0.7443 & 0.1103 \\
IDMN        & SAT & 0.3974 & 0.0096 & \cellcolor{green!10}\textbf{0.9169} & 0.9204 & 0.4012 & 0.5896 & 0.7716 & \cellcolor{green!10}\textbf{0.5781} & 0.7768 & 0.0038 \\

\midrule
\rowcolor{gray!15}
\multicolumn{12}{c}{\textbf{Per-class F1}} \\

\midrule

\rowcolor{blue!10}
\multicolumn{12}{c}{\textbf{General methods}} \\

ML-GCN     & BCE          & 0.5690 & 0.0541 & 0.8321 & 0.8424 & 0.4146 & 0.4206 & 0.6858 & 0.5678 & 0.7827 & 0.0000 \\
ADD-GCN    & BCE          & 0.6339 & 0.1988 & 0.8322 & 0.8466 & 0.5805 & 0.7059 & \cellcolor{blue!10}\textbf{0.7084} & 0.5481 & 0.7763 & 0.0000 \\
CSRA       & BCE          & 0.5554 & 0.3462 & 0.8153 & 0.8102 & 0.4689 & 0.5034 & 0.5799 & \cellcolor{blue!10}\textbf{0.6400} & 0.7490 & 0.0211 \\
C-Tran     & BCE          & 0.7178 & 0.0000 & 0.7668 & 0.8484 & 0.4511 & \cellcolor{blue!10}\textbf{0.7361} & 0.6658 & 0.6218 & 0.8072 & 0.0000 \\
TDRG       & BCE          & 0.7356 & 0.2874 & 0.8161 & 0.8553 & \cellcolor{blue!10}\textbf{0.6037} & 0.7157 & 0.7013 & 0.6129 & \cellcolor{blue!10}\textbf{0.8115} & 0.0211 \\
Q2L        & BCE          & \cellcolor{blue!10}\textbf{0.7604} & \cellcolor{blue!10}\textbf{0.5069} & 0.8088 & 0.8348 & 0.5220 & 0.6741 & 0.6943 & 0.5388 & 0.7809 & \cellcolor{blue!10}\textbf{0.4046} \\
CPCL       & BCE+Triplet  & 0.6473 & 0.1190 & 0.8065 & 0.8160 & 0.5901 & 0.7057 & 0.7040 & 0.5667 & 0.7463 & 0.0000 \\
DualCoOp   & CE           & 0.6616 & 0.2069 & 0.7999 & 0.7905 & 0.1823 & 0.5334 & 0.6280 & 0.5757 & 0.6378 & 0.0200 \\
SALGL      & ASL          & 0.4455 & 0.1970 & 0.8158 & 0.8069 & 0.4767 & 0.6306 & 0.6876 & 0.5672 & 0.7072 & 0.1444 \\
HSVLT      & ASL          & 0.4816 & 0.2386 & 0.8204 & 0.8088 & 0.0890 & 0.3741 & 0.6817 & 0.5713 & 0.7122 & 0.0000 \\
ML-Decoder & ASL          & 0.4701 & 0.2905 & \cellcolor{blue!10}\textbf{0.8536} & \cellcolor{blue!10}\textbf{0.8667} & 0.5409 & 0.6116 & 0.6850 & 0.6034 & 0.7264 & 0.2137 \\
SGRE       & ASL          & 0.6134 & 0.4011 & 0.8085 & 0.8463 & 0.3647 & 0.5771 & 0.6137 & 0.5587 & 0.7582 & 0.2130 \\
DDA-MLIC   & ASL          & 0.6375 & 0.3607 & 0.7990 & 0.8443 & 0.3645 & 0.6625 & 0.6535 & 0.5490 & 0.7405 & 0.2480 \\
DRL        & ResampleLoss & 0.5968 & 0.1471 & 0.7018 & 0.7173 & 0.5398 & 0.5330 & 0.2537 & 0.1671 & 0.5558 & 0.0901 \\
SpliceMix  & BCE          & 0.5702 & 0.3529 & 0.8086 & 0.8530 & 0.5048 & 0.7206 & 0.6434 & 0.6027 & 0.7772 & 0.1695 \\

\rowcolor{green!10}
\multicolumn{12}{c}{\textbf{Remote-sensing-specific methods}} \\

RelationNet & BCE & 0.5135 & 0.2898 & 0.7880 & 0.7947 & \cellcolor{green!10}\textbf{0.6073} & 0.6851 & 0.6775 & 0.6004 & 0.7451 & \cellcolor{green!10}\textbf{0.2599} \\
GRN         & BCE & 0.4798 & \cellcolor{green!10}\textbf{0.3901} & 0.8098 & 0.7896 & 0.4685 & \cellcolor{green!10}\textbf{0.7013} & 0.6349 & 0.6056 & 0.7366 & 0.2426 \\
RSMLC       & BCE & 0.5741 & 0.0575 & \cellcolor{green!10}\textbf{0.8261} & \cellcolor{green!10}\textbf{0.8224} & 0.4482 & 0.5785 & 0.6427 & 0.5024 & \cellcolor{green!10}\textbf{0.7628} & 0.0000 \\
SIGNA       & BCE & \cellcolor{green!10}\textbf{0.6345} & 0.3274 & 0.8258 & 0.8168 & 0.5547 & 0.6805 & 0.6805 & 0.5870 & 0.7469 & 0.1519 \\
IDMN        & SAT & 0.2735 & 0.0000 & 0.7799 & 0.7332 & 0.3481 & 0.3595 & \cellcolor{green!10}\textbf{0.6942} & \cellcolor{green!10}\textbf{0.6058} & 0.7161 & 0.0000 \\

\bottomrule
\end{tabular}%
}
\end{table*}

\begin{table*}[t]
\centering
\scriptsize
\caption{Per-class precision and recall of direct multi-label classification methods. The upper block reports precision and the lower block reports recall. General methods are highlighted in blue and remote-sensing-specific methods in green. Only the 10 classes present in the test split are shown. Best values are boldfaced separately within each method family for each metric block.}
\label{tab:app_mlc_classwise_pr}
\setlength{\tabcolsep}{2.8pt}
\renewcommand{\arraystretch}{1.08}
\resizebox{\textwidth}{!}{%
\begin{tabular}{llcccccccccc}
\toprule
\multirow{2}{*}{Model} &
\multirow{2}{*}{Loss} &
\multirow{2}{*}{Buildings} &
Mining & Primary & Water & Agricultural & Gravel & Type 1 & Type 2 & Bare & \multirow{2}{*}{Sluices} \\
& &
& rafts & forests & bodies & crops & mounds & regeneration & regeneration & ground & \\
\midrule

\rowcolor{gray!15}
\multicolumn{12}{c}{\textbf{Per-class Precision}} \\

\midrule

\rowcolor{blue!10}
\multicolumn{12}{c}{\textbf{General methods}} \\

ML-GCN     & BCE          & 0.8816 & 0.8000 & 0.8722 & 0.9434 & 0.8430 & \cellcolor{blue!10}\textbf{0.9554} & 0.7270 & 0.6150 & 0.8267 & 0.0000 \\
ADD-GCN    & BCE          & 0.7565 & 0.6071 & 0.8439 & 0.9166 & 0.6757 & 0.7374 & 0.7157 & 0.5615 & 0.7769 & 0.0000 \\
CSRA       & BCE          & 0.6595 & 0.5538 & 0.8800 & 0.9451 & 0.7452 & 0.9168 & 0.7630 & 0.5896 & 0.7069 & \cellcolor{blue!10}\textbf{1.0000} \\
C-Tran     & BCE          & 0.7027 & 0.0000 & 0.9085 & 0.9018 & 0.4897 & 0.8013 & 0.7748 & 0.5315 & 0.7687 & 0.0000 \\
TDRG       & BCE          & 0.8007 & \cellcolor{blue!10}\textbf{0.8065} & 0.8651 & 0.9299 & 0.6472 & 0.9044 & 0.7229 & 0.5432 & 0.7776 & 0.0000 \\
Q2L        & BCE          & 0.7752 & 0.7432 & 0.8454 & 0.9113 & 0.5904 & 0.8031 & 0.7419 & 0.5104 & 0.6975 & 0.4430 \\
CPCL       & BCE+Triplet  & 0.6656 & 0.4000 & 0.8705 & 0.9261 & 0.6491 & 0.6798 & 0.6951 & 0.5214 & 0.7148 & 0.0000 \\
DualCoOp   & CE           & 0.6385 & 0.5806 & 0.7104 & 0.7742 & 0.1054 & 0.4074 & 0.4877 & 0.4329 & 0.4876 & 0.1667 \\
SALGL      & ASL          & 0.4427 & 0.3333 & 0.8342 & 0.8040 & 0.4753 & 0.5378 & 0.5983 & 0.4483 & 0.5837 & 0.1512 \\
HSVLT      & ASL          & 0.6327 & 0.2394 & 0.7489 & 0.7468 & 0.2235 & 0.4094 & 0.5513 & 0.4235 & 0.5810 & 0.0000 \\
ML-Decoder & ASL          & 0.4013 & 0.3571 & 0.8024 & 0.8654 & 0.5351 & 0.4743 & 0.5604 & 0.4644 & 0.5919 & 0.3784 \\
SGRE       & ASL          & 0.5718 & 0.3303 & 0.7806 & 0.8260 & 0.2569 & 0.4340 & 0.4706 & 0.4205 & 0.6521 & 0.2400 \\
DDA-MLIC   & ASL          & 0.6355 & 0.2678 & 0.7467 & 0.7900 & 0.2539 & 0.5797 & 0.5415 & 0.4138 & 0.6143 & 0.1661 \\
DRL        & ResampleLoss & 0.6885 & 0.2459 & \cellcolor{blue!10}\textbf{0.9256} & \cellcolor{blue!10}\textbf{0.9742} & \cellcolor{blue!10}\textbf{0.8469} & 0.9107 & \cellcolor{blue!10}\textbf{0.8129} & \cellcolor{blue!10}\textbf{0.6163} & \cellcolor{blue!10}\textbf{0.8300} & 0.2941 \\
SpliceMix  & BCE          & \cellcolor{blue!10}\textbf{0.9167} & 0.7500 & 0.8673 & 0.8994 & 0.6302 & 0.8477 & 0.7769 & 0.5382 & 0.7510 & 0.4167 \\

\rowcolor{green!10}
\multicolumn{12}{c}{\textbf{Remote-sensing-specific methods}} \\

RelationNet & BCE & 0.4101 & 0.2929 & 0.8465 & 0.9395 & \cellcolor{green!10}\textbf{0.5966} & \cellcolor{green!10}\textbf{0.7425} & 0.7083 & 0.5566 & 0.7317 & \cellcolor{green!10}\textbf{0.2771} \\
GRN         & BCE & 0.3587 & 0.3015 & 0.8447 & 0.9132 & 0.4250 & 0.7080 & 0.6879 & 0.5501 & 0.6826 & 0.1754 \\
RSMLC       & BCE & 0.5653 & 0.1613 & 0.8663 & \cellcolor{green!10}\textbf{0.9424} & 0.4446 & 0.4938 & \cellcolor{green!10}\textbf{0.8046} & \cellcolor{green!10}\textbf{0.5679} & \cellcolor{green!10}\textbf{0.7942} & 0.0000 \\
SIGNA       & BCE & \cellcolor{green!10}\textbf{0.6523} & \cellcolor{green!10}\textbf{0.4458} & 0.8689 & 0.9167 & 0.5863 & 0.6126 & 0.6698 & 0.5644 & 0.6936 & 0.1875 \\
IDMN        & SAT & 0.5113 & 0.0000 & \cellcolor{green!10}\textbf{0.9167} & 0.8829 & 0.4547 & 0.5483 & 0.6163 & 0.5437 & 0.6559 & 0.0000 \\

\midrule
\rowcolor{gray!15}
\multicolumn{12}{c}{\textbf{Per-class Recall}} \\

\midrule

\rowcolor{blue!10}
\multicolumn{12}{c}{\textbf{General methods}} \\

ML-GCN     & BCE          & 0.4201 & 0.0280 & 0.7956 & 0.7610 & 0.2749 & 0.2696 & 0.6490 & 0.5274 & 0.7432 & 0.0000 \\
ADD-GCN    & BCE          & 0.5455 & 0.1189 & 0.8207 & 0.7865 & 0.5088 & 0.6770 & 0.7012 & 0.5354 & 0.7758 & 0.0000 \\
CSRA       & BCE          & 0.4796 & 0.2517 & 0.7595 & 0.7089 & 0.3421 & 0.3469 & 0.4676 & 0.6998 & 0.7964 & 0.0106 \\
C-Tran     & BCE          & 0.7335 & 0.0000 & 0.6633 & 0.8010 & 0.4181 & 0.6808 & 0.5838 & 0.7490 & 0.8499 & 0.0000 \\
TDRG       & BCE          & 0.6803 & 0.1748 & 0.7723 & 0.7917 & 0.5658 & 0.5922 & 0.6810 & 0.7031 & 0.8486 & 0.0106 \\
Q2L        & BCE          & \cellcolor{blue!10}\textbf{0.7461} & 0.3846 & 0.7753 & 0.7701 & 0.4678 & 0.5808 & 0.6525 & 0.5707 & 0.8871 & 0.3723 \\
CPCL       & BCE+Triplet  & 0.6301 & 0.0699 & 0.7512 & 0.7293 & 0.5409 & 0.7337 & 0.7132 & 0.6207 & 0.7807 & 0.0000 \\
DualCoOp   & CE           & 0.6865 & 0.1259 & \cellcolor{blue!10}\textbf{0.9151} & 0.8075 & \cellcolor{blue!10}\textbf{0.6754} & 0.7719 & 0.8813 & 0.8592 & 0.9216 & 0.0106 \\
SALGL      & ASL          & 0.4483 & 0.1399 & 0.7981 & 0.8098 & 0.4781 & 0.7623 & 0.8084 & 0.7719 & 0.8971 & 0.1383 \\
HSVLT      & ASL          & 0.3887 & 0.2378 & 0.9070 & 0.8820 & 0.0556 & 0.3444 & \cellcolor{blue!10}\textbf{0.8929} & \cellcolor{blue!10}\textbf{0.8772} & 0.9197 & 0.0000 \\
ML-Decoder & ASL          & 0.5674 & 0.2448 & 0.9118 & 0.8680 & 0.5468 & 0.8606 & 0.8809 & 0.8610 & \cellcolor{blue!10}\textbf{0.9401} & 0.1489 \\
SGRE       & ASL          & 0.6614 & 0.5105 & 0.8384 & 0.8676 & 0.6287 & \cellcolor{blue!10}\textbf{0.8610} & 0.8819 & 0.8322 & 0.9053 & 0.1915 \\
DDA-MLIC   & ASL          & 0.6395 & \cellcolor{blue!10}\textbf{0.5524} & 0.8591 & \cellcolor{blue!10}\textbf{0.9067} & 0.6462 & 0.7728 & 0.8238 & 0.8153 & 0.9318 & \cellcolor{blue!10}\textbf{0.4894} \\
DRL        & ResampleLoss & 0.5266 & 0.1049 & 0.5651 & 0.5676 & 0.3962 & 0.3767 & 0.1503 & 0.0966 & 0.4178 & 0.0532 \\
SpliceMix  & BCE          & 0.4138 & 0.2308 & 0.7573 & 0.8111 & 0.4211 & 0.6266 & 0.5491 & 0.6848 & 0.8053 & 0.1064 \\

\rowcolor{green!10}
\multicolumn{12}{c}{\textbf{Remote-sensing-specific methods}} \\

RelationNet & BCE & 0.6865 & 0.2867 & 0.7371 & 0.6886 & \cellcolor{green!10}\textbf{0.6184} & 0.6359 & 0.6493 & 0.6518 & 0.7590 & 0.2447 \\
GRN         & BCE & \cellcolor{green!10}\textbf{0.7241} & \cellcolor{green!10}\textbf{0.5524} & 0.7776 & 0.6955 & 0.5219 & 0.6947 & 0.5895 & \cellcolor{green!10}\textbf{0.6735} & 0.7998 & \cellcolor{green!10}\textbf{0.3936} \\
RSMLC       & BCE & 0.5831 & 0.0350 & \cellcolor{green!10}\textbf{0.7896} & 0.7296 & 0.4518 & 0.6984 & 0.5350 & 0.4505 & 0.7339 & 0.0000 \\
SIGNA       & BCE & 0.6176 & 0.2587 & 0.7868 & \cellcolor{green!10}\textbf{0.7365} & 0.5263 & \cellcolor{green!10}\textbf{0.7652} & 0.6916 & 0.6114 & \cellcolor{green!10}\textbf{0.8091} & 0.1277 \\
IDMN        & SAT & 0.3542 & 0.0000 & 0.6653 & 0.7195 & 0.3377 & 0.3574 & \cellcolor{green!10}\textbf{0.6982} & 0.6443 & 0.7716 & 0.0000 \\

\bottomrule
\end{tabular}%
}
\end{table*}

Tables~\ref{tab:app_mlc_classwise} and~\ref{tab:app_mlc_classwise_pr} report the per-class results of direct multi-label classification. Table~\ref{tab:app_mlc_classwise} summarizes class-wise AP and F1, while Table~\ref{tab:app_mlc_classwise_pr} further decomposes performance into precision and recall. We restrict the comparison to the 10 classes present in the test split. Together, these tables provide a more detailed view of which categories are reliably recognized and which remain difficult, especially for rare mining-related classes.

A clear pattern across all four blocks is that dominant ecological categories such as primary forests, water bodies, and bare ground are much easier than rare mining-related categories such as mining rafts and sluices. This is consistent across both general and remote-sensing-specific methods. Agricultural crops, Type~1 regeneration, and Type~2 regeneration occupy an intermediate position: they are generally more difficult than the dominant categories, but still more stable than the smallest mining-related targets.

In Table~\ref{tab:app_mlc_classwise}, the strongest AP and F1 results are again distributed across several methods within each family rather than concentrated in a single model. Among the general methods, TDRG gives the best AP on buildings and gravel mounds, C-Tran is strongest on mining rafts, Type~1 regeneration, and bare ground, ML-GCN performs best on primary forests and water bodies, DDA-MLIC gives the best agricultural-crop AP, CSRA gives the best Type~2 regeneration AP, and Q2L yields the best sluice AP. In the F1 block, Q2L gives the best F1 on buildings, mining rafts, and sluices, ML-Decoder performs best on primary forests and water bodies, TDRG gives the best agricultural-crop and bare-ground F1, C-Tran is strongest on gravel mounds, ADD-GCN performs best on Type~1 regeneration, and CSRA gives the best Type~2 regeneration result. Among the remote-sensing-specific methods, SIGNA gives the best AP on buildings and gravel mounds, GRN is strongest on mining rafts, IDMN performs best on primary forests and Type~2 regeneration, RSMLC gives the best AP on water bodies, Type~1 regeneration, and bare ground, and RelationNet performs best on agricultural crops and sluices. In the F1 block, SIGNA performs best on buildings, GRN gives the best mining-raft and gravel-mound F1, RSMLC is strongest on primary forests, water bodies, and bare ground, RelationNet gives the best agricultural-crop and sluice F1, and IDMN performs best on both regeneration classes.

Table~\ref{tab:app_mlc_classwise_pr} further clarifies how these results arise. In the precision block, several methods achieve very strong precision on dominant categories, but precision on mining rafts and sluices is much less stable. Within the general methods, DRL gives the best precision on primary forests, water bodies, agricultural crops, Type~1 regeneration, Type~2 regeneration, and bare ground, SpliceMix gives the best building precision, TDRG is strongest on mining rafts, ML-GCN performs best on gravel mounds, and CSRA gives the highest sluice precision. Within the remote-sensing-specific family, SIGNA gives the best precision on buildings and mining rafts, IDMN performs best on primary forests, RSMLC is strongest on water bodies, Type~1 regeneration, Type~2 regeneration, and bare ground, and RelationNet performs best on agricultural crops, gravel mounds, and sluices. In the recall block, the strongest methods differ. Among the general methods, Q2L gives the best building recall, DDA-MLIC is strongest on mining rafts, water bodies, and sluices, DualCoOp performs best on primary forests and agricultural crops, SGRE gives the best gravel-mound recall, HSVLT performs best on both regeneration classes, and ML-Decoder gives the strongest bare-ground recall. Among the remote-sensing-specific methods, GRN gives the best recall on buildings, mining rafts, Type~2 regeneration, and sluices, RSMLC is strongest on primary forests, SIGNA performs best on water bodies, gravel mounds, and bare ground, RelationNet gives the best agricultural-crop recall, and IDMN gives the best Type~1 regeneration recall.

Overall, the per-class results reinforce the same conclusion as the aggregate metrics: direct multi-label classification is already reliable on dominant and visually consistent categories, but rare mining-related classes remain substantially harder and continue to show larger variation across architectures. The additional precision--recall breakdown also shows that some methods are more conservative and achieve higher precision on dominant classes, while others obtain stronger recall on ambiguous or visually overlapping categories. For ELDOR, both aspects are important, since practical mining monitoring requires not only accurate recognition of dominant land-cover classes but also reliable recovery of small and operationally important mining-related targets.

\subsubsection{Class-wise Grad-CAM Visualization}
\label{app:mlc_gradcam}

\begin{figure*}[t]
\centering
\includegraphics[width=\textwidth]{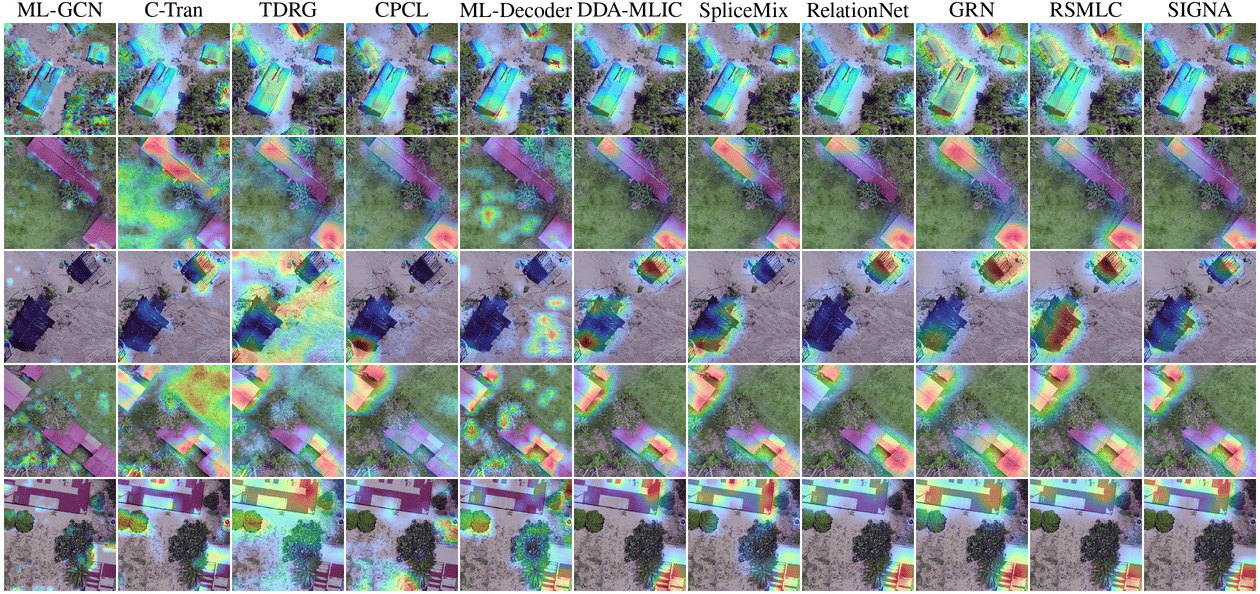}
\caption{Grad-CAM visualizations for \textbf{Buildings} (BU). Each column corresponds to one method and each row to one representative test patch.}
\label{fig:gradcam_buildings}
\end{figure*}

\begin{figure*}[t]
\centering
\includegraphics[width=\textwidth]{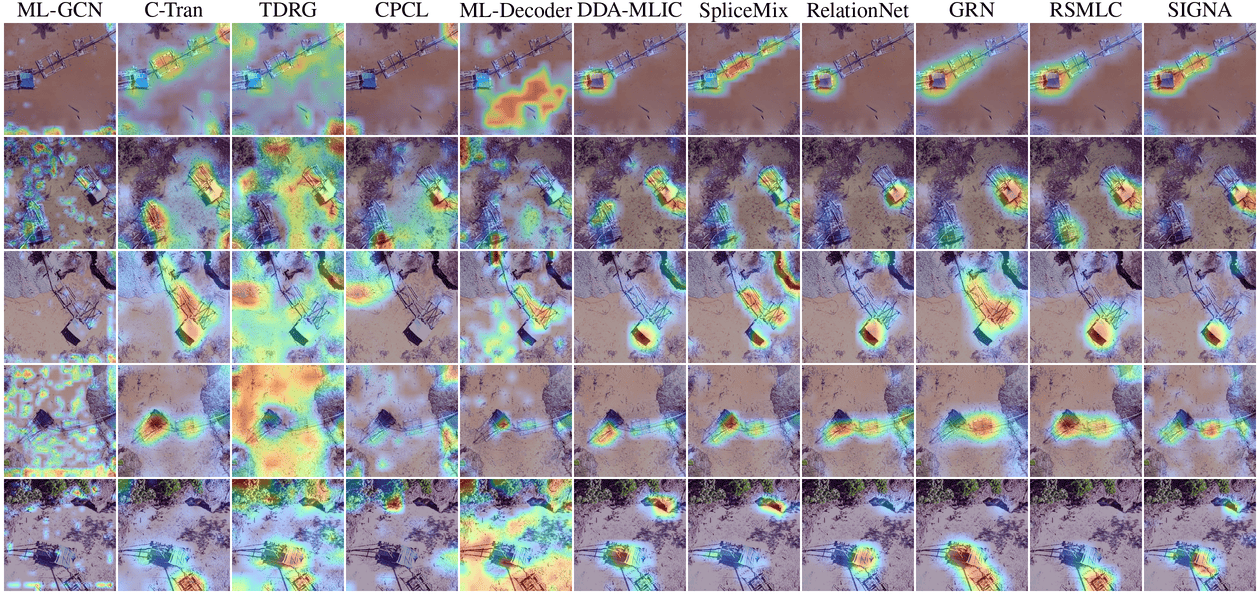}
\caption{Grad-CAM visualizations for \textbf{Mining rafts} (MR). Each column corresponds to one method and each row to one representative test patch.}
\label{fig:gradcam_mining_rafts}
\end{figure*}

\begin{figure*}[t]
\centering
\includegraphics[width=\textwidth]{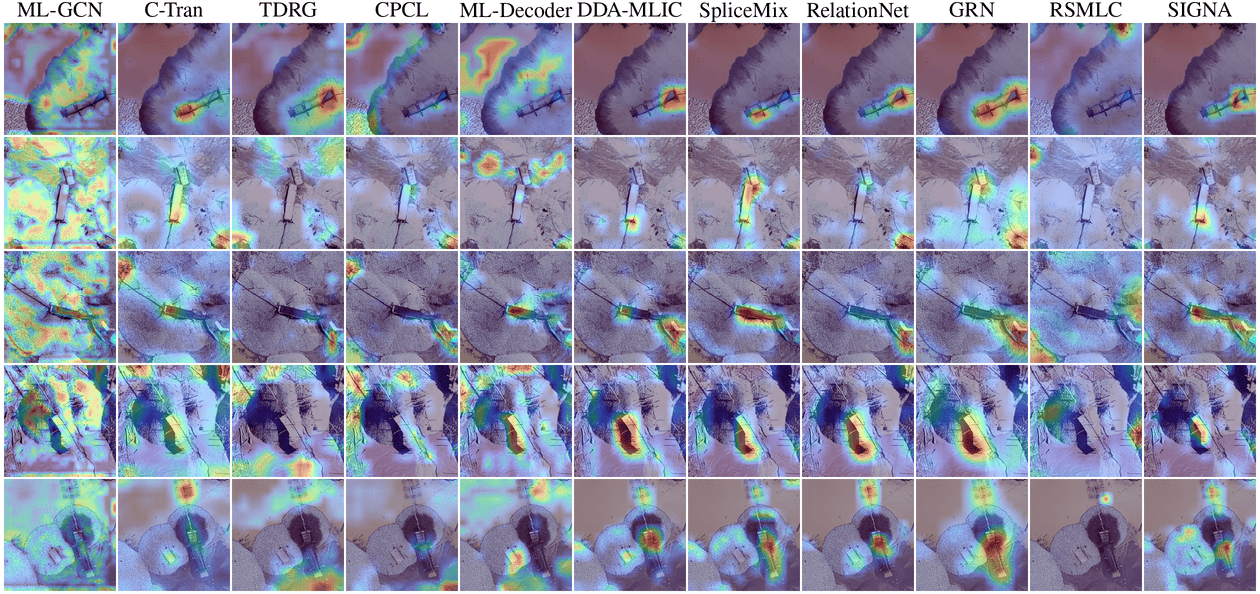}
\caption{Grad-CAM visualizations for \textbf{Sluices} (SL). Each column corresponds to one method and each row to one representative test patch.}
\label{fig:gradcam_sluices}
\end{figure*}

\begin{figure*}[t]
\centering
\includegraphics[width=\textwidth]{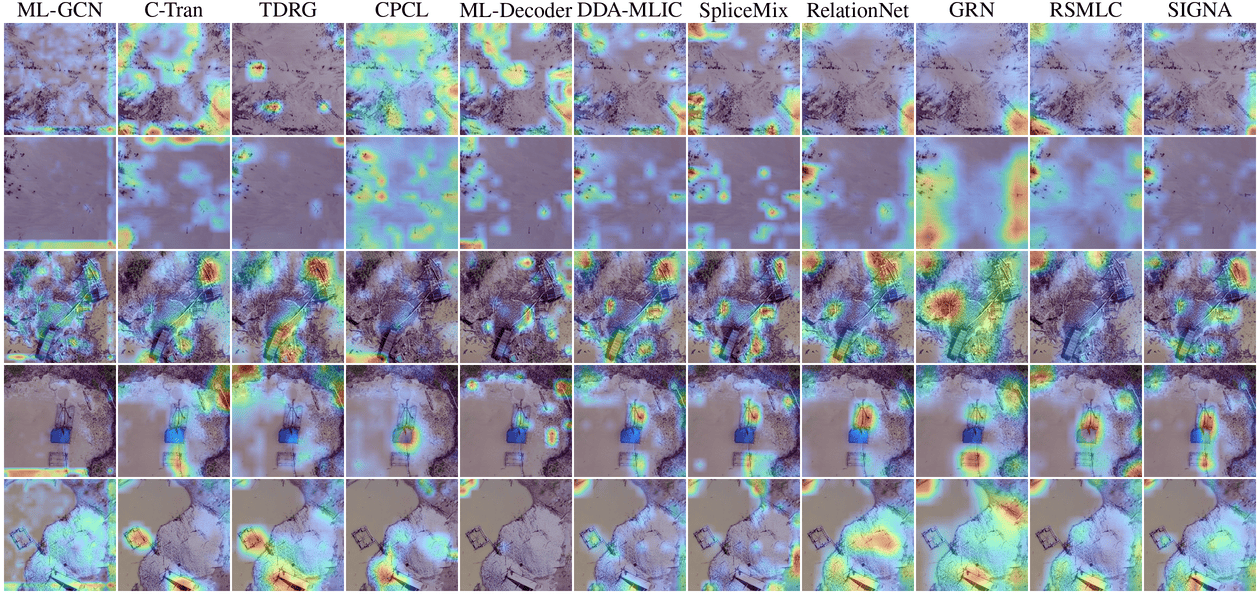}
\caption{Grad-CAM visualizations for \textbf{Bare ground} (BG). Each column corresponds to one method and each row to one representative test patch.}
\label{fig:gradcam_bare_ground}
\end{figure*}

\begin{figure*}[t]
\centering
\includegraphics[width=\textwidth]{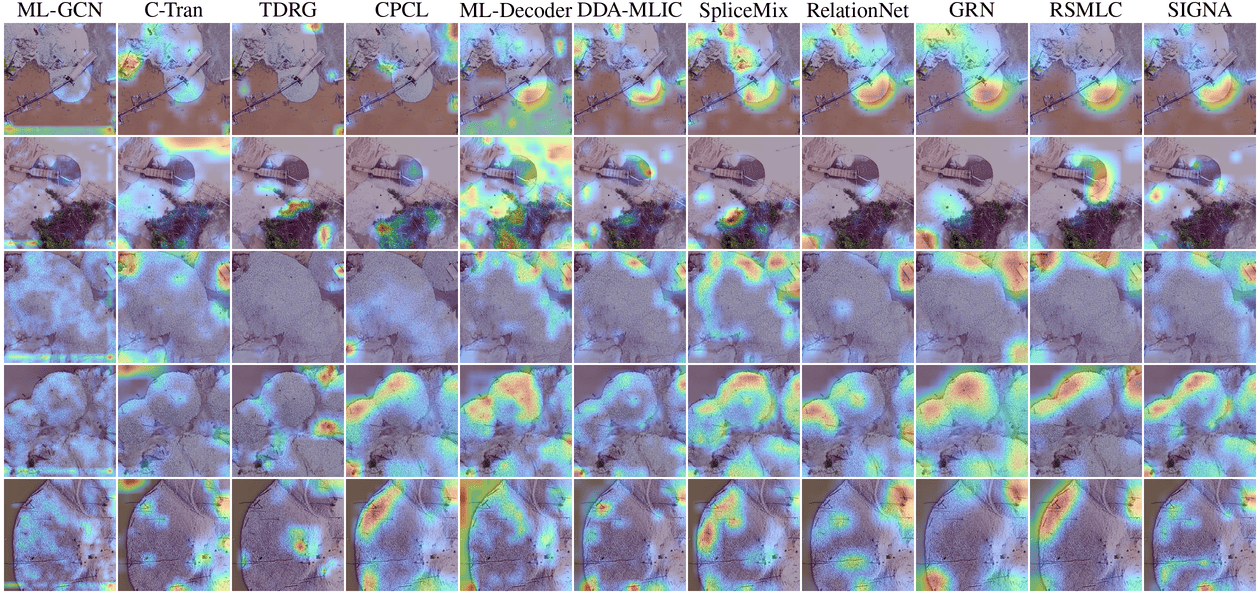}
\caption{Grad-CAM visualizations for \textbf{Gravel mounds} (GM). Each column corresponds to one method and each row to one representative test patch.}
\label{fig:gradcam_gravel_mounds}
\end{figure*}

\begin{figure*}[t]
\centering
\includegraphics[width=\textwidth]{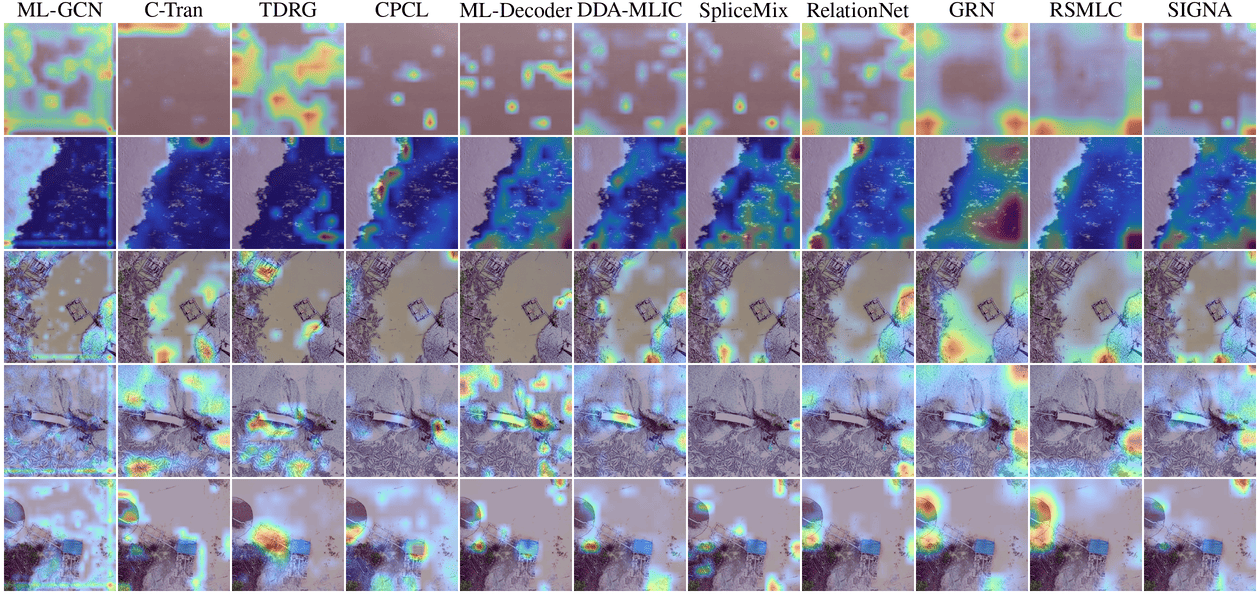}
\caption{Grad-CAM visualizations for \textbf{Water bodies} (WB). Each column corresponds to one method and each row to one representative test patch.}
\label{fig:gradcam_water_bodies}
\end{figure*}

\begin{figure*}[t]
\centering
\includegraphics[width=\textwidth]{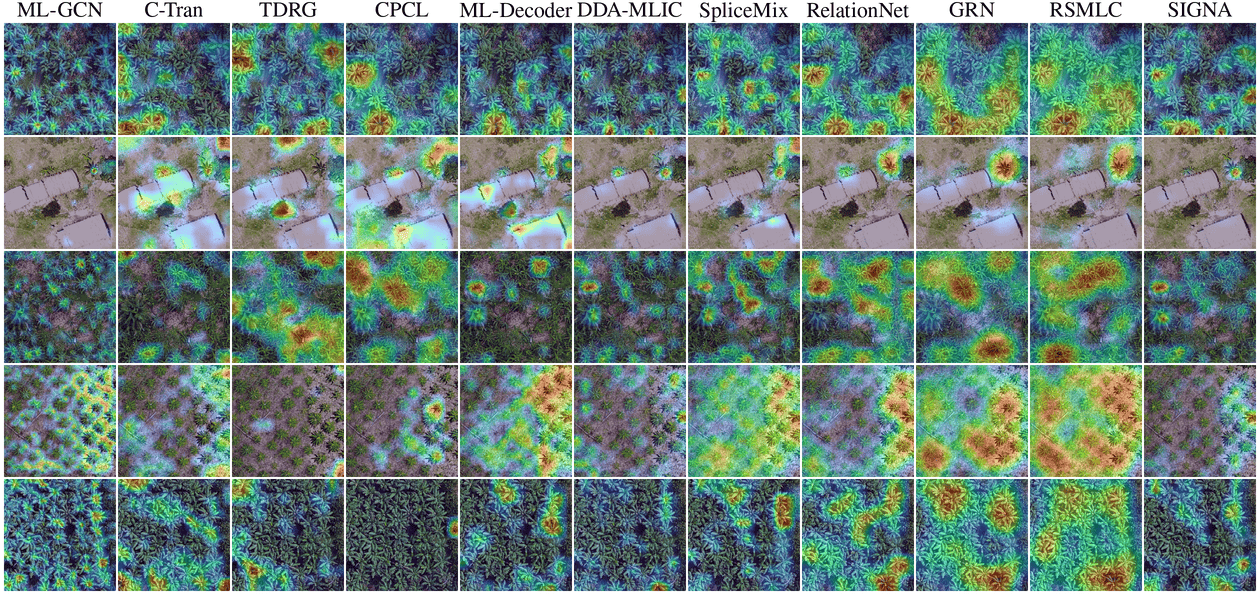}
\caption{Grad-CAM visualizations for \textbf{Agricultural crops} (AC). Each column corresponds to one method and each row to one representative test patch.}
\label{fig:gradcam_agricultural_crops}
\end{figure*}

\begin{figure*}[t]
\centering
\includegraphics[width=\textwidth]{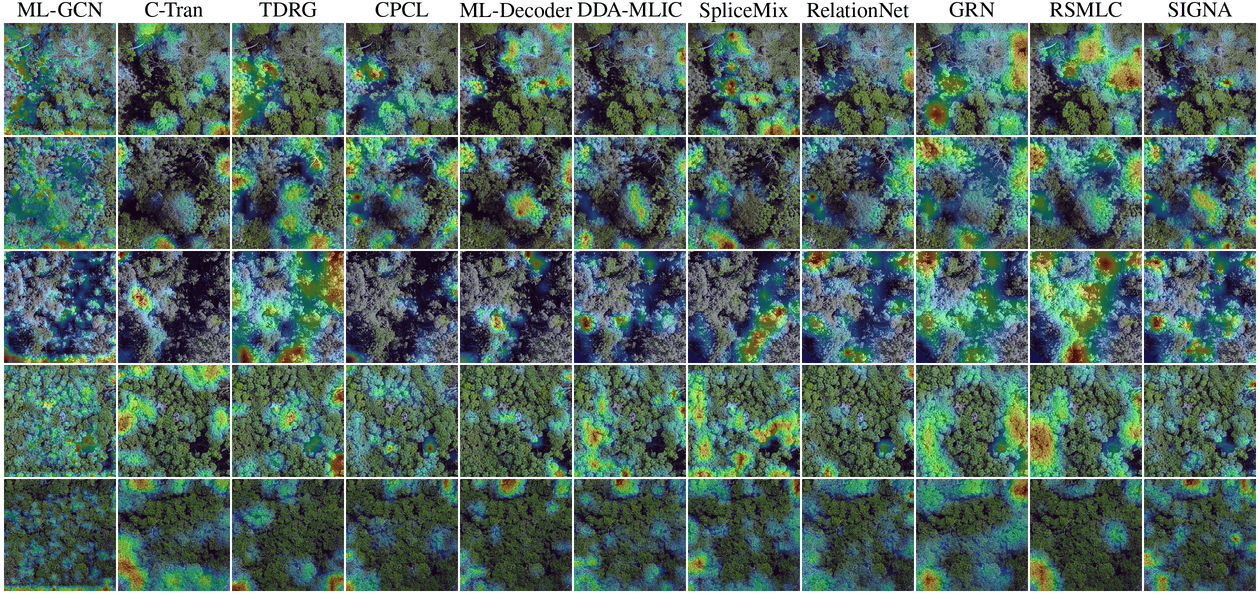}
\caption{Grad-CAM visualizations for \textbf{Primary forests} (PF). Each column corresponds to one method and each row to one representative test patch.}
\label{fig:gradcam_primary_forests}
\end{figure*}

\begin{figure*}[t]
\centering
\includegraphics[width=\textwidth]{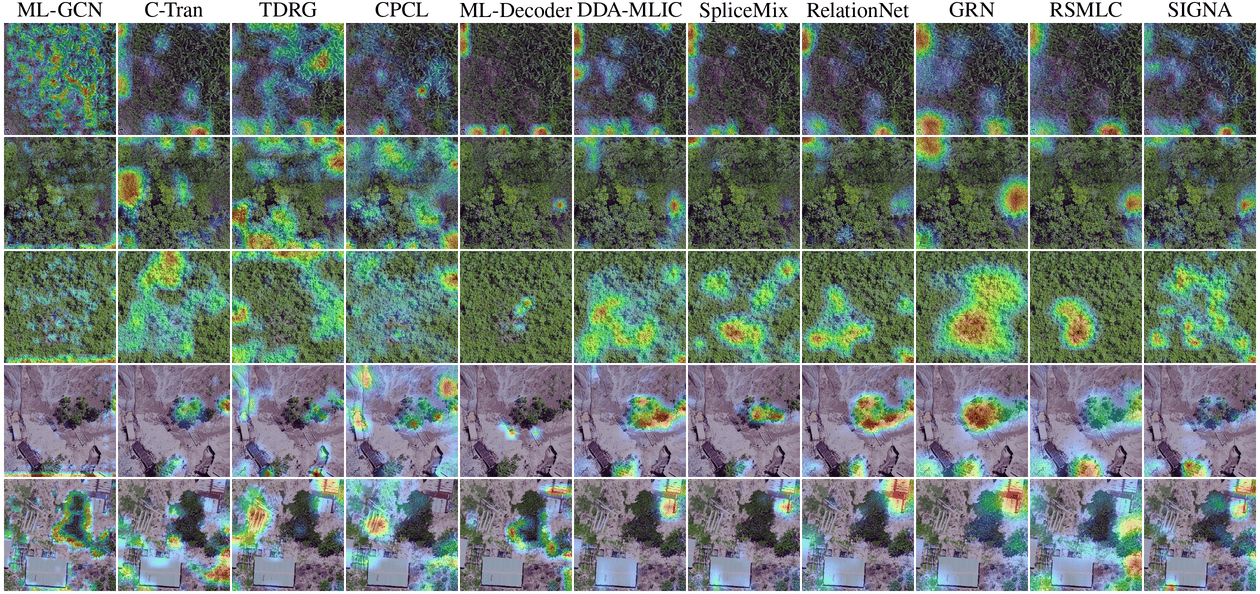}
\caption{Grad-CAM visualizations for \textbf{Type 1 regeneration} (T1R). Each column corresponds to one method and each row to one representative test patch.}
\label{fig:gradcam_type1_regeneration}
\end{figure*}

\begin{figure*}[t]
\centering
\includegraphics[width=\textwidth]{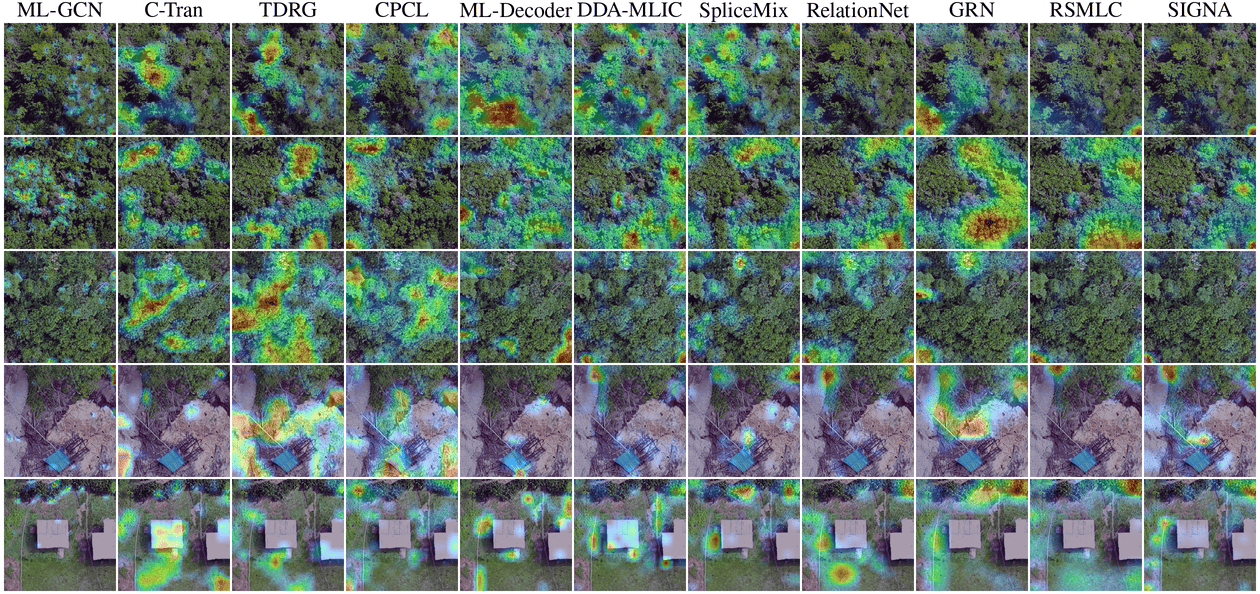}
\caption{Grad-CAM visualizations for \textbf{Type 2 regeneration} (T2R). Each column corresponds to one method and each row to one representative test patch.}
\label{fig:gradcam_type2_regeneration}
\end{figure*}

To complement the aggregate and per-class quantitative results, we further examine class-wise attention patterns using Grad-CAM. Figures~\ref{fig:gradcam_buildings}--\ref{fig:gradcam_type2_regeneration} show five representative test patches for each of the ten classes present in the test split, together with the corresponding activation maps of the selected direct multi-label classification methods. For consistency, these visualizations use the same representative patches as the class-wise qualitative analysis in Appendix~\ref{app:visual_selected_seg}. This makes it easier to compare the spatial evidence learned by direct multi-label classification with the earlier segmentation-based qualitative results.

Overall, the Grad-CAM maps show that many methods can still highlight semantically meaningful regions even though direct multi-label classification uses only image-level class-presence supervision and does not rely on pixel-level annotation. In most cases, the responses are placed on target-related regions rather than on irrelevant background areas, which suggests that weak supervision is still sufficient for learning useful spatial cues. The quality of this evidence, however, varies across methods. Across multiple categories, ML-GCN, CPCL, and ML-Decoder often produce more diffuse or less stable activation patterns. TDRG, despite its strong aggregate recognition performance, also shows broader or less compact responses in some examples. These visual trends are broadly consistent with Tables~\ref{tab:app_mlc_overall}--\ref{tab:app_mlc_classwise_pr}, and indicate that strong image-level metrics do not always correspond to the most spatially concentrated evidence.

For buildings, Figure~\ref{fig:gradcam_buildings} shows that most methods attend to the roof regions, and the roofs usually receive the strongest activations. This is a reasonable cue for the class because the roof is the most visually distinctive building structure in these examples. The differences across methods mainly lie in how compactly the responses are aligned with the roof footprints. ML-GCN, CPCL, and ML-Decoder are less clean on this class, while TDRG, although strong quantitatively, can still activate broader surrounding context in some examples. This class is also one of the more recoverable mining-related categories in the per-class tables, where TDRG gives the best building AP among the general methods, while SIGNA gives the best AP and GRN the best recall within the remote-sensing-specific family.

For mining rafts, the same contrast becomes clearer because this category is more difficult overall. In Figure~\ref{fig:gradcam_mining_rafts}, most methods can still identify the raft structure itself, and only a few methods, mainly ML-GCN, TDRG, CPCL, and ML-Decoder, show clearly weaker and less stable responses. C-Tran also identifies the target region well, but its response often covers a broader raft-related extent and includes more surrounding structure. This visual pattern is also consistent with the quantitative results, where mining rafts remain substantially harder than dominant land-cover categories and the strongest methods differ depending on the metric. Among the general methods, C-Tran gives the best mining-raft AP, TDRG gives the highest precision, and DDA-MLIC gives the highest recall, while among the remote-sensing-specific methods GRN gives the strongest AP and F1. Overall, the visualizations suggest that most methods can already locate the mining-raft region reasonably well, but the weaker methods are noticeably less stable, and the stronger methods are usually those that focus more reliably on the raft structure itself.

For sluices, more methods show unstable responses than in the previous two classes. In Figure~\ref{fig:gradcam_sluices}, SpliceMix is the most visually stable among the selected general methods, and C-Tran can also identify the main sluice-related region reasonably well in several examples. ML-GCN often responds more strongly to the surrounding background than to the sluice itself, which makes the target appear clearer by contrast but does not reflect precise localization; CPCL, TDRG, and ML-Decoder are also less stable on this class. Among the remote-sensing-specific methods, RelationNet, GRN, and SIGNA look broadly similar and all provide usable localization, while GRN appears the most stable overall. RSMLC is visibly weaker than the other three, and this is also supported by the quantitative results: within the remote-sensing-specific family, RSMLC gives the lowest sluice AP, and its sluice F1, precision, and recall are all zero in Tables~\ref{tab:app_mlc_classwise} and~\ref{tab:app_mlc_classwise_pr}. By contrast, RelationNet gives the best sluice AP, F1, and precision in the same family, while GRN gives the best recall.

For bare ground, the Grad-CAM maps show that this category is less visually clean than it first appears. In Figure~\ref{fig:gradcam_bare_ground}, many methods do not place their attention only on the bare-ground surface itself. Instead, the responses are often drawn to other mining-related structures that co-occur with bare ground. This is especially visible in the third and fourth examples, where attention is frequently attracted to the mining-raft regions, and in the fifth example, where some responses shift toward nearby gravel-mound structures. In that sense, bare ground is not recognized purely as a single surface category, but often together with correlated disturbance context. This also fits the broader pattern seen earlier in the segmentation analysis, where bare ground was one of the more ambiguous classes and was often mixed with visually related disturbance categories.

For gravel mounds, most methods tend to respond more strongly to the mound boundaries than to the full mound interior. In Figure~\ref{fig:gradcam_gravel_mounds}, the edges are usually more visually distinctive than the surrounding exposed surface, so many activation maps are concentrated on those boundary regions. TDRG is somewhat different in that its response is often more focused on the mound top. Aside from the few methods that have already looked less stable across several categories, most methods are still reasonably usable on this class, although ML-GCN and C-Tran are less clean in some examples. The quantitative results are partly consistent with this pattern: among the general methods, TDRG gives the best gravel-mound AP and C-Tran gives the best F1, while within the remote-sensing-specific family SIGNA gives the best AP and GRN gives the best F1. This kind of behavior is not unique to gravel mounds, and a similar tendency can also be seen in several major classes, where the models often rely on the most visually distinctive and decision-relevant features rather than responding uniformly over the whole semantic region.

For water bodies, the first two examples in Figure~\ref{fig:gradcam_water_bodies} are almost saturated by the target class, so they mainly verify coarse target awareness rather than fine localization. The more informative cases are the last three rows, where water co-occurs with gravel mounds, bare ground, mining rafts, and sluices. In these mixed patches, many methods do not rely purely on water-body texture, but instead respond strongly to shorelines and nearby mining context, which shows a clear boundary/context bias. The methods also differ in the signals they use: some responses are more diffuse at the scene level, while others are more concentrated on the shoreline or other high-contrast local regions. Water bodies are therefore not always recognized through a uniform focus on the water region itself, especially in these mixed examples where surrounding disturbance patterns provide additional visual information.

For agricultural crops, most methods look reasonably good in Figure~\ref{fig:gradcam_agricultural_crops}. The responses are usually placed on the relatively regular and repetitive crop patterns rather than on unrelated background areas, which is also a reasonable cue for this class because agricultural crops are visually characterized by repeated structures with similar local appearance. Compared with several harder mining-related categories, the gap across methods is smaller here, and no selected method looks clearly unusable from the Grad-CAM maps. This visual impression is also broadly consistent with the per-class tables, where agricultural crops behave more like an intermediate category than an extremely difficult one.

For primary forests, Figure~\ref{fig:gradcam_primary_forests} mainly shows how the models distribute attention within an almost pure canopy patch, rather than how they separate primary forest from complex co-occurring categories. A clear pattern is that many methods do not respond uniformly across the full canopy. Stronger activations often appear around canopy gaps, shadowed regions, and edge-transition areas, which suggests that the models are often using canopy contrast and internal texture variation rather than the forest mass itself as the main evidence. Because these examples are visually dominated by primary forest, this does not necessarily prevent correct classification, but it weakens the localization interpretation. This is also a very major category in ELDOR, which is natural in a tropical rainforest setting, so its quantitative results are generally strong as well. In that sense, the Grad-CAM maps for primary forest are more useful for showing how the models allocate attention within a dominant class than for demonstrating a sharp decision boundary in a complex mixed scene.

For Type~1 regeneration, the first three examples in Figure~\ref{fig:gradcam_type1_regeneration} are dominated by the target class, so they mainly show coarse target awareness rather than fine discrimination. Since this category corresponds to low and early-stage recovering vegetation, these rows mostly suggest that the models are using relatively dense but still low and fairly even canopy patterns as indicators. The more informative cases are the last two rows, where the class appears near bare ground and residual mining disturbance. In the fourth row, many methods can already identify the central and lower Type~1 regeneration region reasonably well. The fifth row is more challenging: the true Type~1 regeneration area is the lower-left region, while the taller vegetation on the right belongs to Type~2 regeneration. In this case, many methods place substantial attention on the right side or on nearby structures instead of focusing cleanly on the lower-left target region. TDRG is noticeably better on this example, and CPCL and C-Tran are also relatively more reasonable. This class therefore remains visually ambiguous, which is also consistent with the earlier analyses where regeneration categories were repeatedly mixed with nearby disturbance and recovery patterns.

For Type~2 regeneration, the main difficulty is that it looks much closer to primary forest than Type~1 regeneration does, while still belonging to the recovery stage. In Figure~\ref{fig:gradcam_type2_regeneration}, the first three examples are dominated by this class, so they mainly show which cues the models rely on within relatively pure Type~2 regeneration patches. The more informative cases are the last two rows, where Type~2 regeneration is no longer the dominant class. In the fourth row, some methods respond more strongly near the boundary of the Type~2 regeneration region, which suggests that local contrast with the surrounding vegetation is one useful cue. In the fifth row, the target is the small Type~2 regeneration patch in the upper part of the image, and most methods can still identify it to some extent. Some responses are placed more on the canopy edges, while others are placed more on the canopy top. This suggests that different methods use somewhat different cues for this class: some rely more on the contrast with nearby regions, while others rely more on the denser and more mature canopy pattern of the Type~2 regeneration itself.

\subsection{VLM-based Recognition}
\label{app:vqa}

This appendix reports the complete results of VLM-based recognition for the evaluated VLMs. We first summarize the overall comparison and then present a more detailed class-wise analysis. Since all methods are evaluated in a fully zero-shot setting, we additionally report results on the train and validation splits together with the standard test split, which provides a fuller view of model behavior across the benchmark.

\subsubsection{Overall VLM-based Recognition Evaluation Performance}
\label{app_vqa_overall}

\begin{table*}[t]
\centering
\scriptsize
\caption{Overall image-level VLM-based recognition results. Protocol~A1 denotes binary QA, Protocol~A2 denotes contrastive scoring, Protocol~A3 denotes positive-threshold scoring, Protocol~A4 denotes model-native multi-label classification, and Protocol~A5 denotes segmentation-derived recognition. SAM3 uses class-name prompts, while SAM3$^\dagger$ uses class-description prompts.}
\label{tab:app_vqa_overall_recognition}
\setlength{\tabcolsep}{2.8pt}
\renewcommand{\arraystretch}{1.06}
\resizebox{0.8\textwidth}{!}{%
\begin{tabular}{lcccccccccc}
\toprule
Model & Protocol & mAP & CP & CR & CF1 & OP & OR & OF1 & Macro-F1 & Sample-F1 \\
\midrule

\rowcolor{gray!15}
\multicolumn{11}{c}{\textbf{Train}} \\
GeoChat    & A1 & 0.3710 & 0.1901 & 0.9083 & 0.3143 & 0.1820 & 0.9265 & 0.3042 & 0.2543 & 0.2921 \\
RS-LLaVA   & A1 & 0.3607 & 0.3548 & 0.5440 & \cellcolor{red!10}\textbf{0.4294} & 0.3900 & 0.7419 & \cellcolor{red!10}\textbf{0.5113} & \cellcolor{red!10}\textbf{0.3643} & \cellcolor{red!10}\textbf{0.5023} \\
VHM        & A1 & \cellcolor{red!10}\textbf{0.4214} & 0.2863 & 0.4485 & 0.3495 & 0.3562 & 0.3047 & 0.3285 & 0.2890 & 0.3479 \\
RemoteCLIP & A2 & 0.2064 & 0.2752 & 0.5248 & 0.3611 & 0.3093 & 0.7798 & 0.4429 & 0.2651 & 0.4256 \\
RemoteCLIP & A3 & 0.3143 & 0.1492 & \cellcolor{red!10}\textbf{1.0000} & 0.2596 & 0.1492 & \cellcolor{red!10}\textbf{1.0000} & 0.2596 & 0.2204 & 0.2523 \\
GeoRSCLIP  & A2 & 0.2818 & 0.3088 & 0.5597 & 0.3980 & 0.2866 & 0.7144 & 0.4091 & 0.2366 & 0.3973 \\
GeoRSCLIP  & A3 & 0.2780 & 0.1492 & \cellcolor{red!10}\textbf{1.0000} & 0.2596 & 0.1492 & \cellcolor{red!10}\textbf{1.0000} & 0.2596 & 0.2204 & 0.2523 \\
DOFA-CLIP  & A2 & 0.3241 & 0.2464 & 0.1110 & 0.1530 & \cellcolor{red!10}\textbf{0.6128} & 0.2105 & 0.3133 & 0.2060 & 0.3570 \\
RemoteSAM  & A4 & 0.1968 & 0.1921 & 0.5864 & 0.2894 & 0.1993 & 0.7355 & 0.3136 & 0.2511 & 0.3015 \\
RemoteSAM & A5 & 0.1759 & 0.1988 & 0.7654 & 0.3157 & 0.1906 & 0.8183 & 0.3092 & 0.2516 & 0.2952 \\
SAM3       & A5 & 0.3076 & 0.3348 & 0.2792 & 0.3045 & 0.4742 & 0.2047 & 0.2860 & 0.2790 & 0.3374 \\
SAM3$^\dagger$ & A5 & 0.3292 & \cellcolor{red!10}\textbf{0.4428} & 0.3260 & 0.3755 & 0.6065 & 0.3978 & 0.4805 & 0.3330 & 0.4960 \\

\midrule
\rowcolor{gray!15}
\multicolumn{11}{c}{\textbf{Validation}} \\
GeoChat    & A1 & 0.3365 & 0.2070 & 0.9070 & 0.3371 & 0.1709 & 0.9383 & 0.2891 & 0.2625 & 0.2764 \\
RS-LLaVA   & A1 & 0.4214 & 0.4424 & 0.5411 & \cellcolor{red!10}\textbf{0.4868} & 0.4055 & 0.8469 & 0.5484 & \cellcolor{red!10}\textbf{0.4326} & 0.5291 \\
VHM        & A1 & \cellcolor{red!10}\textbf{0.4506} & 0.3425 & 0.3952 & 0.3669 & 0.5060 & 0.5015 & 0.5038 & 0.3495 & 0.5359 \\
RemoteCLIP & A2 & 0.1983 & 0.1968 & 0.5668 & 0.2922 & 0.2651 & 0.8150 & 0.4001 & 0.2668 & 0.3813 \\
RemoteCLIP & A3 & 0.3341 & 0.1416 & \cellcolor{red!10}\textbf{1.0000} & 0.2481 & 0.1315 & \cellcolor{red!10}\textbf{1.0000} & 0.2324 & 0.2056 & 0.2254 \\
GeoRSCLIP  & A2 & 0.2967 & 0.2682 & 0.5254 & 0.3551 & 0.3060 & 0.7968 & 0.4422 & 0.2722 & 0.4298 \\
GeoRSCLIP  & A3 & 0.3314 & 0.1416 & \cellcolor{red!10}\textbf{1.0000} & 0.2481 & 0.1315 & \cellcolor{red!10}\textbf{1.0000} & 0.2324 & 0.2056 & 0.2254 \\
DOFA-CLIP  & A2 & 0.3809 & 0.4537 & 0.1360 & 0.2093 & \cellcolor{red!10}\textbf{0.8571} & 0.3589 & 0.5060 & 0.2700 & 0.5586 \\
RemoteSAM  & A4 & 0.1526 & 0.1610 & 0.5555 & 0.2496 & 0.1859 & 0.7674 & 0.2992 & 0.2604 & 0.2889 \\
RemoteSAM & A5 & 0.1460 & 0.1725 & 0.7344 & 0.2794 & 0.1772 & 0.8345 & 0.2924 & 0.2224 & 0.2812 \\
SAM3       & A5 & 0.3151 & 0.4048 & 0.2526 & 0.3111 & 0.8007 & 0.4283 & 0.5581 & 0.3612 & 0.6315 \\
SAM3$^\dagger$ & A5 & 0.3324 & \cellcolor{red!10}\textbf{0.5015} & 0.3343 & 0.4012 & 0.7557 & 0.5462 & \cellcolor{red!10}\textbf{0.6341} & 0.3675 & \cellcolor{red!10}\textbf{0.6946} \\

\midrule
\rowcolor{gray!15}
\multicolumn{11}{c}{\textbf{Test}} \\
GeoChat    & A1 & 0.4675 & 0.2479 & 0.8882 & 0.3876 & 0.1672 & 0.8955 & 0.2818 & 0.3103 & 0.2655 \\
RS-LLaVA   & A1 & 0.4848 & \cellcolor{red!10}\textbf{0.5540} & 0.5195 & \cellcolor{red!10}\textbf{0.5362} & 0.4162 & 0.7532 & 0.5362 & 0.4368 & 0.4861 \\
VHM        & A1 & \cellcolor{red!10}\textbf{0.5159} & 0.4605 & 0.3984 & 0.4272 & 0.4967 & 0.3992 & 0.4426 & \cellcolor{red!10}\textbf{0.4630} & 0.4343 \\
RemoteCLIP & A2 & 0.3064 & 0.2913 & 0.4822 & 0.3632 & 0.2920 & 0.7368 & 0.4183 & 0.3354 & 0.3751 \\
RemoteCLIP & A3 & 0.3880 & 0.1924 & \cellcolor{red!10}\textbf{1.0000} & 0.3227 & 0.1374 & \cellcolor{red!10}\textbf{1.0000} & 0.2417 & 0.2816 & 0.2335 \\
GeoRSCLIP  & A2 & 0.3544 & 0.4282 & 0.6093 & 0.5030 & 0.2961 & 0.6938 & 0.4151 & 0.2893 & 0.3704 \\
GeoRSCLIP  & A3 & 0.3219 & 0.1924 & \cellcolor{red!10}\textbf{1.0000} & 0.3227 & 0.1374 & \cellcolor{red!10}\textbf{1.0000} & 0.2417 & 0.2816 & 0.2335 \\
DOFA-CLIP  & A2 & 0.3986 & 0.3286 & 0.1300 & 0.1863 & \cellcolor{red!10}\textbf{0.7798} & 0.2750 & 0.4066 & 0.3189 & 0.4457 \\
RemoteSAM  & A4 & 0.2580 & 0.2626 & 0.5802 & 0.3616 & 0.1991 & 0.7753 & 0.3168 & 0.3119 & 0.3052 \\
RemoteSAM & A5 & 0.2403 & 0.2795 & 0.7475 & 0.4068 & 0.1913 & 0.8379 & 0.3115 & 0.3416 & 0.2987 \\
SAM3       & A5 & 0.3608 & 0.5127 & 0.1797 & 0.2662 & 0.6483 & 0.2811 & 0.3922 & 0.3251 & 0.4307 \\
SAM3$^\dagger$ & A5 & 0.3786 & 0.4802 & 0.3206 & 0.3845 & 0.7239 & 0.4614 & \cellcolor{red!10}\textbf{0.5636} & 0.3536 & \cellcolor{red!10}\textbf{0.5461} \\

\bottomrule
\end{tabular}%
}
\end{table*}

\begin{table*}[t]
\centering
\scriptsize
\caption{Overall prompted segmentation results for Protocol~B. Metrics follow the same definitions as in the semantic segmentation appendix. SAM3 uses class-name prompts, while SAM3$^\dagger$ uses class-description prompts.}
\label{tab:app_vqa_overall_segmentation}
\setlength{\tabcolsep}{3.2pt}
\renewcommand{\arraystretch}{1}
\resizebox{\textwidth}{!}{%
\begin{tabular}{lcccccccccccc}
\toprule
\multirow{2}{*}{Model} &
\multicolumn{4}{c}{Train} &
\multicolumn{4}{c}{Validation} &
\multicolumn{4}{c}{Test} \\
\cmidrule(lr){2-5} \cmidrule(lr){6-9} \cmidrule(lr){10-13}
& mIoU$_p$ & mIoU & Macro-F1 & OA
& mIoU$_p$ & mIoU & Macro-F1 & OA
& mIoU$_p$ & mIoU & Macro-F1 & OA \\
\midrule
RemoteSAM
& 0.0683 & 0.0683 & 0.1048 & 0.1085
& 0.0468 & 0.0434 & 0.0737 & 0.0983
& 0.0767 & 0.0548 & 0.0854 & 0.1309 \\
SAM3
& 0.1223 & 0.1223 & 0.1852 & 0.3174
& 0.1534 & 0.1425 & 0.2030 & 0.6393
& 0.1502 & 0.1073 & 0.1502 & 0.4353 \\
SAM3$^\dagger$
& \cellcolor{red!10}\textbf{0.1672} & \cellcolor{red!10}\textbf{0.1672} & \cellcolor{red!10}\textbf{0.2475} & \cellcolor{red!10}\textbf{0.4088}
& \cellcolor{red!10}\textbf{0.1736} & \cellcolor{red!10}\textbf{0.1612} & \cellcolor{red!10}\textbf{0.2343} & \cellcolor{red!10}\textbf{0.6634}
& \cellcolor{red!10}\textbf{0.1742} & \cellcolor{red!10}\textbf{0.1244} & \cellcolor{red!10}\textbf{0.1796} & \cellcolor{red!10}\textbf{0.5094} \\
\bottomrule
\end{tabular}%
}
\end{table*}

Tables~\ref{tab:app_vqa_overall_recognition} and~\ref{tab:app_vqa_overall_segmentation} summarize the overall VLM-based recognition results for image-level recognition and prompted segmentation, respectively. For image-level recognition, several consistent patterns can be observed across the three splits. First, the generative models in Protocol~A1 provide the strongest overall zero-shot recognition performance. In particular, VHM achieves the best mAP on all three splits, while RS-LLaVA gives the strongest balanced threshold-based results on several F1-oriented metrics, including CF1 and OF1. This suggests that, among the directly prompted image-level methods, the generative VLMs currently provide the most reliable trade-off between ranking quality and balanced class-presence prediction.

A second important pattern is the difference between Protocol~A2 and Protocol~A3 for the CLIP-style models. Protocol~A3 uses only the positive prompt with threshold-based prediction, and for both RemoteCLIP and GeoRSCLIP it leads to CR=1 and OR=1 across all three splits. This indicates that positive-only scoring is too permissive and does not separate positive from negative labels reliably, so the models tend to over-predict class presence. In contrast, Protocol~A2 introduces an explicit positive--negative comparison, which substantially resolves this issue and yields much more meaningful threshold-based results. The gains are clearly reflected in CF1, OF1, Macro-F1, and Sample-F1, showing that prompt contrast is important for obtaining usable image-level predictions from CLIP-style models. DOFA-CLIP further suggests that alternative scoring schemes can also mitigate this failure mode, although its precision--recall trade-off remains uneven across classes.

The SAM-based image-level results also show that prompt design matters. Across train, validation, and test, SAM3$^\dagger$, which uses detailed class descriptions, consistently improves over SAM3 with class-name prompts in mAP, CF1, OF1, and Sample-F1. The gains are especially clear on validation and test, where SAM3$^\dagger$ gives substantially stronger OF1 and Sample-F1. This suggests that richer semantic descriptions help SAM3 align language prompts with image content more reliably, leading to better image-level recognition after mask aggregation. RemoteSAM is competitive on some metrics, but overall it remains below the strongest generative models and below SAM3$^\dagger$.

Table~\ref{tab:app_vqa_overall_segmentation} shows that a similar effect also appears in zero-shot prompted segmentation. Overall, SAM3 is consistently stronger than RemoteSAM on all three splits, and SAM3$^\dagger$ further improves over SAM3 in mIoU$_p$, mIoU, Macro-F1, and OA on every split. This indicates that stronger promptable segmentation models transfer better to this benchmark, and that detailed class descriptions help produce cleaner and more semantically aligned masks. This trend is also consistent with the image-level recognition results in Table~\ref{tab:app_vqa_overall_recognition}, where using detailed class descriptions likewise improves over class-name prompts.

\begin{table*}[t]
\centering
\scriptsize
\caption{Per-class image-level VLM-based recognition results, organized by split. Within each split, AP is reported in the upper block and F1 in the lower block. Protocol~A1 denotes binary QA, Protocol~A2 denotes contrastive scoring, Protocol~A3 denotes positive-threshold scoring, Protocol~A4 denotes model-native multi-label classification, and Protocol~A5 denotes segmentation-derived recognition. SAM3 uses class-name prompts, while SAM3$^\dagger$ uses class-description prompts.}
\label{tab:app_vqa_classwise_apf1}
\setlength{\tabcolsep}{2.4pt}
\renewcommand{\arraystretch}{1}
\resizebox{\textwidth}{!}{%
\begin{tabular}{lccccccccccccccc}
\toprule
\multirow{2}{*}{Model} &
\multirow{2}{*}{Protocol} &
\multirow{2}{*}{Buildings} &
Mining & Primary & Heavy & Water & Agricultural & Compact & Gravel & \multirow{2}{*}{Grass} & Type 1 & Type 2 & Bare & \multirow{2}{*}{Sluices} & \multirow{2}{*}{Vehicles} \\
& &
& rafts & forests & machinery & bodies & crops & mounds & mounds & & regeneration & regeneration & ground & & \\
\midrule

\rowcolor{gray!15}
\multicolumn{16}{c}{\textbf{Train}} \\
\multicolumn{16}{c}{\textit{AP}} \\
GeoChat & A1 & 0.6185 & 0.1227 & \cellcolor{red!10}\textbf{0.7810} & 0.6542 & 0.7596 & 0.1490 & 0.2740 & 0.0772 & 0.0108 & 0.2236 & 0.4301 & 0.7951 & \cellcolor{red!10}\textbf{0.1532} & 0.1452 \\
RS-LLaVA & A1 & 0.6330 & 0.1442 & 0.6551 & 0.6051 & \cellcolor{red!10}\textbf{0.8494} & 0.0649 & 0.1220 & 0.1596 & 0.0607 & 0.2190 & 0.4576 & 0.8555 & 0.0402 & 0.1828 \\
VHM & A1 & \cellcolor{red!10}\textbf{0.6966} & \cellcolor{red!10}\textbf{0.1536} & 0.7419 & \cellcolor{red!10}\textbf{0.7245} & 0.7970 & 0.1531 & \cellcolor{red!10}\textbf{0.3480} & \cellcolor{red!10}\textbf{0.2253} & 0.1627 & 0.2672 & 0.6012 & \cellcolor{red!10}\textbf{0.8616} & 0.0882 & 0.0788 \\
RemoteCLIP & A2 & 0.0384 & 0.0028 & 0.6204 & 0.0030 & 0.7328 & 0.0606 & 0.1351 & 0.0455 & 0.0141 & 0.2640 & 0.4660 & 0.5059 & 0.0011 & 0.0006 \\
RemoteCLIP & A3 & 0.5437 & 0.0034 & 0.4594 & 0.3647 & 0.8294 & 0.0509 & 0.3256 & 0.0466 & 0.0361 & 0.3308 & 0.4821 & 0.7513 & 0.0011 & 0.1751 \\
GeoRSCLIP & A2 & 0.2000 & 0.0011 & 0.6132 & 0.1141 & 0.7971 & \cellcolor{red!10}\textbf{0.1629} & 0.2425 & 0.0566 & 0.0791 & 0.3444 & \cellcolor{red!10}\textbf{0.6844} & 0.6262 & 0.0011 & 0.0227 \\
GeoRSCLIP & A3 & 0.0740 & 0.1166 & 0.7243 & 0.2562 & 0.7548 & 0.1622 & 0.2299 & 0.0438 & 0.1505 & 0.2967 & 0.4579 & 0.6145 & 0.0041 & 0.0065 \\
DOFA-CLIP & A2 & 0.6176 & 0.0176 & 0.7412 & 0.0675 & 0.7049 & 0.0907 & 0.2685 & 0.0353 & \cellcolor{red!10}\textbf{0.3860} & 0.3148 & 0.4751 & 0.7741 & 0.0099 & 0.0339 \\
RemoteSAM & A4 & 0.5212 & 0.0003 & 0.5639 & 0.0035 & 0.2674 & 0.0177 & 0.0915 & 0.0102 & 0.0084 & 0.2795 & 0.4915 & 0.2682 & 0.0004 & 0.2310 \\
RemoteSAM & A5 & 0.5464 & 0.0004 & 0.3694 & 0.0019 & 0.2787 & 0.0158 & 0.1026 & 0.0107 & 0.0090 & 0.2872 & 0.5087 & 0.3304 & 0.0002 & 0.0013 \\
SAM3 & A5 & 0.3857 & 0.0219 & 0.7225 & 0.5302 & 0.4686 & 0.0281 & 0.1384 & 0.1909 & 0.0710 & \cellcolor{red!10}\textbf{0.4477} & 0.5156 & 0.5612 & 0.0015 & 0.2228 \\
SAM3$^\dagger$ & A5 & 0.3500 & 0.0041 & 0.5786 & 0.4553 & 0.7179 & 0.1426 & 0.1576 & 0.1097 & 0.0824 & 0.3851 & 0.5206 & 0.7332 & 0.0986 & \cellcolor{red!10}\textbf{0.2729} \\
\midrule
\multicolumn{16}{c}{\textit{F1}} \\
GeoChat & A1 & 0.0191 & 0.0039 & 0.5398 & 0.0010 & 0.6794 & 0.1757 & \cellcolor{red!10}\textbf{0.3580} & 0.0413 & 0.0278 & \cellcolor{red!10}\textbf{0.4805} & \cellcolor{red!10}\textbf{0.6825} & 0.5499 & 0.0005 & 0.0008 \\
RS-LLaVA & A1 & 0.4977 & 0.1702 & 0.6357 & 0.5385 & \cellcolor{red!10}\textbf{0.7552} & 0.0760 & 0.1004 & 0.2122 & 0.0423 & 0.3383 & 0.6701 & \cellcolor{red!10}\textbf{0.7969} & 0.1212 & 0.1458 \\
VHM & A1 & \cellcolor{red!10}\textbf{0.6552} & \cellcolor{red!10}\textbf{0.2254} & 0.7210 & 0.3117 & 0.6696 & 0.1400 & 0.0009 & 0.1543 & 0.0368 & 0.0000 & 0.0000 & 0.3278 & 0.0296 & 0.1955 \\
RemoteCLIP & A2 & 0.0676 & 0.0054 & 0.6124 & 0.0020 & 0.4922 & 0.0015 & 0.1994 & 0.0903 & 0.0340 & 0.4703 & 0.6579 & 0.5478 & 0.0000 & 0.0000 \\
RemoteCLIP & A3 & 0.0191 & 0.0010 & 0.5396 & 0.0008 & 0.4932 & 0.0402 & 0.2306 & 0.0351 & 0.0277 & 0.4763 & 0.6773 & 0.5435 & 0.0005 & 0.0008 \\
GeoRSCLIP & A2 & 0.0157 & 0.0018 & 0.5935 & 0.1000 & 0.6303 & \cellcolor{red!10}\textbf{0.1932} & 0.1850 & 0.0575 & 0.0308 & 0.4782 & 0.6787 & 0.3336 & 0.0016 & 0.0119 \\
GeoRSCLIP & A3 & 0.0191 & 0.0010 & 0.5396 & 0.0008 & 0.4932 & 0.0402 & 0.2306 & 0.0351 & 0.0277 & 0.4763 & 0.6773 & 0.5435 & 0.0005 & 0.0008 \\
DOFA-CLIP & A2 & 0.0000 & 0.0496 & 0.7270 & 0.1356 & 0.0411 & 0.0000 & 0.0000 & 0.0544 & 0.0086 & 0.0000 & 0.0000 & 0.4260 & 0.0000 & 0.0000 \\
RemoteSAM & A4 & 0.6274 & 0.0000 & 0.5676 & 0.0101 & 0.4932 & 0.0401 & 0.2306 & 0.0232 & 0.0222 & 0.2357 & 0.5050 & 0.4972 & 0.0009 & 0.0112 \\
RemoteSAM & A5 & 0.6095 & 0.0014 & 0.5568 & 0.0019 & 0.4906 & 0.0403 & 0.2334 & 0.0289 & 0.0252 & 0.3802 & 0.6204 & 0.5290 & 0.0008 & 0.0040 \\
SAM3 & A5 & 0.5387 & 0.0870 & \cellcolor{red!10}\textbf{0.7401} & \cellcolor{red!10}\textbf{0.6471} & 0.2174 & 0.1025 & 0.0039 & \cellcolor{red!10}\textbf{0.2311} & 0.0744 & 0.0000 & 0.0000 & 0.2203 & 0.0000 & 0.2069 \\
SAM3$^\dagger$ & A5 & 0.5023 & 0.0000 & 0.6982 & 0.5909 & 0.6270 & 0.1196 & 0.0592 & 0.0044 & \cellcolor{red!10}\textbf{0.1489} & 0.2836 & 0.0193 & 0.7204 & \cellcolor{red!10}\textbf{0.2727} & \cellcolor{red!10}\textbf{0.2824} \\

\midrule
\rowcolor{gray!15}
\multicolumn{16}{c}{\textbf{Validation}} \\
\multicolumn{16}{c}{\textit{AP}} \\
GeoChat & A1 & 0.3178 & 0.1478 & 0.9568 & 0.6777 & 0.6381 & 0.4071 & 0.1054 & 0.1207 & 0.0771 & 0.1544 & 0.1846 & 0.4469 & 0.1400 & \ph \\
RS-LLaVA & A1 & \cellcolor{red!10}\textbf{0.3792} & \cellcolor{red!10}\textbf{0.3983} & 0.9553 & 0.6802 & 0.7186 & 0.4354 & 0.0673 & 0.1862 & 0.6381 & 0.1508 & 0.1858 & 0.5954 & 0.0871 & \ph \\
VHM & A1 & 0.3438 & 0.3195 & \cellcolor{red!10}\textbf{0.9621} & 0.6839 & 0.6864 & 0.3209 & 0.2107 & \cellcolor{red!10}\textbf{0.2640} & \cellcolor{red!10}\textbf{0.7809} & 0.2124 & 0.2283 & \cellcolor{red!10}\textbf{0.6456} & \cellcolor{red!10}\textbf{0.1992} & \ph \\
RemoteCLIP & A2 & 0.0061 & 0.0231 & 0.9061 & 0.0913 & 0.5655 & 0.1349 & 0.0445 & 0.1428 & 0.0521 & 0.1717 & 0.2058 & 0.2273 & 0.0067 & \ph \\
RemoteCLIP & A3 & 0.1572 & 0.0024 & 0.8906 & 0.6121 & \cellcolor{red!10}\textbf{0.7231} & 0.2075 & \cellcolor{red!10}\textbf{0.3025} & 0.0270 & 0.2938 & 0.3181 & 0.2924 & 0.4799 & 0.0371 & \ph \\
GeoRSCLIP & A2 & 0.0642 & 0.0030 & 0.9372 & 0.1265 & 0.5785 & 0.3337 & 0.1725 & 0.0367 & 0.6346 & 0.2862 & 0.3671 & 0.3106 & 0.0059 & \ph \\
GeoRSCLIP & A3 & 0.0731 & 0.0412 & 0.8905 & 0.3369 & 0.6649 & \cellcolor{red!10}\textbf{0.4673} & 0.1610 & 0.0637 & 0.6056 & \cellcolor{red!10}\textbf{0.3547} & \cellcolor{red!10}\textbf{0.3783} & 0.2594 & 0.0117 & \ph \\
DOFA-CLIP & A2 & 0.2211 & 0.1612 & 0.9081 & 0.5907 & 0.5835 & 0.0955 & 0.3018 & 0.0678 & 0.7545 & 0.3201 & 0.3001 & 0.5588 & 0.0880 & \ph \\
RemoteSAM & A4 & 0.2250 & 0.0006 & 0.8880 & 0.0010 & 0.2091 & 0.0399 & 0.0197 & 0.0129 & 0.0440 & 0.2065 & 0.2106 & 0.1253 & 0.0006 & \ph \\
RemoteSAM & A5 & 0.1857 & 0.0007 & 0.8024 & 0.0007 & 0.1882 & 0.0412 & 0.0283 & 0.0125 & 0.0537 & 0.1977 & 0.2169 & 0.1693 & 0.0008 & \ph \\
SAM3 & A5 & 0.2899 & 0.0286 & 0.9355 & \cellcolor{red!10}\textbf{0.7667} & 0.2594 & 0.0971 & 0.1179 & 0.1372 & 0.5436 & 0.3301 & 0.2283 & 0.3607 & 0.0009 & \ph \\
SAM3$^\dagger$ & A5 & 0.1981 & 0.0009 & 0.8899 & 0.7509 & 0.5436 & 0.2293 & 0.1156 & 0.0249 & 0.3947 & 0.3463 & 0.2386 & 0.5330 & 0.0551 & \ph \\
\midrule
\multicolumn{16}{c}{\textit{F1}} \\
GeoChat & A1 & 0.0128 & 0.0084 & 0.8681 & 0.0015 & 0.6440 & \cellcolor{red!10}\textbf{0.4128} & 0.1560 & 0.0696 & 0.1309 & 0.3782 & 0.4366 & 0.2914 & 0.0022 & \ph \\
RS-LLaVA & A1 & 0.2941 & \cellcolor{red!10}\textbf{0.2857} & 0.8903 & 0.7500 & \cellcolor{red!10}\textbf{0.6666} & 0.3405 & 0.1158 & 0.1986 & 0.2750 & 0.3292 & 0.4294 & 0.6155 & 0.0000 & \ph \\
VHM & A1 & \cellcolor{red!10}\textbf{0.4623} & 0.2581 & 0.9004 & 0.4800 & 0.6157 & 0.3252 & 0.0037 & \cellcolor{red!10}\textbf{0.2772} & 0.1658 & 0.0000 & 0.0000 & 0.2865 & 0.0693 & \ph \\
RemoteCLIP & A2 & 0.0121 & 0.0347 & 0.8443 & 0.0013 & 0.5042 & 0.0000 & 0.0967 & 0.2064 & 0.1411 & 0.3777 & 0.4231 & 0.2928 & 0.0000 & \ph \\
RemoteCLIP & A3 & 0.0128 & 0.0022 & 0.8681 & 0.0011 & 0.3348 & 0.1171 & 0.0630 & 0.0434 & 0.1308 & 0.3776 & 0.4371 & 0.2823 & 0.0022 & \ph \\
GeoRSCLIP & A2 & 0.0000 & 0.0049 & 0.8794 & 0.1471 & 0.4111 & 0.3727 & \cellcolor{red!10}\textbf{0.1566} & 0.0650 & 0.1422 & \cellcolor{red!10}\textbf{0.3815} & \cellcolor{red!10}\textbf{0.4401} & 0.2574 & 0.0083 & \ph \\
GeoRSCLIP & A3 & 0.0128 & 0.0022 & 0.8681 & 0.0011 & 0.3348 & 0.1171 & 0.0630 & 0.0434 & 0.1308 & 0.3776 & 0.4371 & 0.2823 & 0.0022 & \ph \\
DOFA-CLIP & A2 & 0.0000 & 0.2424 & 0.8425 & 0.6154 & 0.0655 & 0.0000 & 0.0115 & 0.0540 & 0.0053 & 0.0000 & 0.0000 & 0.3235 & 0.0000 & \ph \\
RemoteSAM & A4 & 0.3737 & 0.0000 & 0.8813 & 0.0000 & 0.3338 & 0.1169 & 0.0628 & 0.0314 & 0.1131 & 0.1737 & 0.2545 & 0.2624 & 0.0000 & \ph \\
RemoteSAM & A5 & 0.4065 & 0.0026 & 0.8765 & 0.0026 & 0.3346 & 0.1173 & 0.0615 & 0.0342 & 0.1282 & 0.2556 & 0.3908 & 0.2747 & 0.0055 & \ph \\
SAM3 & A5 & 0.4151 & 0.0909 & 0.8901 & 0.5833 & 0.0800 & 0.1353 & 0.0000 & 0.2442 & \cellcolor{red!10}\textbf{0.5234} & 0.0000 & 0.0000 & 0.2888 & 0.0000 & \ph \\
SAM3$^\dagger$ & A5 & 0.2899 & 0.0000 & \cellcolor{red!10}\textbf{0.9056} & \cellcolor{red!10}\textbf{0.7778} & 0.5339 & 0.3849 & 0.0197 & 0.0274 & 0.3686 & 0.2243 & 0.0269 & \cellcolor{red!10}\textbf{0.6633} & \cellcolor{red!10}\textbf{0.1875} & \ph \\

\midrule
\rowcolor{gray!15}
\multicolumn{16}{c}{\textbf{Test}} \\
\multicolumn{16}{c}{\textit{AP}} \\
GeoChat & A1 & 0.6633 & 0.3192 & \cellcolor{red!10}\textbf{0.8727} & \ph & 0.7976 & 0.1351 & \ph & 0.4281 & \ph & 0.2101 & 0.2477 & 0.7073 & \cellcolor{red!10}\textbf{0.2935} & \ph \\
RS-LLaVA & A1 & 0.6187 & 0.3692 & 0.8622 & \ph & 0.8081 & \cellcolor{red!10}\textbf{0.2661} & \ph & 0.5409 & \ph & 0.2087 & 0.2532 & \cellcolor{red!10}\textbf{0.7498} & 0.1710 & \ph \\
VHM & A1 & \cellcolor{red!10}\textbf{0.7309} & \cellcolor{red!10}\textbf{0.4410} & 0.8708 & \ph & 0.8295 & 0.1120 & \ph & \cellcolor{red!10}\textbf{0.6163} & \ph & 0.2779 & 0.3489 & 0.7154 & 0.2165 & \ph \\
RemoteCLIP & A2 & 0.0689 & 0.0112 & 0.7797 & \ph & 0.6984 & 0.1787 & \ph & 0.1968 & \ph & 0.2571 & 0.3079 & 0.5124 & 0.0528 & \ph \\
RemoteCLIP & A3 & 0.5341 & 0.0168 & 0.7718 & \ph & 0.8333 & 0.0895 & \ph & 0.2896 & \ph & 0.2863 & 0.3310 & 0.6465 & 0.0815 & \ph \\
GeoRSCLIP & A2 & 0.0489 & 0.0210 & 0.8335 & \ph & 0.7868 & 0.1822 & \ph & 0.2922 & \ph & 0.3654 & 0.3957 & 0.6029 & 0.0159 & \ph \\
GeoRSCLIP & A3 & 0.0260 & 0.0921 & 0.8086 & \ph & \cellcolor{red!10}\textbf{0.8399} & 0.1852 & \ph & 0.0981 & \ph & 0.4218 & 0.3664 & 0.3650 & 0.0155 & \ph \\
DOFA-CLIP & A2 & 0.4674 & 0.3221 & 0.8642 & \ph & 0.7981 & 0.0602 & \ph & 0.2091 & \ph & 0.2892 & 0.3437 & 0.5160 & 0.1155 & \ph \\
RemoteSAM & A4 & 0.5151 & 0.0021 & 0.7472 & \ph & 0.5751 & 0.0136 & \ph & 0.0345 & \ph & 0.2510 & 0.2233 & 0.2163 & 0.0014 & \ph \\
RemoteSAM & A5 & 0.5016 & 0.0024 & 0.6385 & \ph & 0.4700 & 0.0193 & \ph & 0.0345 & \ph & 0.2356 & 0.2332 & 0.2660 & 0.0016 & \ph \\
SAM3 & A5 & 0.2962 & 0.0827 & 0.8322 & \ph & 0.4383 & 0.1106 & \ph & 0.4805 & \ph & \cellcolor{red!10}\textbf{0.4457} & \cellcolor{red!10}\textbf{0.4085} & 0.5032 & 0.0100 & \ph \\
SAM3$^\dagger$ & A5 & 0.2002 & 0.0189 & 0.8101 & \ph & 0.6146 & 0.1679 & \ph & 0.2307 & \ph & 0.4430 & 0.3759 & 0.7085 & 0.2157 & \ph \\
\midrule
\multicolumn{16}{c}{\textit{F1}} \\
GeoChat & A1 & 0.0158 & 0.0279 & 0.7114 & \ph & \cellcolor{red!10}\textbf{0.6684} & 0.1421 & \ph & 0.1440 & \ph & 0.4521 & 0.4660 & 0.4709 & 0.0047 & \ph \\
RS-LLaVA & A1 & 0.3791 & 0.2540 & 0.8349 & \ph & 0.6597 & 0.1364 & \ph & 0.3508 & \ph & 0.3738 & \cellcolor{red!10}\textbf{0.4939} & 0.6963 & 0.1890 & \ph \\
VHM & A1 & \cellcolor{red!10}\textbf{0.6972} & \cellcolor{red!10}\textbf{0.4049} & \cellcolor{red!10}\textbf{0.8575} & \ph & 0.5971 & 0.1499 & \ph & \cellcolor{red!10}\textbf{0.5465} & \ph & 0.0000 & 0.0000 & 0.2498 & 0.2014 & \ph \\
RemoteCLIP & A2 & 0.1077 & 0.0272 & 0.7627 & \ph & 0.3172 & 0.0000 & \ph & 0.2656 & \ph & \cellcolor{red!10}\textbf{0.4587} & 0.4738 & 0.4909 & 0.1148 & \ph \\
RemoteCLIP & A3 & 0.0158 & 0.0071 & 0.7113 & \ph & 0.5855 & 0.0335 & \ph & 0.1121 & \ph & 0.4441 & 0.4388 & 0.4635 & 0.0047 & \ph \\
GeoRSCLIP & A2 & 0.0063 & 0.0221 & 0.8228 & \ph & 0.4005 & 0.1523 & \ph & 0.2288 & \ph & 0.4581 & 0.4497 & 0.3235 & 0.0293 & \ph \\
GeoRSCLIP & A3 & 0.0158 & 0.0071 & 0.7113 & \ph & 0.5855 & 0.0335 & \ph & 0.1121 & \ph & 0.4441 & 0.4388 & 0.4635 & 0.0047 & \ph \\
DOFA-CLIP & A2 & 0.0000 & 0.2486 & 0.8121 & \ph & 0.1070 & 0.0000 & \ph & 0.1484 & \ph & 0.0000 & 0.0000 & 0.2786 & 0.0000 & \ph \\
RemoteSAM & A4 & 0.6277 & 0.0010 & 0.8014 & \ph & 0.5846 & 0.0337 & \ph & 0.0769 & \ph & 0.2416 & 0.3306 & 0.4201 & 0.0017 & \ph \\
RemoteSAM & A5 & 0.5949 & 0.0093 & 0.7830 & \ph & 0.5843 & 0.0338 & \ph & 0.0928 & \ph & 0.3819 & 0.4358 & 0.4927 & 0.0074 & \ph \\
SAM3 & A5 & 0.4200 & 0.1472 & 0.8359 & \ph & 0.0775 & 0.1337 & \ph & 0.4742 & \ph & 0.0000 & 0.0000 & 0.1874 & 0.0000 & \ph \\
SAM3$^\dagger$ & A5 & 0.2979 & 0.0000 & 0.8468 & \ph & 0.4374 & \cellcolor{red!10}\textbf{0.1652} & \ph & 0.0041 & \ph & 0.3397 & 0.0202 & \cellcolor{red!10}\textbf{0.6997} & \cellcolor{red!10}\textbf{0.3710} & \ph \\

\bottomrule
\end{tabular}%
}
\end{table*}

\begin{table*}[t]
\centering
\scriptsize
\caption{Per-class precision and recall of image-level VLM-based recognition, organized by split. Within each split, precision is reported in the upper block and recall in the lower block. Protocol~A1 denotes binary QA, Protocol~A2 denotes contrastive scoring, Protocol~A3 denotes positive-threshold scoring, Protocol~A4 denotes model-native multi-label classification, and Protocol~A5 denotes segmentation-derived recognition. SAM3 uses class-name prompts, while SAM3$^\dagger$ uses class-description prompts.}
\label{tab:app_vqa_classwise_pr}
\setlength{\tabcolsep}{2.4pt}
\renewcommand{\arraystretch}{1}
\resizebox{\textwidth}{!}{%
\begin{tabular}{lccccccccccccccc}
\toprule
\multirow{2}{*}{Model} &
\multirow{2}{*}{Protocol} &
\multirow{2}{*}{Buildings} &
Mining & Primary & Heavy & Water & Agricultural & Compact & Gravel & \multirow{2}{*}{Grass} & Type 1 & Type 2 & Bare & \multirow{2}{*}{Sluices} & \multirow{2}{*}{Vehicles} \\
& &
& rafts & forests & machinery & bodies & crops & mounds & mounds & & regeneration & regeneration & ground & & \\
\midrule

\rowcolor{gray!15}
\multicolumn{16}{c}{\textbf{Train}} \\
\multicolumn{16}{c}{\textit{Precision}} \\
GeoChat & A1 & 0.0096 & 0.0020 & 0.3697 & 0.0005 & 0.6677 & 0.1258 & 0.2275 & 0.0211 & 0.0141 & 0.3168 & 0.5262 & 0.3793 & 0.0002 & 0.0004 \\
RS-LLaVA & A1 & 0.8765 & 0.2857 & 0.4699 & 0.4038 & 0.8400 & 0.0419 & 0.0827 & 0.1969 & 0.0216 & 0.2412 & 0.5491 & 0.7922 & 0.0800 & 0.0848 \\
VHM & A1 & 0.8285 & 0.2105 & 0.5909 & 0.1875 & 0.8222 & 0.0945 & 0.0800 & 0.0878 & 0.0188 & 0.0000 & 0.0000 & 0.9502 & 0.0152 & 0.1226 \\
RemoteCLIP & A2 & 0.0396 & 0.0027 & 0.4510 & 0.0010 & 0.9225 & 1.0000 & 0.1467 & 0.0517 & 0.0173 & 0.3139 & 0.5228 & 0.3839 & 0.0000 & 0.0000 \\
RemoteCLIP & A3 & 0.0096 & 0.0005 & 0.3694 & 0.0004 & 0.3273 & 0.0205 & 0.1303 & 0.0178 & 0.0140 & 0.3126 & 0.5120 & 0.3732 & 0.0002 & 0.0004 \\
GeoRSCLIP & A2 & 1.0000 & 0.0009 & 0.4234 & 0.0561 & 0.8646 & 0.1220 & 0.2463 & 0.0298 & 0.0157 & 0.3145 & 0.5138 & 0.7293 & 0.0008 & 0.0061 \\
GeoRSCLIP & A3 & 0.0096 & 0.0005 & 0.3694 & 0.0004 & 0.3273 & 0.0205 & 0.1303 & 0.0178 & 0.0140 & 0.3126 & 0.5120 & 0.3732 & 0.0002 & 0.0004 \\
DOFA-CLIP & A2 & 0.0000 & 0.0341 & 0.6194 & 0.1212 & 0.9700 & 0.0000 & 0.0000 & 0.0350 & 0.8000 & 0.0000 & 0.0000 & 0.8698 & 0.0000 & 0.0000 \\
RemoteSAM & A4 & 0.6948 & 0.0000 & 0.3968 & 0.0052 & 0.3278 & 0.0205 & 0.1304 & 0.0118 & 0.0113 & 0.2543 & 0.4833 & 0.3476 & 0.0005 & 0.0057 \\
RemoteSAM & A5 & 0.6124 & 0.0007 & 0.3862 & 0.0010 & 0.3264 & 0.0205 & 0.1323 & 0.0147 & 0.0128 & 0.3732 & 0.5175 & 0.3836 & 0.0004 & 0.0020 \\
SAM3 & A5 & 0.8697 & 0.1538 & 0.6470 & 0.5238 & 0.9579 & 0.1020 & 0.1518 & 0.2686 & 0.0392 & 0.0000 & 0.0000 & 0.8387 & 0.0000 & 0.1348 \\
SAM3$^\dagger$ & A5 & 0.9053 & 0.0000 & 0.5698 & 0.7222 & 0.9493 & 0.0748 & 0.4444 & 0.0166 & 0.3899 & 0.3886 & 0.5869 & 0.7639 & 0.1800 & 0.2069 \\
\midrule
\multicolumn{16}{c}{\textit{Recall}} \\
GeoChat & A1 & 1.0000 & 0.9697 & 1.0000 & 0.9615 & 0.6915 & 0.2915 & 0.8401 & 0.9974 & 1.0000 & 0.9940 & 0.9708 & 0.9996 & 1.0000 & 1.0000 \\
RS-LLaVA & A1 & 0.3476 & 0.1212 & 0.9822 & 0.8077 & 0.6860 & 0.4028 & 0.1278 & 0.2300 & 0.9145 & 0.5662 & 0.8593 & 0.8017 & 0.2500 & 0.5185 \\
VHM & A1 & 0.5419 & 0.2424 & 0.9244 & 0.9231 & 0.5648 & 0.2700 & 0.0005 & 0.6363 & 0.9340 & 0.0000 & 0.0000 & 0.1981 & 0.5625 & 0.4815 \\
RemoteCLIP & A2 & 0.2306 & 0.5758 & 0.9540 & 0.9231 & 0.3356 & 0.0007 & 0.3112 & 0.3560 & 0.8799 & 0.9377 & 0.8870 & 0.9560 & 0.0000 & 0.0000 \\
RemoteCLIP & A3 & 1.0000 & 1.0000 & 1.0000 & 1.0000 & 1.0000 & 1.0000 & 1.0000 & 1.0000 & 1.0000 & 1.0000 & 1.0000 & 1.0000 & 1.0000 & 1.0000 \\
GeoRSCLIP & A2 & 0.0079 & 0.7576 & 0.9919 & 0.4615 & 0.4959 & 0.4636 & 0.1481 & 0.8620 & 0.9870 & 0.9973 & 0.9996 & 0.2162 & 0.1875 & 0.2593 \\
GeoRSCLIP & A3 & 1.0000 & 1.0000 & 1.0000 & 1.0000 & 1.0000 & 1.0000 & 1.0000 & 1.0000 & 1.0000 & 1.0000 & 1.0000 & 1.0000 & 1.0000 & 1.0000 \\
DOFA-CLIP & A2 & 0.0000 & 0.0909 & 0.8799 & 0.1538 & 0.0210 & 0.0000 & 0.0000 & 0.1218 & 0.0043 & 0.0000 & 0.0000 & 0.2821 & 0.0000 & 0.0000 \\
RemoteSAM & A4 & 0.5719 & 0.0000 & 0.9965 & 0.1923 & 0.9952 & 0.9948 & 0.9957 & 0.5954 & 0.7392 & 0.2197 & 0.5287 & 0.8729 & 0.0625 & 0.4444 \\
RemoteSAM & A5 & 0.6066 & 0.7879 & 0.9974 & 0.6923 & 0.9872 & 0.9978 & 0.9895 & 0.7606 & 0.8647 & 0.3875 & 0.7745 & 0.8517 & 0.5000 & 0.5185 \\
SAM3 & A5 & 0.3902 & 0.0606 & 0.8645 & 0.8462 & 0.1226 & 0.1031 & 0.0020 & 0.2027 & 0.7457 & 0.0000 & 0.0000 & 0.1268 & 0.0000 & 0.4444 \\
SAM3$^\dagger$ & A5 & 0.3476 & 0.0000 & 0.9012 & 0.5000 & 0.4681 & 0.2990 & 0.0317 & 0.0026 & 0.0920 & 0.2232 & 0.0098 & 0.6816 & 0.5625 & 0.4444 \\

\midrule
\rowcolor{gray!15}
\multicolumn{16}{c}{\textbf{Validation}} \\
\multicolumn{16}{c}{\textit{Precision}} \\
GeoChat & A1 & 0.0064 & 0.0042 & 0.7669 & 0.0008 & 0.6181 & 0.4173 & 0.0865 & 0.0361 & 0.0700 & 0.2332 & 0.2802 & 0.1706 & 0.0011 & \ph \\
RS-LLaVA & A1 & 0.6061 & 1.0000 & 0.8159 & 0.8571 & 0.6829 & 0.2388 & 0.0755 & 0.2679 & 0.1604 & 0.2073 & 0.2788 & 0.5604 & 0.0000 & \ph \\
VHM & A1 & 0.4495 & 0.3077 & 0.9112 & 0.3750 & 0.6802 & 0.4472 & 0.0667 & 0.1990 & 0.0905 & 0.0000 & 0.0000 & 0.8874 & 0.0376 & \ph \\
RemoteCLIP & A2 & 0.0066 & 0.0178 & 0.7844 & 0.0006 & 0.8034 & 0.0000 & 0.0558 & 0.1291 & 0.0760 & 0.2342 & 0.2778 & 0.1729 & 0.0000 & \ph \\
RemoteCLIP & A3 & 0.0064 & 0.0011 & 0.7669 & 0.0006 & 0.2011 & 0.0622 & 0.0325 & 0.0222 & 0.0700 & 0.2327 & 0.2796 & 0.1644 & 0.0011 & \ph \\
GeoRSCLIP & A2 & 0.0000 & 0.0024 & 0.7927 & 0.0847 & 0.8255 & 0.3713 & 0.3611 & 0.0343 & 0.0765 & 0.2357 & 0.2822 & 0.4156 & 0.0043 & \ph \\
GeoRSCLIP & A3 & 0.0064 & 0.0011 & 0.7669 & 0.0006 & 0.2011 & 0.0622 & 0.0325 & 0.0222 & 0.0700 & 0.2327 & 0.2796 & 0.1644 & 0.0011 & \ph \\
DOFA-CLIP & A2 & 0.0000 & 0.2667 & 0.8831 & 1.0000 & 0.9646 & 0.0000 & 1.0000 & 0.0672 & 1.0000 & 0.0000 & 0.0000 & 0.7167 & 0.0000 & \ph \\
RemoteSAM & A4 & 0.3895 & 0.0000 & 0.7943 & 0.0000 & 0.2006 & 0.0621 & 0.0324 & 0.0161 & 0.0608 & 0.1874 & 0.1959 & 0.1538 & 0.0000 & \ph \\
RemoteSAM & A5 & 0.3497 & 0.0013 & 0.7856 & 0.0013 & 0.2011 & 0.0623 & 0.0317 & 0.0175 & 0.0689 & 0.2888 & 0.2691 & 0.1631 & 0.0028 & \ph \\
SAM3 & A5 & 0.5893 & 0.2500 & 0.8871 & 0.4667 & 0.8438 & 0.6522 & 0.0000 & 0.4833 & 0.3954 & 0.0000 & 0.0000 & 0.6952 & 0.0000 & \ph \\
SAM3$^\dagger$ & A5 & 0.5714 & 0.0000 & 0.8859 & 0.7778 & 0.8735 & 0.3011 & 0.0682 & 0.5000 & 0.8788 & 0.4878 & 0.4662 & 0.5780 & 0.1304 & \ph \\
\midrule
\multicolumn{16}{c}{\textit{Recall}} \\
GeoChat & A1 & 1.0000 & 0.9444 & 1.0000 & 1.0000 & 0.6722 & 0.4085 & 0.7904 & 0.9887 & 1.0000 & 0.9997 & 0.9875 & 0.9992 & 1.0000 & \ph \\
RS-LLaVA & A1 & 0.1942 & 0.1667 & 0.9796 & 0.6667 & 0.6510 & 0.5926 & 0.2481 & 0.1577 & 0.9634 & 0.7982 & 0.9340 & 0.6826 & 0.0000 & \ph \\
VHM & A1 & 0.4757 & 0.2222 & 0.8898 & 0.6667 & 0.5624 & 0.2555 & 0.0019 & 0.4563 & 0.9920 & 0.0000 & 0.0000 & 0.1709 & 0.4444 & \ph \\
RemoteCLIP & A2 & 0.0777 & 0.6667 & 0.9141 & 0.6667 & 0.3673 & 0.0000 & 0.3635 & 0.5155 & 0.9794 & 0.9747 & 0.8873 & 0.9555 & 0.0000 & \ph \\
RemoteCLIP & A3 & 1.0000 & 1.0000 & 1.0000 & 1.0000 & 1.0000 & 1.0000 & 1.0000 & 1.0000 & 1.0000 & 1.0000 & 1.0000 & 1.0000 & 1.0000 & \ph \\
GeoRSCLIP & A2 & 0.0000 & 0.6111 & 0.9873 & 0.5556 & 0.2737 & 0.3742 & 0.1000 & 0.6310 & 1.0000 & 0.9997 & 1.0000 & 0.1865 & 0.1111 & \ph \\
GeoRSCLIP & A3 & 1.0000 & 1.0000 & 1.0000 & 1.0000 & 1.0000 & 1.0000 & 1.0000 & 1.0000 & 1.0000 & 1.0000 & 1.0000 & 1.0000 & 1.0000 & \ph \\
DOFA-CLIP & A2 & 0.0000 & 0.2222 & 0.8055 & 0.4444 & 0.0339 & 0.0000 & 0.0058 & 0.0451 & 0.0027 & 0.0000 & 0.0000 & 0.2089 & 0.0000 & \ph \\
RemoteSAM & A4 & 0.3592 & 0.0000 & 0.9897 & 0.0000 & 0.9947 & 0.9940 & 0.9923 & 0.6676 & 0.8043 & 0.1618 & 0.3635 & 0.8938 & 0.0000 & \ph \\
RemoteSAM & A5 & 0.4854 & 0.6667 & 0.9914 & 0.4444 & 0.9963 & 1.0000 & 0.9654 & 0.7577 & 0.9285 & 0.2292 & 0.7133 & 0.8687 & 0.5000 & \ph \\
SAM3 & A5 & 0.3204 & 0.0556 & 0.8932 & 0.7778 & 0.0420 & 0.0755 & 0.0000 & 0.1634 & 0.7739 & 0.0000 & 0.0000 & 0.1823 & 0.0000 & \ph \\
SAM3$^\dagger$ & A5 & 0.1942 & 0.0000 & 0.9262 & 0.7778 & 0.3844 & 0.5332 & 0.0115 & 0.0141 & 0.2332 & 0.1457 & 0.0139 & 0.7782 & 0.3333 & \ph \\

\midrule
\rowcolor{gray!15}
\multicolumn{16}{c}{\textbf{Test}} \\
\multicolumn{16}{c}{\textit{Precision}} \\
GeoChat & A1 & 0.0080 & 0.0142 & 0.5521 & \ph & 0.8309 & 0.0877 & \ph & 0.0776 & \ph & 0.2929 & 0.3054 & 0.3080 & 0.0023 & \ph \\
RS-LLaVA & A1 & 0.9268 & 0.5217 & 0.7377 & \ph & 0.8608 & 0.0766 & \ph & 0.8100 & \ph & 0.2625 & 0.3391 & 0.6407 & 0.3636 & \ph \\
VHM & A1 & 0.7952 & 0.4808 & 0.8416 & \ph & 0.8460 & 0.1041 & \ph & 0.4659 & \ph & 0.0000 & 0.0000 & 0.9503 & 0.1208 & \ph \\
RemoteCLIP & A2 & 0.0812 & 0.0145 & 0.6551 & \ph & 0.9098 & 0.0000 & \ph & 0.2039 & \ph & 0.3039 & 0.3198 & 0.3313 & 0.0933 & \ph \\
RemoteCLIP & A3 & 0.0080 & 0.0036 & 0.5519 & \ph & 0.4139 & 0.0171 & \ph & 0.0594 & \ph & 0.2854 & 0.2811 & 0.3017 & 0.0023 & \ph \\
GeoRSCLIP & A2 & 1.0000 & 0.0112 & 0.7099 & \ph & 0.9232 & 0.0899 & \ph & 0.1311 & \ph & 0.2984 & 0.2901 & 0.8129 & 0.0153 & \ph \\
GeoRSCLIP & A3 & 0.0080 & 0.0036 & 0.5519 & \ph & 0.4139 & 0.0171 & \ph & 0.0594 & \ph & 0.2854 & 0.2811 & 0.3017 & 0.0023 & \ph \\
DOFA-CLIP & A2 & 0.0000 & 0.5476 & 0.8192 & \ph & 0.9926 & 0.0000 & \ph & 0.3456 & \ph & 0.0000 & 0.0000 & 0.5809 & 0.0000 & \ph \\
RemoteSAM & A4 & 0.7224 & 0.0005 & 0.6726 & \ph & 0.4139 & 0.0171 & \ph & 0.0411 & \ph & 0.2360 & 0.2441 & 0.2773 & 0.0009 & \ph \\
RemoteSAM & A5 & 0.6106 & 0.0047 & 0.6459 & \ph & 0.4135 & 0.0172 & \ph & 0.0494 & \ph & 0.4006 & 0.2989 & 0.3501 & 0.0037 & \ph \\
SAM3 & A5 & 0.8800 & 0.6000 & 0.8244 & \ph & 0.9767 & 0.2107 & \ph & 0.7489 & \ph & 0.0000 & 0.0000 & 0.8866 & 0.0000 & \ph \\
SAM3$^\dagger$ & A5 & 0.9825 & 0.0000 & 0.7990 & \ph & 0.9637 & 0.1002 & \ph & 0.0847 & \ph & 0.5420 & 0.2404 & 0.7903 & 0.2987 & \ph \\
\midrule
\multicolumn{16}{c}{\textit{Recall}} \\
GeoChat & A1 & 1.0000 & 0.9790 & 1.0000 & \ph & 0.5590 & 0.3728 & \ph & 0.9979 & \ph & 0.9904 & 0.9833 & 0.9993 & 1.0000 & \ph \\
RS-LLaVA & A1 & 0.2382 & 0.1678 & 0.9616 & \ph & 0.5347 & 0.6213 & \ph & 0.2239 & \ph & 0.6489 & 0.9084 & 0.7625 & 0.1277 & \ph \\
VHM & A1 & 0.6207 & 0.3497 & 0.8741 & \ph & 0.4613 & 0.2675 & \ph & 0.6606 & \ph & 0.0000 & 0.0000 & 0.1438 & 0.6064 & \ph \\
RemoteCLIP & A2 & 0.1599 & 0.2308 & 0.9126 & \ph & 0.1921 & 0.0000 & \ph & 0.3809 & \ph & 0.9352 & 0.9139 & 0.9474 & 0.1489 & \ph \\
RemoteCLIP & A3 & 1.0000 & 1.0000 & 1.0000 & \ph & 1.0000 & 1.0000 & \ph & 1.0000 & \ph & 1.0000 & 1.0000 & 1.0000 & 1.0000 & \ph \\
GeoRSCLIP & A2 & 0.0031 & 0.9510 & 0.9784 & \ph & 0.2557 & 0.5000 & \ph & 0.8992 & \ph & 0.9854 & 0.9995 & 0.2019 & 0.3191 & \ph \\
GeoRSCLIP & A3 & 1.0000 & 1.0000 & 1.0000 & \ph & 1.0000 & 1.0000 & \ph & 1.0000 & \ph & 1.0000 & 1.0000 & 1.0000 & 1.0000 & \ph \\
DOFA-CLIP & A2 & 0.0000 & 0.1608 & 0.8051 & \ph & 0.0565 & 0.0000 & \ph & 0.0945 & \ph & 0.0000 & 0.0000 & 0.1832 & 0.0000 & \ph \\
RemoteSAM & A4 & 0.5549 & 0.0140 & 0.9911 & \ph & 0.9949 & 1.0000 & \ph & 0.6002 & \ph & 0.2474 & 0.5121 & 0.8662 & 0.0213 & \ph \\
RemoteSAM & A5 & 0.5799 & 0.6573 & 0.9941 & \ph & 0.9955 & 1.0000 & \ph & 0.7690 & \ph & 0.3649 & 0.8042 & 0.8311 & 0.4787 & \ph \\
SAM3 & A5 & 0.2759 & 0.0839 & 0.8476 & \ph & 0.0404 & 0.0980 & \ph & 0.3469 & \ph & 0.0000 & 0.0000 & 0.1048 & 0.0000 & \ph \\
SAM3$^\dagger$ & A5 & 0.1755 & 0.0000 & 0.9008 & \ph & 0.2829 & 0.4693 & \ph & 0.0021 & \ph & 0.2474 & 0.0106 & 0.6277 & 0.4894 & \ph \\

\bottomrule
\end{tabular}%
}
\end{table*}

\begin{table*}[t]
\centering
\scriptsize
\caption{Per-class prompted segmentation results under Protocol~B. Following the semantic segmentation appendix, the table reports class-wise IoU. SAM3 uses class-name prompts, while SAM3$^\dagger$ uses class-description prompts.}
\label{tab:app_vqa_seg_classwise}
\setlength{\tabcolsep}{2.6pt}
\renewcommand{\arraystretch}{1.08}
\resizebox{\textwidth}{!}{%
\begin{tabular}{lccccccccccccccc}
\toprule
\multirow{2}{*}{Model} &
\multirow{2}{*}{Protocol} &
\multirow{2}{*}{Buildings} &
Mining & Primary & Heavy & Water & Agricultural & Compact & Gravel & \multirow{2}{*}{Grass} & Type 1 & Type 2 & Bare & \multirow{2}{*}{Sluices} & \multirow{2}{*}{Vehicles} \\
& &
& rafts & forests & machinery & bodies & crops & mounds & mounds & & regeneration & regeneration & ground & & \\
\midrule

\rowcolor{gray!15}
\multicolumn{16}{c}{\textbf{Per-class IoU (Train)}} \\
RemoteSAM & B & \cellcolor{red!10}\textbf{0.4802} & 0.0000 & 0.1034 & 0.0020 & 0.2548 & 0.0133 & \cellcolor{red!10}\textbf{0.0779} & 0.0001 & 0.0007 & 0.0005 & \cellcolor{red!10}\textbf{0.0041} & 0.0115 & 0.0000 & 0.0081 \\
SAM3      & B & 0.3977 & \cellcolor{red!10}\textbf{0.0992} & \cellcolor{red!10}\textbf{0.5845} & 0.2632 & 0.0683 & 0.0280 & 0.0001 & \cellcolor{red!10}\textbf{0.0978} & \cellcolor{red!10}\textbf{0.0569} & 0.0000 & 0.0000 & 0.0665 & 0.0000 & 0.0503 \\
SAM3$^\dagger$ & B & 0.3838 & 0.0000 & 0.5705 & \cellcolor{red!10}\textbf{0.3027} & \cellcolor{red!10}\textbf{0.4066} & \cellcolor{red!10}\textbf{0.0587} & 0.0187 & 0.0001 & 0.0532 & \cellcolor{red!10}\textbf{0.0506} & 0.0009 & \cellcolor{red!10}\textbf{0.3422} & \cellcolor{red!10}\textbf{0.0841} & \cellcolor{red!10}\textbf{0.0687} \\

\midrule
\rowcolor{gray!15}
\multicolumn{16}{c}{\textbf{Per-class IoU (Validation)}} \\
RemoteSAM & B & \cellcolor{red!10}\textbf{0.2553} & 0.0002 & 0.0637 & 0.0000 & 0.1986 & 0.0558 & \cellcolor{red!10}\textbf{0.0150} & 0.0001 & 0.0048 & 0.0004 & \cellcolor{red!10}\textbf{0.0055} & 0.0087 & 0.0000 & \ph \\
SAM3      & B & 0.2325 & \cellcolor{red!10}\textbf{0.0750} & 0.7765 & 0.1922 & 0.0244 & 0.0237 & 0.0000 & \cellcolor{red!10}\textbf{0.1397} & \cellcolor{red!10}\textbf{0.4220} & 0.0000 & 0.0000 & 0.1086 & 0.0000 & \ph \\
SAM3$^\dagger$ & B & 0.1490 & 0.0000 & \cellcolor{red!10}\textbf{0.8020} & \cellcolor{red!10}\textbf{0.1993} & \cellcolor{red!10}\textbf{0.2706} & \cellcolor{red!10}\textbf{0.2669} & 0.0009 & 0.0000 & 0.1480 & \cellcolor{red!10}\textbf{0.0429} & 0.0016 & \cellcolor{red!10}\textbf{0.2922} & \cellcolor{red!10}\textbf{0.0830} & \ph \\

\midrule
\rowcolor{gray!15}
\multicolumn{16}{c}{\textbf{Per-class IoU (Test)}} \\
RemoteSAM & B & \cellcolor{red!10}\textbf{0.3930} & 0.0006 & 0.0941 & \ph & \cellcolor{red!10}\textbf{0.2494} & 0.0089 & \ph & 0.0005 & \ph & 0.0002 & \cellcolor{red!10}\textbf{0.0044} & 0.0157 & 0.0004 & \ph \\
SAM3      & B & 0.3159 & \cellcolor{red!10}\textbf{0.0903} & 0.7332 & \ph & 0.0163 & 0.0274 & \ph & \cellcolor{red!10}\textbf{0.2581} & \ph & 0.0000 & 0.0000 & 0.0609 & 0.0000 & \ph \\
SAM3$^\dagger$ & B & 0.1735 & 0.0000 & \cellcolor{red!10}\textbf{0.7399} & \ph & 0.1537 & \cellcolor{red!10}\textbf{0.1017} & \ph & 0.0000 & \ph & \cellcolor{red!10}\textbf{0.0940} & 0.0005 & \cellcolor{red!10}\textbf{0.3529} & \cellcolor{red!10}\textbf{0.1255} & \ph \\

\bottomrule
\end{tabular}%
}
\end{table*}

\subsubsection{Per-class VLM-based Recognition Evaluation Performance}
\label{app_vqa_perclass}

Tables~\ref{tab:app_vqa_classwise_apf1} and~\ref{tab:app_vqa_classwise_pr} provide a more detailed class-wise view of image-level VLM-based recognition, while Table~\ref{tab:app_vqa_seg_classwise} reports the class-wise IoU results for prompted segmentation. Overall, these tables show that the benchmark remains difficult at the class level, especially for sparse mining-related categories. Across splits, classes such as primary forest, water bodies, and bare ground are generally easier and more stable, whereas mining rafts, heavy machinery, sluices, and vehicles remain much more challenging. Agricultural crops and the regeneration categories are intermediate: they are more predictable than the rarest object classes, but still much less stable than the dominant land-cover categories. 

In Table~\ref{tab:app_vqa_classwise_apf1}, no single method family dominates all categories. The generative models in Protocol~A1 are generally more balanced on the dominant classes and give many of the strongest AP and F1 results across splits, which is consistent with their stronger overall recognition performance. The CLIP-style models are more uneven: under Protocol~A2 they can be competitive on selected categories, but their strengths are scattered rather than consistent across the label space. The SAM-based recognition variants are also mixed. In particular, SAM3 and SAM3$^\dagger$ become competitive on several semantically specific categories, but their gains are not uniform across all classes. This again shows that the overall ranking in Table~\ref{tab:app_vqa_overall_recognition} does not fully explain the detailed class-level behavior. 

Table~\ref{tab:app_vqa_classwise_pr} further clarifies these differences through the precision--recall decomposition. The clearest failure mode appears in the positive-threshold variants of RemoteCLIP and GeoRSCLIP: recall becomes extremely high across many classes, while precision remains much weaker, indicating that positive-only thresholding tends to over-predict class presence rather than cleanly separate positive and negative labels. By contrast, the contrastive variants in Protocol~A2 are more selective, and the resulting precision--recall trade-off is more meaningful. DOFA-CLIP shows a different profile again: it is often more selective than the softmax-based CLIP variants, but this selectivity sometimes comes with substantially lower recall on difficult categories. The generative models and SAM-based variants also show class-dependent trade-offs, but they do not exhibit the same recall-saturation behavior as the positive-threshold CLIP rows. Overall, the precision--recall table is useful because it shows whether a method is achieving its scores through genuine class discrimination or simply through broad positive prediction. 

The class-wise image-level results also reinforce the role of prompt quality. Although the improvement is not uniform for every class, SAM3$^\dagger$ often performs better than SAM3 on categories that benefit from richer semantic description, especially on several operationally relevant classes. This is consistent with the overall VQA results, where detailed prompts improved the SAM3-derived recognition metrics. In other words, better prompts do not guarantee improvement everywhere, but they do help the model align language and visual evidence more effectively on a meaningful subset of categories. 

Table~\ref{tab:app_vqa_seg_classwise} shows that the prompted segmentation results follow a similar but not identical pattern. Overall, the SAM3-based methods are stronger than RemoteSAM on more classes, but the advantage is not universal. RemoteSAM remains competitive, and sometimes stronger, on a few categories such as buildings and water bodies, while SAM3 and especially SAM3$^\dagger$ provide clearer advantages on several semantically specific classes, including primary forest, some regeneration-related categories, bare ground, sluices, and, depending on the split, agricultural crops or vehicles. At the same time, the two SAM3 variants do not behave identically: class-name prompts and detailed description prompts emphasize different categories, which again suggests that prompt formulation materially affects zero-shot segmentation quality. In summary, these class-wise results support the same conclusion as the earlier appendices: aggregate metrics are informative, but detailed per-class analysis is necessary to determine whether the gains come from dominant land-cover classes or from the smaller and operationally important mining-related categories.

\FloatBarrier

\section{Interactive Explorer Implementation}
\label{app:system}

\subsection{Overview}

\begin{figure*}[t]
    \centering
    \includegraphics[width=\textwidth]{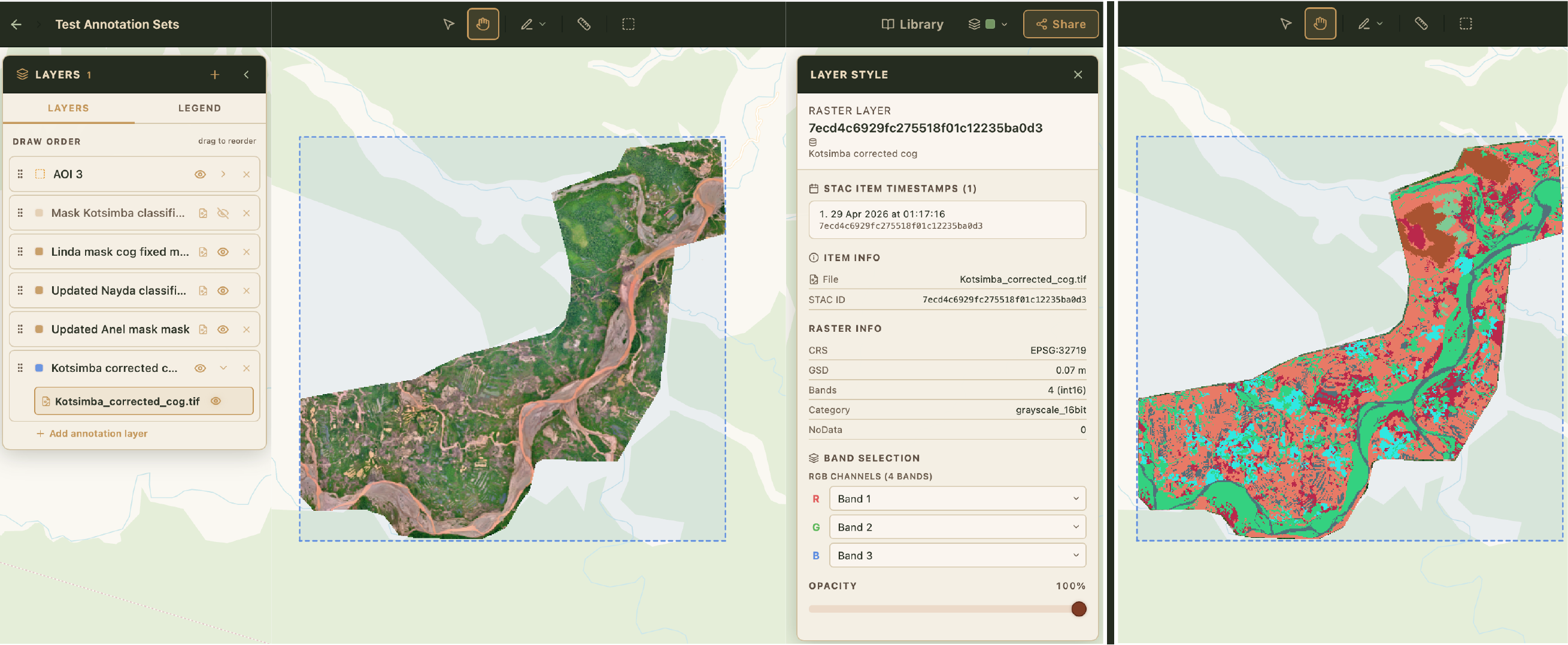}
    \caption{Interactive explorer interface for the Kotsimba site ($57{,}331 \times 53{,}961$ pixels). \textit{Left}: the main workspace, showing layer controls on the left, drawing tools at the top, and site information on the right. \textit{Right}: the same view with the ground-truth layer enabled, illustrating how annotations can be overlaid directly on the orthomosaic.}
    \label{fig:explorer-ui}
\end{figure*}

High-resolution UAV orthomosaics in ELDOR create substantial practical barriers for ecologists and other domain experts. For example, a single site such as Kotsimba, with size $57{,}331 \times 53{,}961$ pixels, can easily exceed the memory capacity of a typical local device, making direct loading and browsing difficult. Even when parts of the data can be viewed in GIS software such as ArcGIS, comparative analysis across sites, inspection of arbitrary sub-regions, and integration of annotations and predictions remain cumbersome. A second barrier is model use: running pretrained models on selected regions usually requires separate environments, data preprocessing, and custom inference scripts, which makes these models difficult to use in practice for non-technical domain experts. The interactive explorer (Figure~\ref{fig:explorer-ui}) addresses both challenges through a browser-based platform that integrates ELDOR orthomosaics, pixel-level annotations, and benchmark models within a single geographic interface. Users can browse large imagery at native resolution without downloading the data, define one or more regions of interest (ROIs) using standard drawing tools, run asynchronous inference with registered models directly on those ROIs, and overlay the resulting predictions as geospatial layers. In this way, the explorer reduces both data-access and model-use barriers for ELDOR, and supports flexible ROI-driven analysis of mining-related landscape change.

The explorer is implemented with a lightweight three-tier design for ELDOR data browsing and ROI-based inference. The frontend is implemented in Next.js~\cite{nextjs14} with Leaflet~\cite{leaflet} for interactive map rendering, while the backend is a FastAPI~\cite{fastapi} service responsible for metadata management, annotation retrieval, model registration, and inference orchestration. Persistent storage is divided between PostgreSQL with PostGIS~\cite{postgis} for vector geometries and catalog metadata, and MinIO~\cite{minio} for raster assets. Orthomosaics, ground-truth rasters, and prediction outputs are stored as Cloud-Optimized GeoTIFFs (COGs)~\cite{cog} and indexed as SpatioTemporal Asset Catalog (STAC)~\cite{stacspec} items in pgSTAC~\cite{pgstac}, allowing TiTiler-pgSTAC~\cite{titiler} to serve them dynamically as web map tiles. Model inference is handled asynchronously through Celery~\cite{celery} workers, which keeps the interface responsive even when large ROIs are being processed.

\subsection{Data Pipeline}
To support efficient browsing and ROI-based analysis of extremely large ELDOR orthomosaics, the explorer organizes ELDOR assets using COG-based storage, STAC indexing, a spatial database, and dynamic tile serving. These components make the orthomosaics, annotations, and prediction outputs addressable through a unified geospatial catalog.

\textbf{Catalog.} Each orthomosaic is registered as a STAC item in pgSTAC together with its spatial metadata, including bounding box, coordinate reference system, and ground sampling distance, while the underlying COG is stored in MinIO. The catalog acts as the main entry point for ELDOR raster assets, including the original orthomosaics, ground-truth rasters, and saved prediction outputs. Assets are then accessed through their STAC identifiers for tile serving, ROI clipping, and prediction registration.

\textbf{Ingestion.} The 12 ELDOR sites are preloaded into the explorer, but the platform also supports custom data ingestion. Users can upload new orthomosaics through the browser interface, after which the source GeoTIFFs are stored in MinIO and registered in PostGIS and pgSTAC. Once registered, these new assets can be accessed through the same catalog and visualization pipeline as the preloaded sites.

\textbf{Annotations.} Pixel-level annotations are organized through an annotation schema, which defines class names, styles, and descriptions, together with annotation sets that bind the schema to one or more datasets. The vector annotations themselves are stored as PostGIS geometries with spatial indexes, which enables efficient retrieval of the features intersecting the current map viewport. Figure~\ref{fig:explorer-single-roi} shows how the annotation schema is presented in the interface, including class names and visualization colors.

\textbf{Tile serving.} All raster layers, including orthomosaics, prediction overlays, and satellite basemaps, are served as web map tiles through TiTiler-pgSTAC. The frontend requests only the tiles needed for the current viewport, and TiTiler reads the corresponding COG ranges from MinIO through HTTP range requests. This design supports responsive pan and zoom even for multi-gigapixel rasters.

\subsection{Model Pipeline}

The model pipeline connects the interactive interface to registered inference services and returns the outputs as geospatial layers that can be visualized and reused within the same workspace. In practice, this pipeline covers four main components: model registration, user-triggered inference on selected ROIs, backend execution of the inference request, and persistent storage of the resulting predictions. Together, these steps allow the explorer to support interactive model use without requiring users to manage preprocessing, inference scripts, or output alignment themselves.

\textbf{Registration.} The backend maintains a model registry in which each entry records the model name, task type (such as semantic segmentation or multi-label classification), external inference endpoint, input specification (including patch size, normalization, band order, and overlap strategy), and output specification (including class indices and color mapping). Because the explorer relies on this registry rather than a fixed built-in model list, new model endpoints can be added and exposed through the same interface.

\begin{figure*}[t]
    \centering
    \includegraphics[width=\textwidth]
    {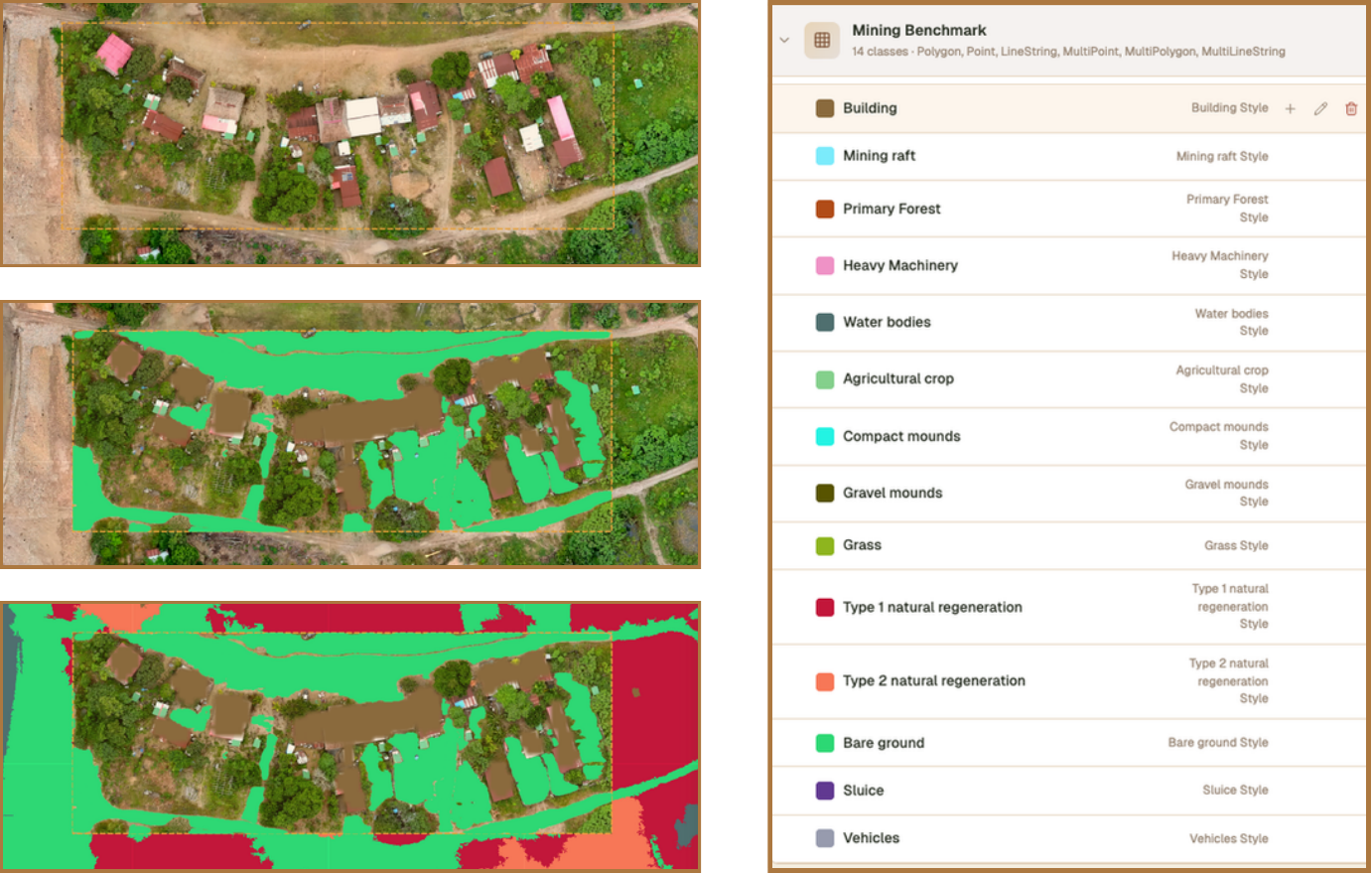}
    \caption{Single-ROI inference workflow in the interactive explorer. \textit{Left}: a three-step example showing (top) selection of a region of interest on the orthomosaic, (middle) the prediction generated for that selected region, and (bottom) comparison with the surrounding ground-truth layer. \textit{Right}: the editable annotation schema used by the explorer, including class names and visualization colors.}
    \label{fig:explorer-single-roi}
\end{figure*}

\begin{figure*}[t]
    \centering
    \includegraphics[width=\textwidth]{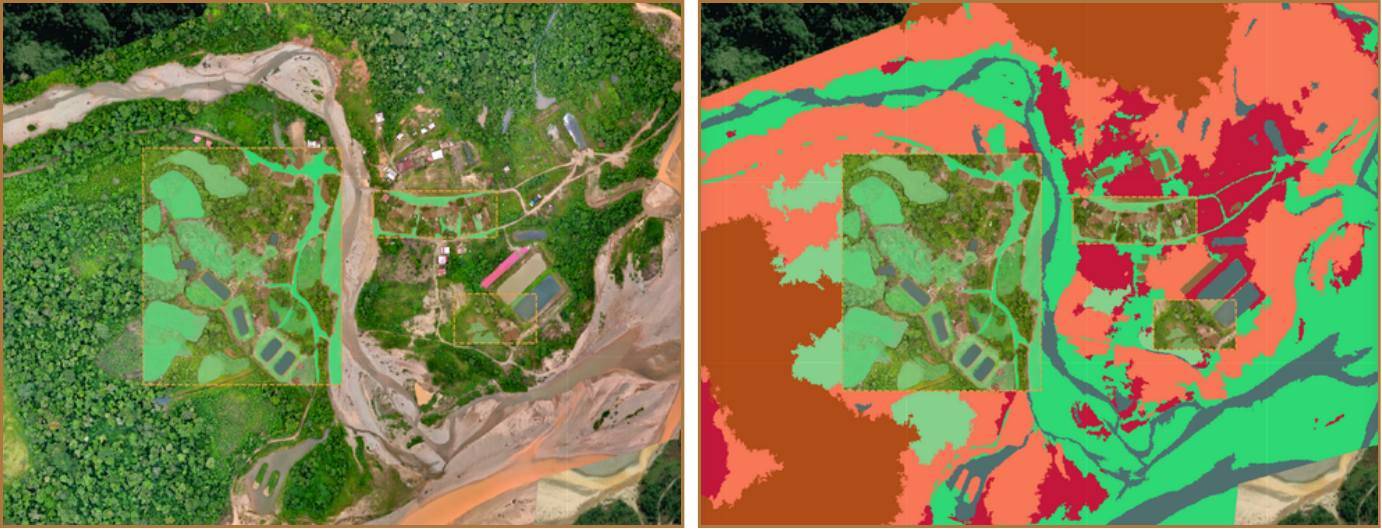}
    \caption{Multi-ROI inference in the interactive explorer. \textit{Left}: the interface with three separate user-defined regions of interest, each with its own prediction mask. \textit{Right}: the same view with the ground-truth layer enabled, allowing direct comparison between predictions inside the selected regions and annotations in the surrounding area.}
    \label{fig:explorer-multi-roi}
\end{figure*}

\textbf{On-demand inference workflow.} From the user’s perspective, analyzing a region follows a simple interactive workflow (Figures~\ref{fig:explorer-single-roi} and~\ref{fig:explorer-multi-roi}). First, the user selects an ELDOR site from the catalog, and the orthomosaic is streamed into the interface at native resolution (Figure~\ref{fig:explorer-ui}, left). Next, the user draws one or more regions of interest (ROIs) using the interface tools. Figure~\ref{fig:explorer-single-roi} (left, top) shows an example of a single ROI, while Figure~\ref{fig:explorer-multi-roi} (left) shows multiple disjoint ROIs within the same view. The user then selects a registered model and starts inference. A dedicated job panel reports the current status of the task, such as queued, running, or completed. When inference finishes, the prediction masks are rendered as geospatial overlays, which allow the user to inspect the model output directly within the selected ROI and compare it with the surrounding ground-truth layer by toggling visibility (Figure~\ref{fig:explorer-single-roi}, left, middle and bottom). The same workflow naturally extends to multiple ROIs, enabling the user to evaluate several separate areas in the same workspace (Figure~\ref{fig:explorer-multi-roi}, right).

\textbf{Inference execution.} When inference is initiated, the FastAPI backend serializes the request, including the ROI geometries, model identifier, and tile parameters, and dispatches it as a Celery task. A worker then performs targeted reads from MinIO, retrieving only the COG tiles that intersect the requested ROIs. These data are assembled into input patches according to the model specification, forwarded to the external inference endpoint, and then reassembled into a single output mosaic aligned with each ROI geometry.

\textbf{Persistent predictions.} The output mosaic is saved as a new COG in MinIO and registered as a STAC item in a prediction collection, and indexed with its spatial extent and history information, including the source model, input ROI, and timestamp. This allows ELDOR predictions to be reloaded in later sessions, overlaid with different basemaps, compared with other model outputs, or downloaded for offline analysis. In this way, the explorer supports not only interactive inspection, but also the reuse, sharing, and comparison of model outputs in larger collaborative workflows.

\subsection{Visualization and Interaction}
The visualization layer is designed to support smooth exploration of large geospatial data together with annotations and model outputs. The map is rendered in Leaflet using canvas-based drawing, which keeps interaction responsive even when multiple raster and vector layers are displayed at the same time. A left-side panel (e.g., Figures~\ref{fig:explorer-ui}--\ref{fig:explorer-multi-roi}) lists all loaded sources, including orthomosaics, ground-truth annotation sets, satellite basemaps, and prediction overlays, with separate controls for visibility, opacity, and drawing order. Vector layers are styled by class using the annotation schema, so the same legend can be applied consistently to both ground truth and model predictions. A right-side panel provides contextual information for the current selection, including raster metadata and band information, class composition and area statistics for annotation sets, and per-class summaries for model outputs.

\section{Responsible Release and Intended Use}
\label{app:ethics}

Since ELDOR contains high-resolution UAV imagery of mining-affected landscapes, the release follows a responsible-use protocol approved by the data providers and domain collaborators. The dataset is intended for research, ecological monitoring, restoration-oriented analysis, and remote sensing benchmark development, rather than direct enforcement attribution or standalone decisions about illegal activity. Model predictions should be interpreted as decision-support outputs and require expert review and local contextual knowledge before operational use.

\section{Broader Impacts and Limitations}
\label{app:impact_limitations}

ELDOR is designed to support machine learning research on a high-impact environmental monitoring problem: illegal gold mining and its associated landscape disturbance in the Amazon rainforest. By providing fine-grained UAV annotations, benchmark tasks, baseline results, and an interactive explorer, this work can help the community study fine-scale but socially important environmental targets that are often missed by coarse satellite imagery. The interactive explorer also lowers the technical barrier for ecologists and domain experts, allowing them to inspect large orthomosaics, visualize annotations, run pretrained models, and examine predictions without directly managing the full machine learning pipeline.

Beyond binary mined-versus-unmined mapping, fine-grained object and land-cover predictions may help characterize the modality of land-use change. Land-use change occurs across wide landscapes, but it is often produced through local actions on individual parcels using different tools, materials, and extraction practices. In alluvial gold mining, structures such as heavy machinery, sluices, mining rafts, and mound-like processed material can indicate different modes and intensities of disturbance. Identifying such features may help domain experts reason about how a landscape is being modified, what types of impacts are likely to follow, and what forms of remediation or post-mining land use may be appropriate.

These predictions may also support environmental science analyses when combined with additional measurements or domain assumptions. For example, tailings-like mounds and other processed-material features can provide evidence about the amount of earth moved, the intensity of mining activity, and potential downstream sediment mobilization. If paired with elevation information or surface models, their spatial extent could further support estimates of processed volume, which may help assess impacts on drainage networks, river health, and related ecological or human-health outcomes. Because mercury is commonly used in alluvial gold processing, processed-material features may also provide indirect cues about mercury-related exposure, although such estimates require external calibration and expert validation. In this sense, object-level segmentation can provide information about not only where mining occurs, but also how disturbance is produced and how recovery may proceed.

The dataset and system may support practical applications such as rapid screening of mining disturbance, restoration assessment, field planning, conservation monitoring, and collaboration between computer scientists and environmental practitioners. For conservation, fine-grained UAV monitoring may help inspect large or difficult-to-access areas such as conservation easements, protected areas, and park buffer zones. For planning and regulatory contexts, aggregated predictions may help estimate areas under active disturbance, compare disturbance intensity across parcels or sites, and prioritize field verification, remediation, or restoration actions.

At the same time, ELDOR should be used as a decision-support resource rather than as definitive evidence of illegal activity or direct attribution to specific actors. Fine-grained object predictions may suggest activity modes or disturbance mechanisms, but they should not be used alone to infer responsibility, legal status, or practitioner identity. False positives may incorrectly flag non-mining disturbance, while false negatives may miss small or ambiguous mining-related structures. Fine-grained geospatial predictions may also expose sensitive site information if released or used without appropriate access control, aggregation, or local review. Therefore, downstream use should involve expert interpretation, local context, and safeguards for sensitive locations and stakeholders.

Several limitations remain. Although ELDOR covers more than 2,500 hectares and uses a site-level split, the current data are from the Madre de Dios region and do not yet establish generalization to other mining regions, forest types, seasons, or UAV acquisition settings. The benchmark is based on RGB UAV orthomosaics, so it does not yet use temporal, multispectral, SAR, LiDAR, or other complementary signals. The annotations are expensive to produce, and some classes remain visually ambiguous. The evaluation is also affected by severe class imbalance and rare mining-related objects, where partial localization may already be useful for rapid screening even if the full object mask is incomplete.

A natural future direction is to connect ELDOR with broader multi-scale and multimodal monitoring workflows. Public satellite data, such as Sentinel-1 SAR and Sentinel-2 multispectral imagery, could first be used to learn regional indicators of possible mining disturbance and identify candidate areas for closer inspection~\cite{cao2025superpixel, gupta2025mosaic, cao2025spatial, liu2023adaptive, cao2026integrating, camalan2022detecting}. UAV surveys could then be prioritized over these areas to obtain centimeter-level imagery and detect specific mining-related structures and disturbance patterns. VLM-based recognition and interactive human-in-the-loop tools may further reduce the effort required to adapt monitoring systems to new sites, although the present results show that zero-shot models still remain below the strongest fine-tuned methods. In this way, satellite-scale screening, UAV-scale fine-grained interpretation, and expert-guided validation can become complementary components of a practical monitoring pipeline.

The practical value of ELDOR for illegal gold mining monitoring will further increase if its models can be deployed directly on resource-constrained UAV platforms. Because UAVs can provide centimeter-level imagery over remote, difficult-to-access, and rapidly changing mining sites, efficient onboard processing could support rapid screening of active disturbance and help prioritize follow-up inspection in the field~\cite{zhu2025orthomosaics, cui2024palmprobnet, di2025toward}. To make this feasible, future work may combine memory-efficient streaming inference with model compression techniques such as pruning, quantization, and low-rank compression~\cite{shao2026flashsvd, wu2026flashsvd} to handle high-resolution imagery under limited memory and compute budgets. When real-time image restoration or related preprocessing is required, adaptive acceleration~\cite{jiang2025sada} and second-order diffusion speedups~\cite{wang2026zeus,wang2026accelerating} may further reduce latency. Structured reasoning frameworks~\cite{wang2026t2s} could then help organize model outputs into concise environmental reports, making ELDOR more useful as a practical decision-support tool for monitoring illegal gold mining.




\end{document}